%% file: main.tex
\documentclass[runningheads]{llncs}

\newcommand{\finalcopy}{\cvprfinalcopy}
\input{tex/plots}

\usepackage{graphicx}
\usepackage{comment}
\usepackage{amsmath,amssymb} 
\usepackage{mathtools}
\usepackage{xspace}
\usepackage{color}
\usepackage{float}
\usepackage{cite}
\usepackage{colortbl}
\usepackage{bbm}
\usepackage{enumitem}
\usepackage{booktabs}
\usepackage{multirow}
\usepackage{capt-of}

\usepackage{times}
\usepackage[normalem]{ulem}
\usepackage[colorlinks, citecolor=blue, urlcolor=pink]{hyperref}
\usepackage[caption=false,position=top]{subfig}

\usepackage[accsupp]{axessibility}  

\usepackage{lipsum}

\newcommand\blfootnote[1]{
  \begingroup
  \renewcommand\thefootnote{}\footnote{#1}
  \addtocounter{footnote}{-1}
  \endgroup
}

\input{tex/abbrev}
\input{tex/macro}

\begin{document}
\pagestyle{headings}
\mainmatter

\title{What to Hide from Your Students: 

Attention-Guided Masked Image Modeling} 

\titlerunning{Attention-Guided Masked Image Modeling}

\author{Ioannis Kakogeorgiou\inst{1} \and
        Spyros Gidaris\inst{2} \and
        Bill Psomas\inst{1} \and
        Yannis Avrithis\inst{3,4} \and \\
        Andrei Bursuc\inst{2} \and
        Konstantinos Karantzalos\inst{1} \and
        Nikos Komodakis\inst{5,6}}

\authorrunning{I. Kakogeorgiou \etal}

\institute{\textsuperscript{1}National Technical University of Athens \enspace 
\textsuperscript{2}valeo.ai\\
\textsuperscript{3}Institute of Advanced Research in Artificial Intelligence (IARAI) \enspace 
\textsuperscript{4}Athena RC\\
\textsuperscript{5}University of Crete \enspace 
\textsuperscript{6} IACM-Forth
}
\maketitle

\begin{abstract}
Transformers and masked language modeling are quickly being adopt\-ed and explored in computer vision as \emph{vision transformers} and \emph{masked image modeling} (MIM). In this work, we argue that image token masking differs from token masking in text, due to the amount and correlation of tokens in an image. In particular, to generate a challenging pretext task for MIM, we advocate a shift from random masking to informed masking.
We develop and exhibit this idea in the context of distillation-based MIM, where a teacher transformer encoder generates an attention map, which we use to guide masking for the student. 

We thus introduce a novel masking strategy, called \Ours, and we demonstrate its effectiveness over random masking for dense distillation-based MIM as well as plain distillation-based self-supervised learning on classification tokens. We confirm that \ours accelerates the learning process and improves the performance on a variety of downstream tasks. We provide the implementation code at \href{https://github.com/gkakogeorgiou/attmask}{https://github.com/gkakogeorgiou/attmask}.

\end{abstract}

{\blfootnote{{Correspondence: \email{gkakogeorgiou@central.ntua.gr}}}}

\input{tex/intro}

\input{tex/related}
\input{tex/method}
\input{tex/exp-setup}
\input{tex/exp-analysis}
\input{tex/exp-bench}
\input{tex/exp-ablation}
\input{tex/conclusion}
\input{tex/acknowledgments}

\bibliographystyle{splncs04}
\bibliography{egbib}

\input{tex/supp}

\end{document}

%% file: tex/plots.tex
\usepackage[dvipsnames,svgnames,x11names]{xcolor}
\usepackage{tikz}
\usetikzlibrary{arrows.meta,shapes,calc,matrix,fit,positioning,backgrounds,decorations.markings}
\usepackage{pgfplots}
\usepackage{pgfplotstable}
\pgfplotsset{compat=1.9}
\usepackage{xstring}

\usepgfplotslibrary{external}
\newcommand{\extfig}[2]{\tikzsetnextfilename{#1}{#2}}

\IfBeginWith*{\jobname}{fig/extern/}{\finalcopy}{}

\tikzstyle{every picture}+=[
	remember picture,
	every text node part/.style={align=center},
	every matrix/.append style={ampersand replacement=\&},
]
\tikzstyle{tight} = [inner sep=0pt,outer sep=0pt]
\tikzstyle{node}  = [draw,circle,tight,minimum size=12pt,anchor=center]
\tikzstyle{op}    = [draw,circle,tight]
\tikzstyle{dot}   = [fill,draw,circle,inner sep=1pt,outer sep=0]
\tikzstyle{pt}    = [fill,draw,circle,inner sep=1.5pt,outer sep=.2pt]
\tikzstyle{box}   = [draw,rectangle,inner sep=3pt]
\tikzstyle{high}  = [black!60]
\tikzstyle{group} = [high,box,opacity=.5]
\tikzstyle{dim1}  = [fill opacity=.3,text opacity=1]
\tikzstyle{dim2}  = [fill opacity=.5,text opacity=1]
\tikzstyle{dim3}  = [fill opacity=.7,text opacity=1]
\tikzstyle{rectc} = [tight,transform shape]
\tikzstyle{rect}  = [rectc,anchor=south west]

\newcommand{\leg}[1]{\addlegendentry{#1}}

\tikzset{every mark/.append style={solid}}
\pgfplotsset{
	grid=both, width=\columnwidth, try min ticks=5,
	every axis/.append style={font=\small},
	every axis plot/.append style={thick,mark=none,mark size=1.8,tension=0.18},
	legend cell align=left, legend style={fill opacity=0.8},
	xticklabel={\pgfmathprintnumber[assume math mode=true]{\tick}},
	yticklabel={\pgfmathprintnumber[assume math mode=true]{\tick}},
	nodes near coords math/.style={
		nodes near coords={\pgfmathprintnumber[assume math mode=true]{\pgfplotspointmeta}},
	},
}

\pgfplotsset{
	dash/.style={mark=o,dashed,opacity=0.6},
	dott/.style={mark=o,dotted,opacity=0.6},
	nolim/.style={enlargelimits=false},
	plain/.style={every axis plot/.append style={},nolim,grid=none},
}

\pgfdeclarelayer{bg4}
\pgfdeclarelayer{bg3}
\pgfdeclarelayer{bg2}
\pgfdeclarelayer{bg1}
\pgfdeclarelayer{fg1}
\pgfdeclarelayer{fg2}
\pgfdeclarelayer{fg3}
\pgfdeclarelayer{fg4}
\pgfsetlayers{bg4,bg3,bg2,bg1,main,fg1,fg2,fg3,fg4}

\pgfkeys{/tikz/.cd, aspect/.store in=\aspect, aspect=1}
\pgfkeys{/tikz/.cd, depth/.store in=\depth, depth=.5}
\pgfkeys{/tikz/.cd, stepx/.store in=\step, stepx=1}

\tikzstyle{geom} = [line join=bevel,aspect=1,depth=.5,z={(\depth*\aspect,\depth)}]
\tikzstyle{wire} = [geom,draw,thick]

\def\cx[#1,#2,#3]{#1}
\def\cy[#1,#2,#3]{#2}
\def\cz[#1,#2,#3]{#3}
\def\ex[#1,#2,#3]{#1,0,0}
\def\ey[#1,#2,#3]{0,#2,0}
\def\ez[#1,#2,#3]{0,0,#3}



%% file: tex/abbrev.tex
\newcommand{\eq}[1]{(\ref{eq:#1})}

\newcommand{\Th}[1]{\textsc{#1}}
\newcommand{\mr}[2]{\multirow{#1}{*}{#2}}
\newcommand{\mc}[2]{\multicolumn{#1}{c}{#2}}

\newcommand{\red}[1]{{\textcolor{red}{#1}}}

\newcommand{\citeme}[1]{\red{[XX]}}
\newcommand{\refme}[1]{\red{(XX)}}

\newcommand{\fig}[2][1]{\includegraphics[width=#1\linewidth]{fig/#2}}

\newcommand{\tran}{^\top}

\newcommand{\ind}{\mathbbm{1}}

\newcommand{\real}{\mathbb{R}}

\newcommand{\mif}{\textrm{if}\ }
\newcommand{\other}{\textrm{otherwise}}

\newcommand{\softmax}{\operatorname{softmax}}

\newcommand{\floor}[1]{\left\lfloor{#1}\right\rfloor}

\newcommand{\wt}[1]{\widetilde{#1}}

\newcommand{\cF}{\mathcal{F}}

\newcommand{\cJ}{\mathcal{J}}

\newcommand{\cR}{\mathcal{R}}

\newcommand{\va}{\mathbf{a}}

\newcommand{\vm}{\mathbf{m}}

\newcommand{\vz}{\mathbf{z}}

\makeatletter
\newcommand*\bdot{\mathpalette\bdot@{.7}}
\newcommand*\bdot@[2]{\mathbin{\vcenter{\hbox{\scalebox{#2}{$\m@th#1\bullet$}}}}}
\makeatother

\makeatletter
\DeclareRobustCommand\onedot{\futurelet\@let@token\@onedot}
\def\@onedot{\ifx\@let@token.\else.\null\fi\xspace}

\def\eg{\emph{e.g}\onedot} 
\def\ie{\emph{i.e}\onedot} 
 
 \def\vs{\emph{vs}\onedot}
  
\def\etal{\emph{et al}\onedot}
\makeatother

%% file: tex/macro.tex
\newcommand{\Note}[2]{}

\newcommand{\cls}{\textsc{[cls]}\xspace}
\newcommand{\mask}{\textsc{[mask]}\xspace}

\newcommand{\ours}{AttMask\xspace}
\newcommand{\athigh}{AttMask-High\xspace}
\newcommand{\athint}{AttMask-Hint\xspace}
\newcommand{\atlow}{AttMask-Low\xspace}
\newcommand{\Ours}{\emph{attention-guided masking} (\ours)\xspace}

\newcommand{\vit}{ViT\xspace}
\newcommand{\mim}{MIM\xspace}

\definecolor{ForestGreen}{RGB}{34,139,34}
\definecolor{Orange}{RGB}{1.0, 0.55, 0.0}
\definecolor{Purple}{RGB}{0.58, 0.44, 0.86}

\newcommand{\gp}[1]{{\color{ForestGreen}#1}}

\definecolor{TableColor}{rgb}{0.835, 0.894, 0.968}

%% file: tex/intro.tex
\section{Introduction}
\label{sec:intro}

Self-supervised learning (SSL) has attracted significant attention over the last years. Recently, several studies are shifting towards adapting SSL to transformer architectures. Originating in natural language processing, where self-supervised transformers~\cite{attention, bert} have revolutionized the field, these architectures were introduced to computer vision with the \emph{vision transformer} (ViT)~\cite{vit} as an alternative to convolutional neural networks \cite{alexnet, googlenet, resnet}. ViT formulates an image as a sequence of tokens obtained directly from raw patches and then follows a pure transformer architecture. Despite the absence of image-specific inductive bias, ViT shows strong image representation learning capacity.

Considering that transformers are data-hungry, many studies advocate pre-training them on unsupervised pretext tasks, determined only by raw data. A prominent paradigm is to mask a portion of the input tokens---words in text or patches in images---and train the transformer to predict these missing tokens~\cite{bert, beit, ibot, mae, simmim}. This paradigm, called \emph{masked language modeling} (MLM) in the language domain~\cite{bert}, is remarkably successful and extends to the vision domain as \emph{masked image modeling} (MIM)~\cite{beit, simmim, ibot}.

MIM-based self-supervised methods have already shown impressive results on images. However, an important aspect that has not been well explored so far is how to choose which image tokens to mask. Typically, the selection is random, as has been the norm for text data. In this work, we argue that random token masking for image data is not as effective.

\begin{figure}[t]
\input{tex/fig_strategies}
\caption{Different than random masking strategies (b-d), our \Ours uses the attention map arising in the encoder (e) to mask the most highly attended by default (f), or the low-attended (g) patches. (b) is used by SimMIM~\cite{simmim}, (c) by MAE~\cite{mae}, (d) by BEiT~\cite{beit} and (g) by MST~\cite{mst}.}
\label{fig:strategies}
\end{figure}
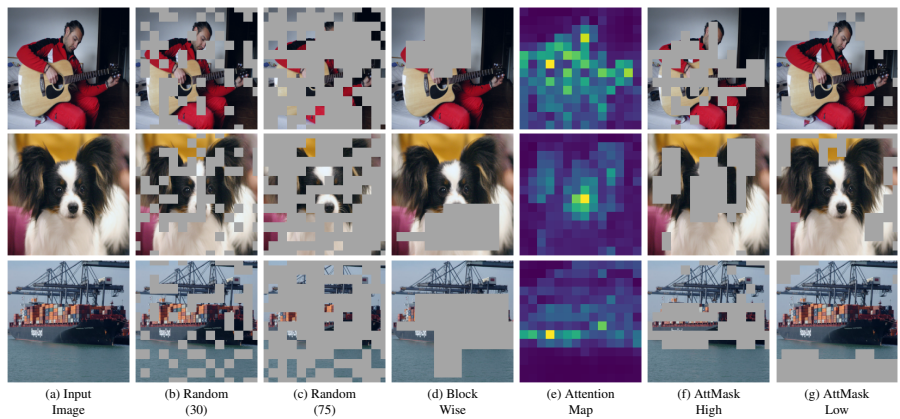

In text, random word masking is likely to hide high-level concepts that describe entire semantic entities such as objects (nouns) and actions (verbs). By contrast, an image has much more tokens than a sentence, which are highly redundant, and random masking is less likely to hide ``interesting'' parts; or when it does, the remaining parts still easily reveal the identity of the visual concepts. As shown in \autoref{fig:strategies}(b-d), unless masking is very aggressive, this is thus less likely to form challenging token reconstruction examples that would allow the transformer to develop strong comprehension skills.

The question we ask is this: \emph{Can we develop a masking strategy that addresses this limitation and makes informed decisions on which tokens to mask?}

To this end, we propose to exploit the intrinsic properties of ViT and in particular its self-attention mechanism. Given an input sequence of image patches, we forward it through the transformer encoder, thereby obtaining an attention map in its output. We then mask the most attended tokens. As shown in \autoref{fig:strategies}(f-g), the motivation is that highly-attended tokens form more coherent image regions that correspond to more discriminative cues comparing with random tokens, thus leading to a more challenging MIM task.

This strategy, which we call \Ours, is an excellent fit to popular distillation-based self-supervised objectives, because it is the teacher encoder that sees the entire image and extracts the attention map, and the student encoder that sees the masked image and solves the reconstruction task. \ours thus incurs zero additional cost.

We make the following contributions:
\begin{enumerate}[itemsep=2pt, parsep=0pt, topsep=0pt]
	\item We introduce a novel masking strategy for self-supervised learning, called \ours, that exploits the intrinsic properties of \vit by leveraging its self-attention maps to guide token masking (\autoref{sec:attmask}).
	\item	We show how to efficiently incorporate this above masking strategy into teacher-student frameworks that use a \mim reconstruction objective and demonstrate significant performance improvements over random masking.
	\item Through extensive experimental evaluation, we confirm that \ours offers several benefits: it accelerates the learning process; it improves performance on a data-limited regime (\autoref{sec:analysis}) and on a variety of downstream tasks (\autoref{sec:bench});
	it increases the robustness against background changes, thus revealing that it reduces background dependency.
\end{enumerate}

%% file: tex/fig_strategies.tex
\tiny
\centering
\setlength{\tabcolsep}{1.14pt}
\begin{tabular}{cccccccc}
	\fig[.133]{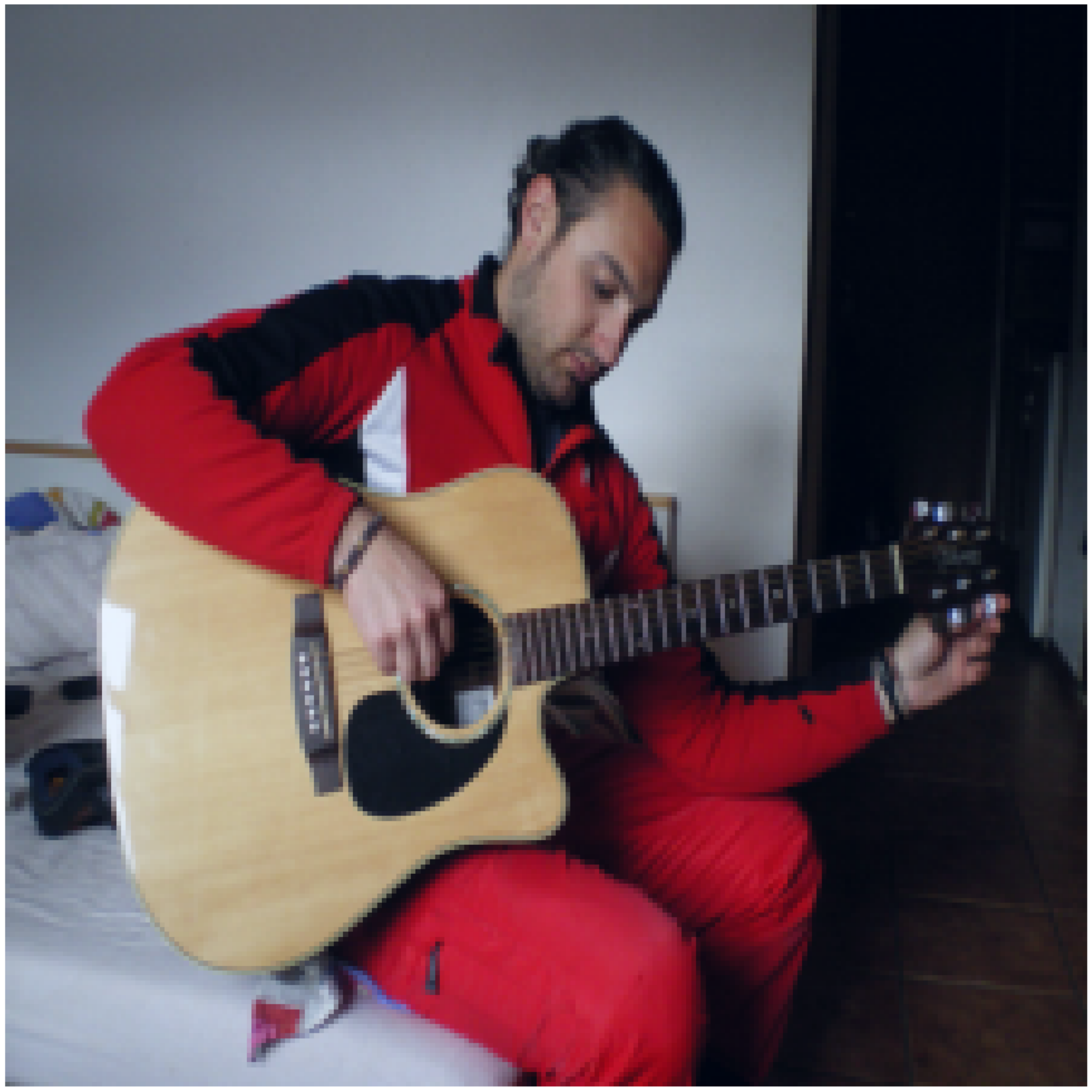}               &
	\fig[.133]{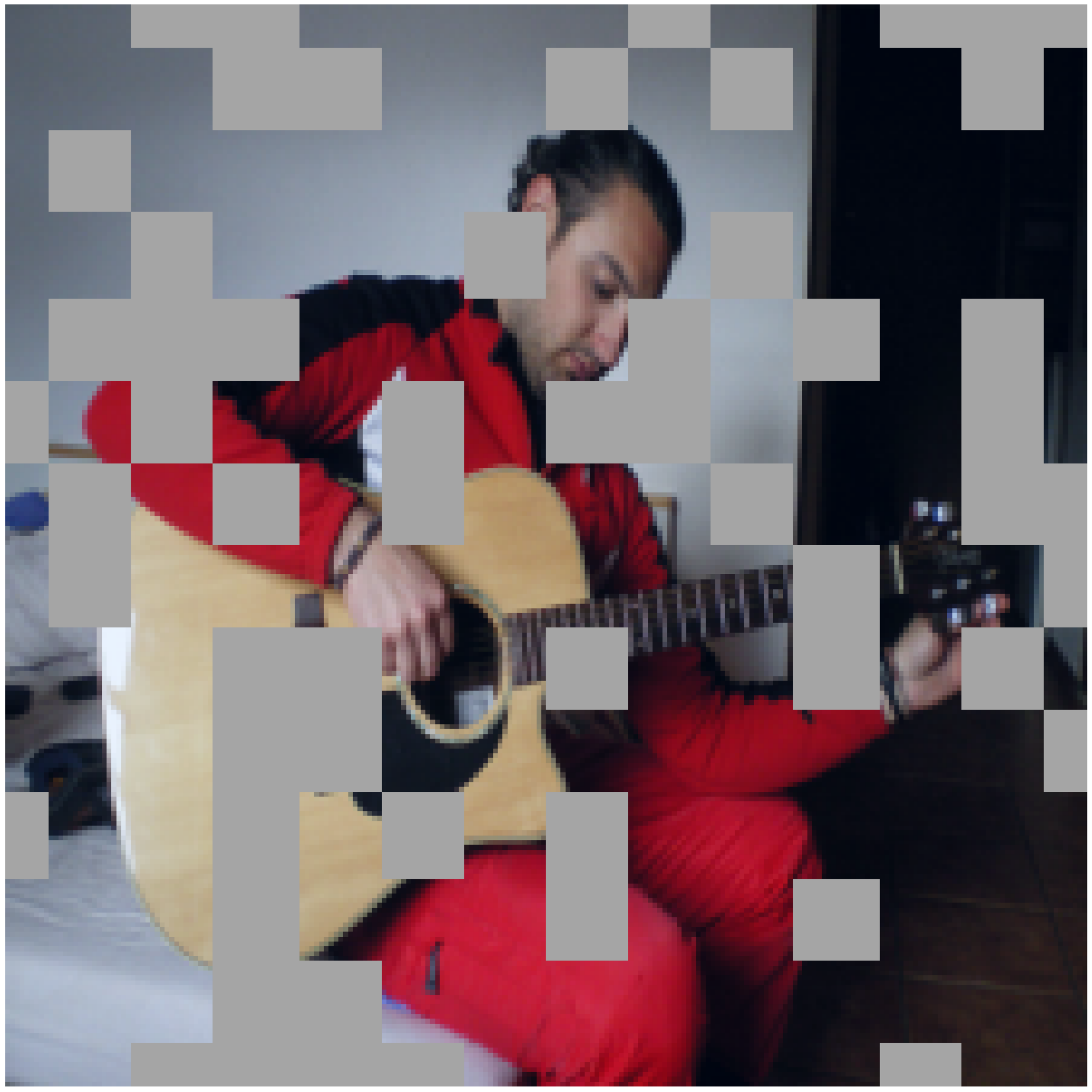}              &
	\fig[.133]{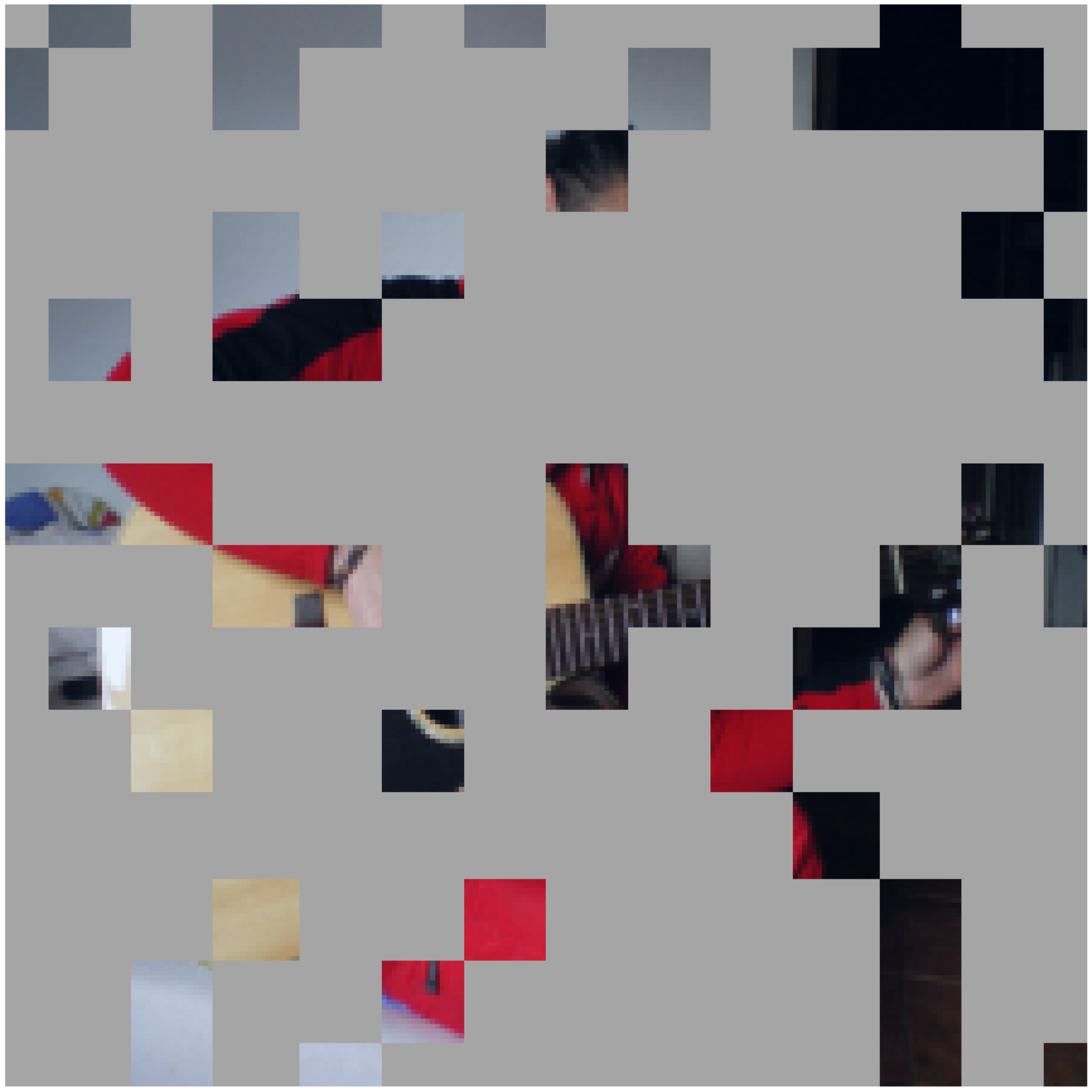}              &
	\fig[.133]{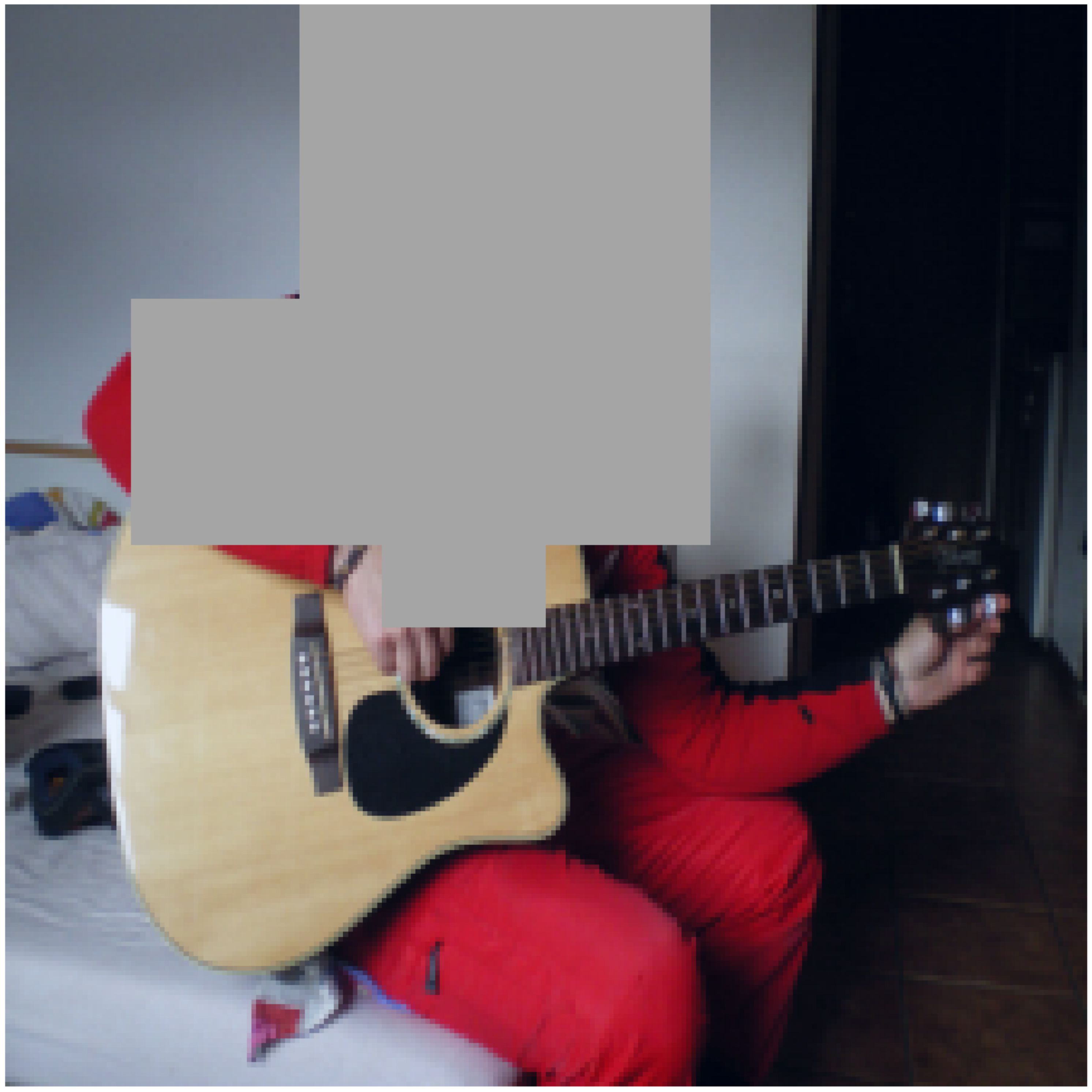}      &
	\fig[.133]{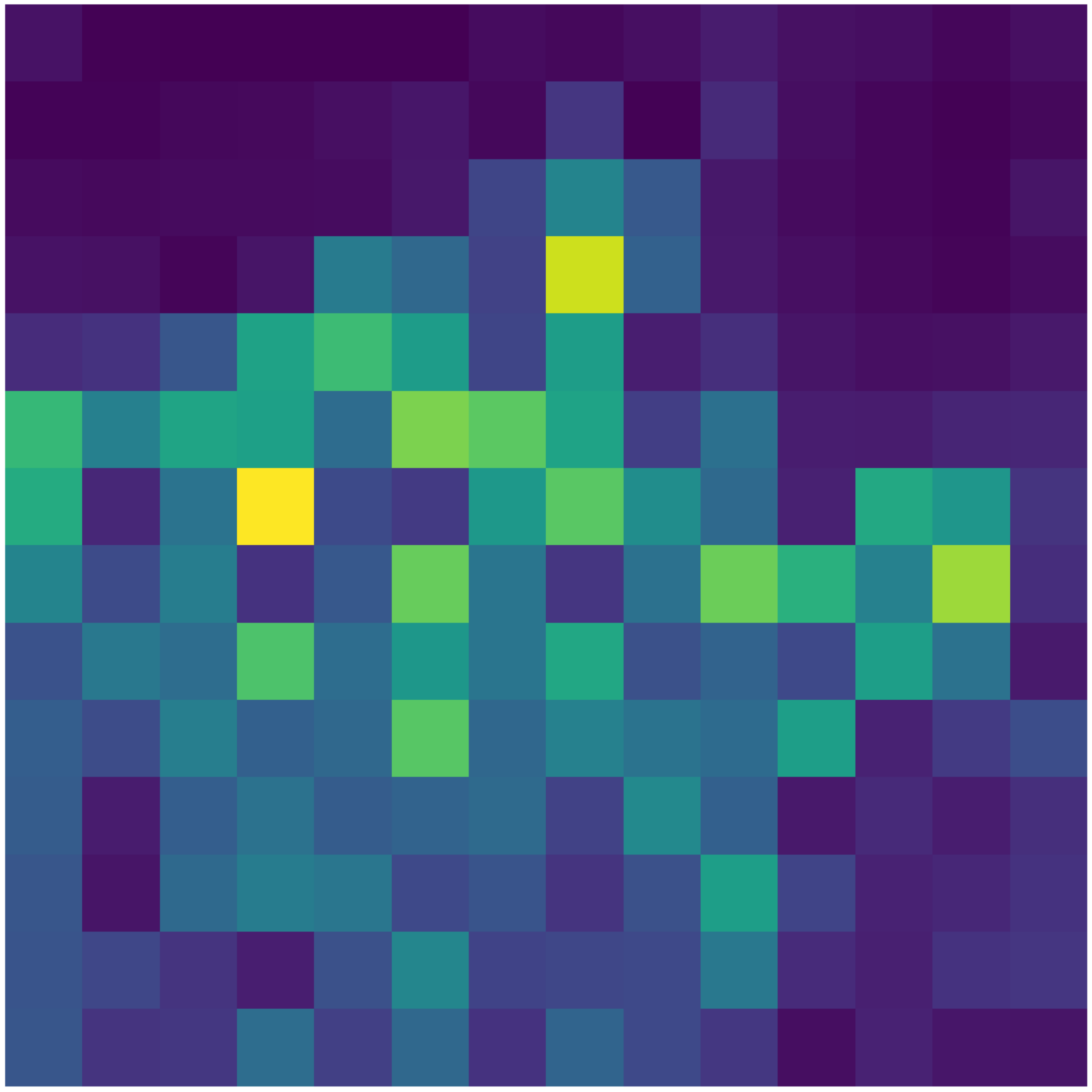}                &
	\fig[.133]{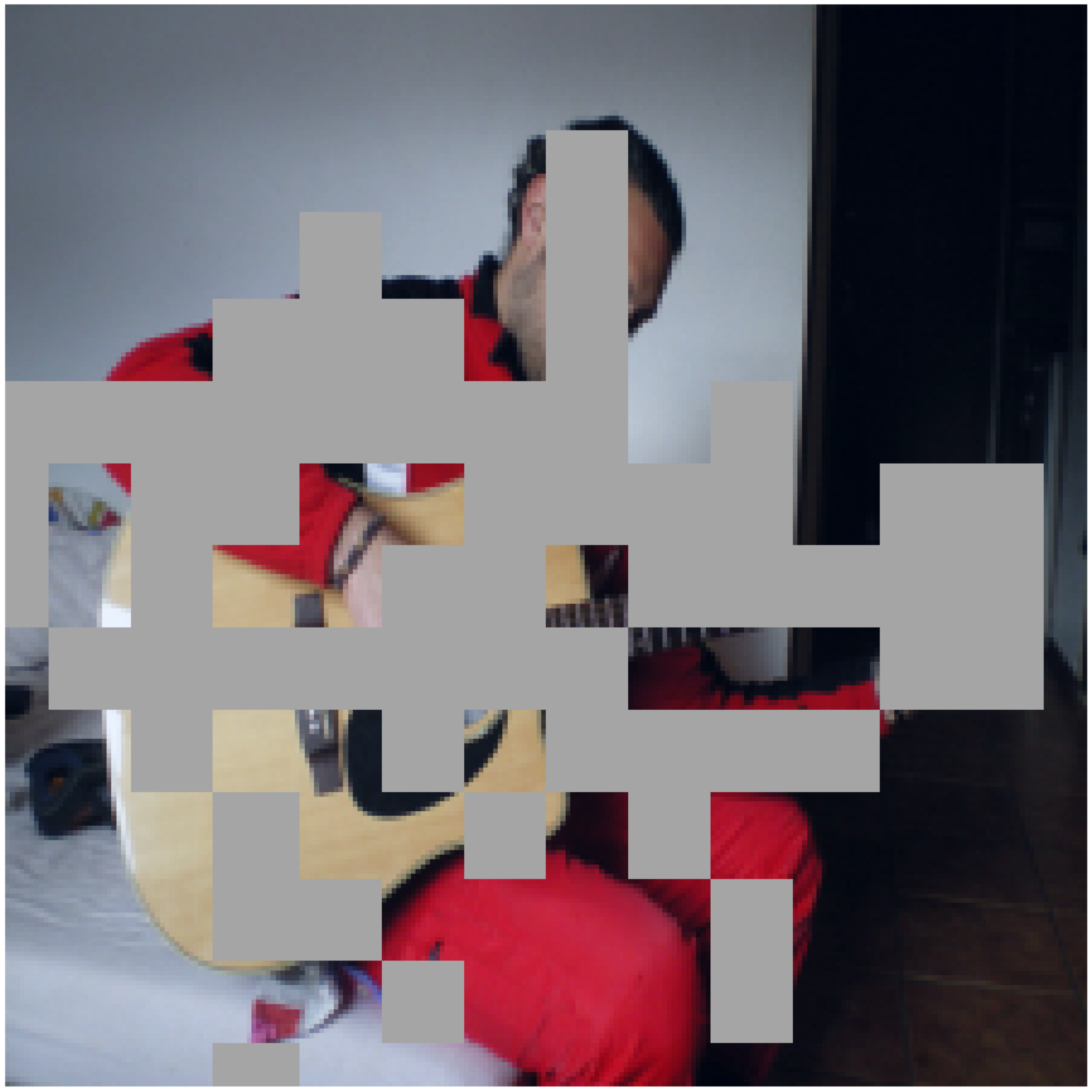}               &
	\fig[.133]{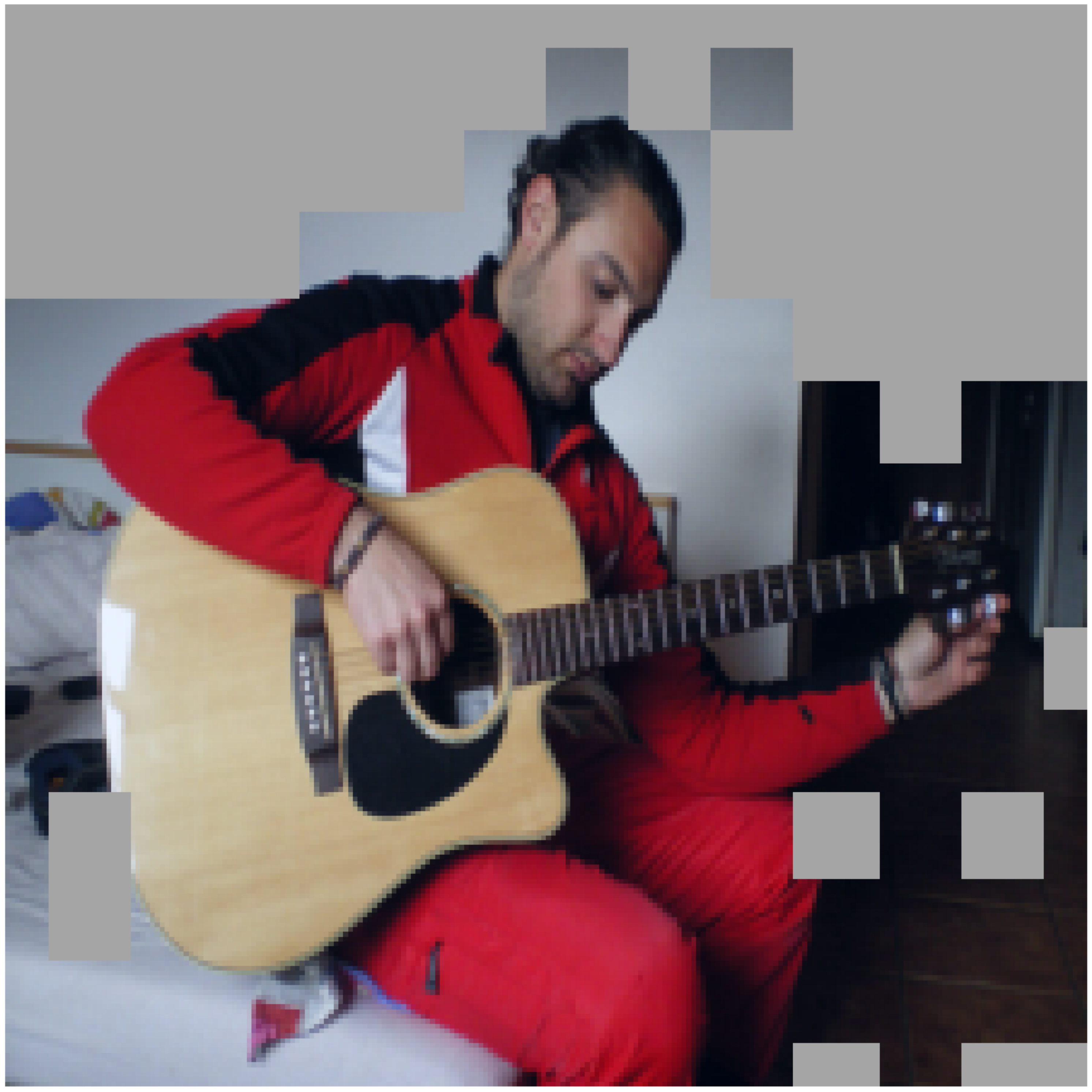}                \\

	\fig[.133]{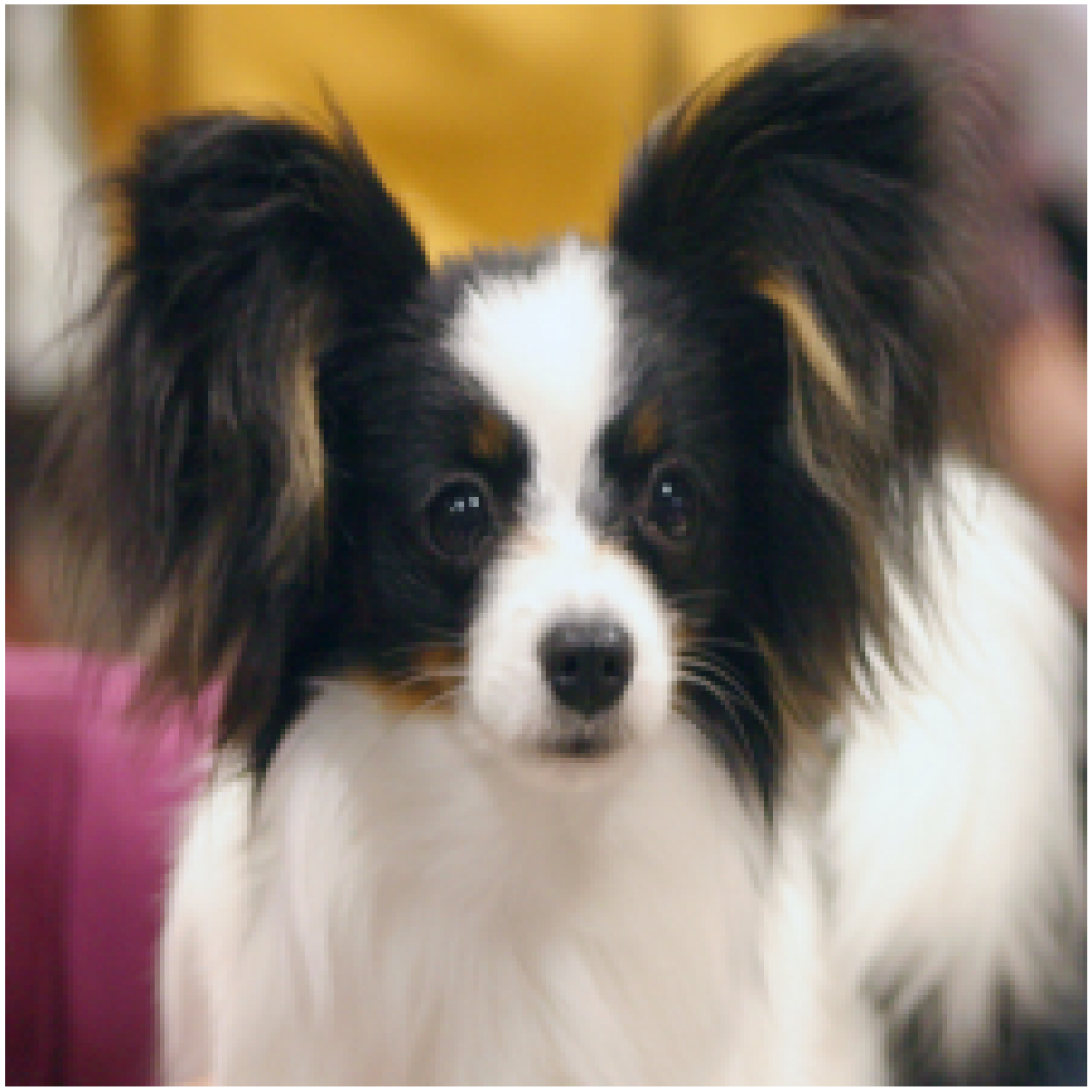}                 &
	\fig[.133]{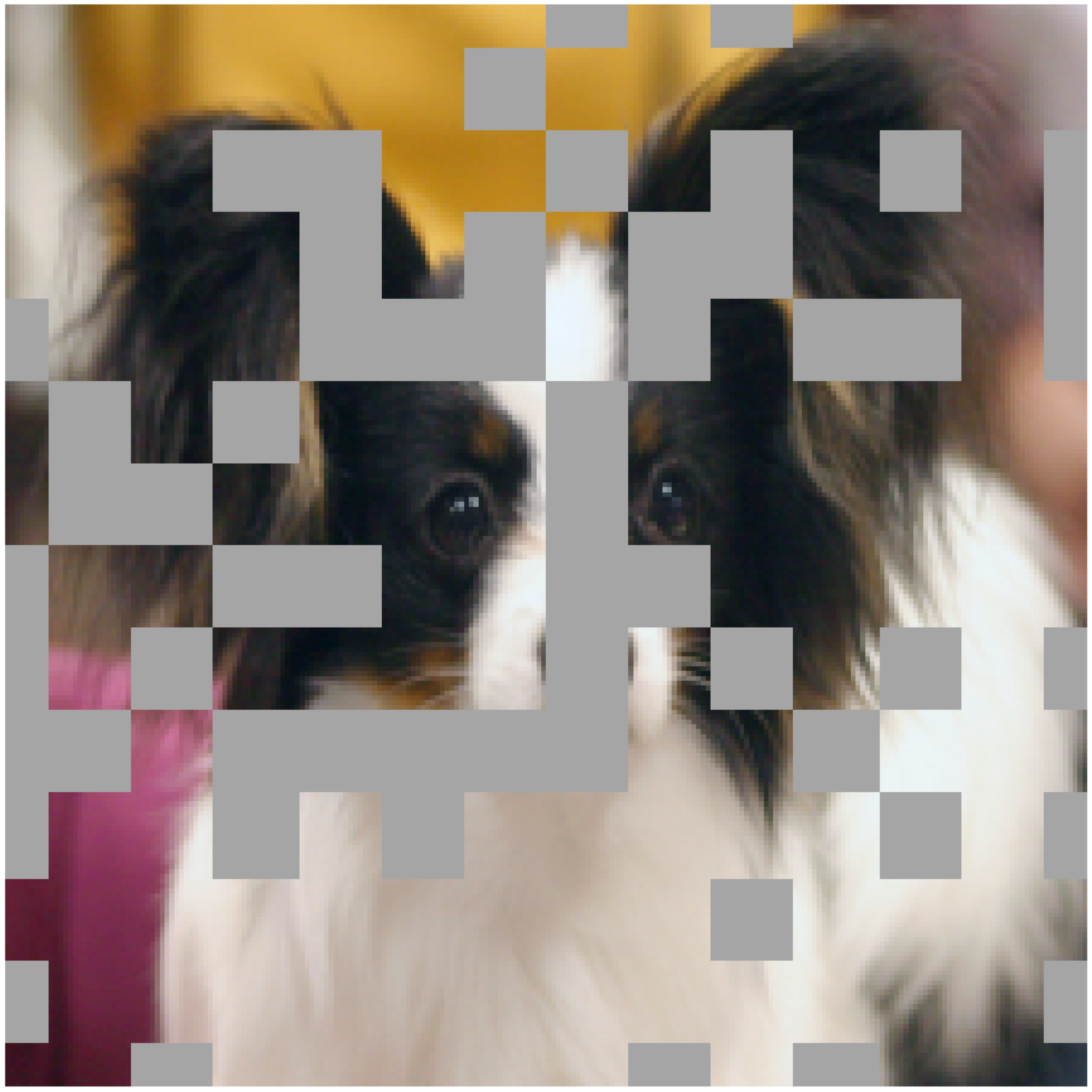}                &
	\fig[.133]{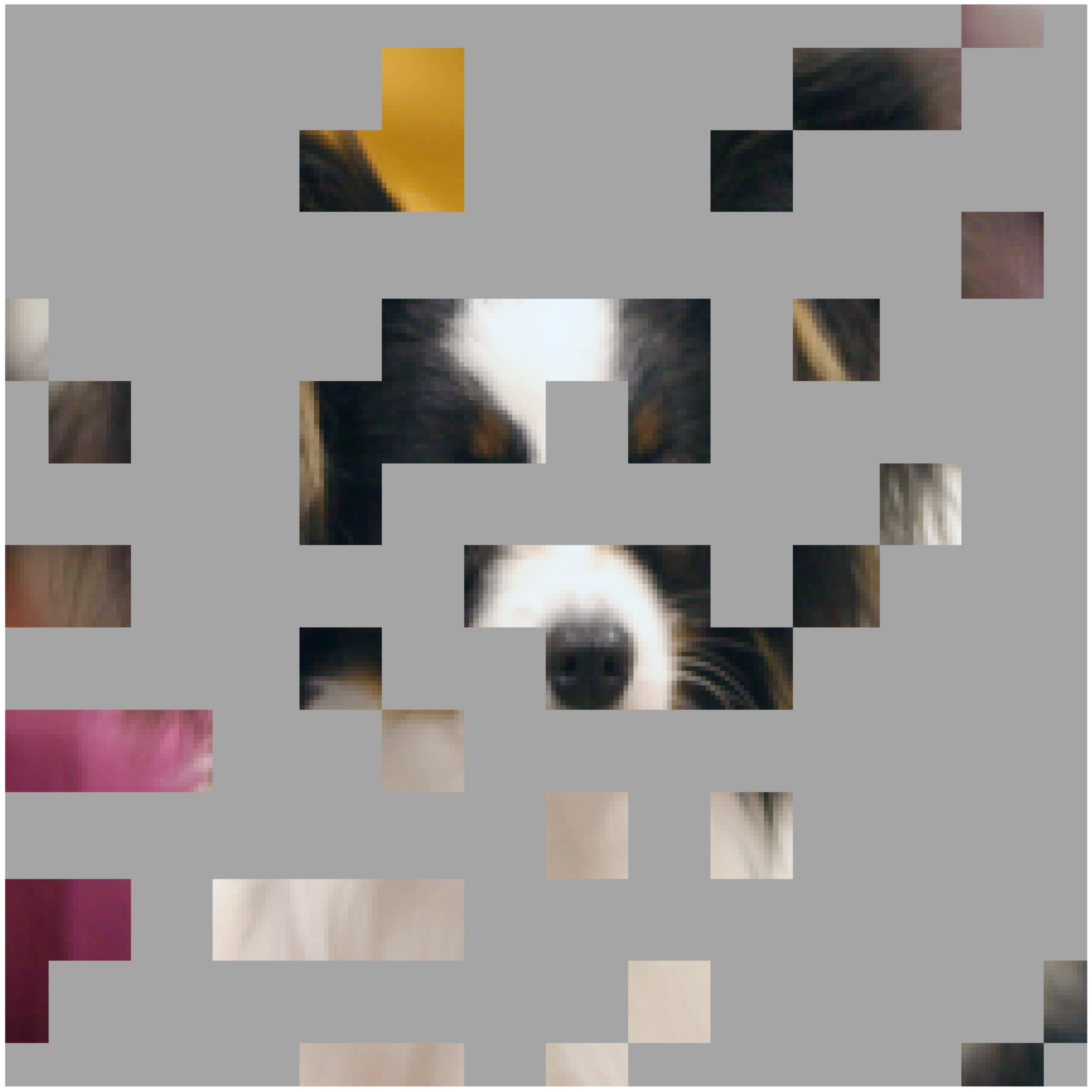}                &
	\fig[.133]{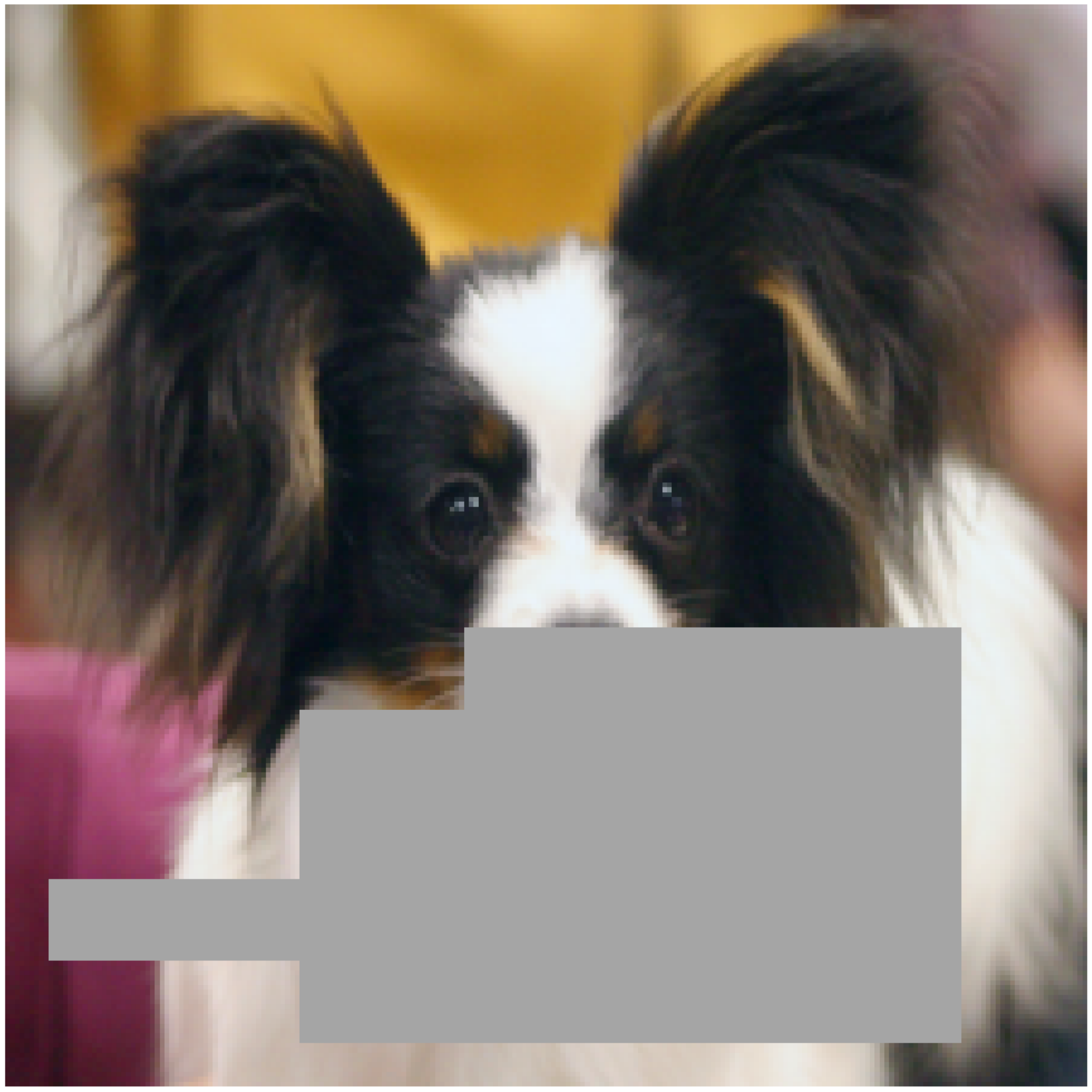}        &
	\fig[.133]{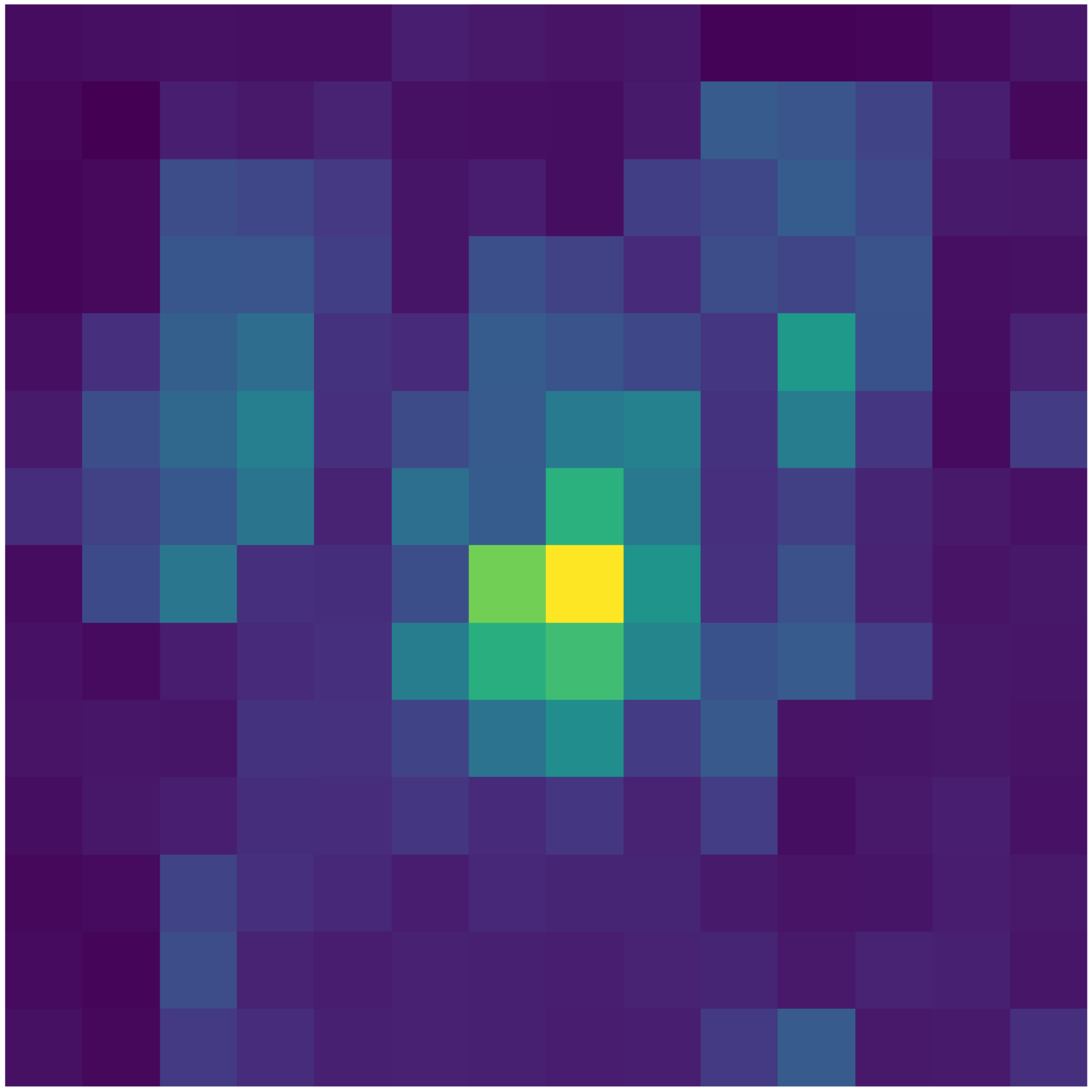}                  &
	\fig[.133]{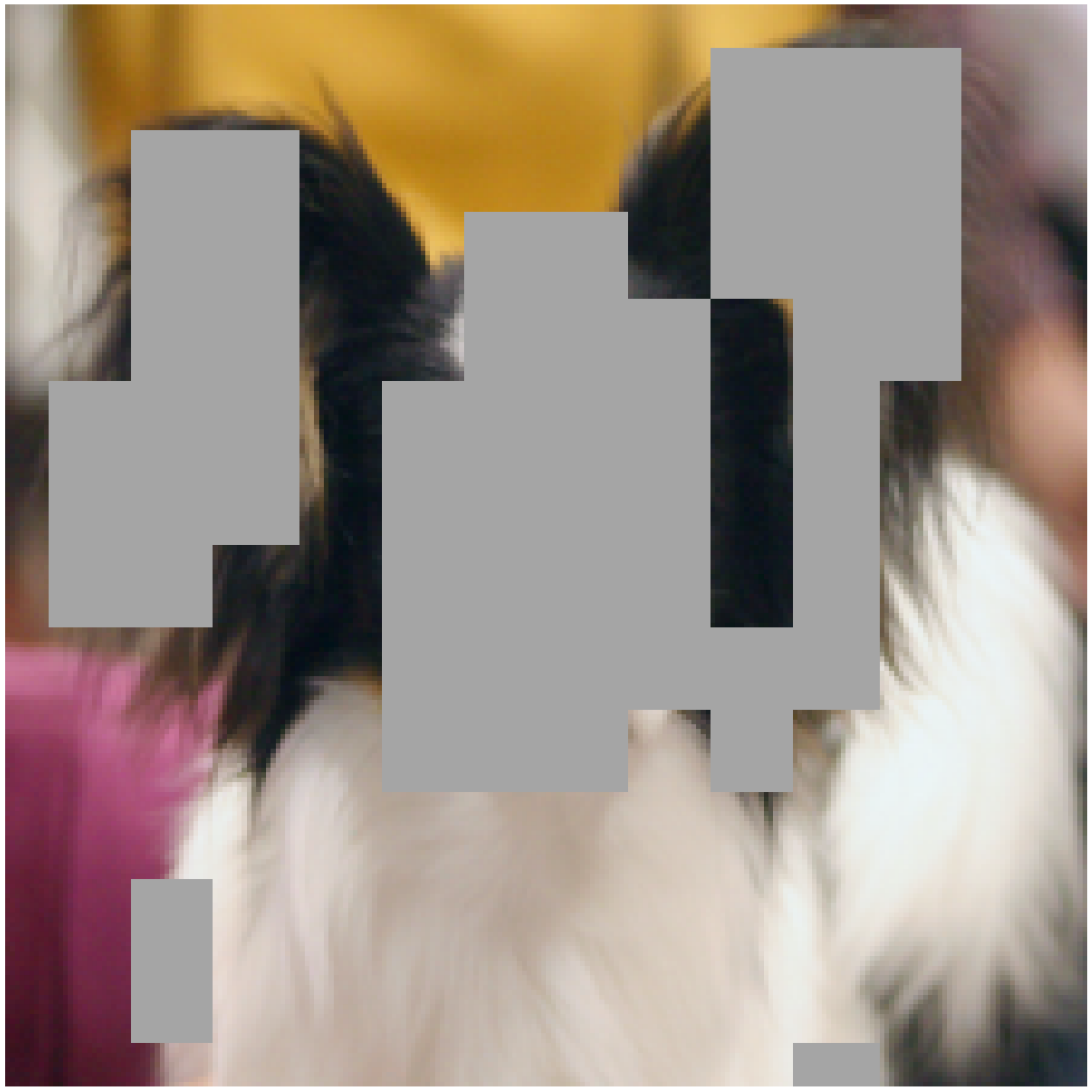}                 &
	\fig[.133]{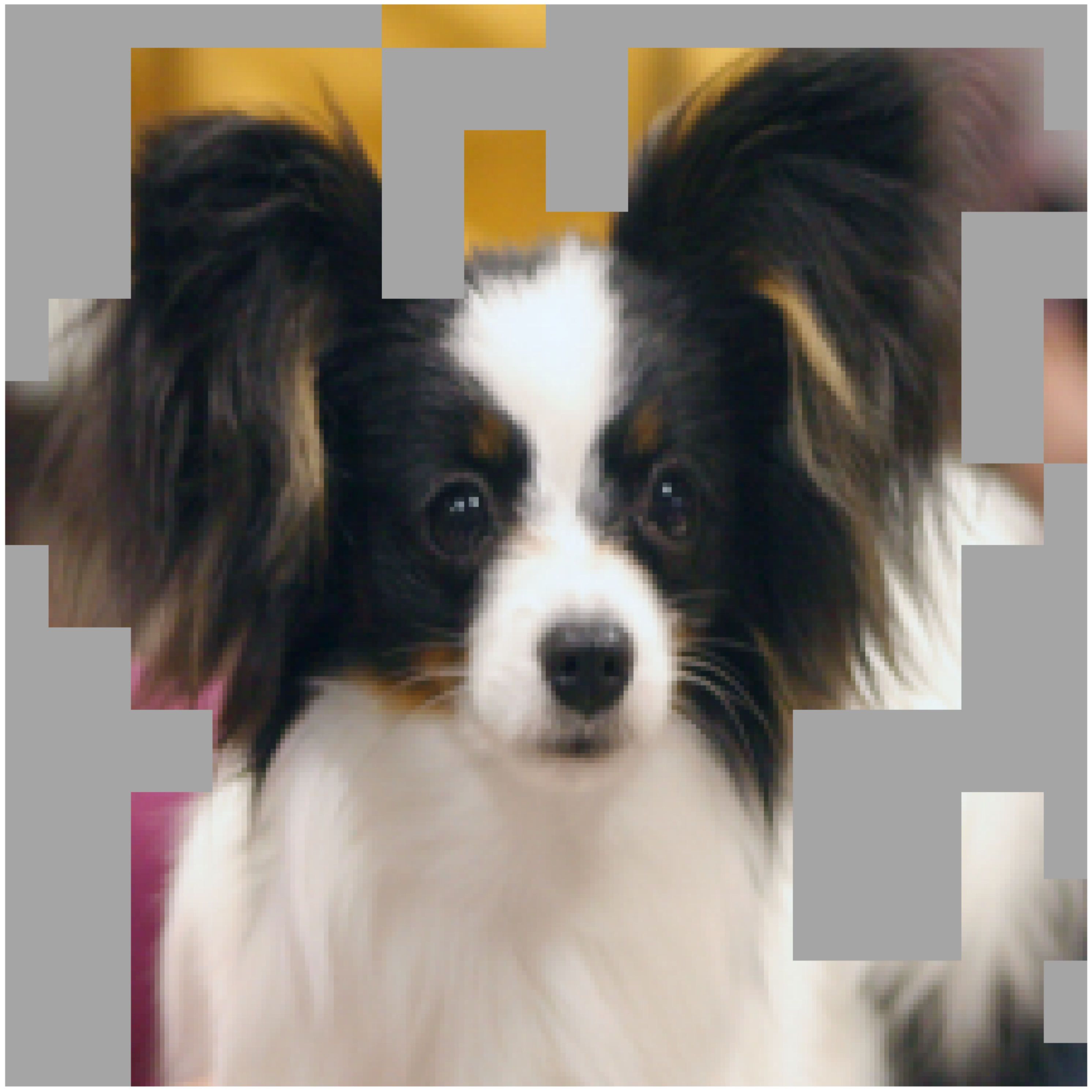}                  \\

	\fig[.133]{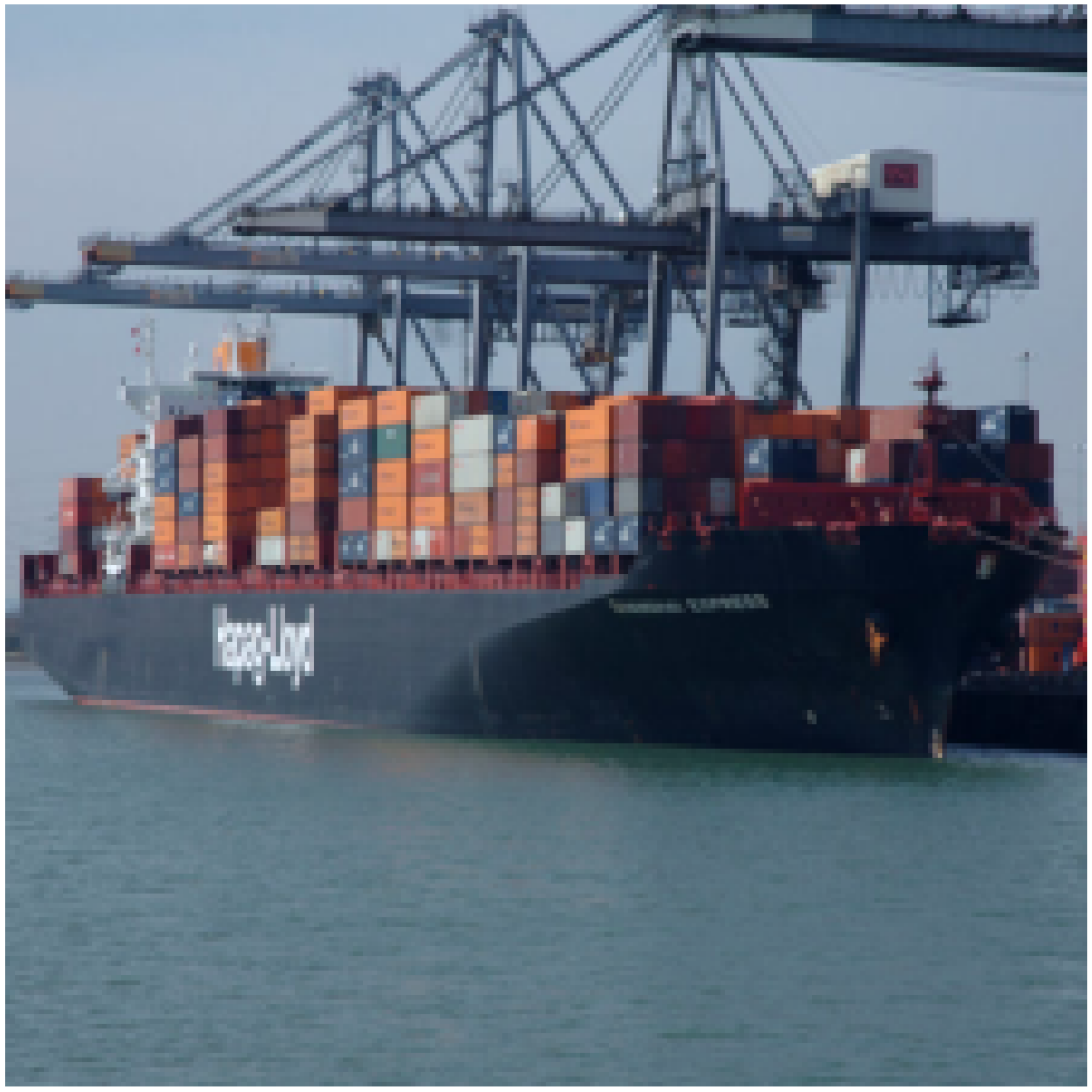}                 &
	\fig[.133]{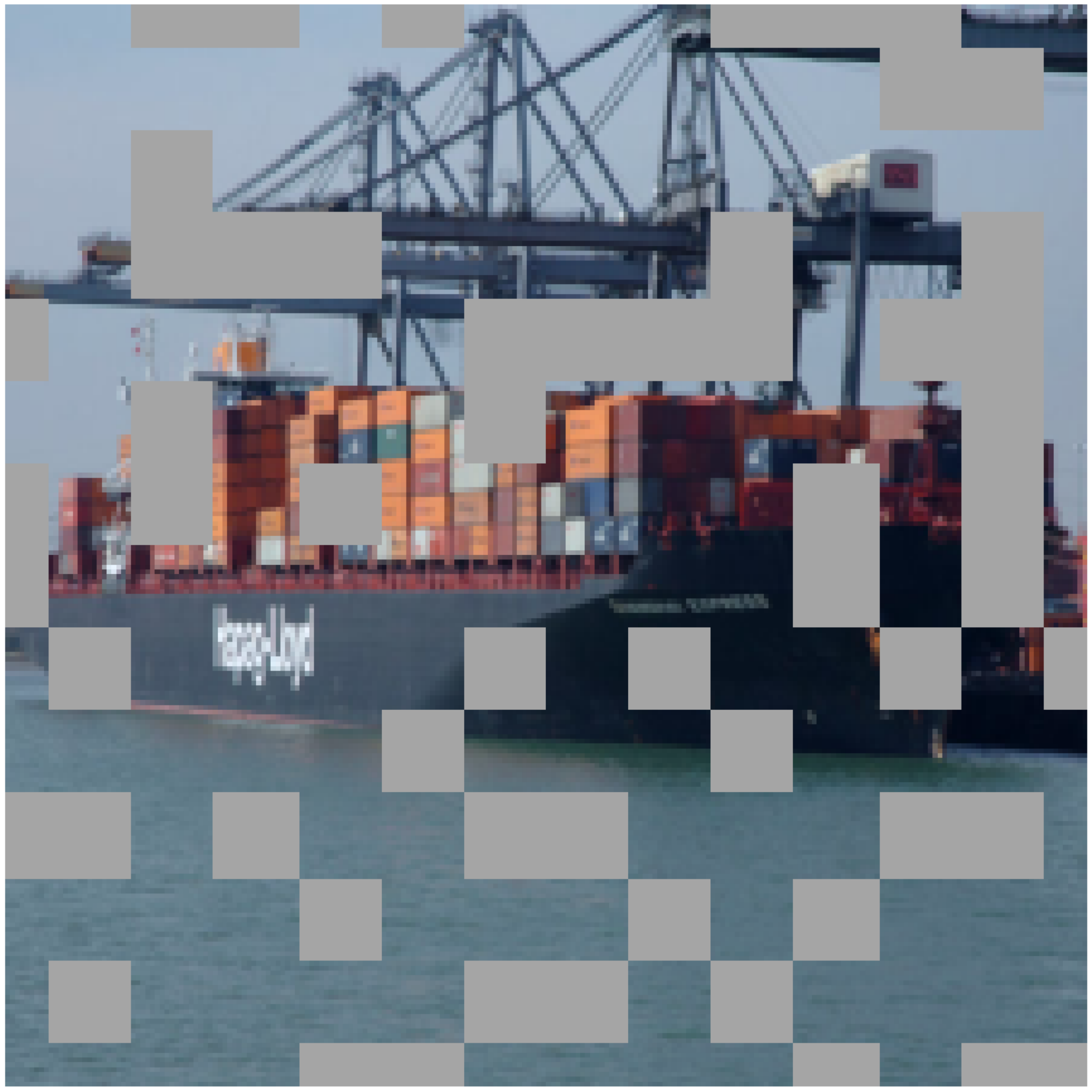}                &
	\fig[.133]{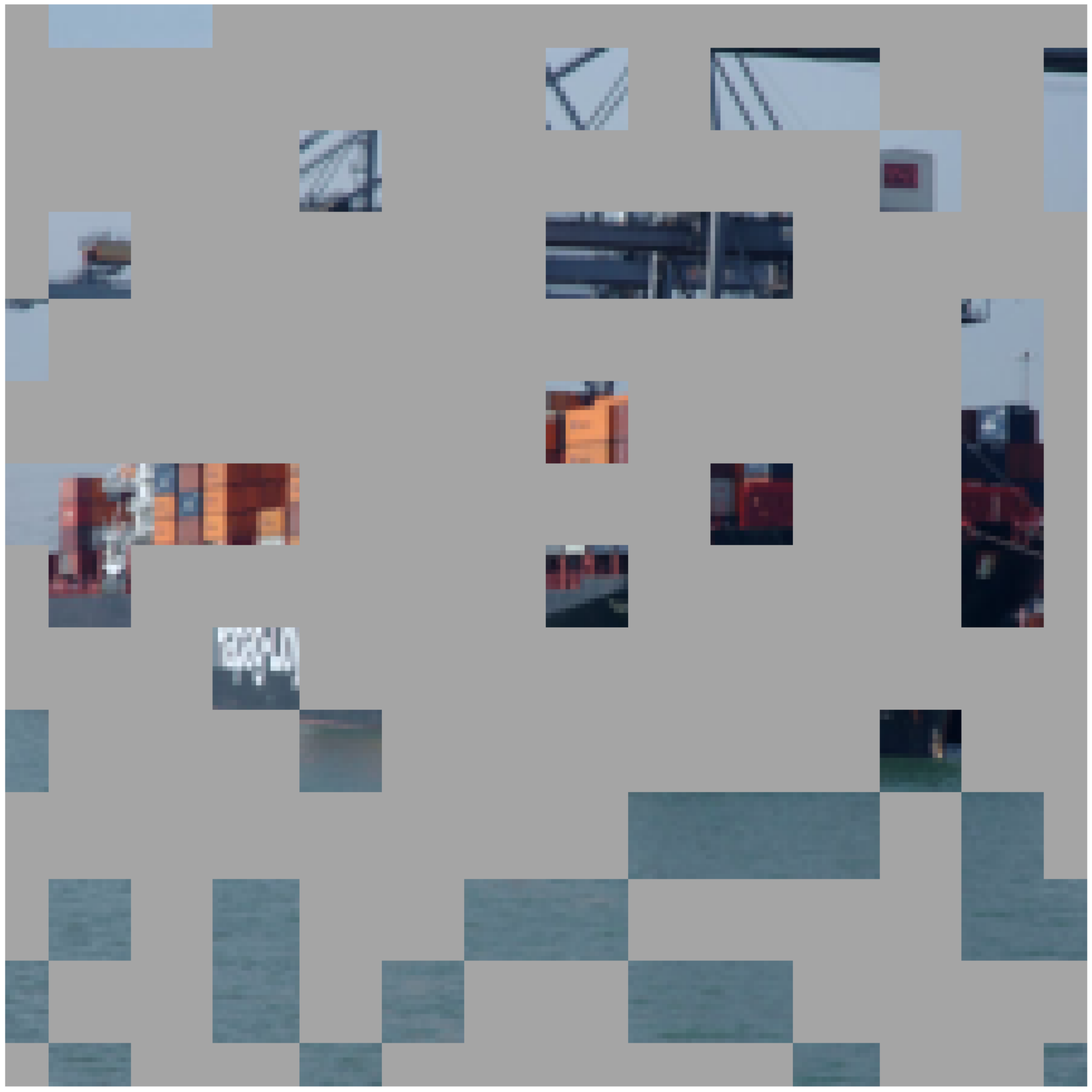}                &
	\fig[.133]{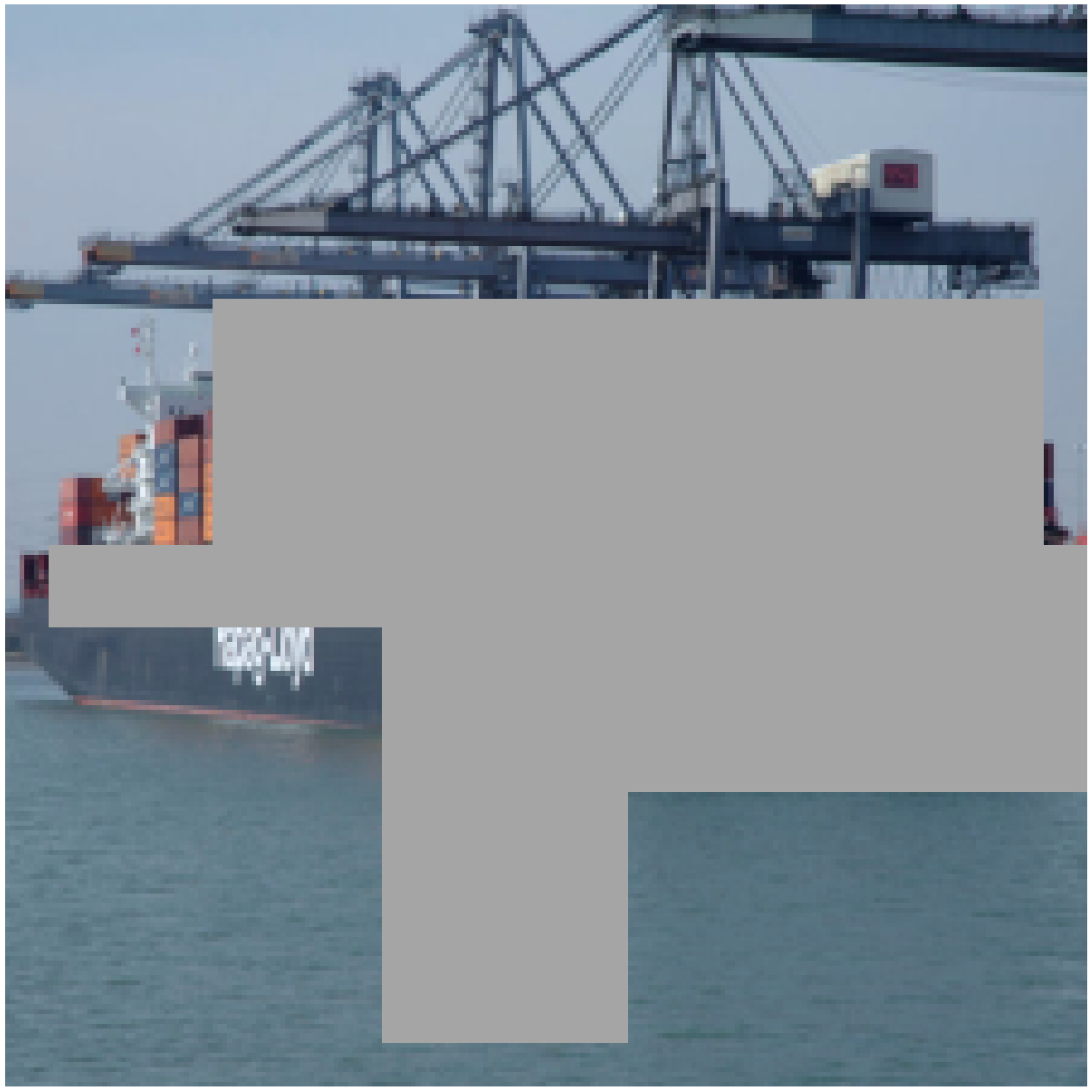}        &
	\fig[.133]{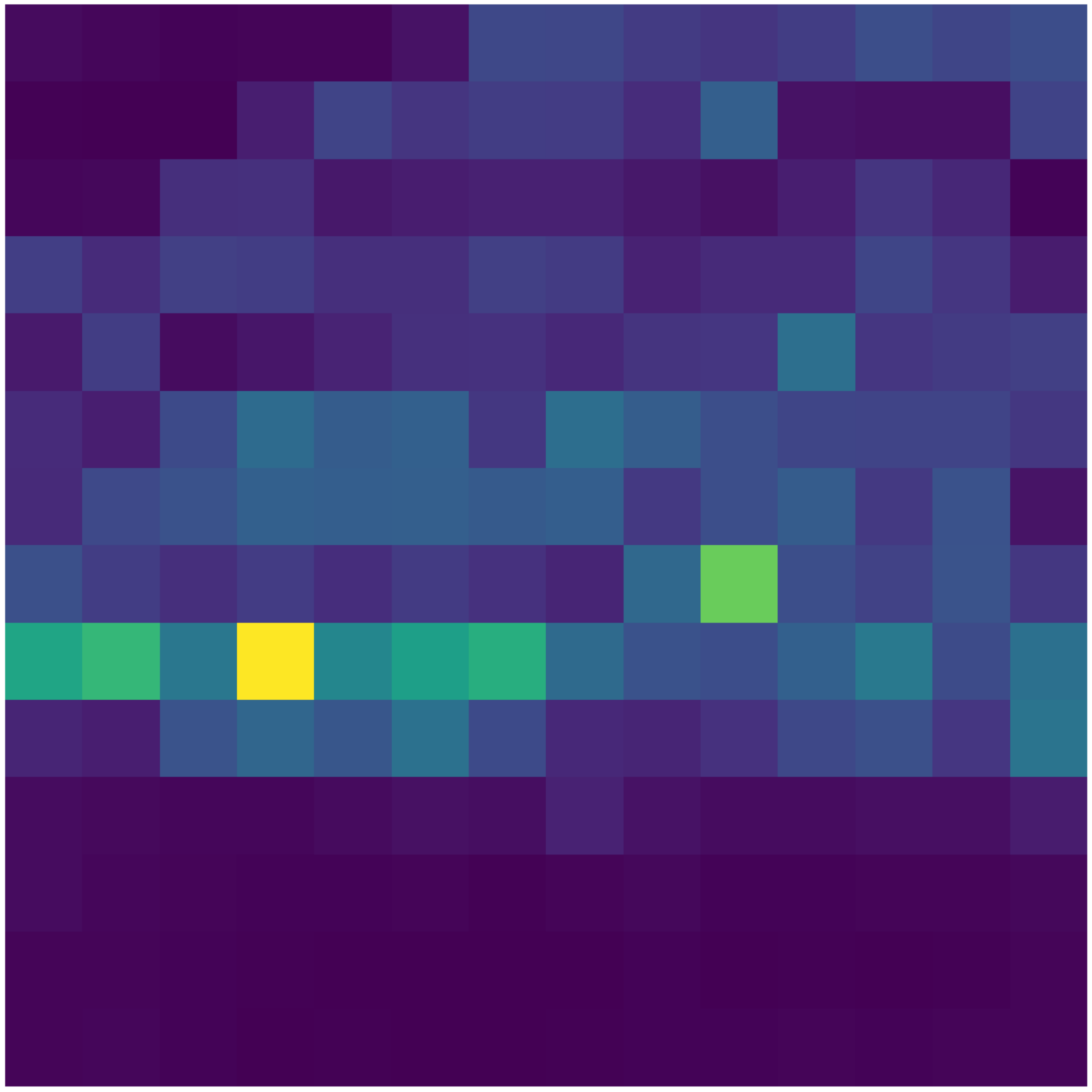}                  &
	\fig[.133]{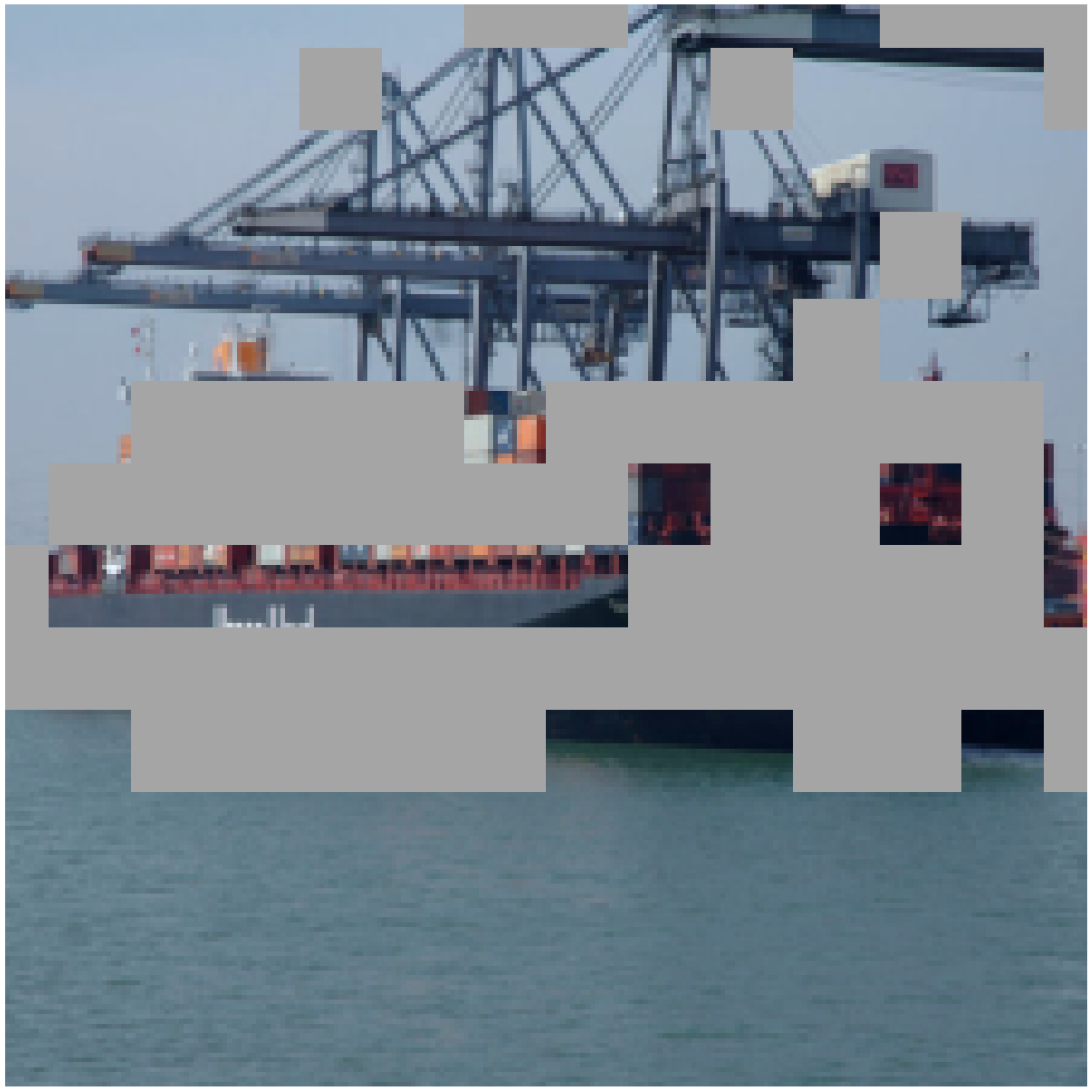}                 &
	\fig[.133]{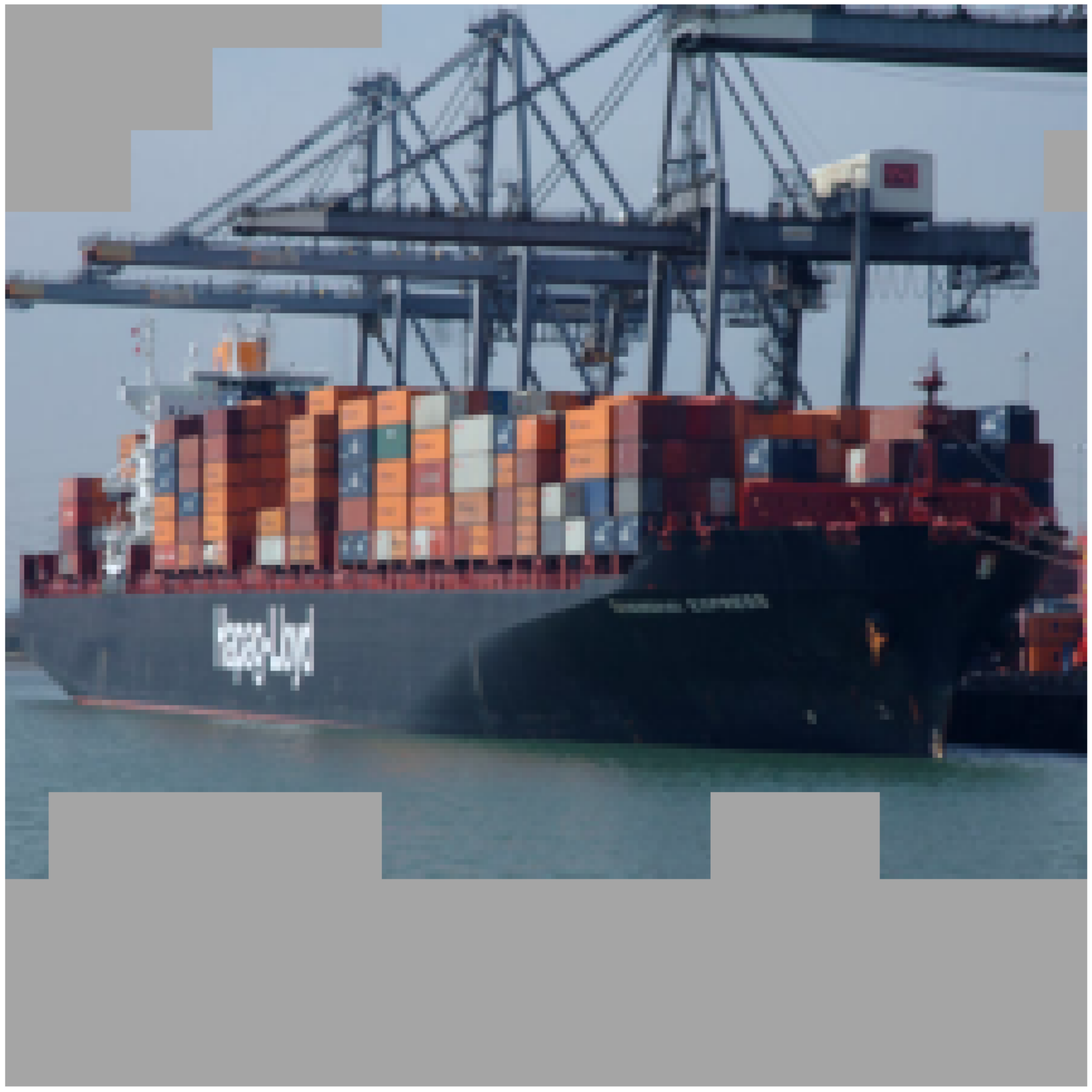}                  \\

	(a) Input                                                &
	(b) Random                                               &
	(c) Random                                               &
	(d) Block                                                &
	(e) Attention                                            &
	(f) \ours                                                &
	(g) \ours                                                \\

	Image                                                    &
	(30)                                                     &
	(75)                                                     &
	Wise                                                     &
	Map                                                      &
	High                                                     &
	Low                                                      \\
\end{tabular}

%% file: tex/related.tex
\section{Related Work}
\label{sec:related}

\paragraph{Vision Transformers.}

Transformers are based on self-attention~\cite{attention} and require pretraining on large unlabelled corpora~\cite{bert}. Their adaptation to vision tasks is not straightforward. Representing pixels by tokens is impractical due to the quadratic complexity of self-attention, giving rise to approximations~\cite{sparse_transformers, image_transformer, scaling_autoregressive, axial, axial_deeplab}.
The idea of representing image patches by tokens is proposed in~\cite{on_the_relationship}, where patches are of size $2\times2$, and is further studied in \vit~\cite{vit}, where patches are $16\times16$. Despite the absence of image-specific inductive bias, \vit is competitive to convolutional neural networks for ImageNet~\cite{imagenet} and other smaller benchmark datasets~\cite{cifar, oxford_flowers}. Since it is pretrained on a large and private dataset~\cite{jft_300m}, authors of DeiT \cite{deit} question its efficiency and propose an improved data-efficient version, which however is based on a strong teacher instead~\cite{Radosavovic_2020_CVPR}.

\paragraph{Self-supervised Learning.}

Early self-supervised learning methods follow the paradigm of training on an annotation-free \emph{pretext task}, determined only by raw data~\cite{context_prediction, rotations, look_listen_learn, sorting_sequences, shuffle_learn, jigsaw, aet, revisiting}. This task can be \eg the prediction of patch orderings~\cite{jigsaw} or rotation angles~\cite{rotations}. Starting from instance discrimination~\cite{instance_discrimination} and contrastive predictive coding \cite{infonce}, \emph{contrastive learning} has become very popular~\cite{simclr, falcon2020framework, cai2020all, hard_negative, pirl, contrastive_multiview_coding, understanding}. These methods pull positives together and push negatives apart, where positives are typically determined by different views of the same example.
Alternatively, contrastive learning often relies on clustering~\cite{asano2019self, caron2018deep, caron2019unsupervised, swav, li2021prototypical, zhuang2019local, bownet}.
The requirement of negatives is eliminated in BYOL~\cite{byol}, OBoW~\cite{obow}, SimSiam~\cite{simsiam} and DINO~\cite{dino}, where the challenge is to avoid representation collapse, most notably by a form of \emph{self-distillation}~\cite{tarvainen2017mean}.

Using transformers, MIM as a pretext 
task is proposed in BEiT~\cite{beit}, which maps the images to discrete patch tokens and recovers tokens for masked patches, according to a block-wise random strategy. Other than that, MIM methods use continuous representations: SimMIM~\cite{simmim} randomly masks large patches and predicts the corresponding pixels by direct regression; MAE~\cite{mae} randomly masks a large portion of patches and predicts the corresponding pixels using an autoencoder; MST~\cite{mst} masks low-attended patches and reconstructs the entire input with a decoder; iBOT~\cite{ibot} extends the self-distillation loss of DINO to dense features corresponding to block-wise masked patches. Here, we advocate masking of \emph{highly-attended} patches, in a sense the opposite of MST, and we exhibit this idea in the context of DINO and iBOT.


\paragraph{Regularization and Augmentation.}

As the complexity of a task increases, networks with more and more parameters are introduced. But with increased representational power comes increased need for more data or risk of overfitting. Several regularization and data augmentation methods have been proposed in this direction~\cite{imagenet, verydeep, batch_normalization, dropout}, combined with standard supervised tasks.

In this context, feature masking is introduced by Dropout~\cite{dropout}, which randomly drops hidden neuron activations. To address the strong spatial correlation in convolutional feature maps, SpatialDropout~\cite{spatialdropout} randomly drops entire channels.
DropBlock~\cite{dropblock} generalizes Dropout---or constrains SpatialDropout---by dropping features in a block, \ie, a square region of a feature map. Attention Dropout~\cite{attention_dropout} makes use of self-attention to mask the most discriminative part of an image.
Feature-space masking, guided by attention from another network or branch, has been extensively studied as a mechanism to explore beyond the most discriminative object parts for weakly-supervised object detection~\cite{Kim_2017_ICCV, hou2018self, ZWF+18}.
Our work is a natural evolution of these ideas, where attention is an intrinsic mechanism of transformers; and the task becomes that of densely reconstructing the masked features. This is a pretext task, without need for supervision.

%% file: tex/method.tex
\section{Method}
\label{sec:method}

\begin{figure}[t]
\centering
\fig[1.]{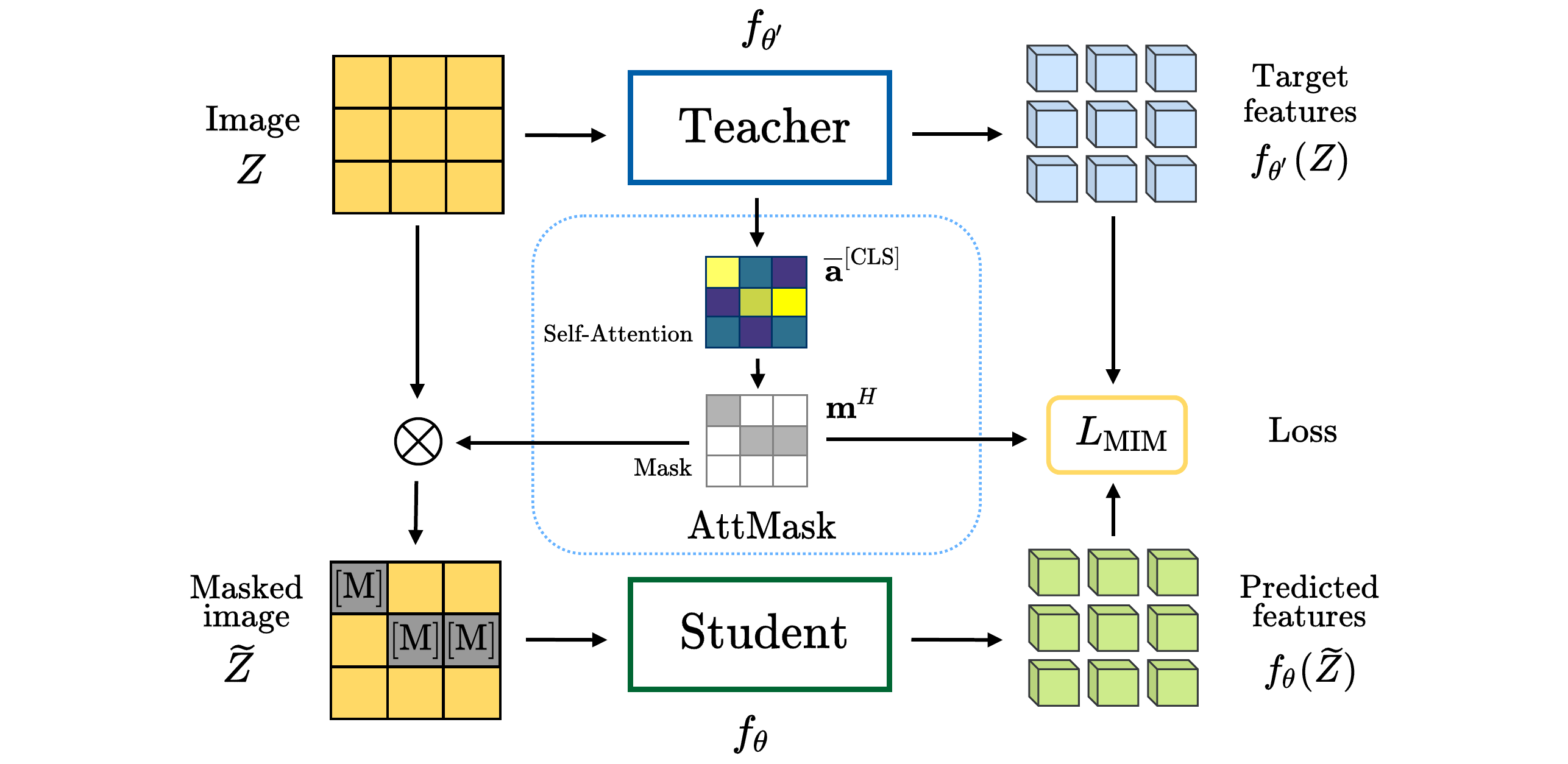}
\caption{Simplified overview of \ours as incorporated in the
masked image modelling (MIM) objective of iBOT~\cite{ibot}. A tokenized image $Z$~\eq{token} is given as input to a teacher encoder $f_{\theta'}$, generating target features $f_{\theta'}(Z)$ and an attention map $\overline{\va}^\cls$~\eq{att-cls}. We then generate a mask $\vm^H$~\eq{high} on the most attended tokens and accordingly a masked version $\wt{Z}$~\eq{mask-high} of the image, which is given as input to a student encoder $f_\theta$ to generate the predicted features $f_\theta(\wt{Z})$. Using $\vm^H$, loss $L_{\textsc{MIM}}$~\eq{mim} is a dense distillation loss between predicted and target features of the masked tokens. Additionally, a global loss $L_\textsc{g}$~\eq{g} between \cls tokens is applied (not shown here).
}
\label{fig:overview}
\end{figure}

A simplified overview of the method is shown in \autoref{fig:overview}. We first discuss in \autoref{sec:prelim} preliminaries and background on vision transformers and self-supervision with distillation-based masked image modeling. In \autoref{sec:attmask}, we then detail our attention-guided token masking strategy, called \ours, and how we incorporate it into masked image modeling.


\subsection{Preliminaries and Background}
\label{sec:prelim}

\paragraph{Vision Transformer~\cite{vit}.}

We are given an input image $X \in \real^{h \times w \times c}$, where $h \times w$ is the spatial resolution and $c$ is the number of channels. The first step is to tokenize it, \ie, convert it to a sequence of token embeddings. The image is divided into $n = hw / p^2$ non-overlapping patches $P_i \in \real^{p \times p \times c}$ for $i = 1, \dots, n$, where $p \times p$ is the patch resolution. Each patch is flattened into a vector in $\real^{p^2 c}$ and projected to an embedding vector $\vz_i \in \real^d$ using a linear layer, where $d$ is the embedding dimension. A learnable embedding $\vz^\cls \in \real^{d}$ of a ``classification'' token \cls is then prepended to form the \emph{tokenized image}
\begin{align}
	Z = (\vz^\cls; \vz_1; \dots; \vz_n) \in \real^{(n+1) \times d},
\label{eq:token}
\end{align}
where ``;'' denotes row-wise stacking. The role of this special token is to represent the image at the output. A sequence of position embeddings is added to $Z$ to retain positional information. The resulting sequence is the input to the \emph{transformer encoder}. Each layer of the encoder consists of a \emph{multi-head self-attention} (MSA) block followed by a \emph{multi-layer perceptron} (MLP) block. Through all of its layers, the encoder uses a sequence of fixed length $n+1$ of token embeddings of fixed dimension $d$, represented by a $(n+1) \times d$ matrix. The embedding of the \cls token at the output layer serves as the image representation.

An MSA block consists of a number $H$ of heads, each computing a \emph{scaled dot-product self-attention}~\cite{attention}, \ie, the relevance of each image patch to others, encoded as an $(n+1) \times (n+1)$ \emph{attention matrix}. As discussed in \autoref{sec:mask}, we average attention matrices over all the heads of the last encoder layer and we use the row corresponding to the \cls token to generate token masks.


\paragraph{Distillation-based Masked Image Modeling.}

\emph{Self-distillation}, using a moving average of the student as teacher~\cite{tarvainen2017mean}, is studied for self-supervision in BYOL~\cite{byol} and extended to vision transformers in DINO \cite{dino}, which applies the distillation loss globally on the \cls token. iBOT~\cite{ibot} turns this task into \emph{masked image modeling} (MIM) by applying the loss densely on masked tokens.

Given an input image $X$ tokenized as $Z = (\vz^\cls; \vz_1; \dots; \vz_n)$, a \emph{mask vector} $\vm = (m_1, \dots, m_n) \in \{0, 1\}^n$ is generated, giving rise to a \emph{masked tokenized image} $\wt{Z} = (\vz^\cls; \tilde{\vz}_1; \dots; \tilde{\vz}_n)$, with
\begin{align}
	\tilde{\vz}_i = (1-m_i) \cdot \vz_i + m_i \cdot \vz^\mask
\label{eq:mask}
\end{align}
for $i = 1, \dots, n$, where $\vz^\mask \in \real^d$ is a learnable embedding of a ``mask'' token $\mask$.
Following the strategy of BEiT~\cite{beit}, the mask vector is generated with random \emph{block-wise} token sampling,
that is, defined in terms of random rectangles in the 2D layout of the $n$ tokens as a $(h/p) \times (w/p)$ matrix.

Following DINO~\cite{dino}, the transformer encoder is followed by a head that includes an MLP and scaled softmax, such that output token embeddings can be interpreted as probabilities. We denote by $f_\theta$ the mapping that includes the addition of the position embeddings, the encoder and the head, while $\theta$ is the set of learnable parameters. Given a tokenized image $Z$, masked or not, we denote by $f_\theta(Z) \in \real^{(n+1) \times d}$ the output token sequence and by $f_\theta(Z)_i, f_\theta(Z)^\cls \in \real^d$ the embedding of the $i$-th and \cls token respectively. The teacher parameters $\theta'$ are obtained from the student parameters $\theta$ by \emph{exponential moving average} (EMA) according to $\theta' \gets \alpha \theta' + (1-\alpha) \theta$.

For each input image, two standard resolution augmented \emph{global views} are generated, with tokenized images $Z^a, Z^b$ and mask vectors $\vm^a, \vm^b$. For each view $v$ in $V = \{a,b\}$ and for each masked token, the MIM objective is to minimize the reconstruction loss between the student $f_\theta$ output for the masked input $\wt{Z}^v$ and the teacher $f_{\theta'}$ output for the non-masked input $Z^v$:
\begin{align}
	L_{\textsc{mim}} = - \sum_{v \in V} \sum_{i=1}^n
		m_i^v f_{\theta'}(Z^v)_i \log({f_{\theta}(\wt{Z}^v)_i}).
\label{eq:mim}
\end{align}
Following DINO~\cite{dino}, a similar loss is applied globally on the \cls tokens between the student output for one masked view $\wt{Z}^v$ and the teacher output for the other non-masked view $Z^u$:
\begin{align}
    L_{\textsc{g}} = - \sum_{(u,v) \in V^2}
		\ind_{u \ne v} f_{\theta'}(Z^u)^\cls \log(f_{\theta}(\wt{Z}^v)^\cls).
\label{eq:g}
\end{align}
Finally, as detailed in the Appendix \autoref{sec:more-setup}, a \emph{multi-crop} strategy applies, giving rise to a loss $L_{\textsc{lc}}$~\eq{lc} between local crops and global views. The overall loss of iBOT~\cite{ibot} is a weighted sum of $L_{\textsc{mim}}$~\eq{mim} and $L_\textsc{g}$~\eq{g} $+$ $L_\textsc{lc}$~\eq{lc}. DINO itself uses the sum $L_\textsc{g}$~\eq{g} $+$ $L_\textsc{lc}$~\eq{lc} without masking.


\subsection{\ours: Attention-guided Token Masking} \label{sec:attmask}
\label{sec:mask}

Prior MIM-based self-supervised methods use random or block-wise random token masking. In this section we describe our attention-guided token masking strategy, which hides tokens that correspond to the salient regions of an image
and thus 
define a more challenging MIM objective.

\paragraph{Attention Map Generation.}

Given an input sequence $Y \in \real^{(n+1) \times d}$, a \emph{multi-head self-attention} (MSA) layer uses three linear layers to map $Y$ to the \emph{query} $Q_j$, \emph{key} $K_j$ and \emph{value} $V_j$ sequences for $j = 1, \dots, H$, where $H$ is the number of heads, $Q_j,K_j,V_j \in \real^{(n+1) \times d'}$ and $d' = d / H$. Then, it forms the $(n+1) \times (n+1)$ \emph{attention matrix}, where softmax is row-wise:
\begin{align}
	A_j = \softmax \left( Q_j K_j\tran / \sqrt{d'} \right).
\label{eq:att-mat}
\end{align}
To generate token masks from any layer of the transformer encoder, we average the attention matrices over all heads:
\begin{align}
  \overline{A} = \frac{1}{H}\sum_{j=1}^{H} A_j.
\label{eq:mean-att-mat}
\end{align}
Now, each row of an attention matrix is a vector in $\real^{n+1}$, that corresponds to one token and, excluding the diagonal elements, determines an \emph{attention vector} in $\real^n$ over all other tokens. We focus on the attention vector of the \cls token, which comprises all but the first elements of the first row of $\overline{A}$:
\begin{align}
	\overline{\va}^\cls & =\left(\overline{a}_{1,2}, \overline{a}_{1,3}, \dots ,\overline{a}_{1,n+1} \right),
\label{eq:att-cls}
\end{align}
where $\overline{a}_{i,j}$ is the element $i,j$ of $\overline{A}$. This vector can be reshaped to $(h/p) \times (w/p)$ \emph{attention map}, to be visualized as a 2D image, indicating the regions of the input image that the \cls token is attending.


\paragraph{Mask Generation: Highly-attended Tokens.}

There is a permutation $\sigma_{\downarrow} : \{ 1,\dots, n \} \rightarrow \{ 1,\dots, n \}$ that brings the elements of $\overline{\va}^\cls$ in descending order, such that $\overline{a}^\cls_{\sigma_{\downarrow}(i)} \ge \overline{a}^\cls_{\sigma_{\downarrow}(j)}$ for $i<j$, where $\overline{a}^\cls_i$ is the $i$-th element of $\overline{\va}^\cls$. Choosing a number $k = \floor{rn}$ that is proportional to the total number $n$ of tokens with \emph{mask ratio} $r \in [0,1]$, we define
\begin{align}
	M^H \coloneqq \{ \sigma_{\downarrow}(i), \dots ,\sigma_{\downarrow}(k) \}
\label{eq:top}
\end{align}
as the set of indices of the top-$k$ most attended tokens. We thus define the \emph{high-attention mask vector} $\vm^H$ with elements
\begin{align}
	m_i^H \coloneqq \ind_{M^H}(i) = \left\{
		\begin{array}{cc}
			1 & \quad \mif i \in M^H \\
			0 & \quad \other \\
		\end{array}
	\right.
\label{eq:high}
\end{align}
for $i=1, \dots, n$.
This masking strategy, which we call \athigh, essentially hides the patches that correspond to the most discriminative or salient regions of an image.
By \ours we shall refer to this strategy as default.


\paragraph{Low-attended Tokens.}

We also examine the opposite approach of \athigh that masks the least attended tokens. In particular, we define the set of indices of the bottom-$k$ least attended tokens $M^L = \{ \sigma_{\uparrow}(i), \dots ,\sigma_{\uparrow}(k) \}$ and the \emph{low-attention mask vector} $\vm^L$ with $m_i^L \coloneqq \ind_{M^L}(i)$ based on the permutation $\sigma_{\uparrow}$ that brings the elements of $\overline{\va}^\cls$ in ascending order, that is, $\overline{a}^\cls_{\sigma_{\downarrow}(i)} \le \overline{a}^\cls_{\sigma_{\downarrow}(j)}$ for $i<j$. 
This strategy, which we call \atlow and is similar to the masking strategy of MST~\cite{mst}, hides patches of the image background.
Our experiments show that \atlow does not work well with the considered MIM-based loss.

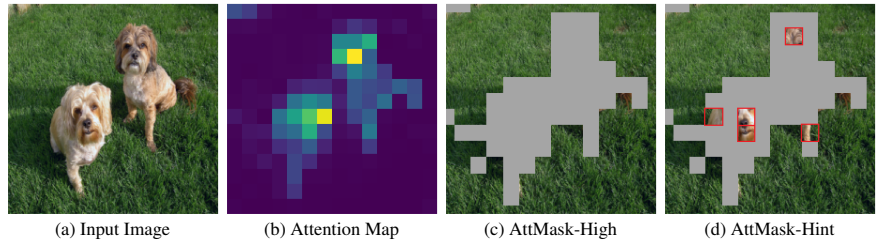
\begin{figure}[t]
\centering
\input{tex/fig_high_vs_hints}
\caption{Given image (a), the mean attention map (b) is averaged over heads \eq{mean-att-mat},\eq{att-cls}.
The \athigh strategy (c) masks the most attended patches, while \athint (d) reveals few of them to leave hints about the identity
of the masked object. }
\label{fig:high_vs_hints}
\end{figure}

\paragraph{Highly-attended with Hints.}

Finally, because \athigh may be overly aggressive in hiding the foreground object of an image, especially when the mask ratio $r$ is high, we also examine an alternative strategy that we call \athint: While still masking highly attended tokens, we allow a small number of the most highly attended ones to be revealed, so as to leave hints about the identity of the masked object. In particular, we remove from the initial set $M^H$ a small number $m = \floor{sn}$ of tokens with \emph{show ratio} $s < r$.
These $m$ tokens are randomly selected from the $\floor{s_{\max}n}$ most attended tokens in $M^H$, where $s_{\max} > s$.
An example comparing \athint with \athigh is illustrated in \autoref{fig:high_vs_hints}.


\paragraph{Incorporating \ours into Self-supervised Methods.}
\label{sec:plug}

Because the embedding of the \cls token at the output layer of the transformer encoder serves as the image representation, we generate token masks based on the attention vector precisely of the \cls token of the output layer. In particular, given a global view tokenized as $Z^v = (\vz^\cls; \vz_1; \dots; \vz_n)$, we obtain the attention vector $\overline{\va}^\cls$~\eq{att-cls} and the corresponding high-attention mask vector $\vm^H$~\eq{high} at the output layer of the teacher. Then, similarly to~\eq{mask}, we give as input to the student the masked version $\wt{Z}^v = (\vz^\cls; \tilde{\vz}_1; \dots; \tilde{\vz}_n)$ with
\begin{align}
	\tilde{\vz}_i = (1-m_i^H) \cdot \vz_i + m_i^H \cdot \vz^\mask.
\label{eq:mask-high}
\end{align}
We argue that masking highly attended regions using $\vm^H$ helps in learning powerful representations. In \autoref{sec:exp}, we also experiment with low-attended regions using $\vm^L$, supporting further our argument.

\ours can be incorporated into different methods to either replace the block-wise strategy of BEiT~\cite{beit} or introduce masking. For iBOT~\cite{ibot}, we use $\wt{Z}^v$ in $L_{\textsc{mim}}$~\eq{mim} and $L_\textsc{g}$~\eq{g}. For DINO~\cite{dino}, we introduce masking by using $\wt{Z}^v$ for global views in $L_\textsc{g}$~\eq{g}, but not for local crops in the $L_\textsc{lc}$~\eq{lc} loss (see Appendix \autoref{sec:more-setup}).

%% file: tex/fig_high_vs_hints.tex
\scriptsize
\centering
\setlength{\tabcolsep}{1.5pt}
\begin{tabular}{cccc}
	\fig[.23]{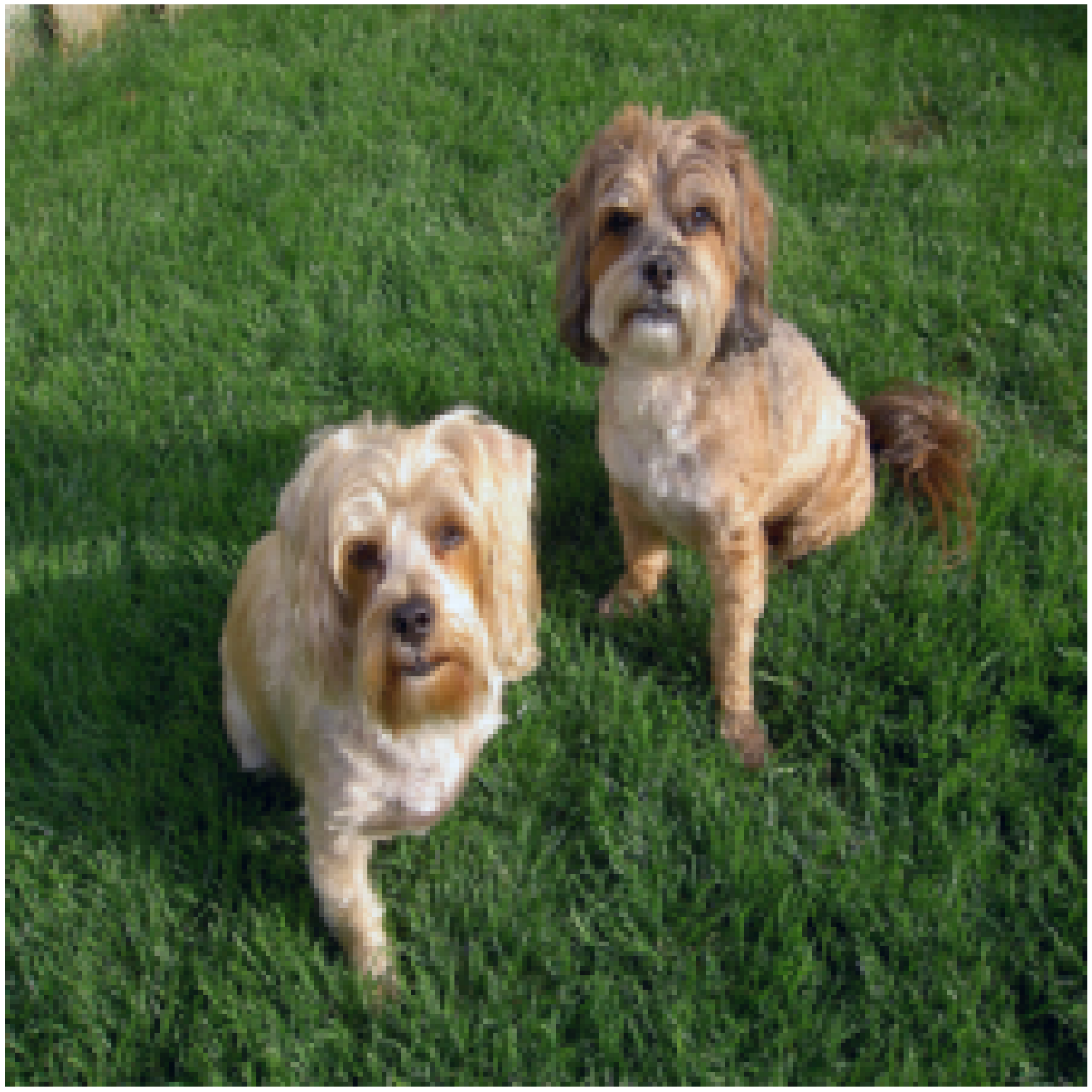}                           &
	\fig[.23]{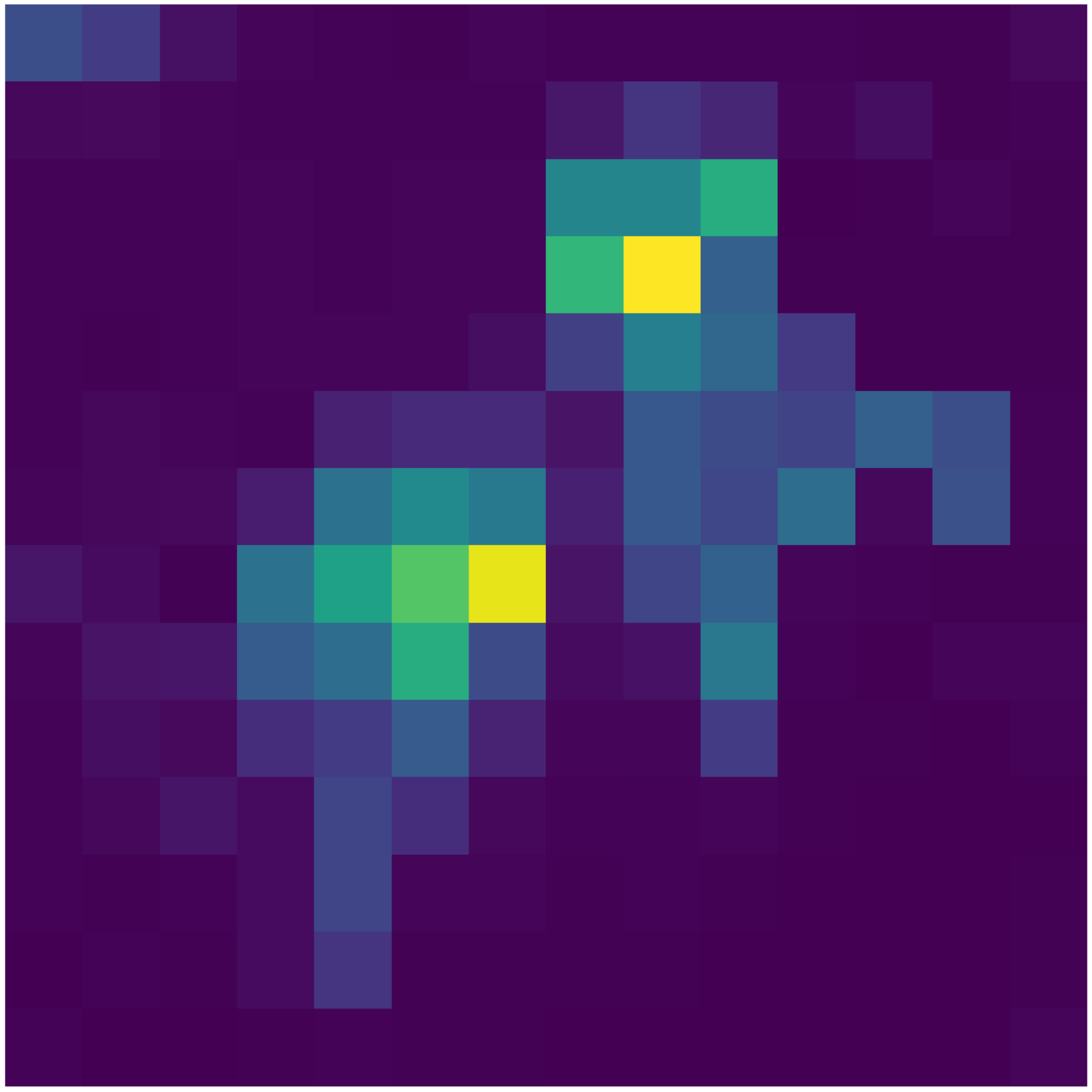}                 &
	\fig[.23]{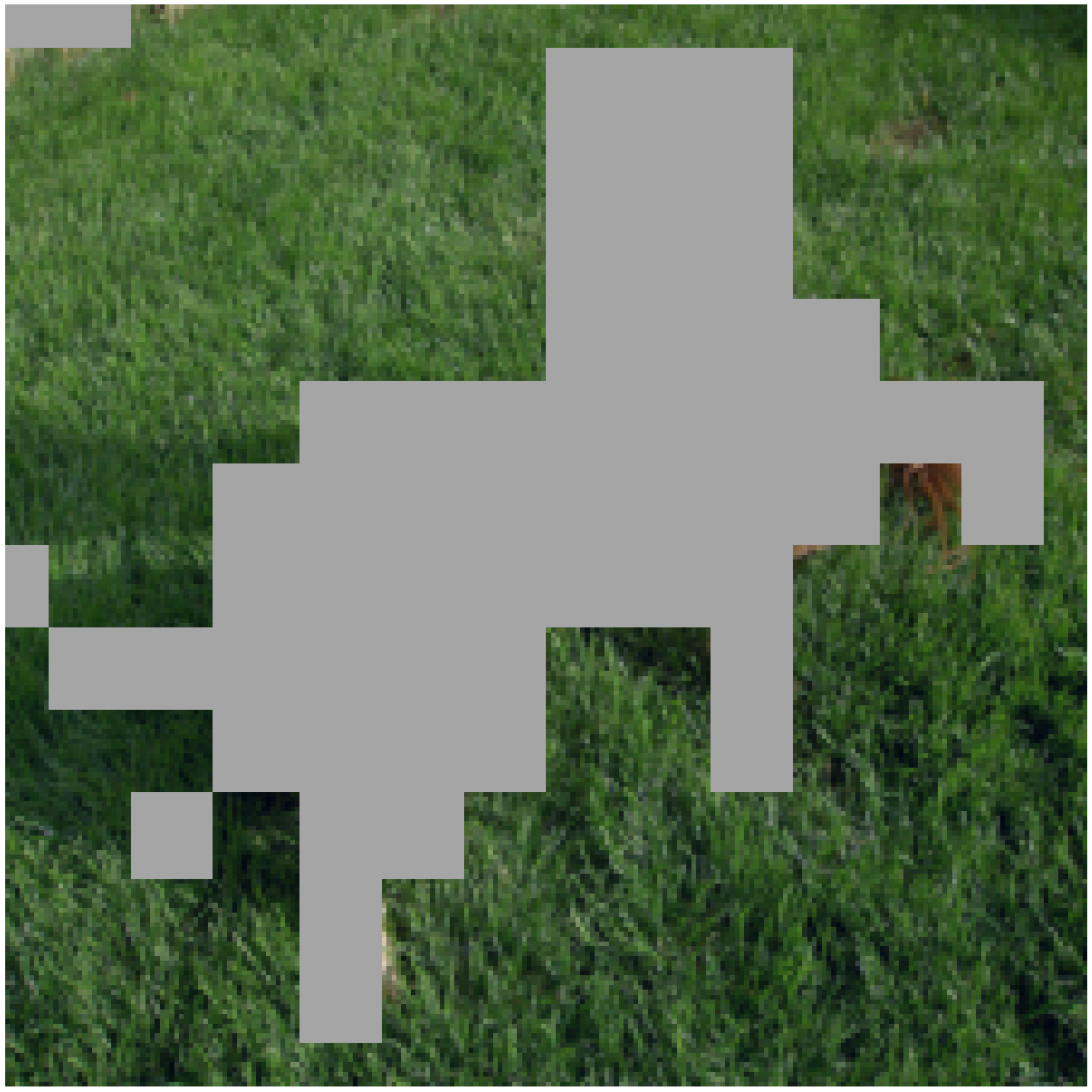}                             &
	\fig[.23]{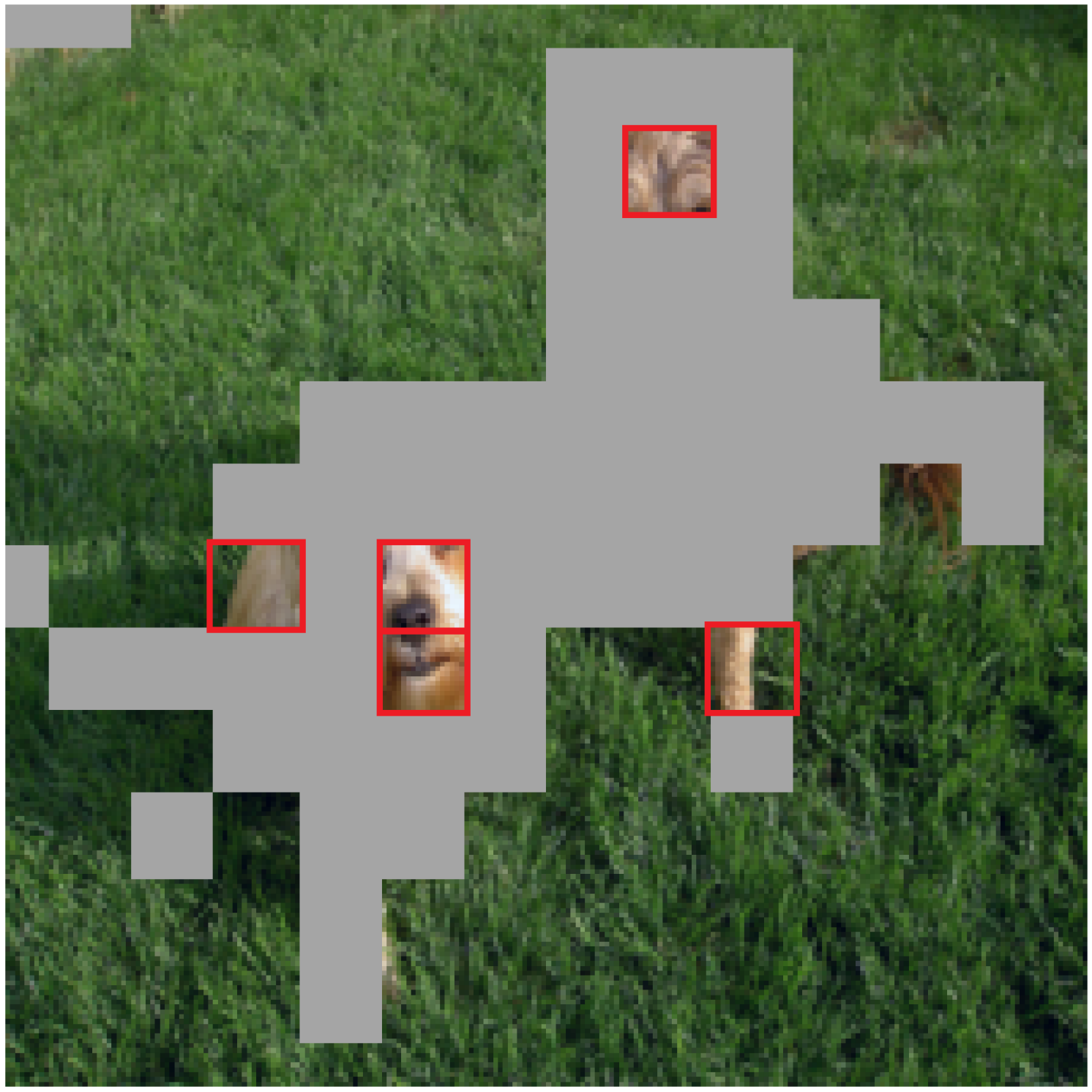}                            \\

	(a) Input Image                                                 &
	(b) Attention Map                                              &
	(c) \athigh                                                     &
	(d) \athint                                                     \\
	
\end{tabular}

%% file: tex/exp-setup.tex
\section{Experiments}
\label{sec:exp}

\subsection{Setup}
\label{sec:setup}


\paragraph{Datasets and Evaluation Protocol.}

We pretrain iBOT and DINO on 20\% and 100\% of the ImageNet-1k~\cite{imagenet} training set. For 20\%, we select the first 20\% of training samples per class. We evaluate on ImageNet-1k validation set by $k$-NN or \emph{linear probing}. For linear probing, we train a linear classifier on top of features using the same training protocol as in DINO~\cite{dino}. With linear probing, we also validate robustness against background changes on ImageNet-9 (IN-9)~\cite{xiao2021noise}. For $k$-NN~\cite{instance_discrimination}, we freeze the pretrained model and extract features of training images, then use a $k$-nearest neighbor classifier with $k = 20$. 
We also perform the same $k$-NN experiment, now extracting features only from $\nu \in \{1,5,10,20\}$ examples per class. This task is more challenging and is similar to few-shot classification, only the test classes are the same as in representation learning. 

We downstream to other tasks either with or without \emph{finetuning}. We finetune on CIFAR10~\cite{cifar}, CIFAR100~\cite{cifar} and Oxford Flowers~\cite{flowers} for \emph{image classification} measuring accuracy; on COCO~\cite{coco} for \emph{object detection} and \emph{instance segmentation} measuring mean average precision (mAP); and on ADE20K~\cite{ade20k} for \emph{semantic segmentation} measuring mean Intersection over Union (mIoU).
Without finetuning, we extract features as with $k$-NN and we evaluate using dataset-specific evaluation protocol and metrics. We test on revisited $\cR$Oxford and $\cR$Paris~\cite{radenovic2018revisiting} for \emph{image retrieval} measuring mAP~\cite{radenovic2018revisiting}; on Caltech-UCSD Birds (CUB200)~\cite{cub}, Stanford Cars (CARS196)~\cite{cars}, Stanford Online Products (SOP)~\cite{sop} and In-Shop Clothing Retrieval (In-Shop)~\cite{in_shop} for \emph{fine-grained classification} measuring Recall@$k$~\cite{reality_check}; and on DAVIS 2017~\cite{davis} for \emph{video object segmentation} measuring mean region similarity $\cJ_m$ and contour-based accuracy $\cF_m$~\cite{davis}.

In Appendix \autoref{sec:more-experiments}, we provide more benchmarks, visualizations and ablations.

\paragraph{Implementation Details.}

As transformer encoder, we use ViT-S/16~\cite{vit}. The attention map~\eq{att-cls} is generated from the last layer of the teacher encoder by default, \ie, layer 12. We mask the input with probability $p=0.5$, while the mask ratio $r$ is sampled uniformly as $r \sim U(a, b)$ with $[a, b] = [0.1, 0.5]$ by default.
For \athint, we set $s_{\max} = 0.1$ and the show ratio $s$ is sampled uniformly from $[s_{\max} a, s_{\max} b]=[0.01, 0.05]$. 
Following~\cite{dino,ibot}, we apply \emph{multi-crop}~\cite{swav} scheme, as detailed in Appendix \autoref{sec:more-setup}.
The overall loss of iBOT~\cite{ibot} is a weighted sum of $L_{\textsc{mim}}$~\eq{mim}, with weight $\lambda$, and $L_\textsc{g}$~\eq{g} $+$ $L_\textsc{lc}$~\eq{lc} (DINO~\cite{dino}), with weight 1, where $L_\textsc{lc}$~\eq{lc} is the multi-crop loss. By default, $\lambda = 1$. Hyperparameters are ablated in \autoref{sec:ablation}.
Training details are given in the Appendix \autoref{sec:more-setup}.

%% file: tex/exp-analysis.tex
\subsection{Experimental Analysis}
\label{sec:analysis}

We provide an analysis on 20\% of ImageNet-1k training samples, incorporating \ours into distillation-based MIM~\cite{ibot} or self-distillation only~\cite{dino}. We also provide results on robustness against background changes.


\paragraph{Masking Strategies: Distillation-based MIM.}

We explore a number of masking strategies using distillation-based MIM, by incorporating \ours into iBOT~\cite{ibot}. We compare \ours with random block-wise masking~\cite{beit}, which is the default in iBOT, random patch masking with the same ratio, as well as with a more aggressive ratio, following MAE~\cite{mae}. \ours masks the most attended tokens (\athigh) by default, but we also consider the least attended (\atlow) and the most attended with hints (\athint).

\begin{table}[t]
\small
\centering
\setlength{\tabcolsep}{3pt}
\caption{\emph{Different masking strategies} for iBOT~\cite{ibot} pre-training on 20\% of ImageNet.
Top-1 accuracy for $k$-NN, linear probing on ImageNet validation set; fine-tuning on CIFAR10/100. $\dagger$: default iBOT masking strategy from BEiT~\cite{beit}. $\ddagger$: aggressive random masking strategy from MAE~\cite{mae}.}
\label{tab:compAttMask}
\begin{tabular}{lccccc} \toprule
\mr{2}{\Th{iBOT Masking}} & \mr{2}{\Th{Ratio (\%)}} & \mc{2}{\Th{ImageNet-1k}} & CIFAR10 & CIFAR100 \\ \cmidrule{3-6}
& & \Th{$k$-NN} & \Th{Linear} & \mc{2}{\Th{Fine-tuning}} \\ \midrule
Random Block-Wise${}^{\dagger}$  & 10-50 & 46.7 & 56.4 & 98.0 & 86.0 \\ 
Random$^\ddagger$ & 75 & 47.3 & 55.5 & 97.7 & 85.5 \\
Random & 10-50 & 47.8 & 56.7 & 98.0 & 86.1 \\ \midrule
\atlow (ours) & 10-50 & 44.0 & 53.4 & 97.6 & 84.6 \\
\athint (ours) & 10-50 & 49.5 & 57.5 & 98.1 & \bf{86.6} \\
\rowcolor{TableColor}
\athigh (ours) & 10-50 & \bf{49.7} & \bf{57.9} & \bf{98.2} & \bf{86.6} \\ 
\bottomrule
\end{tabular}
\end{table}

We evaluate performance using $k$-NN and linear probing evaluation protocol on the validation set, along with a fine-tuning evaluation on CIFAR10 and CIFAR100. As shown in \autoref{tab:compAttMask}, the \athigh outperforms all other masking strategies on all the evaluation metrics. In particular, \athigh achieves an improvement of +3.0\% on $k$-NN and +1.5\% on linear probing compared with the default iBOT strategy (random block-wise).

Interestingly, random patch masking outperforms the default iBOT strategy, while the more aggressive MAE-like strategy is inferior and \atlow performs the lowest. Intuitively, this means that masking and reconstruction of non-salient regions does not provide a strong supervisory signal under a MIM objective. By contrast, our \ours creates the more aggressive task of reconstructing the most salient regions and guides the model to explore the other regions. In this setup, \athint is slightly lower than \athigh{}.

\begin{table}[t]
\raisebox{-15pt}{
\begin{minipage}[b]{0.5\linewidth}

\small
\centering
\setlength{\tabcolsep}{1.5pt}
\caption{Top-1 $k$-NN accuracy on ImageNet-1k validation for iBOT pre-training on different percentage (\%) of ImageNet-1k. $\dagger$: default iBOT masking strategy from BEiT~\cite{beit}.}
\label{tab:INPercAttMask}
\begin{tabular}{lcccc} \toprule
\Th{\% ImageNet-1k} & 5 & 10 & 20 & 100 \\ \midrule
Random Block-Wise$^\dagger$ & 15.7 & 31.9 & 46.7 & 71.5 \\
\rowcolor{TableColor}
\athigh (ours) & \bf{17.5} & \bf{33.8} & \bf{49.7} & \bf{72.5} \\
\bottomrule
\end{tabular}

\end{minipage}\hfill
}
\hfill
\begin{minipage}[t]{0.45\linewidth}

\small
\centering
\pgfplotstableread{
	epoch rand    att
	1     1.464   1.57
	6     8.274   9.11
	11    15.994  17.896
	16    21.036  24.756
	21    25.564  30.096
	26    29.706  34.124
	31    33.012  37.27
	36    35.782  40.038
	41    38.286  41.832
	46    39.938  43.74
	51    41.272  45
	56    42.63   46.146
	61    43.568  47.346
	66    44.508  48.058
	71    45.228  48.676
	76    45.84   48.952
	81    46.278  49.468
	86    46.68   49.648
	91    46.7    49.694
	96    46.67   49.676
	100   46.704  49.708
}{\epoch}
\extfig{epoch}{
\begin{tikzpicture}[%
	lab/.style={fill=white,fill opacity=.8},
]
\begin{axis}[%
	width=1.15\linewidth,
	height=0.8\linewidth,
	font=\tiny,
	xlabel={epoch},
	ylabel={$k$-NN},
	legend pos=south east,
]
	\draw[green,dashed,very thick] (axis cs:0,46.704)--(axis cs:110,46.704);
	\addplot[blue!70] table[x=epoch,y=rand] \epoch; \leg{Random Block-Wise$^\dagger$};
	\addplot[red!70]  table[x=epoch,y=att]  \epoch; \leg{\athigh (ours)};
	\node[dot] (x) at (axis cs:58,46.704){};
	\node[font=\small] (l) at (axis cs:80,30){42\% fewer \\ epochs};
	\draw[->] (l)--(x);
\end{axis}
\end{tikzpicture}
}
\vspace{-10pt}

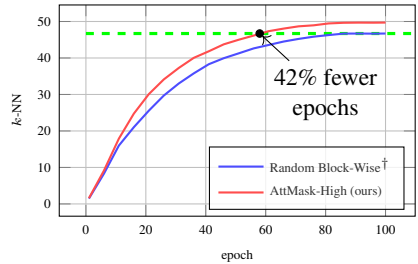
\captionof{figure}
{Top-1 $k$-NN accuracy on ImageNet-1k validation for iBOT training \vs training epoch on 20\% ImageNet training set. $\dagger$: default iBOT masking strategy from BEiT~\cite{beit}.}
\label{fig:perc}

\end{minipage}
\end{table}

\paragraph{Data and Training Efficiency.}

Self-supervised methods on vision transformers typically require millions of images, which is very demanding in computational resources. We advocate that being effective on less data and fast training are good properties for a self-supervised method. In this direction, we assess efficiency on less data and training time, still with iBOT training. In \autoref{tab:INPercAttMask} we observe that our \athigh consistently outperforms the default random block-wise masking strategy of iBOT at lower percentage of ImageNet-1k training set. In addition, in \autoref{fig:perc}, \athigh achieves the same performance as random block-wise with 42\% fewer training epochs.

\begin{table}[t]
\small
\centering
\setlength{\tabcolsep}{3pt}
\caption{Top-1 $k$-NN accuracy on ImageNet-1k validation for DINO~\cite{dino} pre-training on 20\% of the ImageNet-1k training set using mask ratio of 10-50\%. $\dagger$: default DINO.}
\label{tab:DINOAttMask}
\begin{tabular}{ccccc} \toprule
No Masking${}^{\dagger}$ & Random & \atlow & \athint & \athigh \\ \midrule
43.0 & 43.4 & 42.7 & \bf{43.6} & 43.5\\ \bottomrule
\end{tabular}
\end{table}

\paragraph{Masking Strategies: Self-distillation Only.}

Here, we compare masking strategies using distillation only, without MIM reconstruction loss, by incorporating \ours into DINO~\cite{dino}. That is, we apply only the cross-view cross-entropy loss on the \cls token~\eq{g}. In \autoref{tab:DINOAttMask}, \athigh improves $k$-NN by +0.5 compared with the default DINO (no masking), while \atlow is inferior. This reveals that \ours is effective even without a MIM loss. Moreover, \athint is slightly better than \athigh in this setting.

\begin{table}[t]
\small
\centering
\setlength{\tabcolsep}{0.8pt}
\caption{\emph{Background robustness}: Linear probing of iBOT model on IN-9~\cite{xiao2021noise} and its variations, when pre-trained on 20\% ImageNet-1k under different masking strategies. $\dagger$: default iBOT masking strategy from BEiT~\cite{beit}. $\ddagger$: aggressive random masking.}
\label{tab:BGChalAttMask}
\begin{tabular}{lccccccccc}
    {}                                                    &
    {}                                                    &
	\fig[.072]{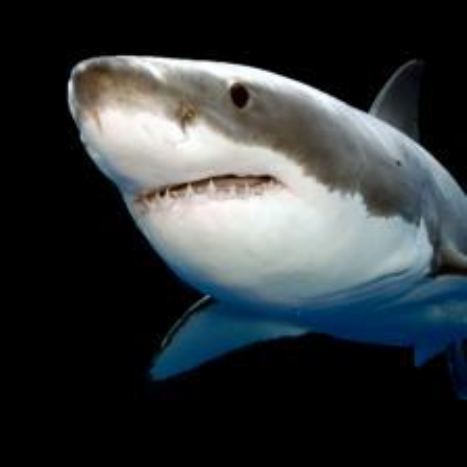}              &
	\fig[.072]{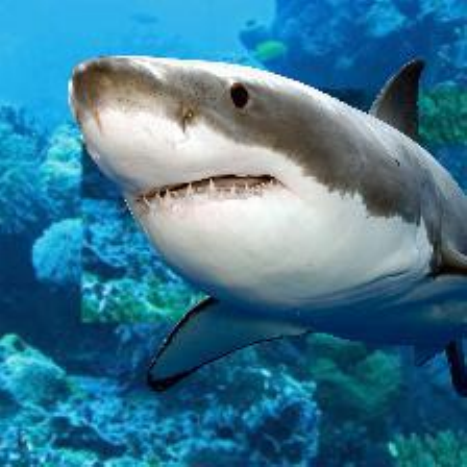}           &
	\fig[.072]{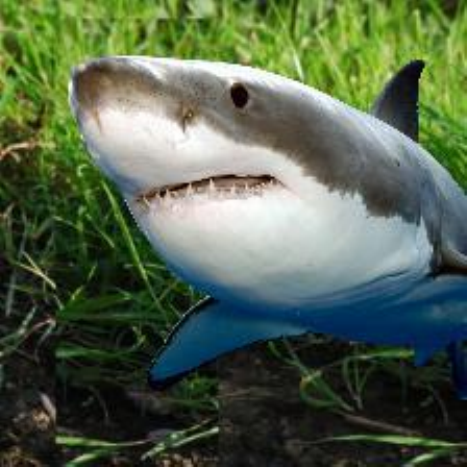}           &
	\fig[.072]{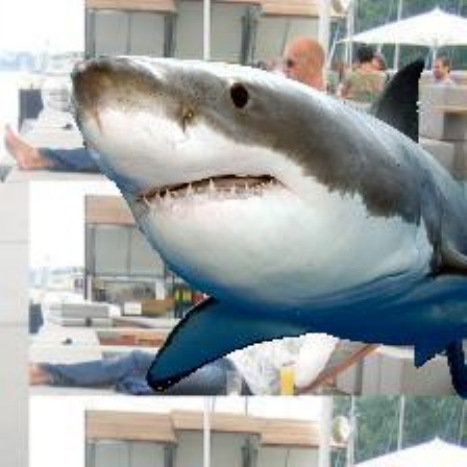}           &
	\fig[.072]{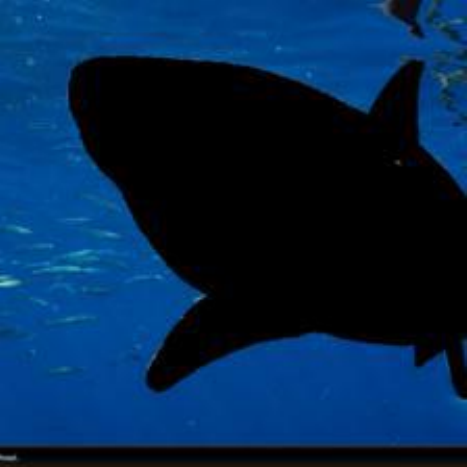}                &
	\fig[.072]{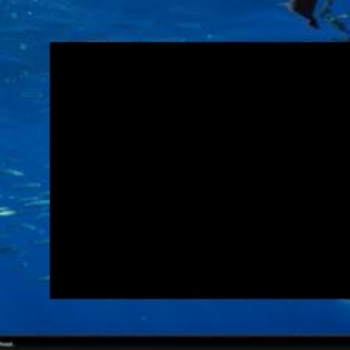}             &
	\fig[.072]{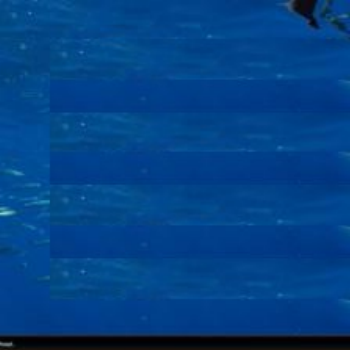}             &
	\fig[.072]{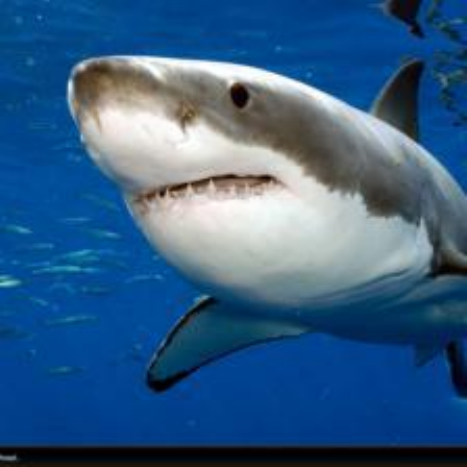}             \\

\toprule
\Th{iBOT Masking} & \Th{Ratio (\%)} & OF & MS & MR & MN & NF & OBB & OBT & IN-9 \\ \midrule
Random Block-wise${}^{\dagger}$ & 10-50 & 72.4 & 74.3 & 59.4 & 56.8 & 36.3 & 14.4 & 15.0 & 89.1 \\
Random$^\ddagger$ & 75 & 73.1 & 73.8 & 58.8 & 55.9 & 35.6 & 13.7 & 14.5 & 87.9 \\
Random & 10-50 & 72.8 & 75.3 & 60.4 & 57.5 & 34.9 & 10.3 & 14.4 & 89.3 \\ \midrule
\atlow (ours) & 10-50 & 66.0 & 71.1 & 55.2 & 52.2 & 32.4 & 12.5 & 14.0 & 86.6 \\

\rowcolor{TableColor}
\athint (ours) & 10-50 & 74.4 & 75.9 & 61.7 & 58.3 & 39.6 & \textbf{16.7} & \textbf{15.7} & 89.6 \\

\rowcolor{TableColor}
\athigh (ours) & 10-50 & \textbf{75.2} & \textbf{76.2} & \textbf{62.3} & \textbf{59.4} & \textbf{40.6} & 15.2 & 15.3 & \textbf{89.8} \\ \bottomrule
\end{tabular}
\end{table}

\paragraph{Robustness Against Background Changes.}

Deep learning models tend to depend on image background. However, to generalize well, they should be robust against background changes and rather focus on foreground. To analyze this property, we use ImageNet-9 (IN-9) dataset~\cite{xiao2021noise}, which includes nine coarse-grained classes with seven background/foreground variations.
In four datasets, the background is altered: Only-FG (OF), Mixed-Same (MS), Mixed-Rand (MR), and Mixed-Next (MN). In another three, the foreground is masked: No-FG (NF), Only-BG-B (OBB), and Only-BG-T (OBT).

In \autoref{tab:BGChalAttMask}, we evaluate the impact of background changes on IN-9 and its variations, training iBOT  under different masking strategies. We observe that, except for O.BB. and O.BT, \athigh is the most robust. On OBB and OBT where the foreground object is completely missing, \athint exploits slightly better the background correlations with the missing object.

In the Appendix \autoref{sec:more-vis}, we provide visualizations of attention maps in \autoref{fig:attmaps} and masking examples in \autoref{fig:MoreStrategies3}.

%% file: tex/exp-bench.tex
\subsection{Benchmark}
\label{sec:bench}

We pre-train iBOT with \athigh and \athint on 100\% of ImageNet-1k and compare it with baseline iBOT and other distillation-based methods.

\begin{table}[t]
\small
\centering
\setlength{\tabcolsep}{3pt}
\caption{Top-1 accuracy on ImageNet validation set. (a) $k$-NN and linear probing using the full ImageNet training set; (b) k-NN using only $\nu \in \{1,5,10,20\}$ examples per class. Pre-training on 100\% ImageNet-1k for 100 epochs.}
\label{tab:100percImageNetAND100percFewShot}
\begin{tabular}{lcc|cccc} \toprule
\mr{2}{\Th{Method}} & \mc{2}{(a) \Th{Full}} & \mc{4}{(b) \Th{Few Examples}} \\  \cmidrule{2-7}
&  $k$-NN   & \Th{Linear}& $\nu=1$     & 5         & 10        & 20 \\
\midrule
DINO~\cite{dino}    &  70.9     & 74.6       &   &     &  &      \\
MST~\cite{mst}    &  72.1     & 75.0        &   &    &  &      \\
iBOT~\cite{ibot}    &  71.5     & 74.4     &    32.9    & 47.6    & 52.5      & 56.4      \\
iBOT+\athigh  & 72.5 & 75.7 & 37.1 & 51.3 & 55.7 & 59.1 \\
iBOT+\athint  & \bf{72.8} & \bf{76.1} & \bf{37.6} & \bf{52.2} & \bf{56.4} & \bf{59.6} \\
\bottomrule
\end{tabular}
\end{table}

\paragraph{ImageNet Classification.}

As shown in \autoref{tab:100percImageNetAND100percFewShot}(a), \athigh brings an improvement of 1\% $k$-NN and 1.3\% linear probing over baseline iBOT~\cite{ibot} and is better than prior methods. \athigh is thus effective for larger datasets too. 
\autoref{tab:100percImageNetAND100percFewShot}(b) shows results of the more challenging task where only $\nu \in \{1,5,10,20\}$ training examples per class are used for the $k$-NN classifier. In this case, \athigh is very effective, improving the baseline iBOT masking strategy by 3-4\%, demonstrating the quality of the learned representation.
In this setup, \athint offers a further small improvement over \athigh. For simplicity though, we use \athigh by default as \ours. 

More results are given in the Appendix. In particular, in \autoref{tab:sup-100percImageNet300epANDFewShot}, we provide results similar to \autoref{tab:100percImageNetAND100percFewShot} but with pre-training for 300 epochs. Also, in \autoref{tab:SupervisedFinetuning} we report further supervised finetuning on ImageNet-1k. In \autoref{tab:GAP}, we investigate the quality of the patch features by using \emph{global average pooling} (GAP) rather than the [CLS] token embeddings. In \autoref{tab:MaskingAcc}, we study the effect of masking salient image parts at inference.

\begin{table}[t]
\small
\centering
\setlength{\tabcolsep}{5pt}
\caption{Fine-tuning for \emph{image classification} on CIFAR10~\cite{cifar}, CIFAR100~\cite{cifar} and Oxford Flowers~\cite{flowers}; \emph{Object detection} 
(AP$^b$, \%) and \emph{instance segmentation} (AP$^m$, \%) on COCO~\cite{coco}; and \emph{semantic segmentation} on ADE20K~\cite{ade20k} (mIoU, \%). Models pre-trained on 100\% ImageNet-1k training set for 100 epochs.}

\label{tab:100percDownStream}
\begin{tabular}{lccc|cc|c} \toprule
\mr{2}{\Th{Method}} & CIFAR10 & CIFAR100 & \Th{Flowers} & \multicolumn{2}{c|}{COCO} & ADE20K \\ \cmidrule{2-7}
 & \multicolumn{3}{c|}{Accuracy} & AP$^b$ & AP$^m$ & mIoU \\ \midrule
iBOT & \bf{98.8} & 89.5 & 96.8 &  48.2 & 41.8 & 44.9 \\
iBOT+\ours & \bf{98.8} & \bf{90.1} & \bf{97.7} & \bf{48.8} & \bf{42.0} & \bf{45.3} \\ \bottomrule
\end{tabular}
\end{table}

\paragraph{Downstream Tasks with Fine-tuning.}

We fine-tune the pre-trained models with iBOT and iBOT with \ours for \emph{image classification} on CIFAR10~\cite{cifar}, CIFAR100~\cite{cifar} and Oxford Flowers~\cite{flowers}, \emph{object detection} and \emph{instance segmentation} on COCO~\cite{coco}, and \emph{semantic segmentation} on ADE20K~\cite{ade20k}.
In \autoref{tab:100percDownStream}, we observe that \ours brings small improvement on the baseline iBOT masking strategy on \emph{image classification} fine-tuning in all cases. Furthermore, we observe that \ours improves clearly the scores by 0.6\% AP$^b$ on object detection and 0.4\% mIoU on semantic segmentation.

\begin{table}[t]
\small
\centering
\setlength{\tabcolsep}{3pt}
\caption{\emph{Image retrieval} (mAP, \%) on (a) $\cR$Oxford and (b) $\cR$Paris~\cite{radenovic2018revisiting} and \emph{video object segmentation} (mean region similarity $\cJ_m$ and contour-based accuracy $\cF_m$, \%) on (c) DAVIS 2017~\cite{davis}, without fine-tuning. Models pre-trained on 100\% ImageNet-1k training set for 100 epochs.}
\label{tab:100percDownStreamRetVid}
\begin{tabular}{lcc|cc|ccc} \toprule
\mr{2}{\Th{Method}} & \mc{2}{(a) \Th{$\cR$Oxford}} & \mc{2}{(b) \Th{$\cR$Paris}} & \mc{3}{(c) DAVIS 2017}                     \\ \cmidrule{2-8}
                    & \Th{Medium} & \Th{Hard}  & \Th{Medium} & \Th{Hard} & $(\cJ\&\cF)_m$ & $\cJ_m$   & $\cF_m$   \\ \midrule
iBOT                & 31.0        & 11.7       & 56.2        & 28.9      & 60.5           & 59.5      & 61.4      \\
iBOT+\ours          & \bf{33.5}   & \bf{12.1}  & \bf{59.0}   & \bf{31.5} & \bf{62.1}      & \bf{60.6} & \bf{63.5} \\
\bottomrule
\end{tabular}
\end{table}

\begin{table}[t]
\small
\centering
\setlength{\tabcolsep}{2pt}
\caption{Fine-grained classification (R@$k$: Recall@$k$, \%)~\cite{reality_check} without fine-tuning. Models pre-trained on 100\% ImageNet-1k training set for 100 epochs.}
\label{tab:100percDownStreamMetric}
\begin{tabular}{lccc|ccc|ccc|ccc} \toprule
\mr{2}{\Th{Method}} & \mc{3}{CUB200} & \mc{3}{CARS196} & \mc{3}{SOP} & \mc{3}{\Th{In-Shop}} \\ \cmidrule{2-13}
& R@1 & 2 & 4 & R@1 & 2 & 4 & R@1 & 10 & 100 & R@1 & 10 & 20 \\ \midrule
iBOT &  51.4 & 63.8 & 75.0 & 35.6 & 46.0 & 56.3 & 57.4 & 72.2 & 84.0 & 39.1 & 61.9 & 68.2 \\
iBOT+\ours & \bf{57.2} & \bf{69.4} & \bf{80.3} & \bf{39.8} & \bf{50.4} & \bf{61.4} & \bf{59.0} & \bf{73.9} & \bf{85.4} & \bf{40.7} & \bf{63.7} & \bf{70.3} \\
\bottomrule
\end{tabular}
\end{table}

\paragraph{Downstream Tasks without Fine-tuning.}

Without finetuning, we use the pretrained models with iBOT and iBOT with \ours to extract features as with $k$-NN and we evaluate using dataset-specific evaluation protocol and metrics. As shown in \autoref{tab:100percDownStreamRetVid}(a,b), \ours is very effective on image retrieval, improving by 1-3\% mAP the baseline iBOT masking strategy on $\cR$Oxford and $\cR$Paris~\cite{radenovic2018revisiting}, on both medium and hard protocols. More impressive the performance on fine-grained classification, improving by 2-6\% R@1 on all datasets, as shown in \autoref{tab:100percDownStreamMetric}. Finally, \ours improves on video object segmentation on DAVIS 2017~\cite{davis} on all metrics, as shown in \autoref{tab:100percDownStreamRetVid}(c). These experiments are very important because they evaluate the quality of the pretrained features as they are, without fine-tuning and without even an additional layer, on datasets of different distribution than the pretraining set. \ours improves performance by a larger margin in this type of tasks, compared with ImageNet.

In \autoref{tab:100percMoreDownStream} in the Appendix, we additionally provide results for \emph{scene classification} with linear probing on Places205~\cite{NIPS2014_5349}.

%% file: tex/exp-ablation.tex
\subsection{Ablation Study}
\label{sec:ablation}

We provide an ablation for the main choices and hyperparameters of our masking strategy and loss function, incorporating \ours into iBOT~\cite{ibot} and pre-training on 20\% of ImageNet-1k training samples. We provide additional ablations in the Appendix. In \autoref{tab:LossAttMask}, we examine the MIM loss weight. In \autoref{tab:RandomMaskThreshold}, we ablate both the masking strategy and the mask ratio $r$.

\begin{table}[t]
\small
\centering
\setlength{\tabcolsep}{3pt}
\caption{\ours $k$-NN top-1 accuracy on ImageNet-1k validation for iBOT pre-training on 20\% of ImageNet-1k \vs (a) layer from which the attention map~\eq{att-cls} is generated; (b) masking probability $p$ (using batch size 180); and (c) mask ratio $r$.}
\label{tab:ParamAttMask}
\begin{tabular}{cccc|ccccc|cccc} \toprule
\mc{4}{(a) \Th{Layer}} & \mc{5}{(b) \Th{Masking Prob $p$}} & \mc{4}{(c) \Th{Mask Ratio $r$ (\%)}} \\ \midrule
6    & 9    & 11        & 12   & 0    & 0.25 & 0.50      & 0.75      & 1    & 10-30 & 10-50     & 10-70 & 30   \\ \midrule
48.1 & 48.1 & \bf{49.8} & 49.7 & 43.4 & 47.3 & \bf{49.4} & \bf{49.4} & 44.2 & 49.5  & \bf{49.7} & 48.5  & 49.1 \\ \bottomrule
\end{tabular}
\end{table}

\paragraph{Layer for Attention Map Generation.}

The attention map~\eq{att-cls} is generated from the last layer of the teacher encoder by default, that is, layer 12 of ViT-S. In \autoref{tab:ParamAttMask}(a), we aim to understand the impact of other layer choices on \ours. We observe that the deeper layers achieve the highest $k$-NN performance. Although layer 11 works slightly better, we keep the choice of layer 12 for simplicity, since layer 12 embeddings are used anyway in the loss function.

\paragraph{Masking Probability and Mask Ratio.}

We mask the global views with probability $p = 0.5$ by default. \autoref{tab:ParamAttMask}(b) reports on other choices and confirms that this choice is indeed best. Therefore, it is useful that student network sees both masked and non-masked images.

The mask ratio $r$ is sampled uniformly as $r \sim U(a, b)$ with $[a, b] = [0.1, 0.5]$ by default.
\autoref{tab:ParamAttMask}(c) shows the sensitivity of \ours with respect to the upper bound $b$, along with a fixed ratio $r = 0.3$. \ours is relatively stable, with the default interval $[0.1, 0.5]$ working best and the more aggressive choice $[0.1, 0.7]$ worst. This is possibly due to the foreground objects being completely masked and confirms that  masking the most attended patches is an effective strategy. The added variation around the fixed ratio $r = 0.3$ is beneficial.

%% file: tex/conclusion.tex
\section{Conclusion}
\label{sec:conclusion}

By leveraging the self-attention maps of ViT for guiding token masking, our \ours is able to hide from the student network discriminative image cues and thus lead to more challenging self-supervised objectives. We empirically demonstrate that \ours offers several benefits over random masking when used in self-supervised pre-training with masked image modeling. Notably, it accelerates the learning process, achieves superior performance on a variety of downstream tasks, and it increases the robustness against background changes, thus revealing that it reduces background dependency. The improvement is most pronounced in more challenging downstream settings, like using the pretrained features without any additional learning or finetuning, or working with limited data. This reveals the superior quality of the learned representation.

%% file: tex/acknowledgments.tex
\paragraph{Acknowledgments.}
We thank Shashanka Venkataramanan for his valuable contribution to certain experiments. This work was supported by computational time granted from GRNET in the Greek HPC facility ARIS under projects PR009017, PR011004 and PR012047 and by the HPC resources of GENCI-IDRIS in France under the 2021 grant AD011012884. NTUA thanks NVIDIA for the support with the donation of GPU hardware. This work has been supported by RAMONES and iToBos projects, funded by the EU Horizon 2020 research and innovation programme, under grants 101017808 and 965221, respectively.

%% file: tex/supp.tex
\clearpage

\appendix

\renewcommand{\theequation}{A\arabic{equation}}
\renewcommand{\thetable}{A\arabic{table}}
\renewcommand{\thefigure}{A\arabic{figure}}


\section{More Experiments}
\label{sec:more-experiments}

We provide more benchmarks (\autoref{sec:more-bench}), more ablations (\autoref{sec:more-ablation}), and more visualizations (\autoref{sec:more-vis}).


\subsection{More Benchmarks}
\label{sec:more-bench}

\paragraph{How Does \ours Affect the Patch Features?}

In contrast with the DINO objective that is applied only on the output [CLS] token embeddings, the MIM objective is directly applied to the output features of the patch tokens. \autoref{tab:GAP} shows that using \emph{global average pooling} (GAP) over patch features instead of the [CLS] token embeddings, \ours outperforms baseline iBOT~\cite{ibot} by 9.0\% $k$-NN accuracy.
This indicates that \ours leads to a more challenging MIM objective, which in turn forces the ViT to learn more discriminative patch features.

\begin{table}
\small
\centering
\setlength{\tabcolsep}{3pt}
\caption{$k$-NN top-1 accuracy on ImageNet-1k validation using global average pooling (GAP) over patch features \vs the [CLS] token embeddings. Models are pre-trained on 100\% of ImageNet-1k for 100 epochs.}
\label{tab:GAP}
\begin{tabular}{lcc} \toprule
 & \Th{CLS} & \Th{GAP}  \\ \midrule
iBOT & 71.5 & 49.0  \\
iBOT + \ours & 72.5 & 58.0  \\ \midrule
Gain  & \gp{+1.0} & \gp{+9.0} \\ \bottomrule
\end{tabular}
\end{table}

\begin{table}
\small
\centering
\setlength{\tabcolsep}{3pt}
\caption{Linear probing
top-1 accuracy on a more challenging \emph{masked version} of ImageNet-1k validation set.
Salient parts are gradually masked using the attention maps of the official pre-trained DINO ViT-Base model and setting the corresponding masked pixel values to zero (black).
Models pre-trained on 100\% of ImageNet-1k for 100 epochs.}
\label{tab:MaskingAcc}
\begin{tabular}{lccccccc} \toprule
\Th{Mask Ratio (\%)} & 0 & 10 & 30 & 50 & 70 \\ \midrule
iBOT & 74.4 & 64.8 & 47.6 & 31.4 & 17.0  \\
iBOT + \ours & 75.7 & 66.9 & 50.0 & 34.2 & 20.5 \\ \midrule
Gain  & \gp{+1.3} & \gp{+2.1} & \gp{+2.4} & \gp{+2.8} & \gp{+3.5} \\ \bottomrule
\end{tabular}
\end{table}

\paragraph{Does \ours Lead to Better Exploitation of Non-Salient Parts?}

We examine the performance of the models pre-trained on 100\% of ImageNet-1k on a more challenging ImageNet-1k validation set. In particular, we gradually mask the salient parts using the attention maps of the official pre-trained DINO ViT-Base model and setting the corresponding masked pixel values to zero. Our assumption is that a more robust model should be less sensitive when salient parts of an object are missing. In \autoref{tab:MaskingAcc}, we observe that as more parts of the images are hidden, a larger gain occurs by using \ours with iBOT. This indicates that \ours leads to less sensitive models that exploit better the non-salient parts or even background context.

\begin{table}[t]
\small
\centering
\setlength{\tabcolsep}{2pt}
\caption{\emph{Scene classification} measuring accuracy (\%) using linear probing on Places205~\cite{NIPS2014_5349}. Models pre-trained on 100\% ImageNet-1k training set for 100 epochs.}

\label{tab:100percMoreDownStream}
\begin{tabular}{lcc} \toprule
 & iBOT & iBOT+\ours \\ \midrule
\mc{1}{Places205} & \mc{1}{55.9} &  \mc{1}{\bf{56.7}} \\ \bottomrule
\end{tabular}
\end{table}

\paragraph{Downstream Tasks using Linear Probing.}

We experiment on \emph{scene classification} on Places205~\cite{NIPS2014_5349}, measuring classification accuracy, using linear probing evaluation on models pre-trained on 100\% of ImageNet-1k for 100 epochs. In \autoref{tab:100percMoreDownStream}, we observe that \ours improves scores by 0.8\% accuracy.

\begin{table}[t]
\small
\centering
\setlength{\tabcolsep}{3pt}
\caption{Top-1 accuracy on ImageNet validation set. (a) $k$-NN and linear probing using the full ImageNet training set; (b) k-NN using only $\nu \in \{1,5,10,20\}$ examples per class. Pre-training on 100\% ImageNet-1k for 300 epochs.}
\label{tab:sup-100percImageNet300epANDFewShot}
\begin{tabular}{lcc|cccc} \toprule
\mr{2}{\Th{Method}} & \mc{2}{(a) \Th{Full}} & \mc{4}{(b) \Th{Few Examples}} \\  \cmidrule{2-7}
&  $k$-NN   & \Th{Linear}& $\nu=1$     & 5         & 10        & 20 \\
\midrule
SimCLR \cite{simclr} &  -  & 69.0 & & & &\\
BYOL \cite{byol} &  66.6  & 71.4 & & & &\\
MoBY \cite{moby} &  -  & 72.8 & & & &\\
DINO \cite{dino} &  72.8  & 76.1 & & & &\\
MST \cite{mst} &  \bf{75.0}  & 76.9  & & & &\\
iBOT \cite{ibot} &  74.6  & 77.4      & 38.9     &    54.1    & 58.5    & 61.9\\
iBOT+\ours (Ours) &  \bf{75.0} & \bf{77.5} & \bf{40.4} & \bf{55.5} & \bf{59.9} & \bf{63.1} \\ \bottomrule
\end{tabular}
\end{table}

\paragraph{Training for More Epochs.}

We train iBOT with \ours on 100\% of ImageNet-1k for 300 epochs. AttMask not only accelerates the learning process and has better performance on data-limited regimes as explained in the main paper, but as we see in \autoref{tab:sup-100percImageNet300epANDFewShot}(a), even when trained for many epochs and with many data, it still brings an improvement of 0.4\% $k$-NN and 0.1\% linear probing over baseline iBOT~\cite{ibot}.
Also, \ours outperforms all other state-of-the-art frameworks on linear probing evaluation on ImageNet-1k validation set. We highlight that MST~\cite{mst} employs an additional CNN decoder, while \ours achieves improved linear probing performance with fewer learnable parameters.

We argue that the higher improvement of \ours $k$-NN compared with linear probing indicates higher quality of learned embeddings, since linear probing amounts to supervised classification on higher-dimensional embeddings\interfootnotelinepenalty=10000\footnote{We remind that, following the evaluation setups of DINO~\cite{dino} for ViT-S, for linear probing we use the concatenated features from the last 4 layers of ViT while for $k$-NN the feature from only the last layer. So, linear probing uses 4 times higher-dimensional features} 
and on the same dataset that was used for self-supervised pre-training. To validate this, we experiment with a more challenging variant of $k$-NN where only $\nu \in \{1,5,10,20\}$ examples per class of the training set are used. \autoref{tab:sup-100percImageNet300epANDFewShot}(b) shows that using \ours for self-supervised pre-training and then using only simple $k$-NN classifier with only one example per class, achieves an accuracy improvement of 1.5\% compared with the default iBOT.
This highlights the superiority of \ours in low-shot learning regimes, which are of great practical interest.

\begin{table}[t]
\small
\centering
\setlength{\tabcolsep}{3pt}
\caption{Top-1 accuracy on ImageNet validation set after supervised fine-tuning for 100 epochs on ImageNet-1k training set. Models pre-trained on 100\% ImageNet-1k training set for 300 epochs.}

\label{tab:SupervisedFinetuning}
\begin{tabular}{l c c} \toprule
 & iBOT & iBOT+\ours \\ \midrule
\mc{1}{Fine-tuning on ImageNet} & \mc{1}{81.1} &  \mc{1}{\bf{81.3}} \\ \bottomrule
\end{tabular}
\end{table}

\paragraph{Full fine-tuning on ImageNet-1k.} 

For iBOT and iBOT+\ours pre-trained on ImageNet-1k for 300 epochs, we also experiment with further supervised fine-tuning on ImageNet-1k, training for 100 epochs. We report results in \autoref{tab:SupervisedFinetuning}. \ours improves iBOT by 0.2\% (81.1\% $\rightarrow$ 81.3\%), providing a better network initialization for supervised finetuning.


\subsection{More Ablations}
\label{sec:more-ablation}

\begin{table}
\small
\centering
\setlength{\tabcolsep}{3pt}
\caption{$k$-NN top-1 accuracy on ImageNet-1k validation \vs MIM Loss Weight $\lambda$, while the weight of DINO loss is fixed to 1. Pre-training on 20\% of ImageNet-1k for 100 epochs.}
\label{tab:LossAttMask}
\begin{tabular}{lcccc} \toprule
\Th{MIM Loss Weight $\lambda$} & 0.0 & 0.5 & 1.0 & 2.0 \\ \midrule
iBOT & 43.4 & 46.5 & 46.7 & 41.9 \\
iBOT+\ours & 43.5 & 47.3 &  \bf{49.7} & 48.3 \\ \midrule
Gain  & \gp{+0.1} &  \gp{+0.8} & \gp{+3.0} & \gp{+6.4} \\ \bottomrule
\end{tabular}
\end{table}

\paragraph{MIM Loss Weight.}

The overall loss of iBOT~\cite{ibot} is a weighted sum of $L_{\textsc{mim}}$~\eq{mim}, with weight $\lambda$, and $L_\textsc{g}$~\eq{g} $+$ $L_\textsc{lc}$~\eq{lc} (DINO), with weight 1. Table \ref{tab:LossAttMask} shows that \ours is superior to the default block-wise random masking of iBOT in all cases, while the default $\lambda = 1$ works best for both and yields the greatest gain of $3\%$ $k$-NN accuracy for \ours. In particular, increasing the weight of the MIM loss leads to a larger gain in $k$-NN accuracy. This shows that \ours boosts the MIM task.



\begin{table}
\small
\centering
\setlength{\tabcolsep}{3pt}
\caption{\emph{\athigh \vs random masking strategies}: $k$-NN top-1 accuracy on ImageNet-1k validation for iBOT pre-training on 20\% of ImageNet-1k for different mask ratio $r$. $\dagger$: default iBOT masking strategy from BEiT~\cite{beit}.}
\label{tab:RandomMaskThreshold}
\begin{tabular}{lccccc} \toprule
\Th{Mask Ratio $r$ (\%)} & 10-30 & 10-50 & 10-70 & 30 \\ \midrule
Random Block-Wise & 46.5 & 46.7${}^{\dagger}$ & 47.1 & 46.9   \\
Random & 47.6 & 47.8 & 47.8 & 48.2  \\
\athigh & 49.5  & \bf{49.7} & 48.5  & 49.1  \\ 
\bottomrule
\end{tabular}
\end{table}


\paragraph{Masking strategy and mask ratio.}

We ablate both the masking strategy (random block-wise, random or \athigh) and the mask ratio $r$  in Table \ref{tab:RandomMaskThreshold}. \athigh with 10-50 mask ratio gives the best results.


\subsection{More Visualizations}
\label{sec:more-vis}

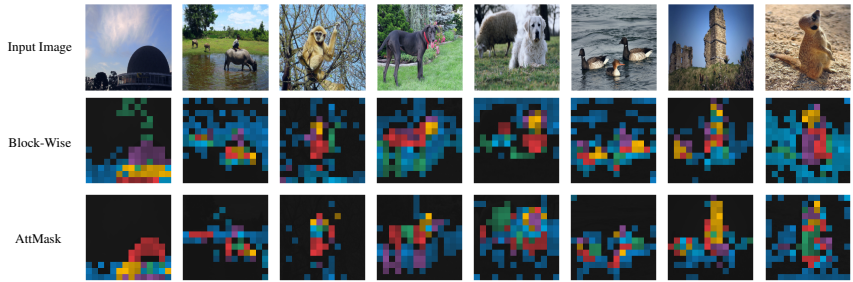
\begin{figure}[t]
\input{tex/fig_attention_maps}
\caption{Multi-head attention maps from the last layer, training iBOT with the default block-wise strategy from BEiT~\cite{beit} and with our \ours.
From the attention matrix~\eq{att-mat} of each head, we extract the attention map of the \cls token and display in different color per head the patch tokens that are included in the top 60\% of the attention mass.}
\label{fig:attmaps}
\end{figure}

\paragraph{Visualization of Attention Maps.}

In \autoref{fig:attmaps}, we utilized the pre-trained models on 20\% of ImageNet and observe that, when training iBOT with the default block-wise random masking strategy, there is at least one head (in blue) that attends the background to a great extent. By contrast, with our \ours, all heads mostly attend salient objects or object parts. It appears that by focusing on reconstructing highly-attended masked tokens, the network learns to focus more on foreground objects.

\input{tex/sup_fig_various_strategies}

\paragraph{Visualization of Masking Examples.}

We illustrate the effect of mask ratio $r$ (\%) to various masking strategies in \autoref{fig:MoreStrategies2} and \autoref{fig:MoreStrategies3}. While random Block-Wise and Random masking fail to consistently mask informative parts of an image, \ours-High and \ours-Hint make use of attention to hide salient and all but very salient parts respectively. This gives rise to a more challenging MIM task.


\section{Experimental Setup}
\label{sec:more-setup}

We provide more details on the experimental setup, including multi-crop, training details and evaluation details.

\paragraph{Multi-Crop.}

Following~\cite{dino,ibot}, we apply the \emph{multi-crop} strategy~\cite{swav} to generate a set of $m$ low-resolution \emph{local crops}, which cover only small parts of the image, tokenized as $Z^c_1, \dots, Z^c_{m}$. Similar to $L_{\textsc{g}}$~\eq{g}, the loss is applied globally on the \cls tokens, in particular between the student output for a local crop $Z^c_j$ and the teacher output for a global view $Z^v$, both of which are non-masked:
\begin{align}
	L_{\textsc{lc}} = - \sum_{v \in V} \sum_{j=1}^m
		f_{\theta'}(Z^v)^\cls \log(f_{\theta}(Z^c_j)^\cls).
\label{eq:lc}
\end{align}
The overall loss is a weighted sum of $L_{\textsc{mim}}$~\eq{mim}, $L_\textsc{g}$~\eq{g} and $L_\textsc{lc}$~\eq{lc}.


\paragraph{Training Details.}

For our \emph{analysis} and \emph{ablation} (\autoref{sec:analysis}, \autoref{sec:ablation} and \autoref{sec:more-ablation}), we pre-train models on 20\% of ImageNet-1k for 100 epochs. For both iBOT and DINO we use AdamW~\cite{adamw} as optimizer. Unless otherwise stated, we use the ViT-S/16 architecture and a batch size of 240. We warm-up learning rate $\eta$ for 10 epochs following the linear scaling rule $\eta = 5 \times 10^{-4} \times \texttt{bs} / 256$ where $\texttt{bs}$ is the batch size and then decay using a cosine schedule. We also use a cosine schedule from $0.04$ to $0.4$ for weight decay. We set teacher momentum to $0.99$ and student temperature to $0.1$. We use a linear warm-up for teacher temperature from $0.04$ to $0.07$ for the first 30 epochs following DINO.

All methods in \autoref{sec:analysis}, \autoref{sec:ablation} and \autoref{sec:more-ablation} use the multi-crop scheme with two $224^2$ global crops and six $96^2$ local crops that approximately scale the training time by a factor of $\gamma = 2 + 6 \times (96/224)^2 = 3.10$. We use color jittering, Gaussian blur and solarization as data augmentations. Local crops scales are sampled from $(0.05, s)$ and global crop scales from $(s, 1)$. We set $s$ to $0.4$ for DINO and $0.25$ for iBOT. We set the dimensionality of the head output to $65536$ for DINO, while for iBOT, we use a shared projection head for [CLS] and patch tokens, of dimensionality $8192$. We do not perform weight normalization on the last layer of the MLP heads.

For our \emph{benchmark} (\autoref{sec:bench} and \autoref{sec:more-bench}), we pre-train models on 100\% of ImageNet-1k for 100 and 300 epochs. For the 100-epoch experiments, the setup is the same as on 20\% of ImageNet-1k except for increasing the teacher momentum to $0.996$ and the number of local crops to ten. 
The scaling factor of the training time in this case is $\gamma = 2 + 10 \times (96/224)^2 = 3.84$.
For the 300 epochs experiments, we increase the batch size to 800 and set $s$ to $0.32$, similar to the iBOT default scale.


\paragraph{Evaluation Details.}

For the ImageNet-1k evaluation, we use $k$-NN and linear probing as in DINO~\cite{dino} and iBOT~\cite{ibot}. We evaluate on ImageNet-1k validation set.
For $k$-NN, we use the [CLS] feature from the last ViT layer and set $k$ to 20.
For linear probing, we train a linear classifier using SGD with a batch size of 1024 for 100 epochs. We set learning rate to $0.003$ and do not apply weight decay. We apply random resized crops and horizontal flips as data augmentations and keep the central crop. Following DINO~\cite{dino} and iBOT~\cite{ibot}, we use the concatenation of the [CLS] features from the last four layers as input to the linear classifier.

For the evaluation of downstream tasks \emph{with finetuning}, we train models on CIFAR10, CIFAR100~\cite{cifar} for 500 epochs and on Oxford Flowers~\cite{flowers} for 1000 epochs. We set learning rate to $7.5 \times 10^{-6}$, weight decay to $0.05$ and use a batch size of 900.

On COCO~\cite{coco}, we evaluate the performance of object detection and instance segmentation downstream tasks. We consider the COCO 2017 set, which contains $118$K training images, $5$k validation and $20$ test-dev. We consider the Cascade Mask R-CNN~\cite{cai2019cascade,he2017mask} as task layer and follow the setup from \cite{liu2021swin}. We use the hyper-parameter configuration from \cite{ibot}: multi-scale training (resizing image with shorter size between $480$ and $800$, with the longer side no larger than $1333$). We use AdamW~\cite{adamw} with initial learning rate $10^{-4}$, the $1\times$ schedule (12 epochs with the learning rate decayed by $10\times$ at epochs $9$ and $11$) and weight decay $0.05$. Unlike~\cite{ibot}, where training is on $8$ GPUs with $4$ images per GPU, we use $2$ images per GPU due to hardware limitations. For a fair and direct comparison, we fine-tune iBOT baseline with the same configuration.

We evaluate on ADE20K~\cite{ade20k} for the semantic segmentation downstream task. It consists of 25k images in 150 classes, with 20k for training, 2k for validation and 3k for testing. We rely on UperNet~\cite{xiao2018unified} as task layer and fine-tune the entire network following the setup from \cite{liu2021swin}: $160$k iterations with $512\times512$ images. We do not perform multi-scale training and testing. We adopt the same hyper-parameters as in \cite{ibot}. We use the AdamW~\cite{adamw} optimizer with an initial learning rate of $7 \times 10^{-4}$ with poly-scheduling, layer decay rate $0.65$ and weight decay $0.05$. We train on $8$ GPUs with $2$ images per GPU.

For the evaluation of downstream tasks \emph{without finetuning}, we follow the protocol of DINO on $\cR$Oxford, $\cR$Paris~\cite{radenovic2018revisiting} and DAVIS 2017~\cite{davis}. On Caltech-UCSD Birds (CUB200)~\cite{cub}, Cars (CARS196)~\cite{cars}, Stanford Online Products (SOP)~\cite{sop} and In-Shop Clothing Retrieval (In-Shop)~\cite{in_shop}, we extract features from test set images and directly apply nearest neighbor search to measure Recall@$k$~\cite{sop}. On Places205~\cite{NIPS2014_5349}, we train a $205$-way linear classifier on pre-cached features, using only horizontal flip as augmentation. Training is with SGD for $50$ epochs using an initial learning rate of $0.01$ that is decreased to $0$ with cosine schedule, a batch-size of $1024$, and no weight decay.

%% file: tex/fig_attention_maps.tex
\tiny
\centering
\setlength{\tabcolsep}{1pt}
\begin{tabular}{cccccccccc}

	\raisebox{15pt}{Input Image}                                                        &&
	\fig[.093]{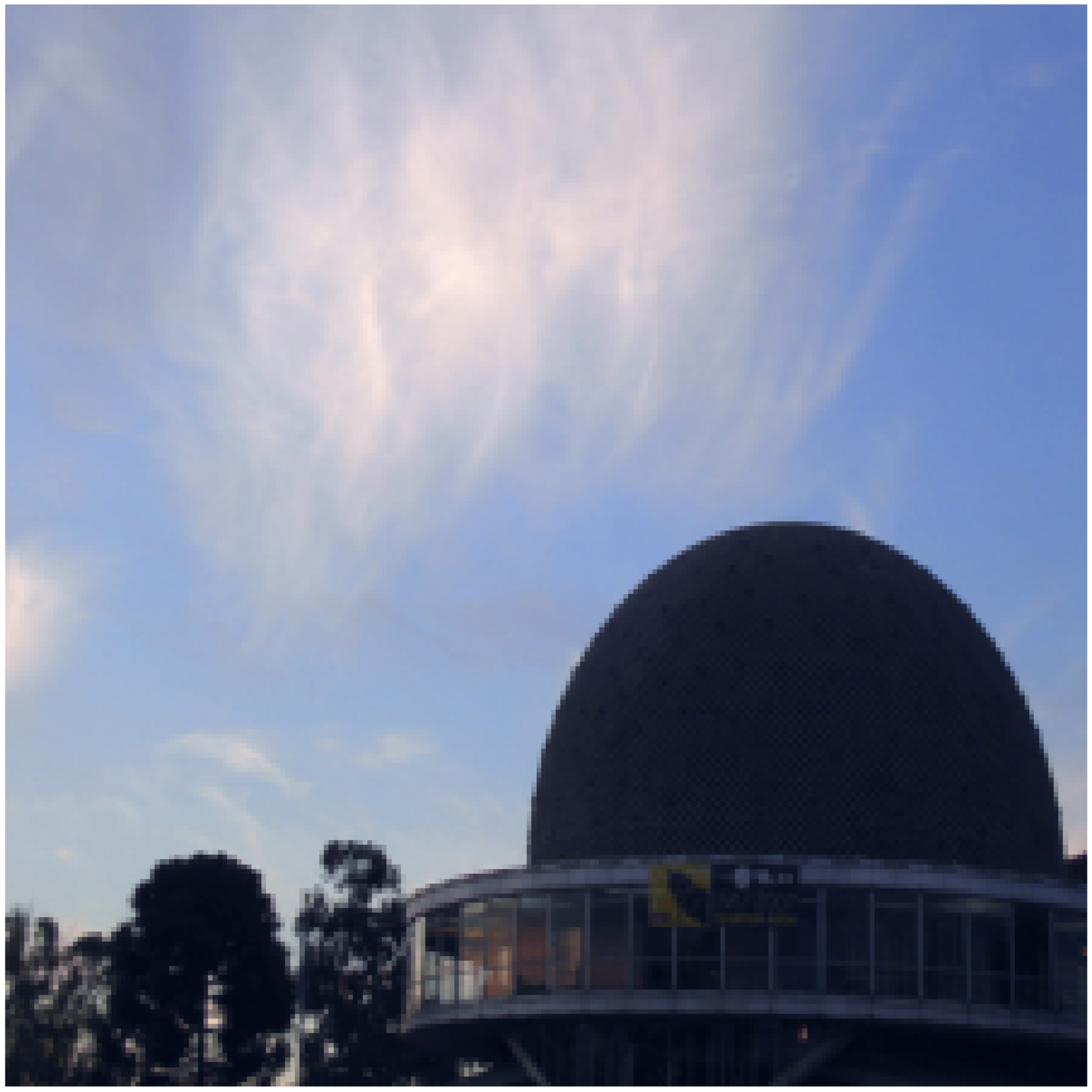}                                                     &
	\fig[.093]{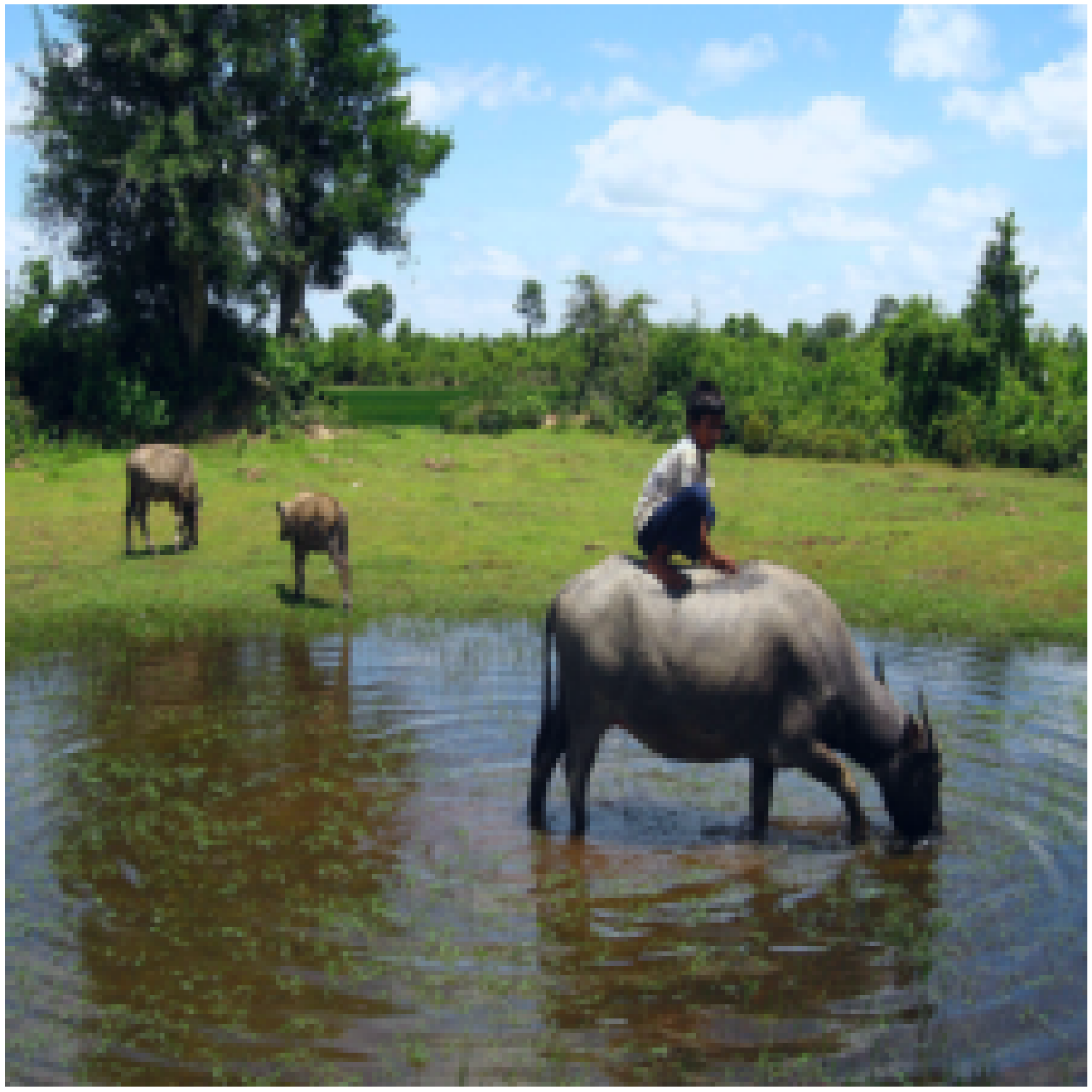}                                                     &
	\fig[.093]{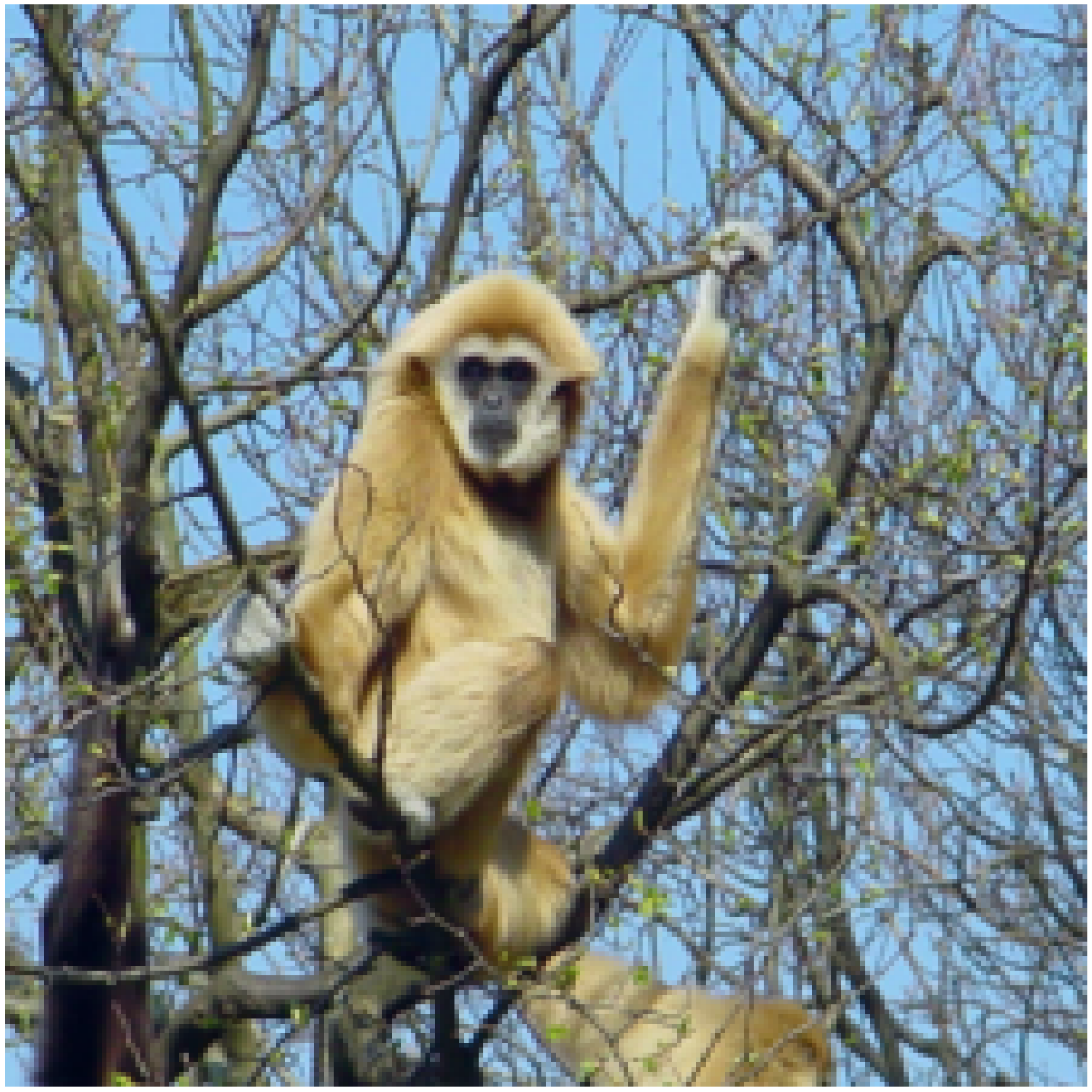}                                                    &
	\fig[.093]{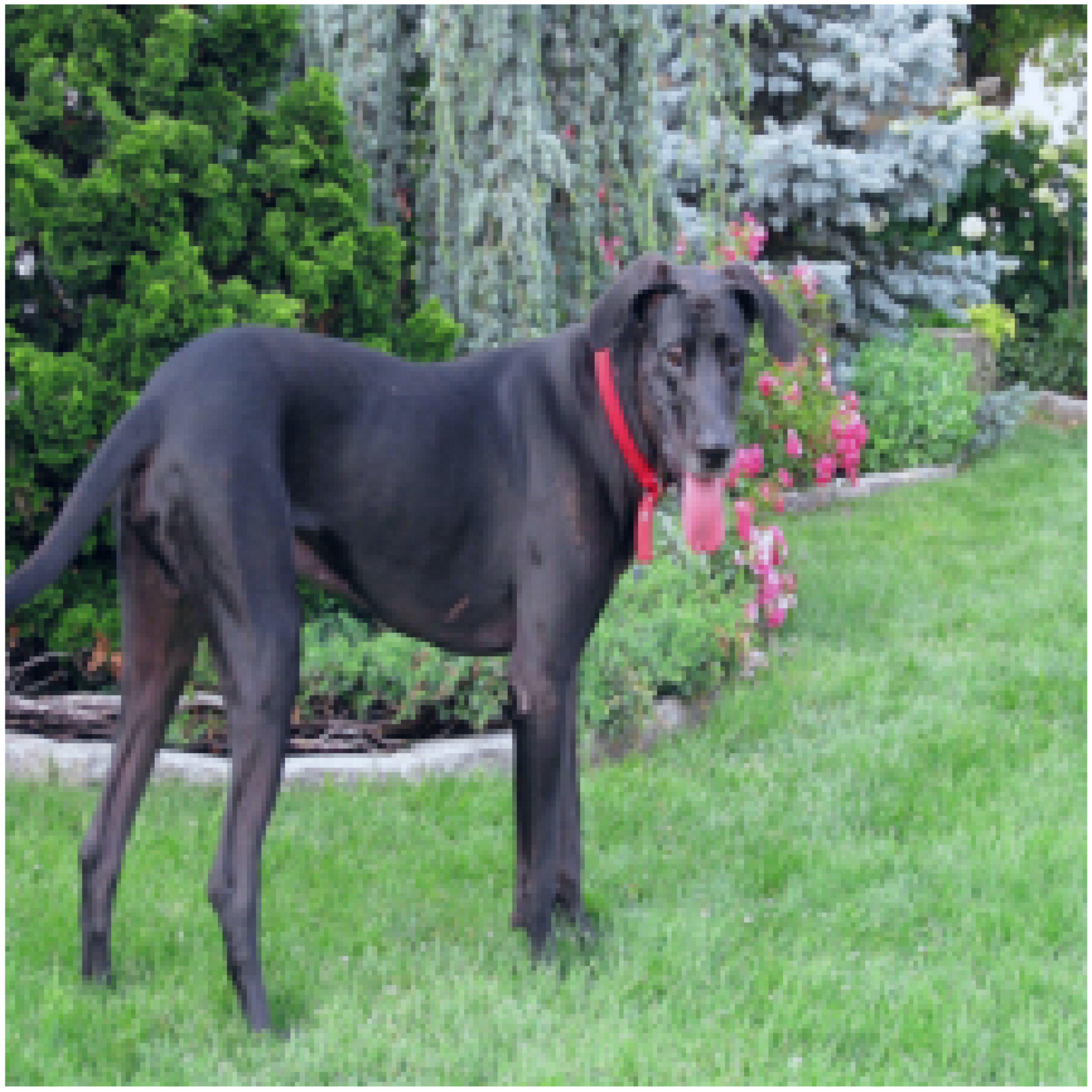}                                                    &
	\fig[.093]{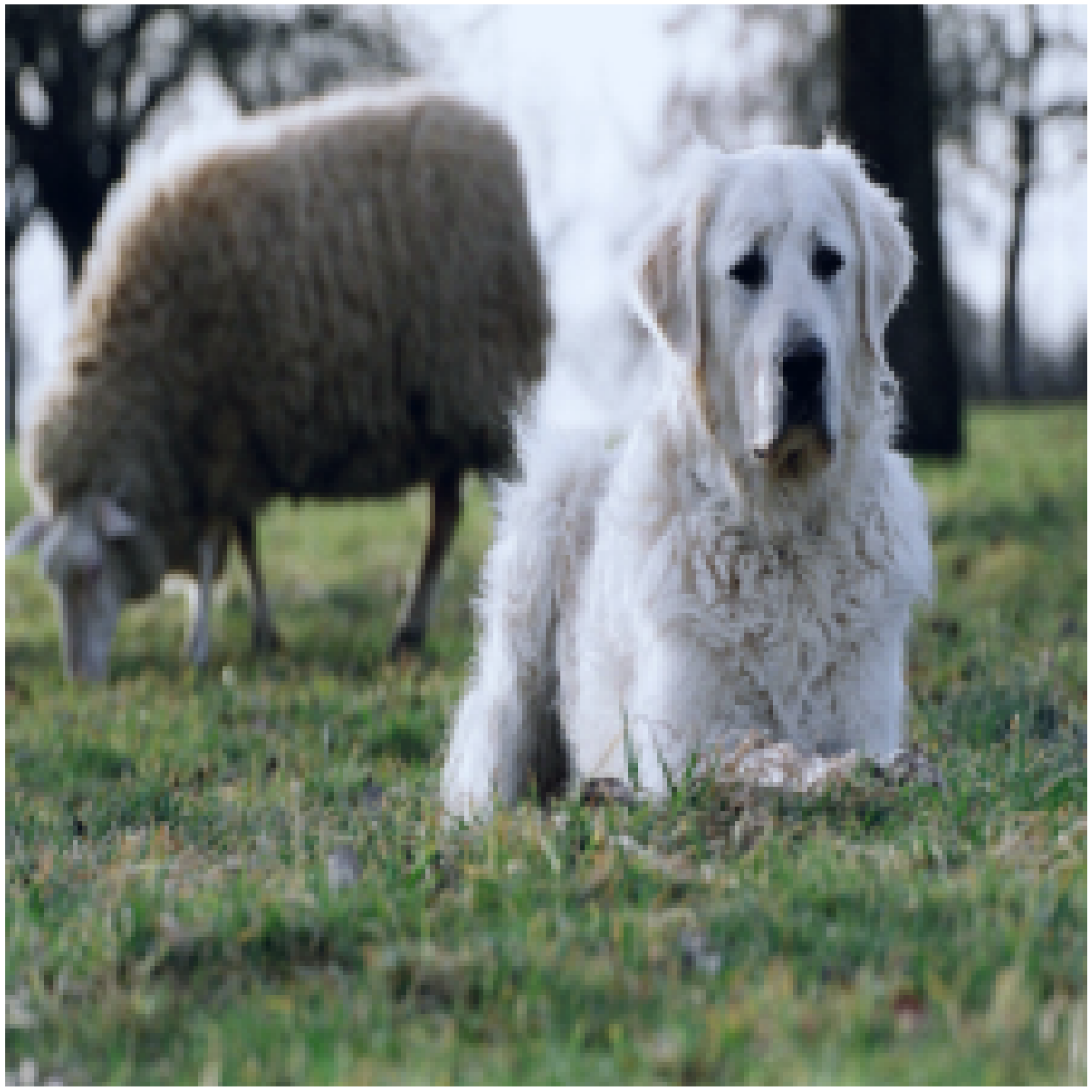}                                                    &
	\fig[.093]{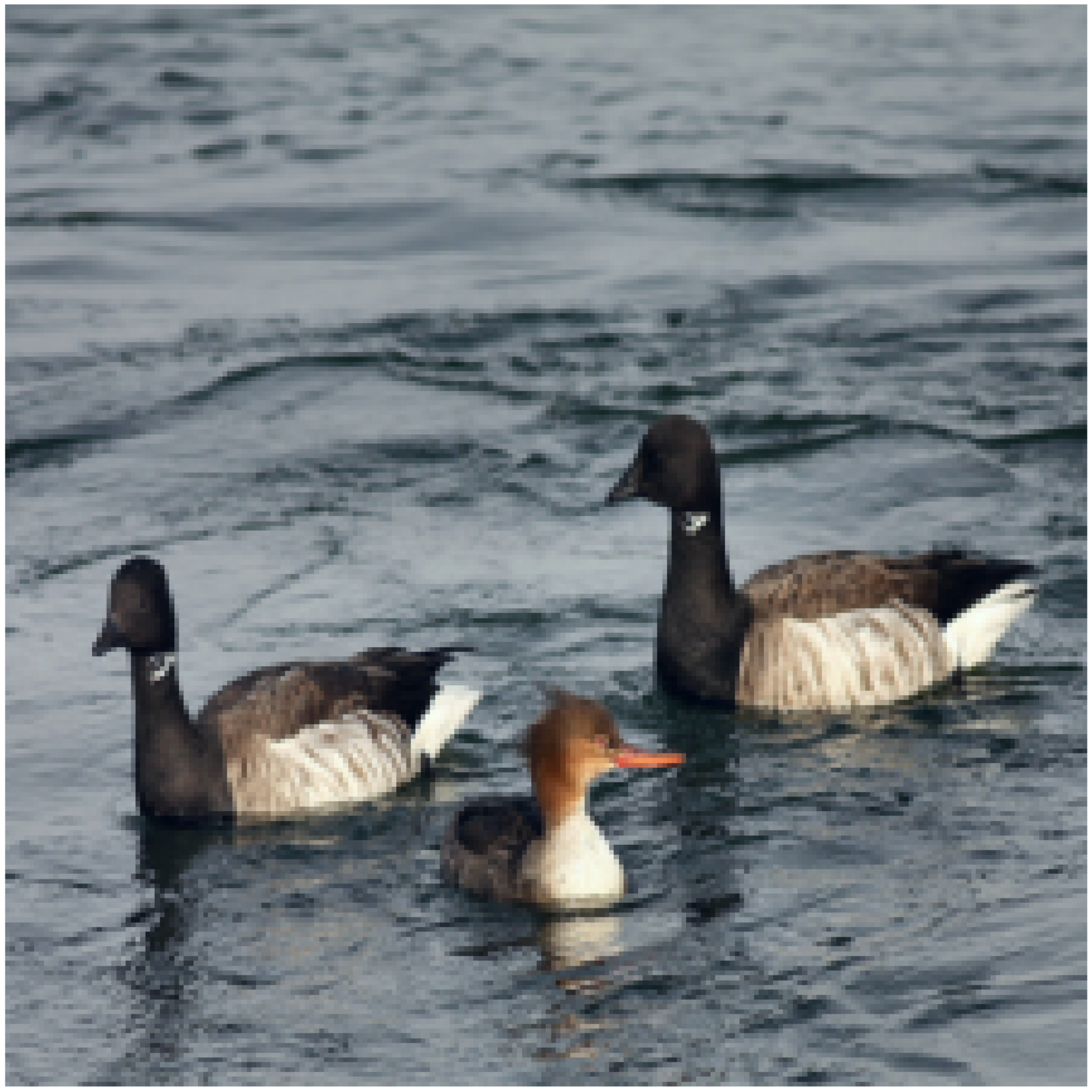}                                                    &
	\fig[.093]{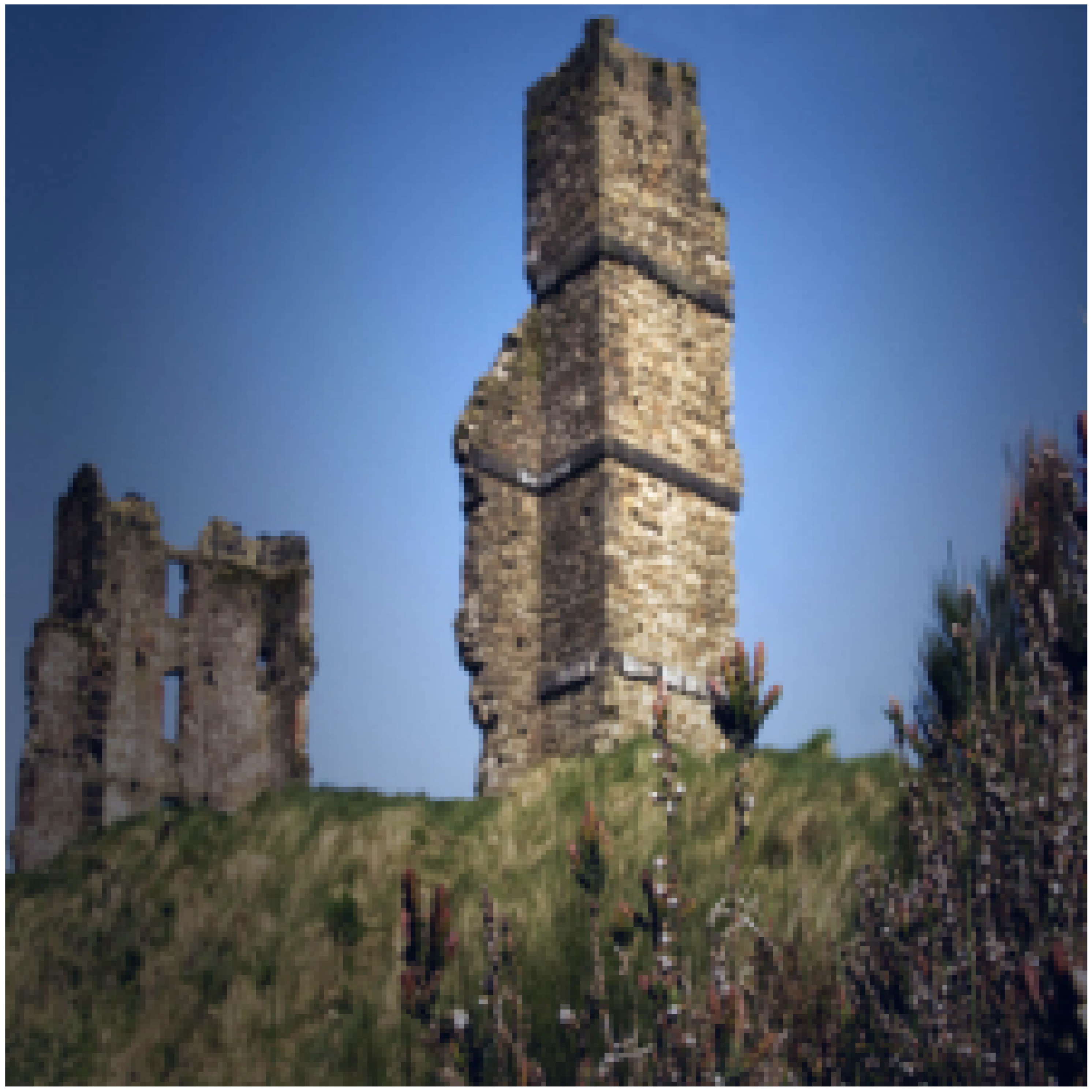}                                                    &
	\fig[.093]{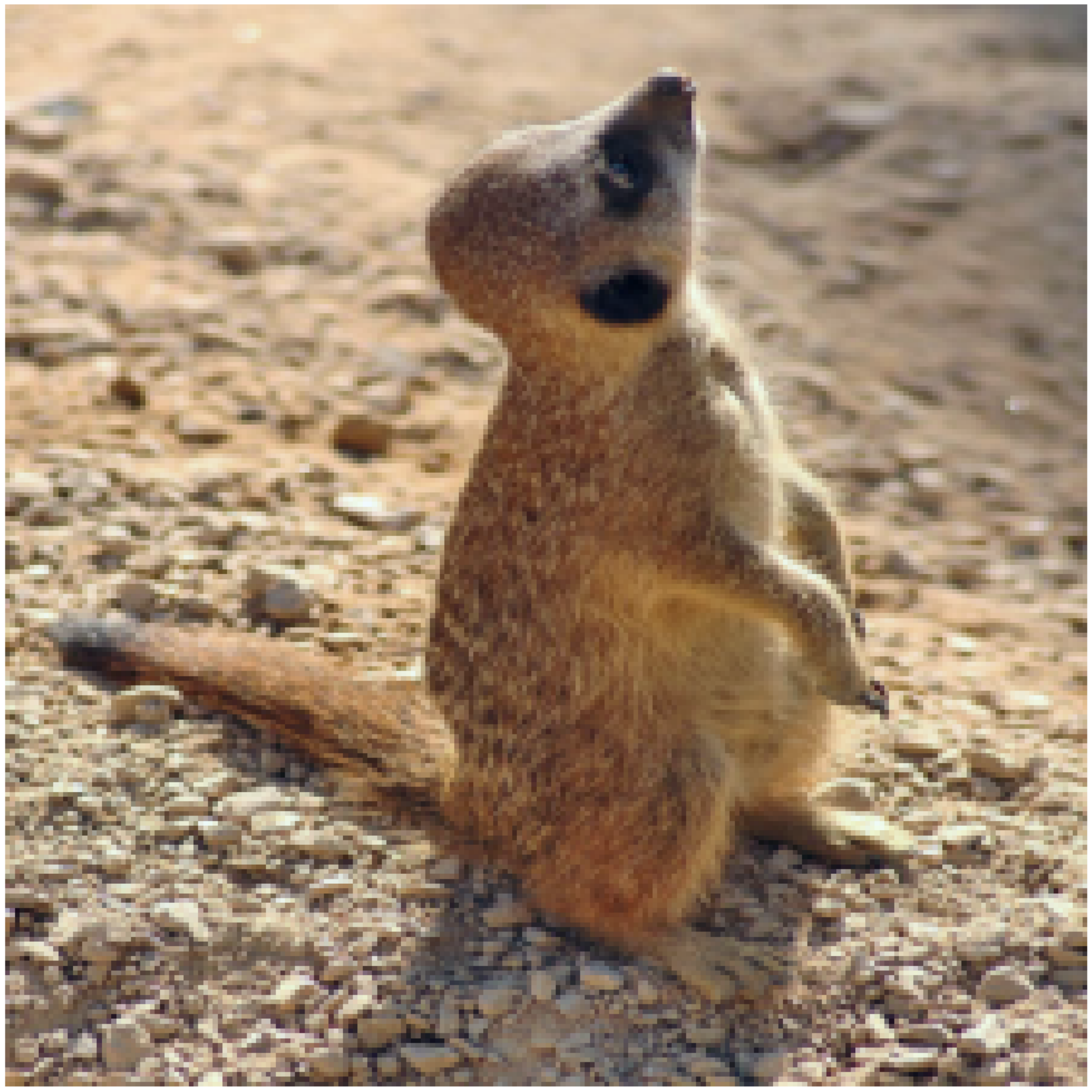}                                                   \\

	\raisebox{15pt}{Block-Wise}                                                         &&
	\fig[.10]{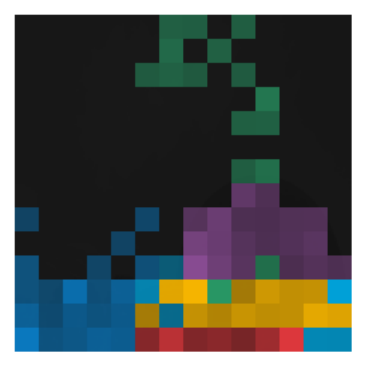}                              &
	\fig[.10]{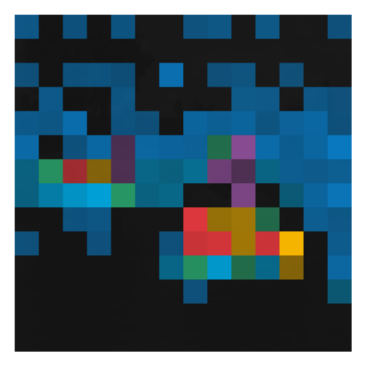}                              &
	\fig[.10]{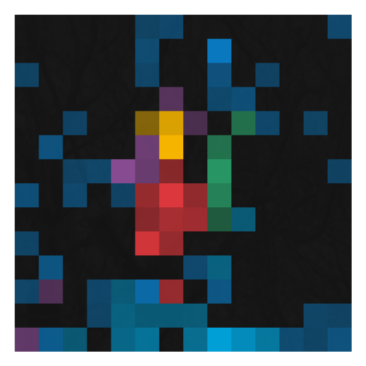}                             &
	\fig[.10]{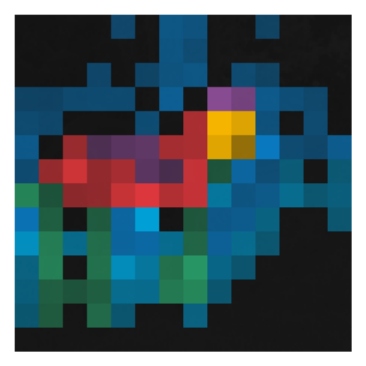}                             &
	\fig[.10]{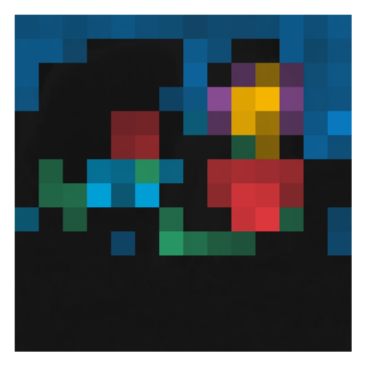}                             &
	\fig[.10]{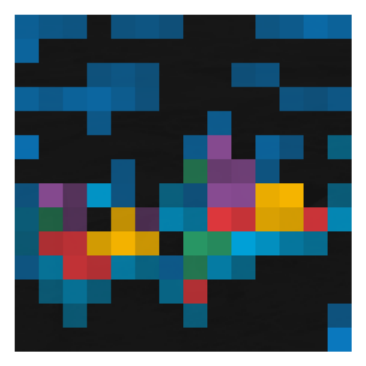}                             &
	\fig[.10]{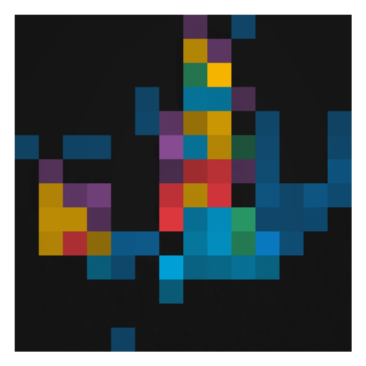}                             &
	\fig[.10]{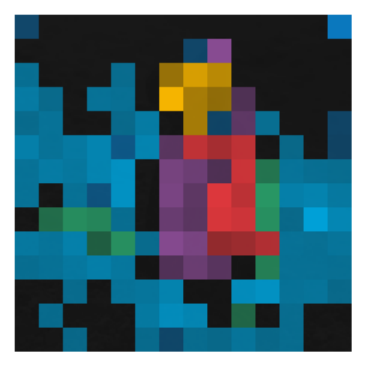}                            \\

	\raisebox{15pt}{\ours}                                                              &&
	\fig[.10]{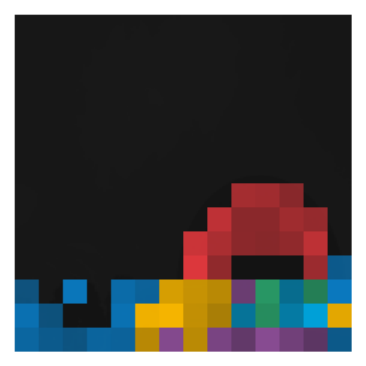}    &
	\fig[.10]{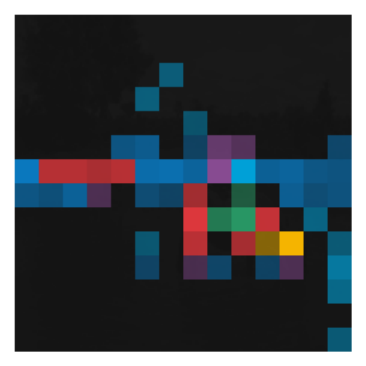}    &
	\fig[.10]{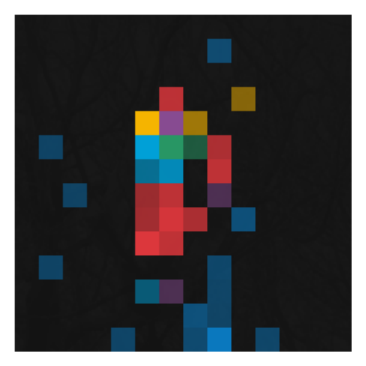}   &
	\fig[.10]{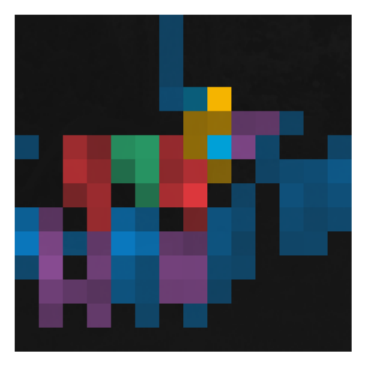}   &
	\fig[.10]{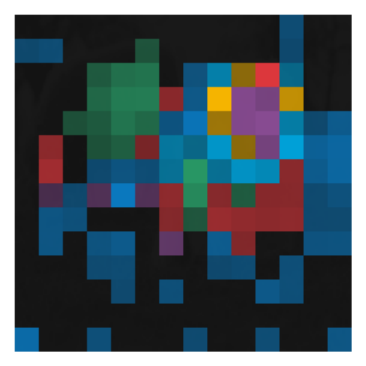}   &
	\fig[.10]{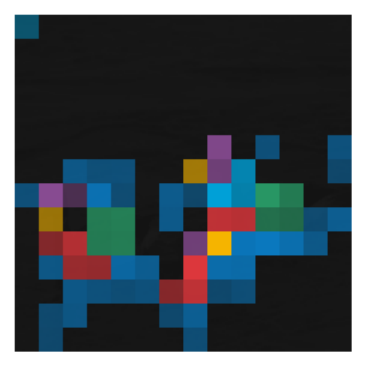}   &
	\fig[.10]{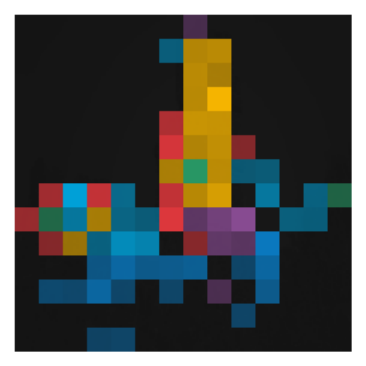}   &
	\fig[.10]{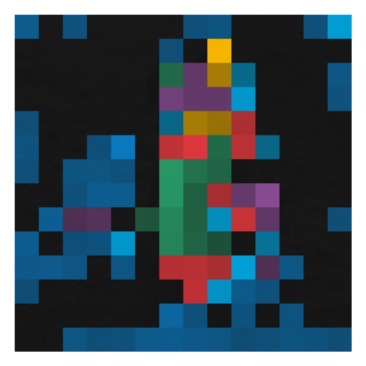}  \\

\end{tabular}

%% file: tex/sup_fig_various_strategies.tex
\begin{figure}[t]
\centering
\tiny
\centering
\setlength{\tabcolsep}{1.5pt}
\begin{tabular}{ccccp{0.15cm}|p{0.15cm}ccc}
    &
    10\%               &
	30\%               &
	50\%               &

    & &

    10\%               &
	30\%               &
	50\%               \\

	\raisebox{15pt}{Block-Wise}  &
	\fig[.120]{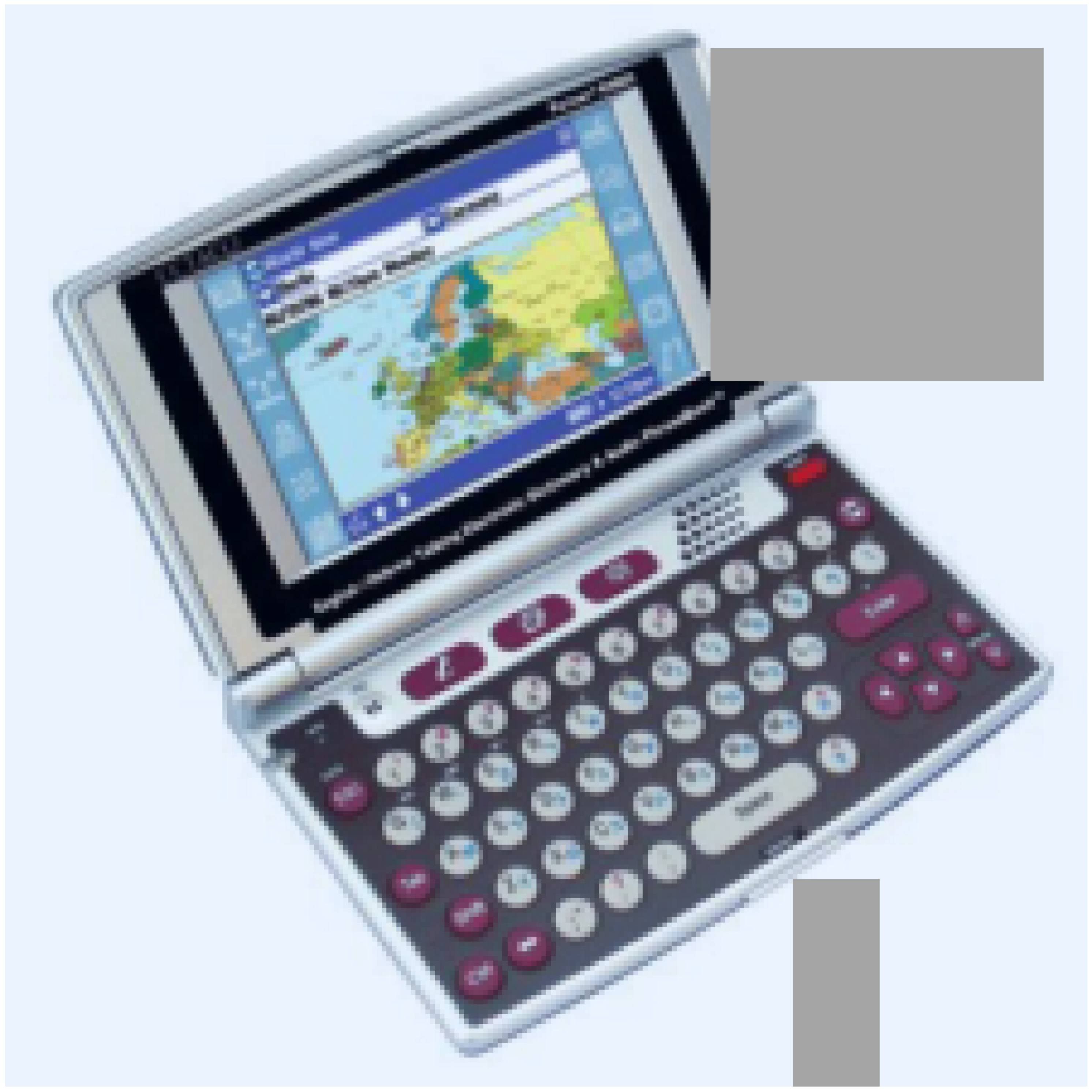}               &
	\fig[.120]{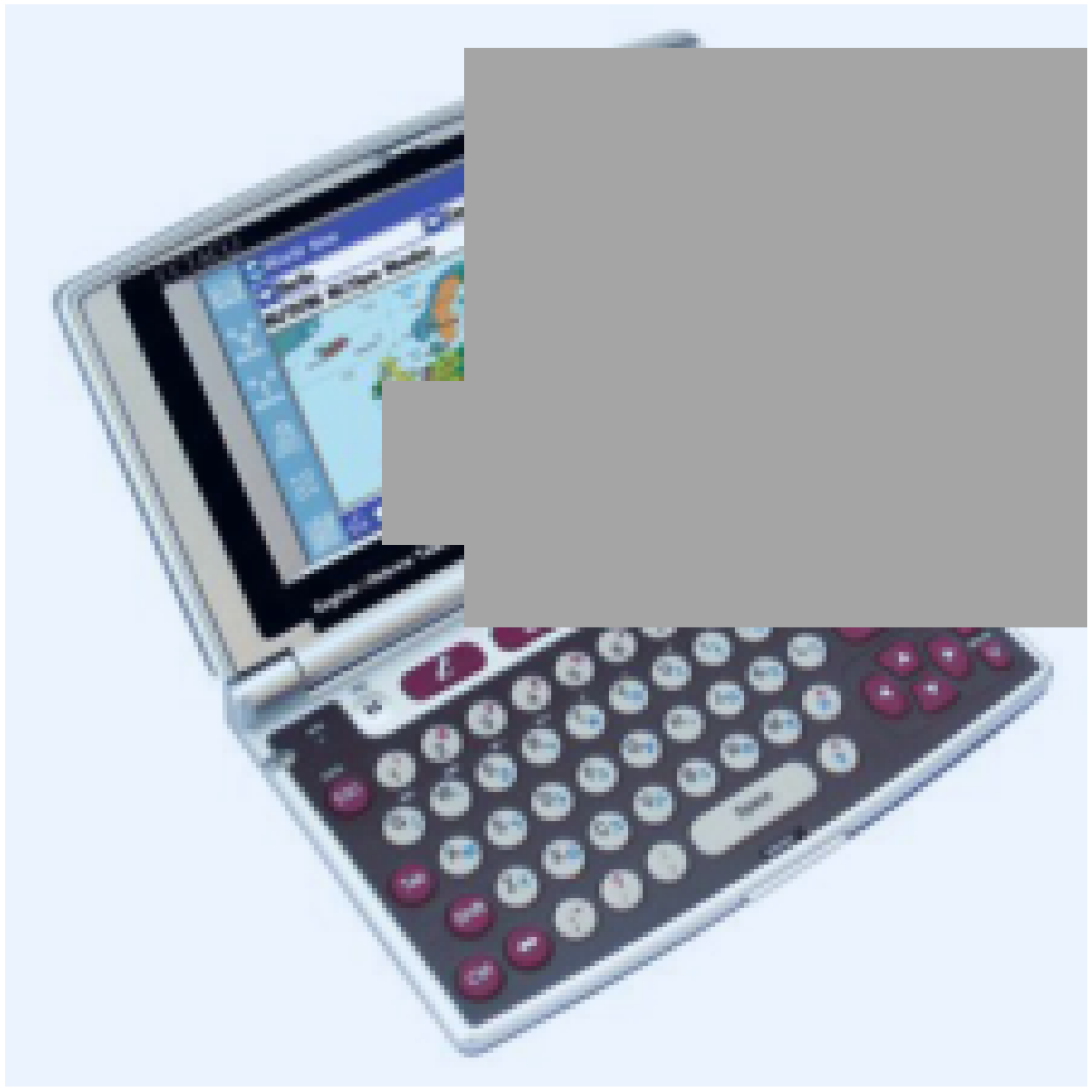}               &
	\fig[.120]{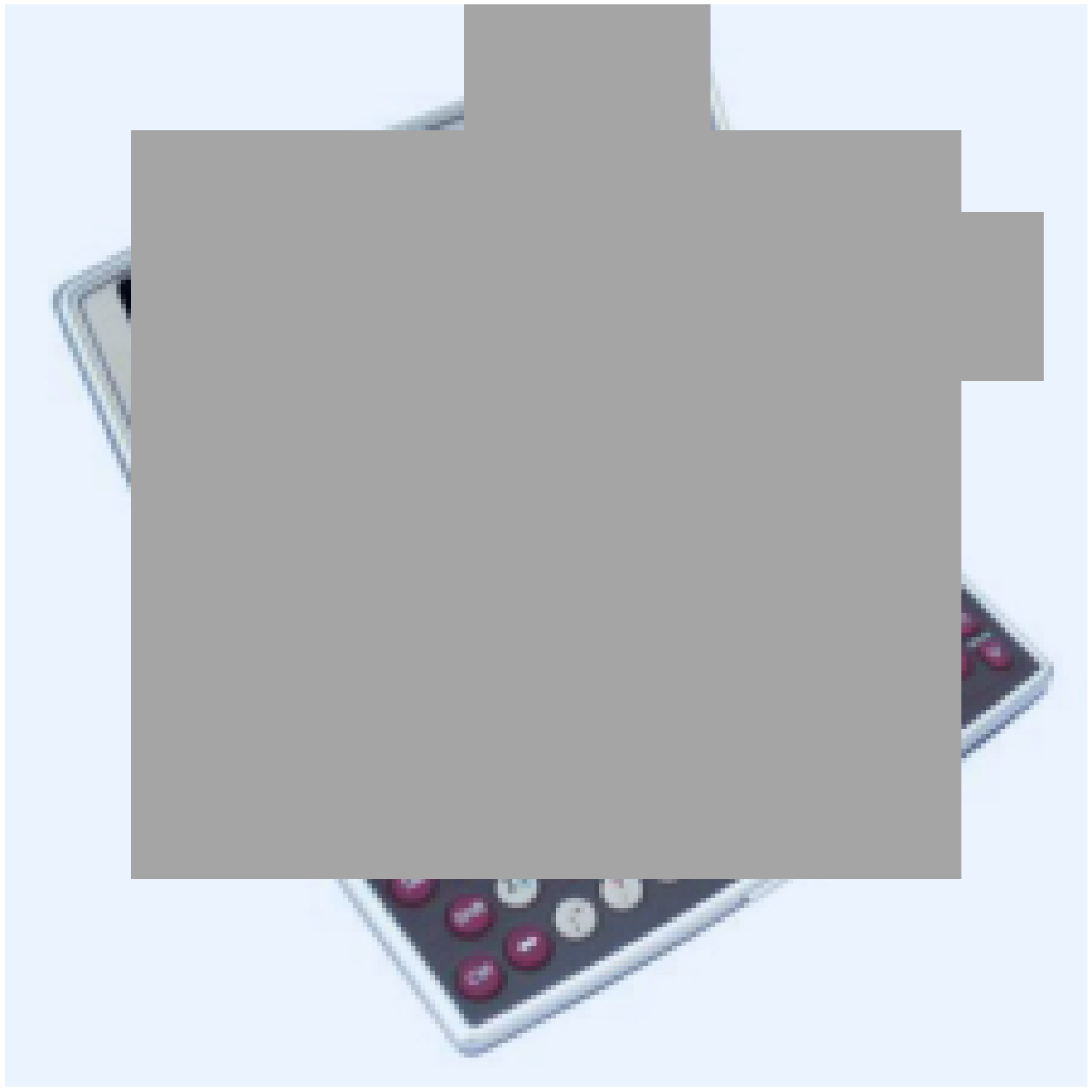}               &

    & &

	\fig[.120]{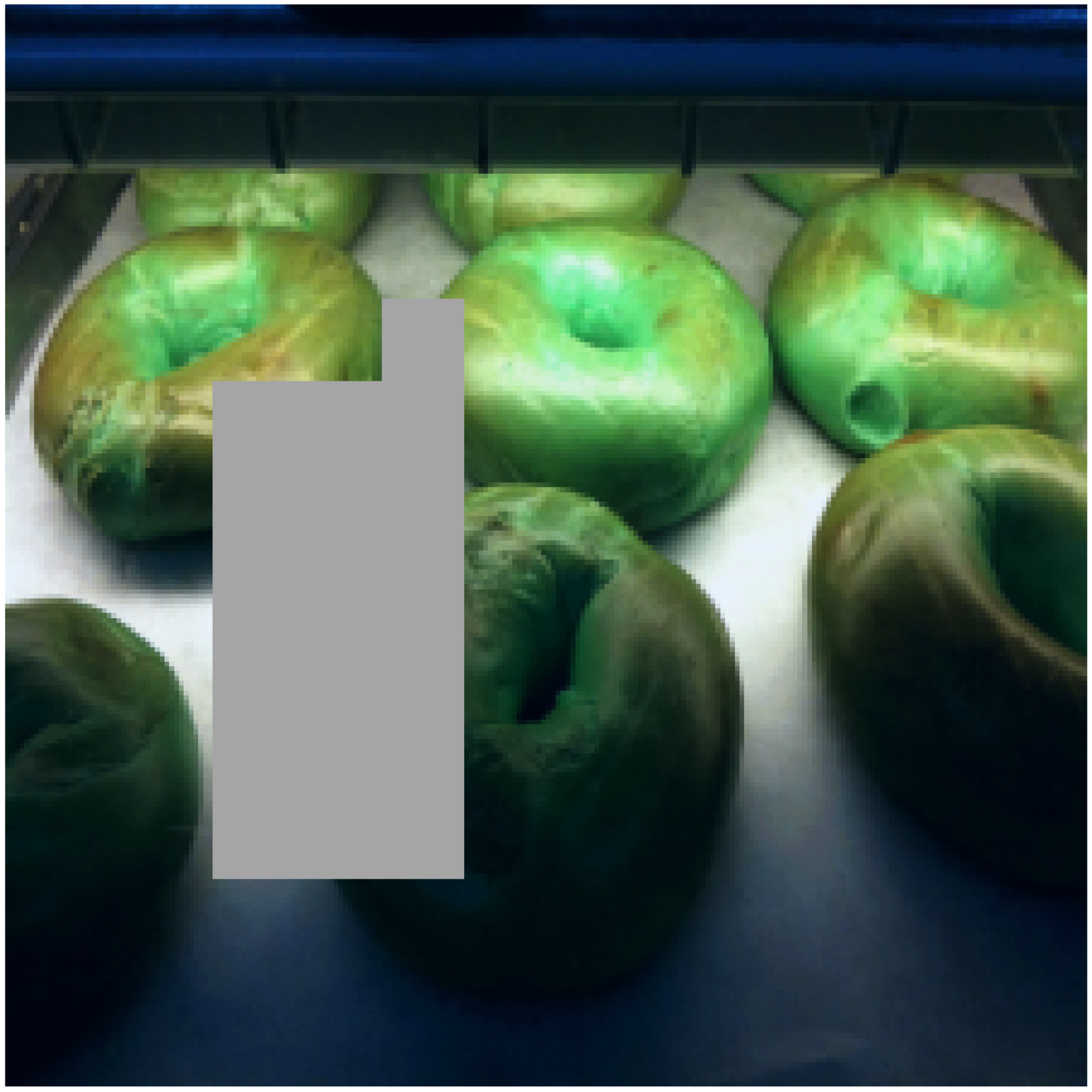}               &
	\fig[.120]{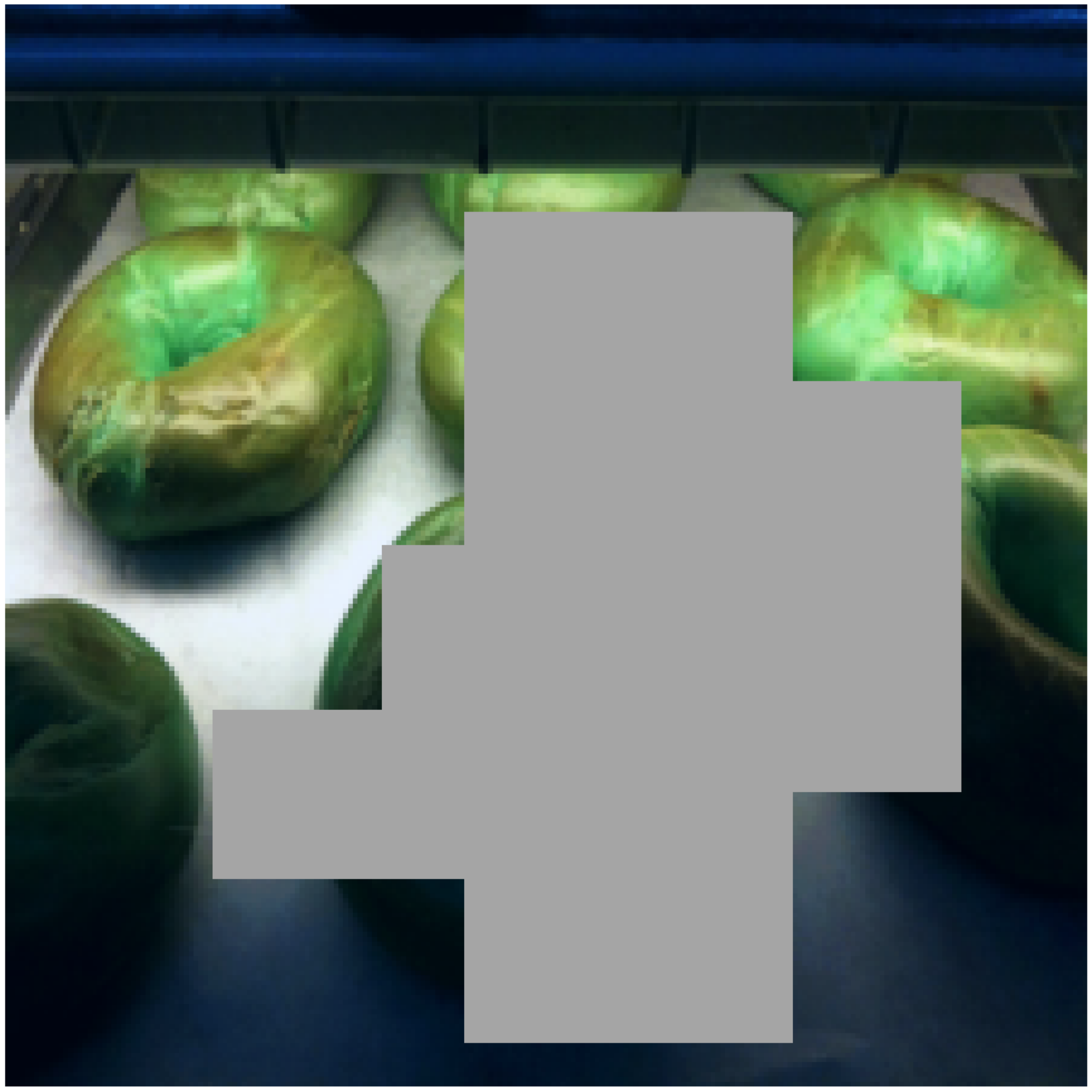}               &
	\fig[.120]{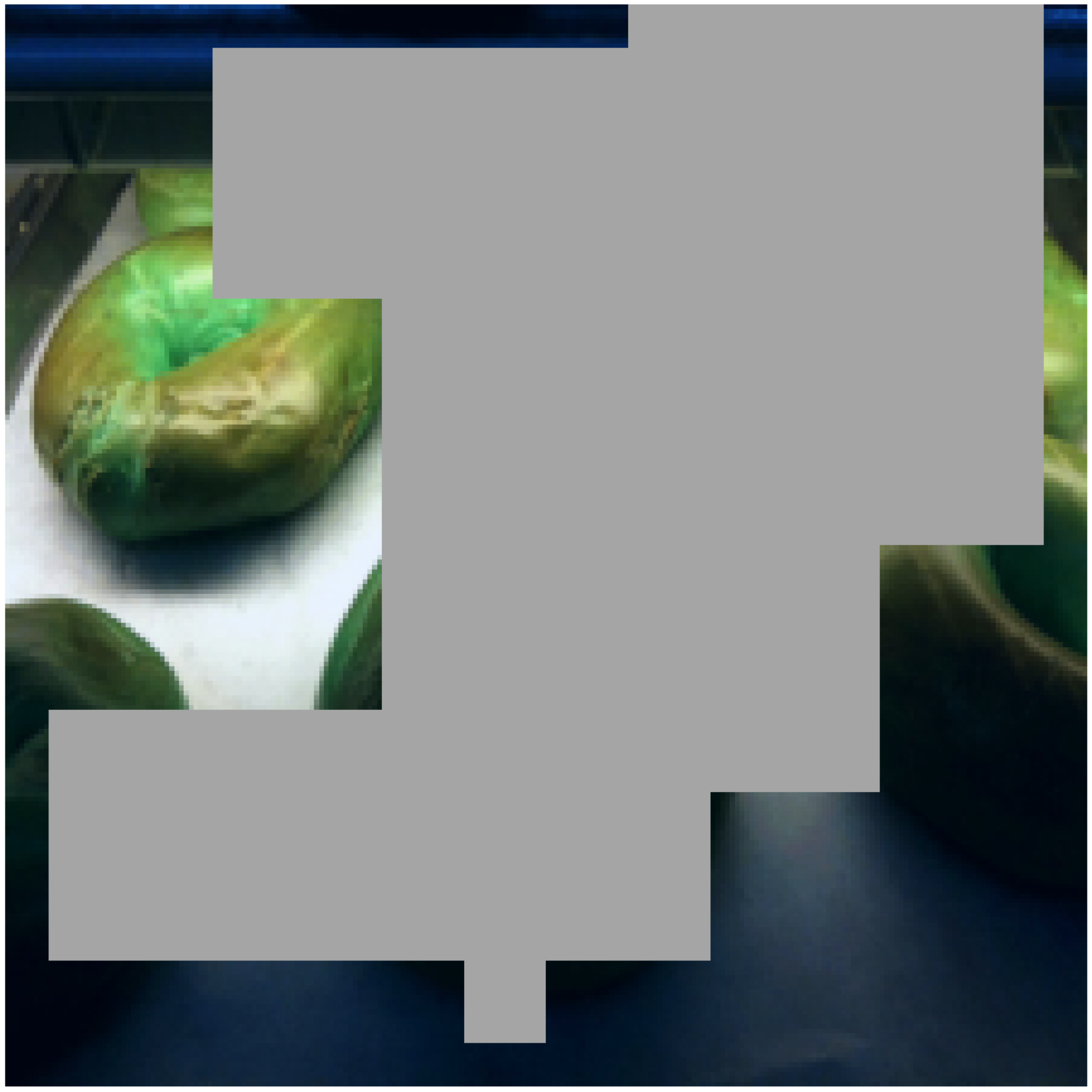}               \\

	\raisebox{15pt}{Random} &
	\fig[.120]{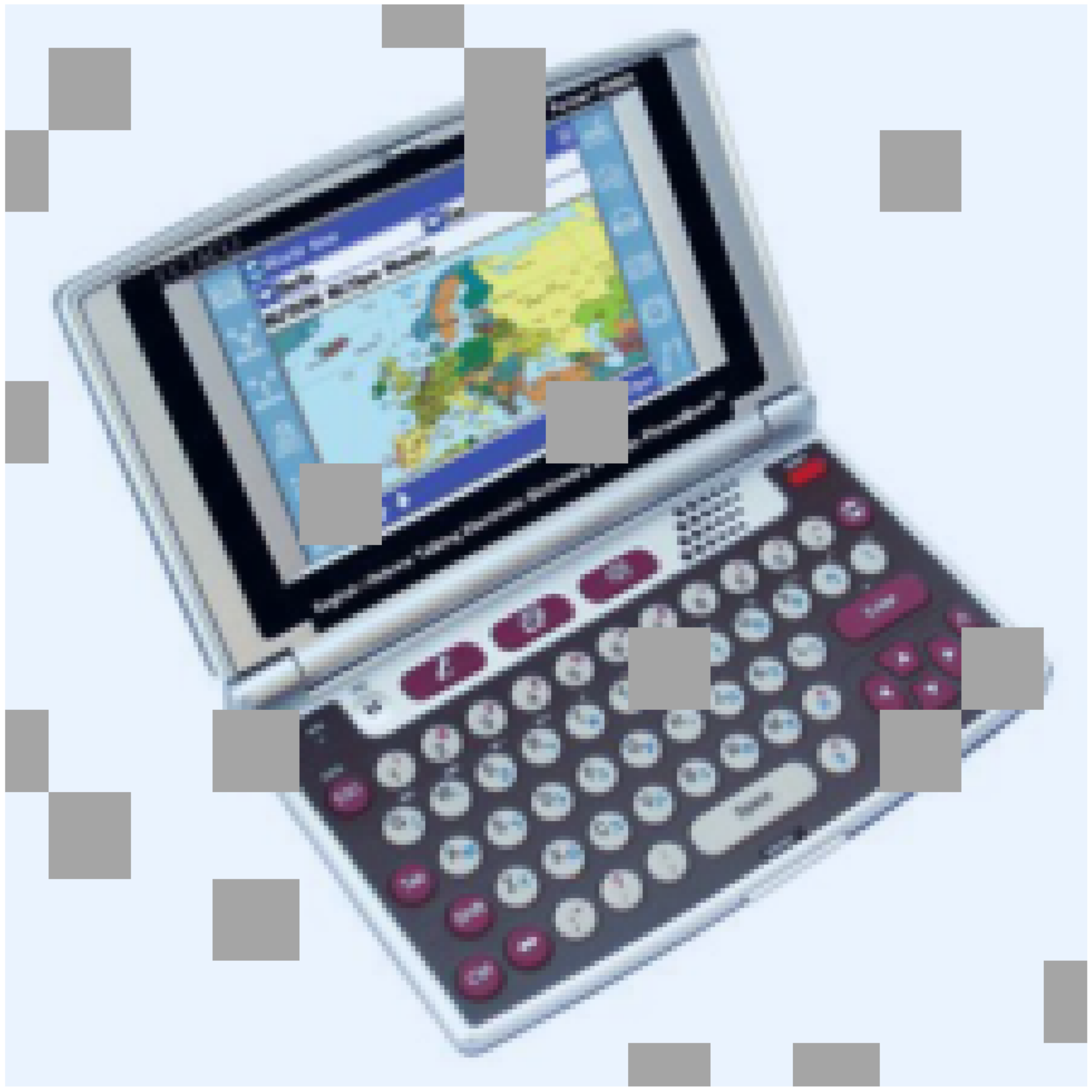}                          &
	\fig[.120]{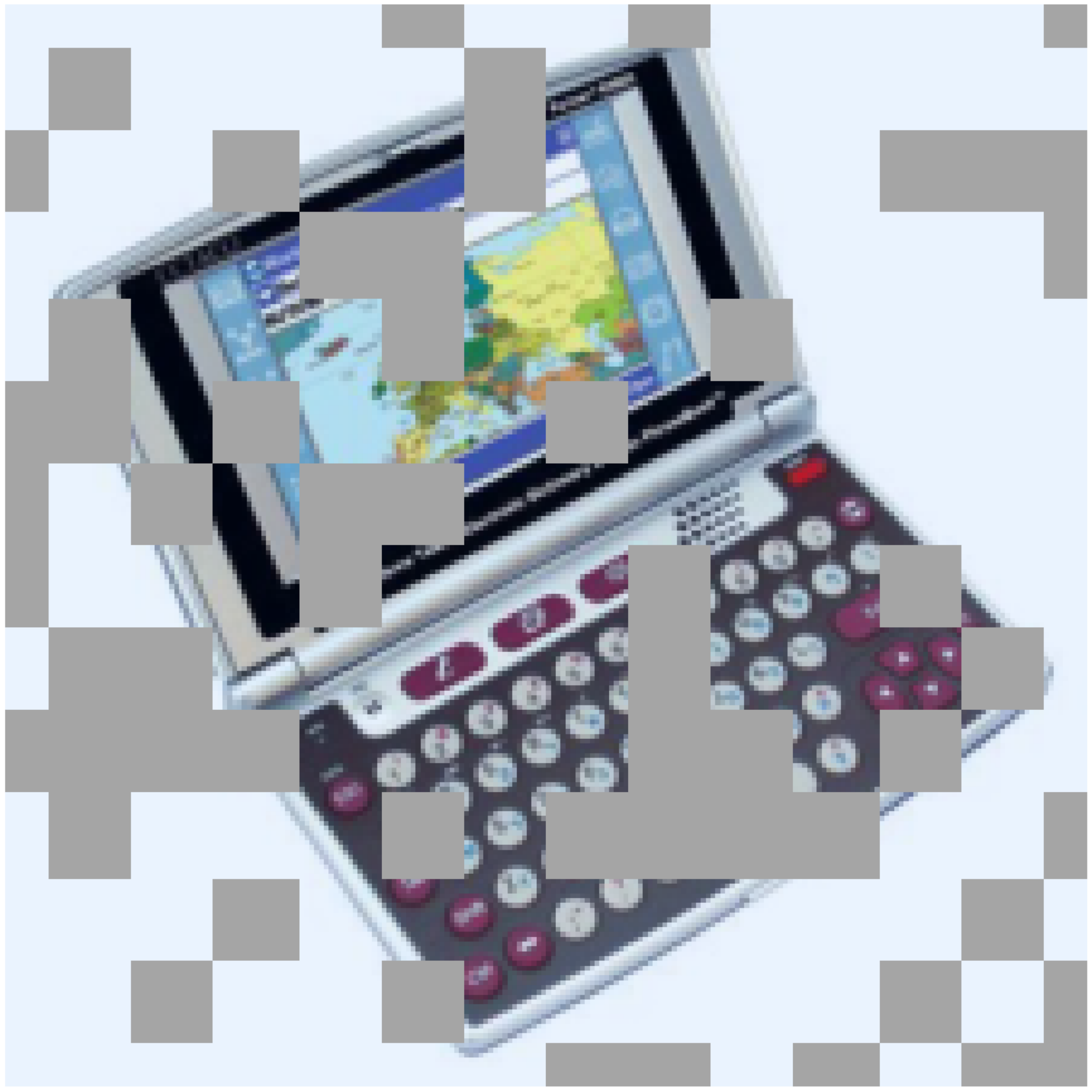}                          &
	\fig[.120]{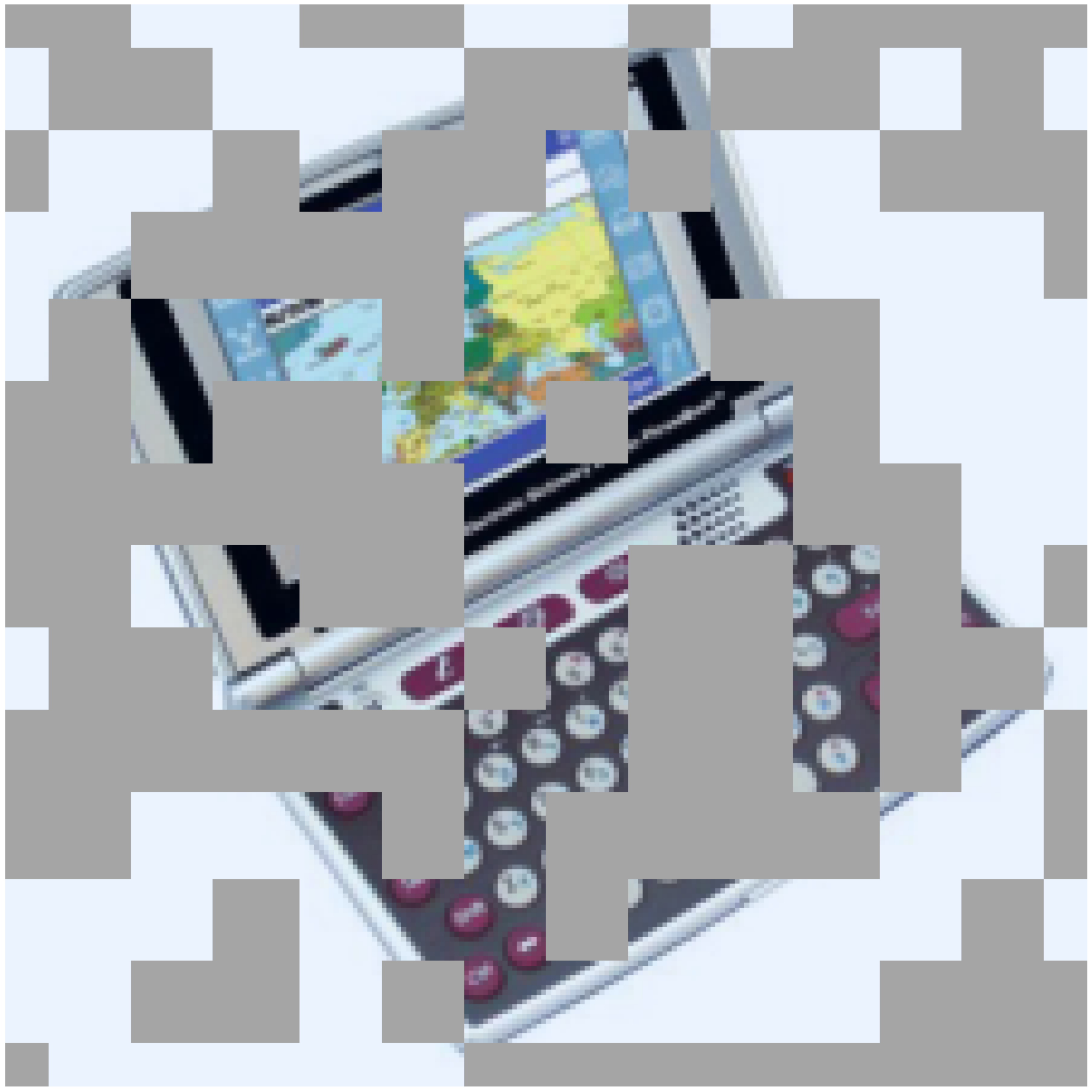}                          &

	& &

	\fig[.120]{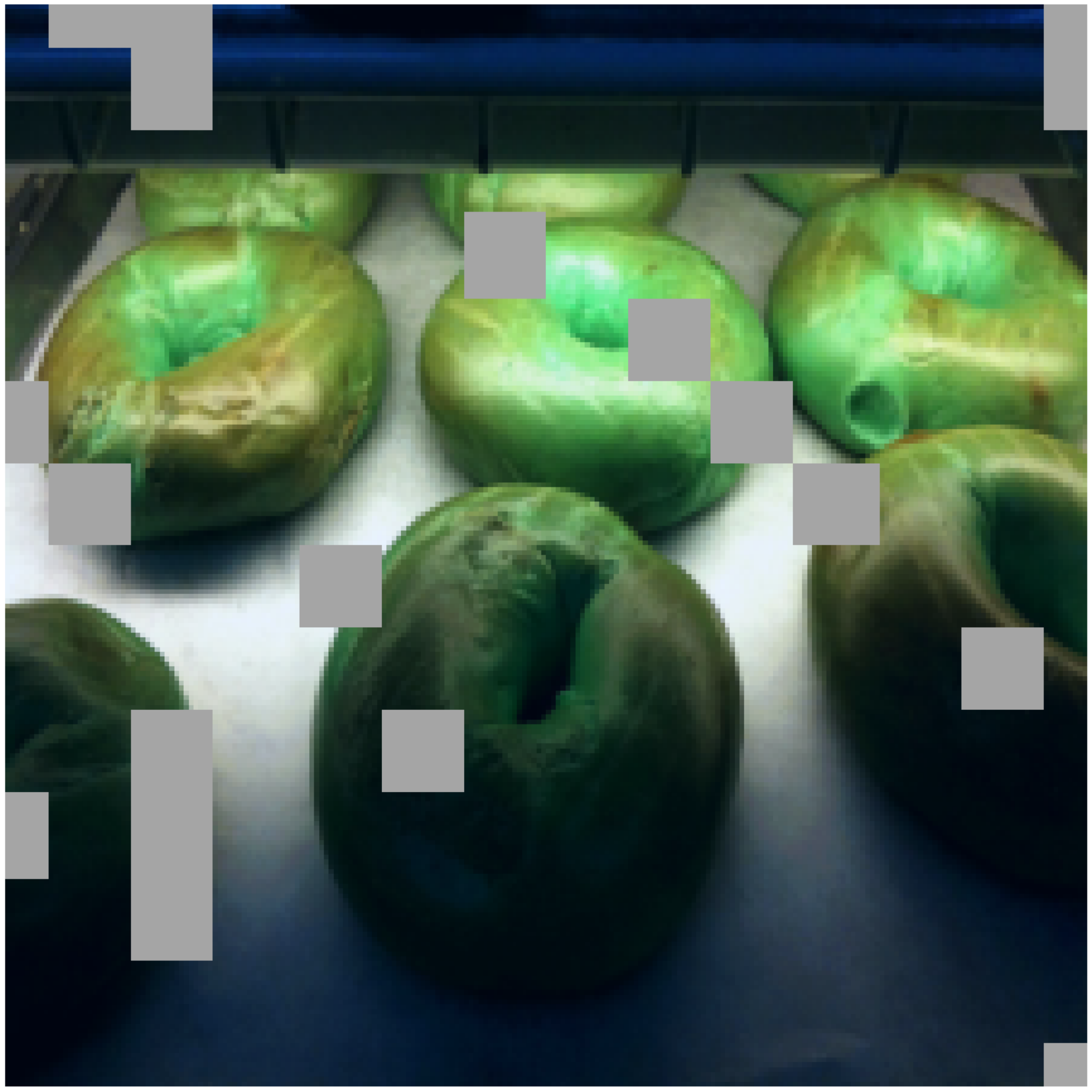}                          &
	\fig[.120]{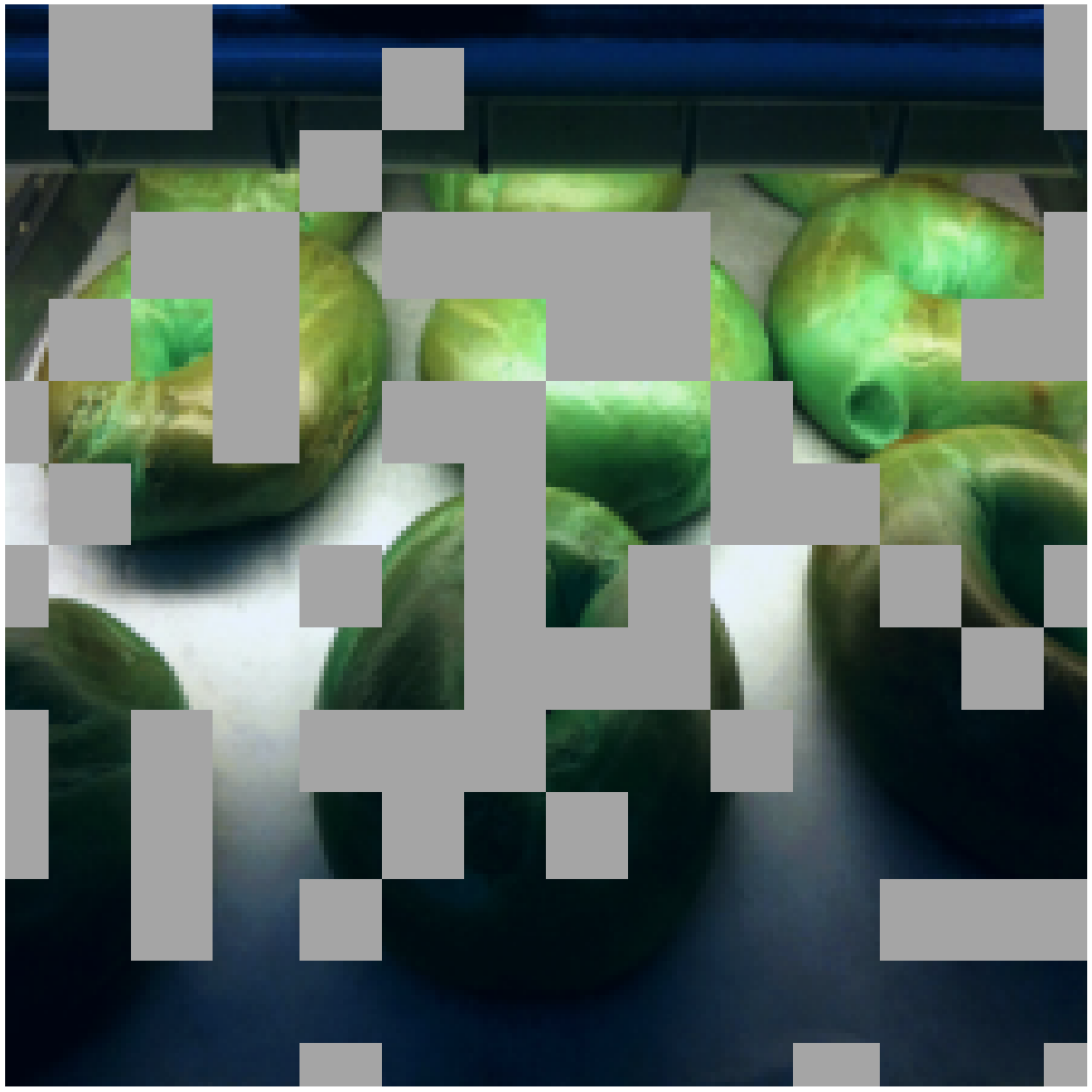}                          &
	\fig[.120]{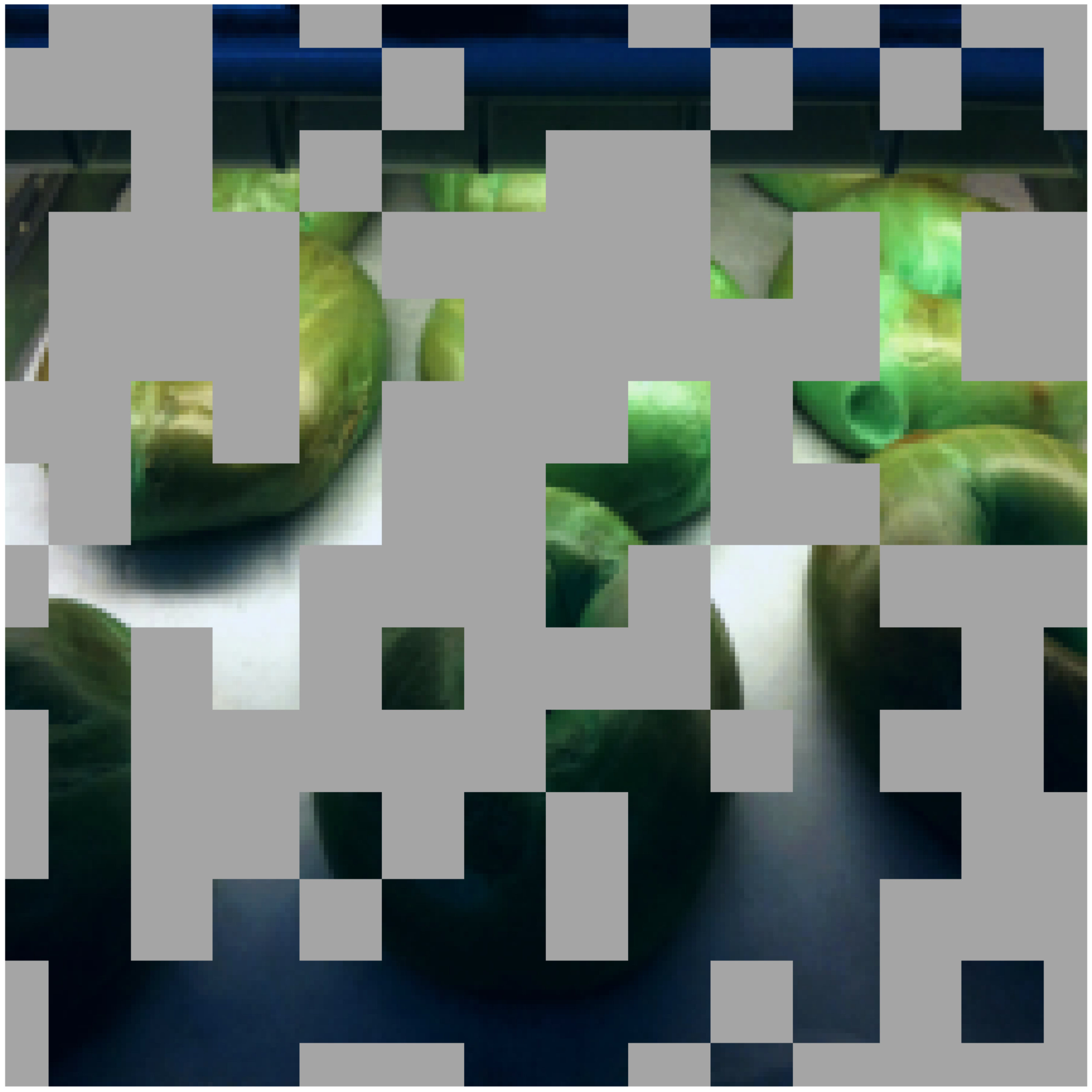}                          \\

	\raisebox{15pt}{\ours-High} &
	\fig[.120]{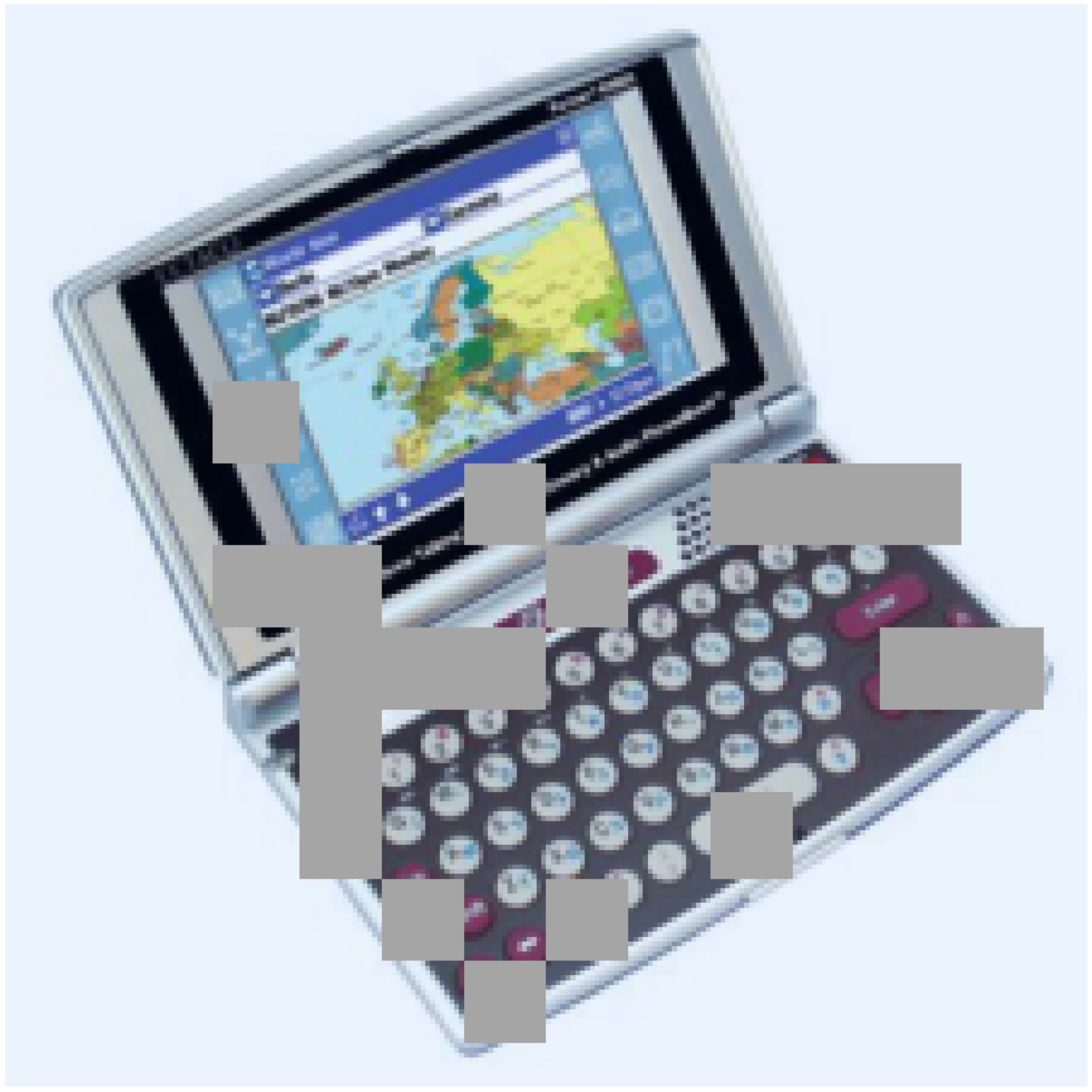}                          &
	\fig[.120]{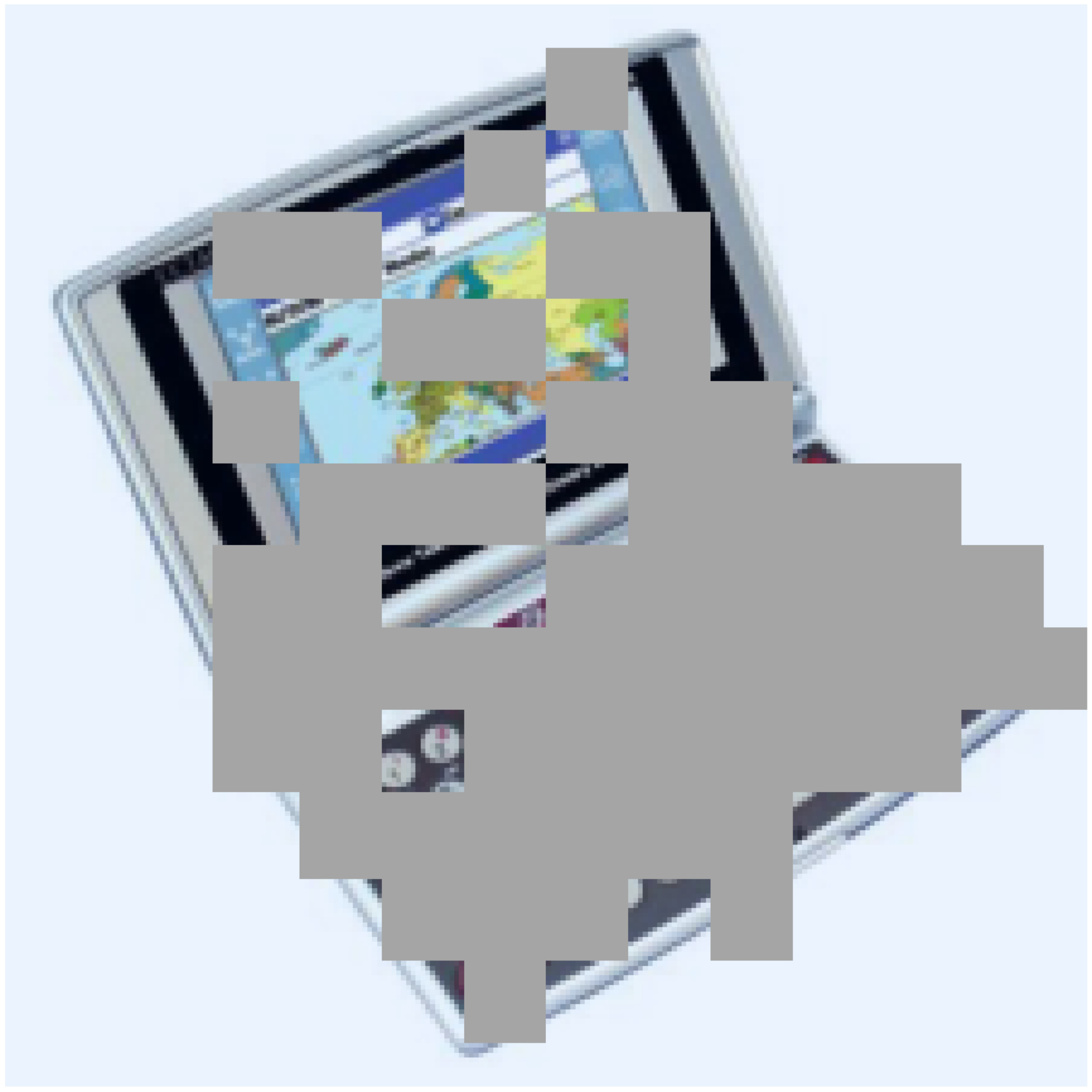}                          &
	\fig[.120]{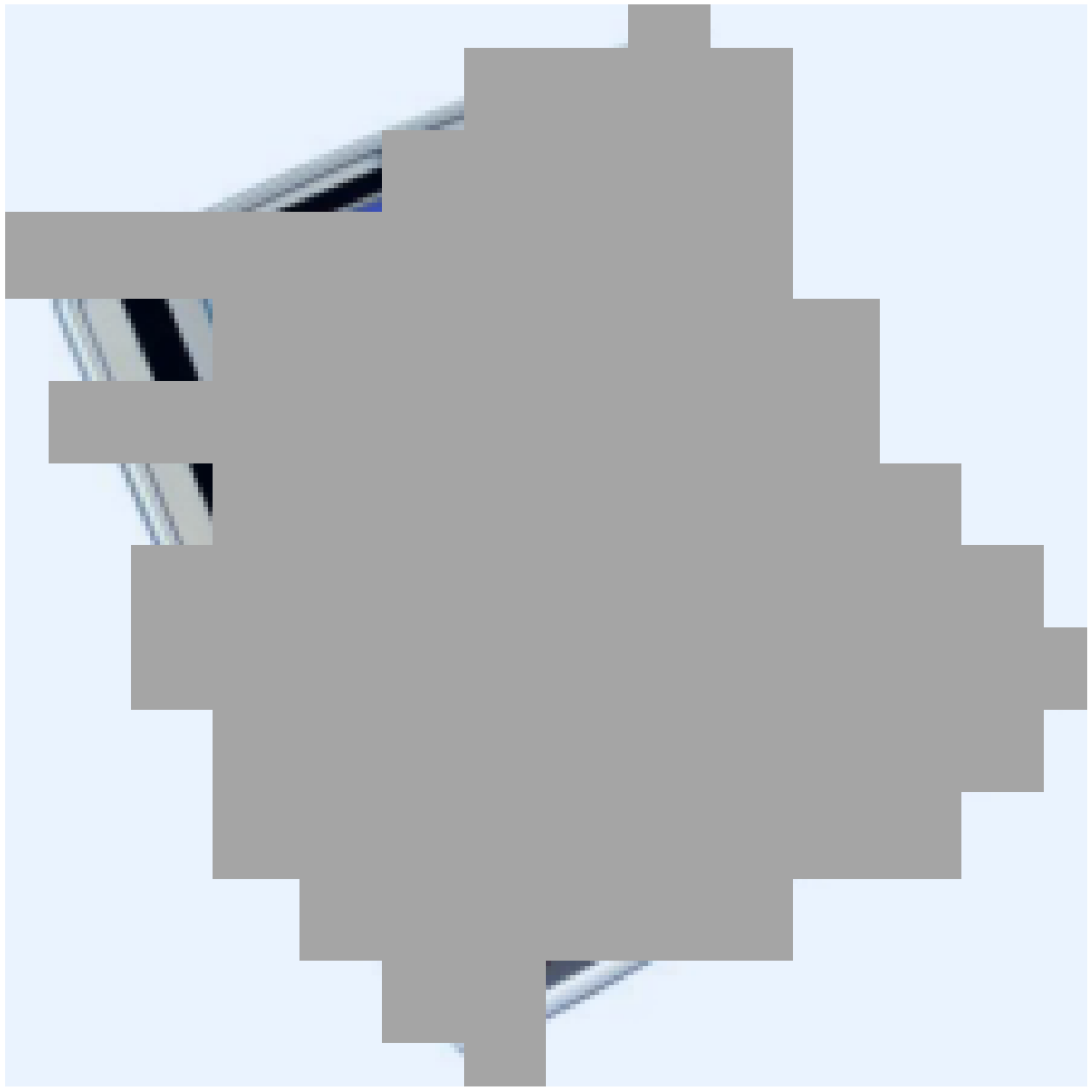}                          &

	& &

	\fig[.120]{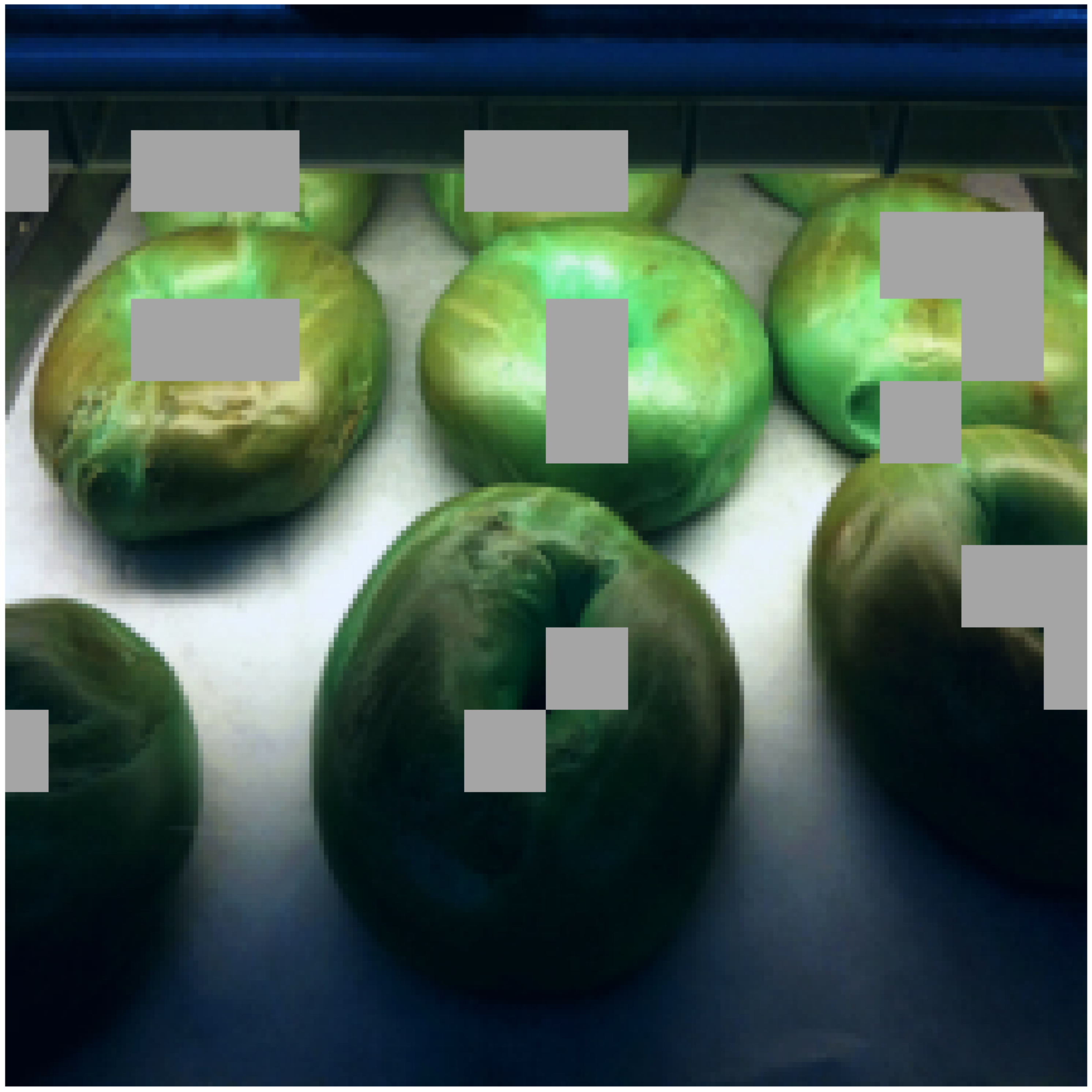}                          &
	\fig[.120]{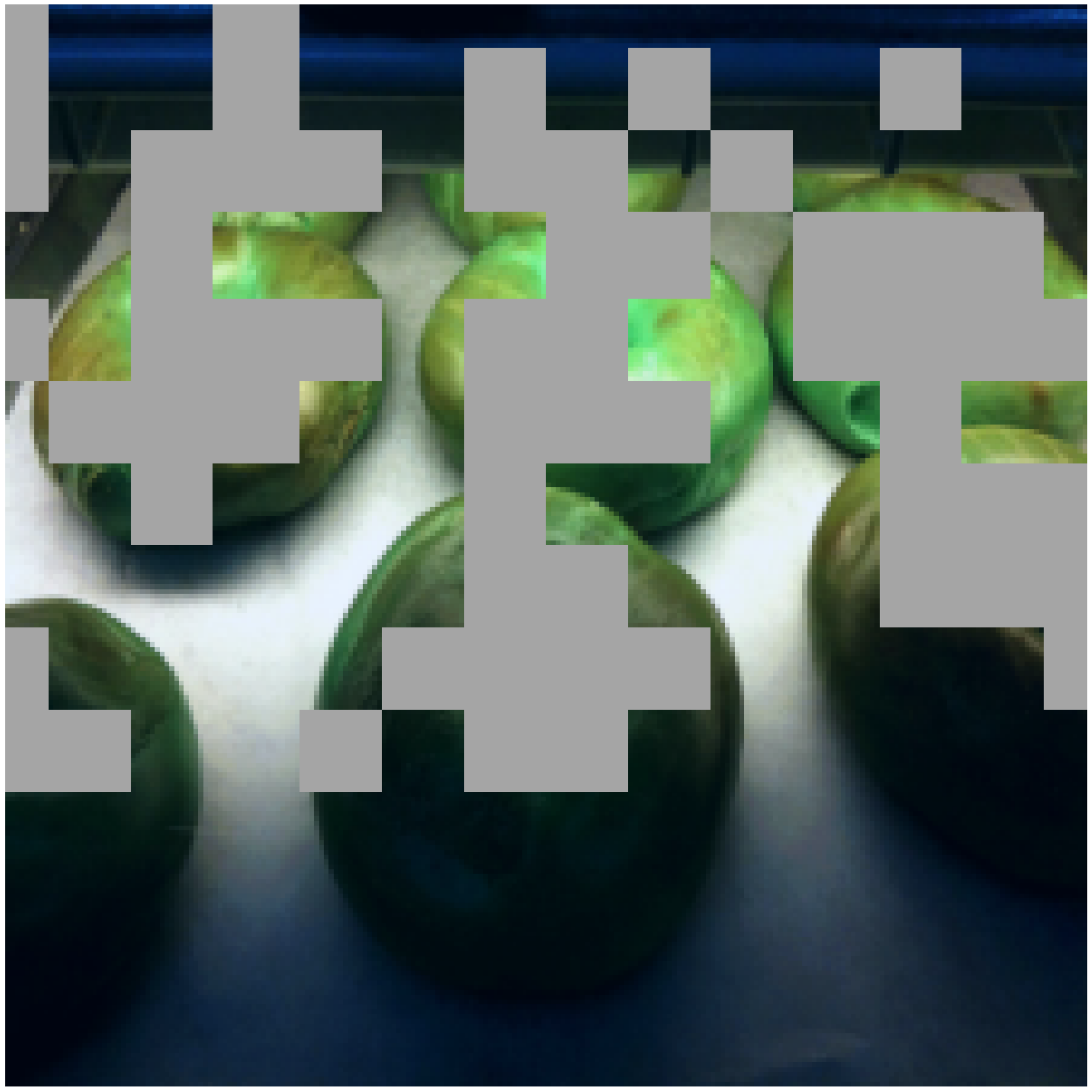}                          &
	\fig[.120]{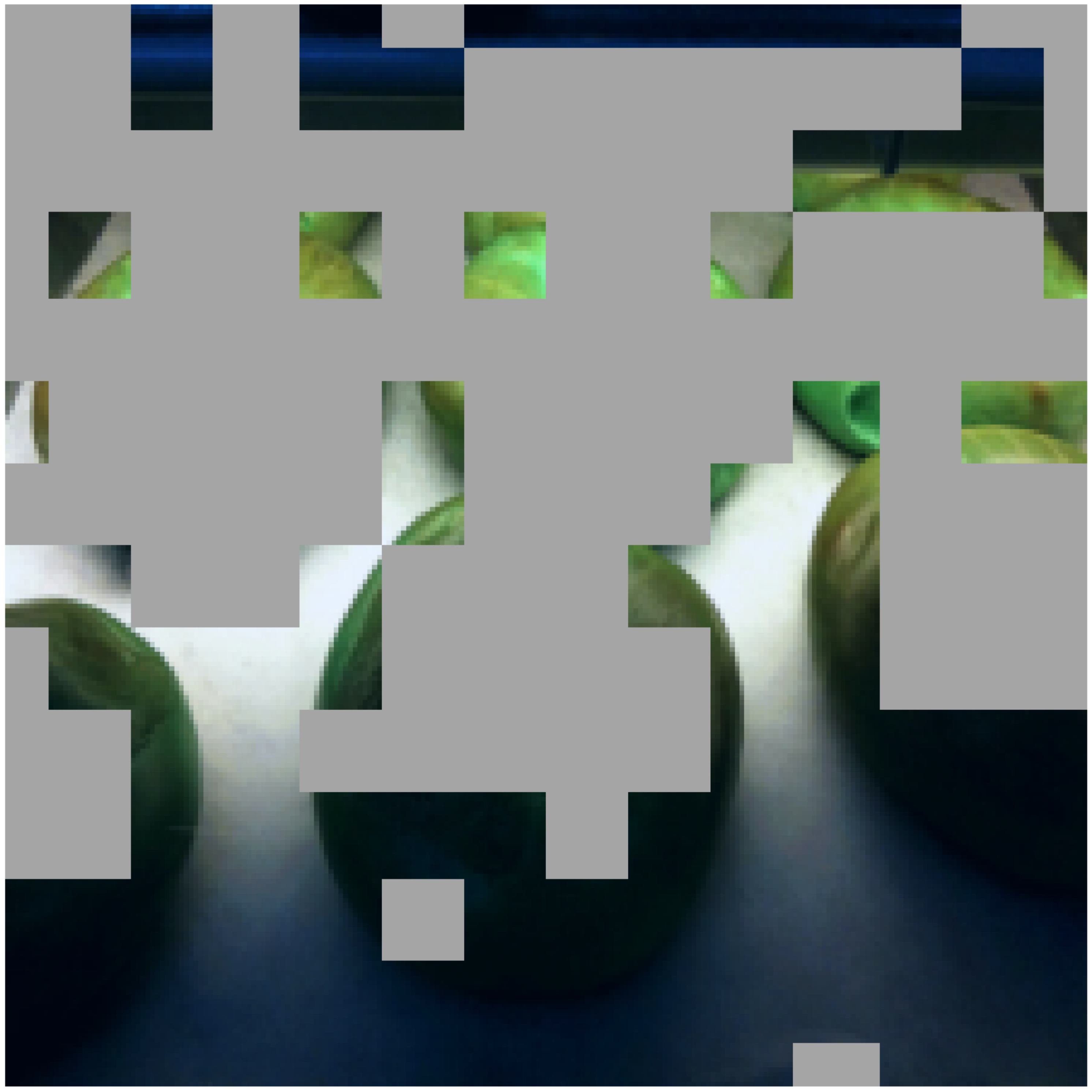}                          \\

	\raisebox{15pt}{\ours-Hint} &
	\fig[.120]{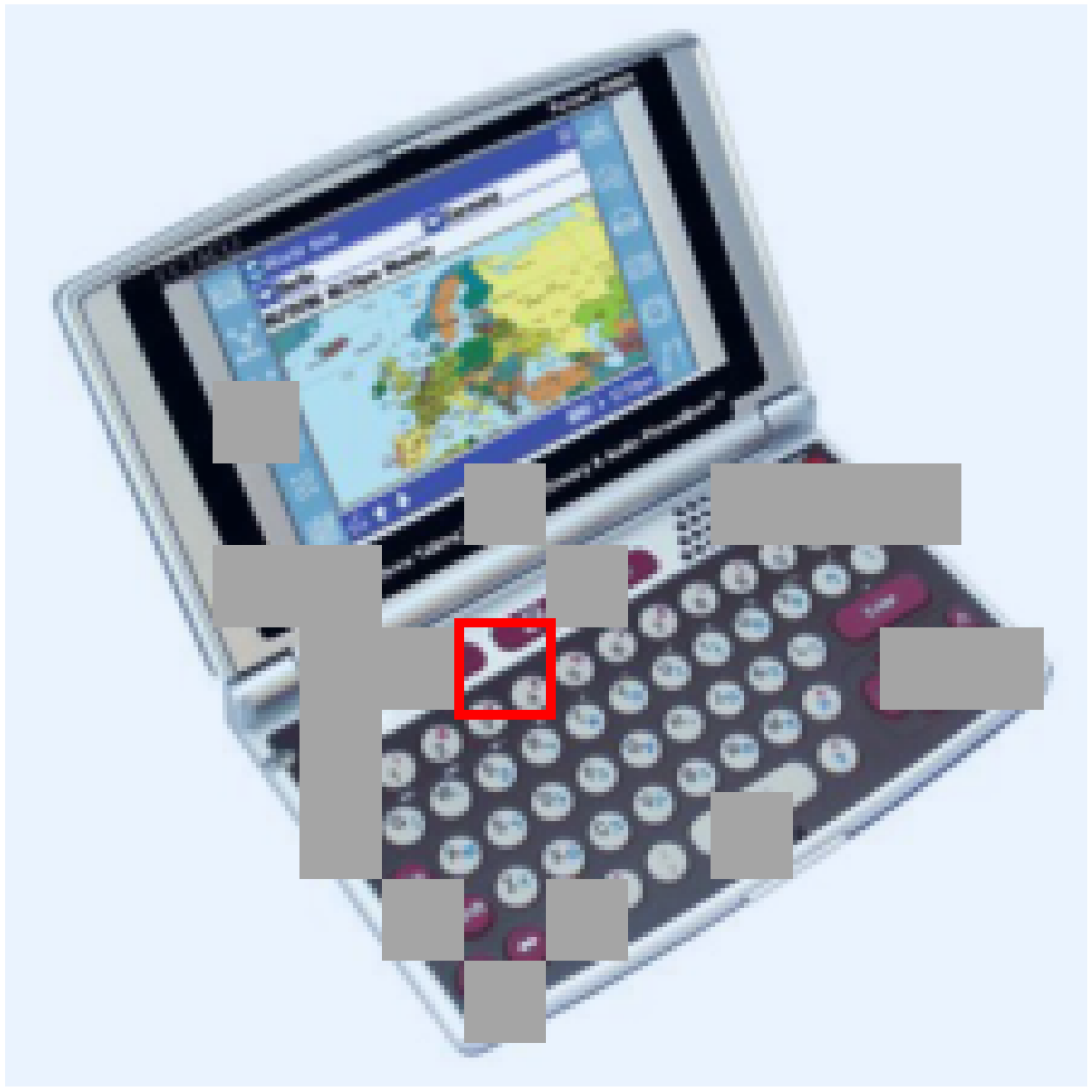}                          &
	\fig[.120]{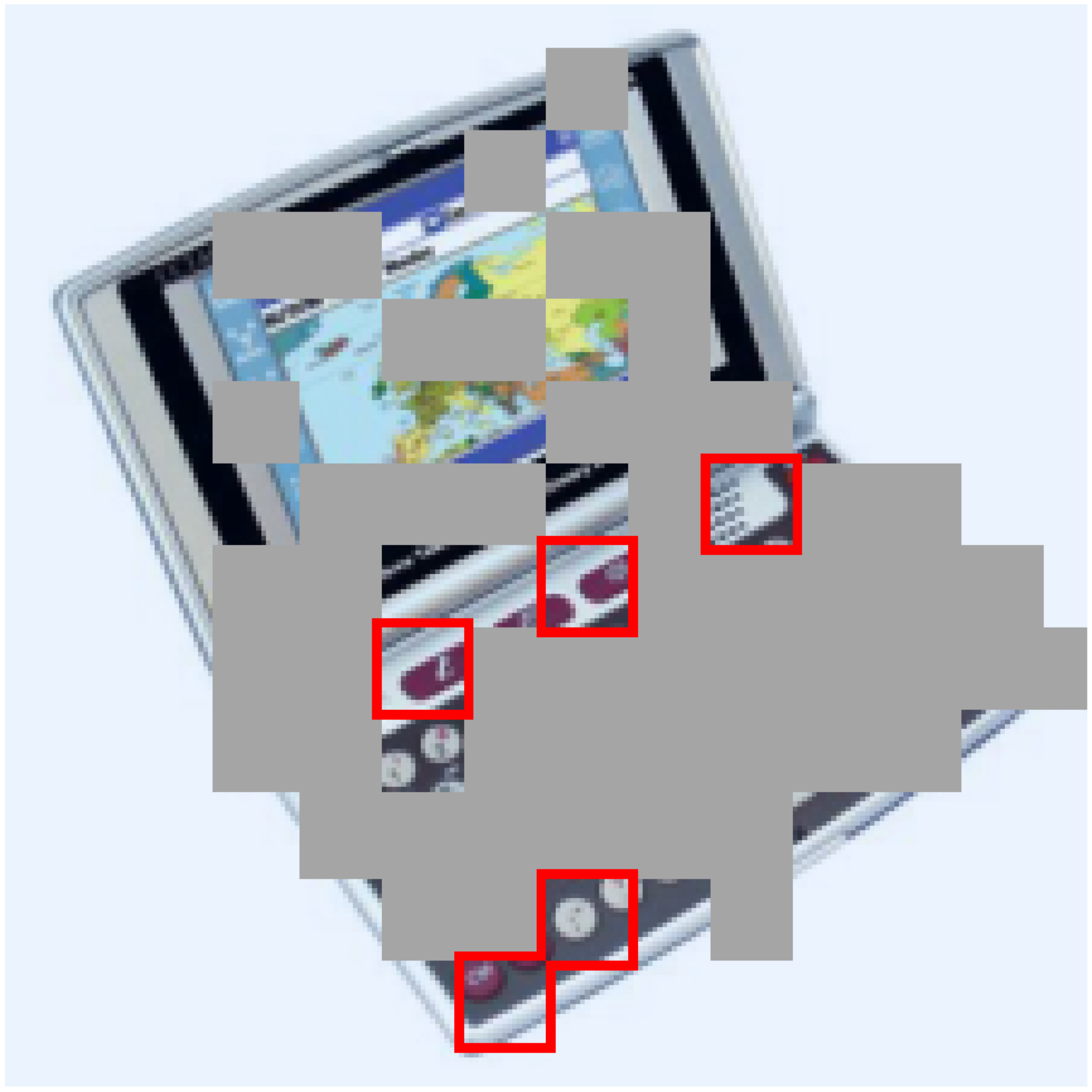}                          &
	\fig[.120]{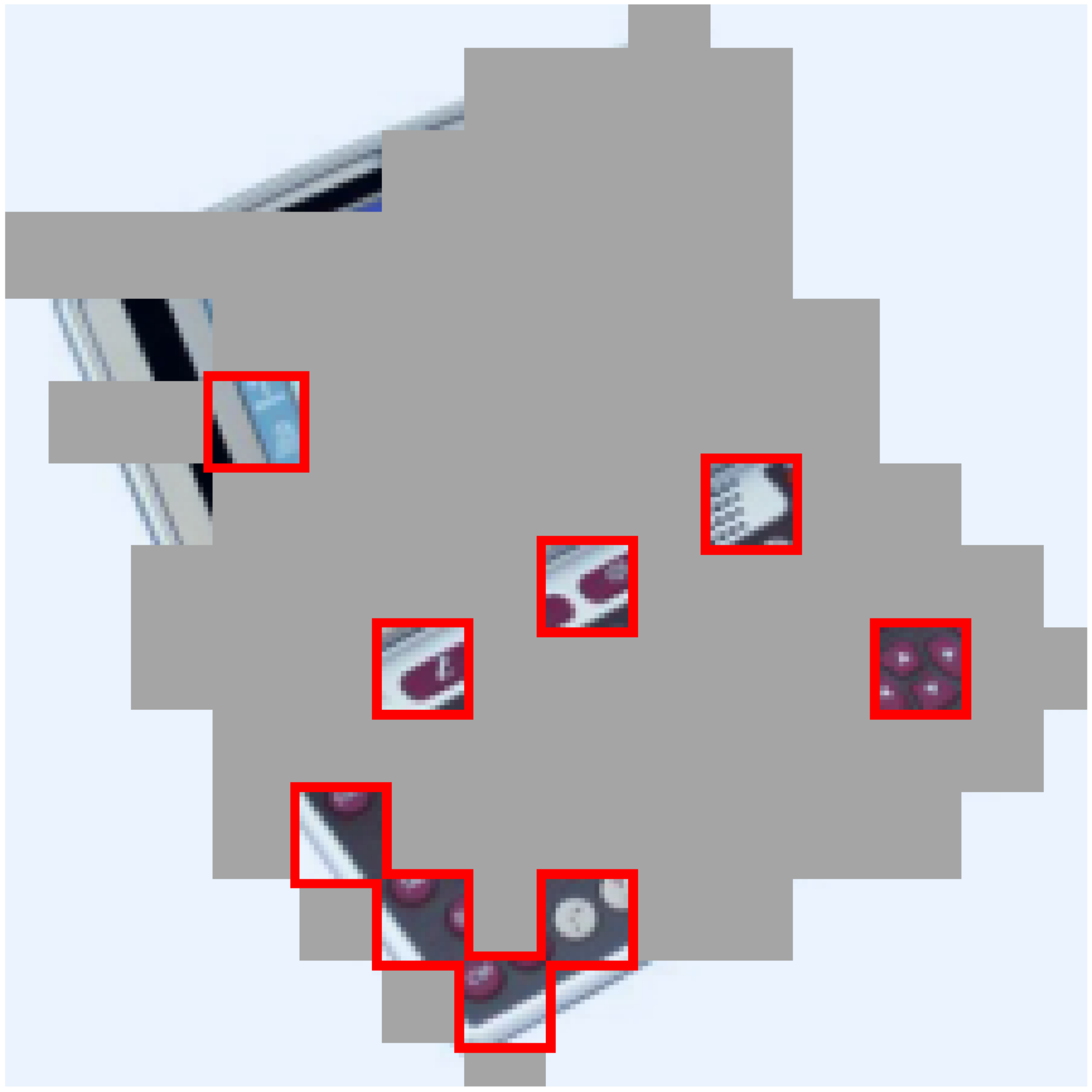}                          &

	&  &

	\fig[.120]{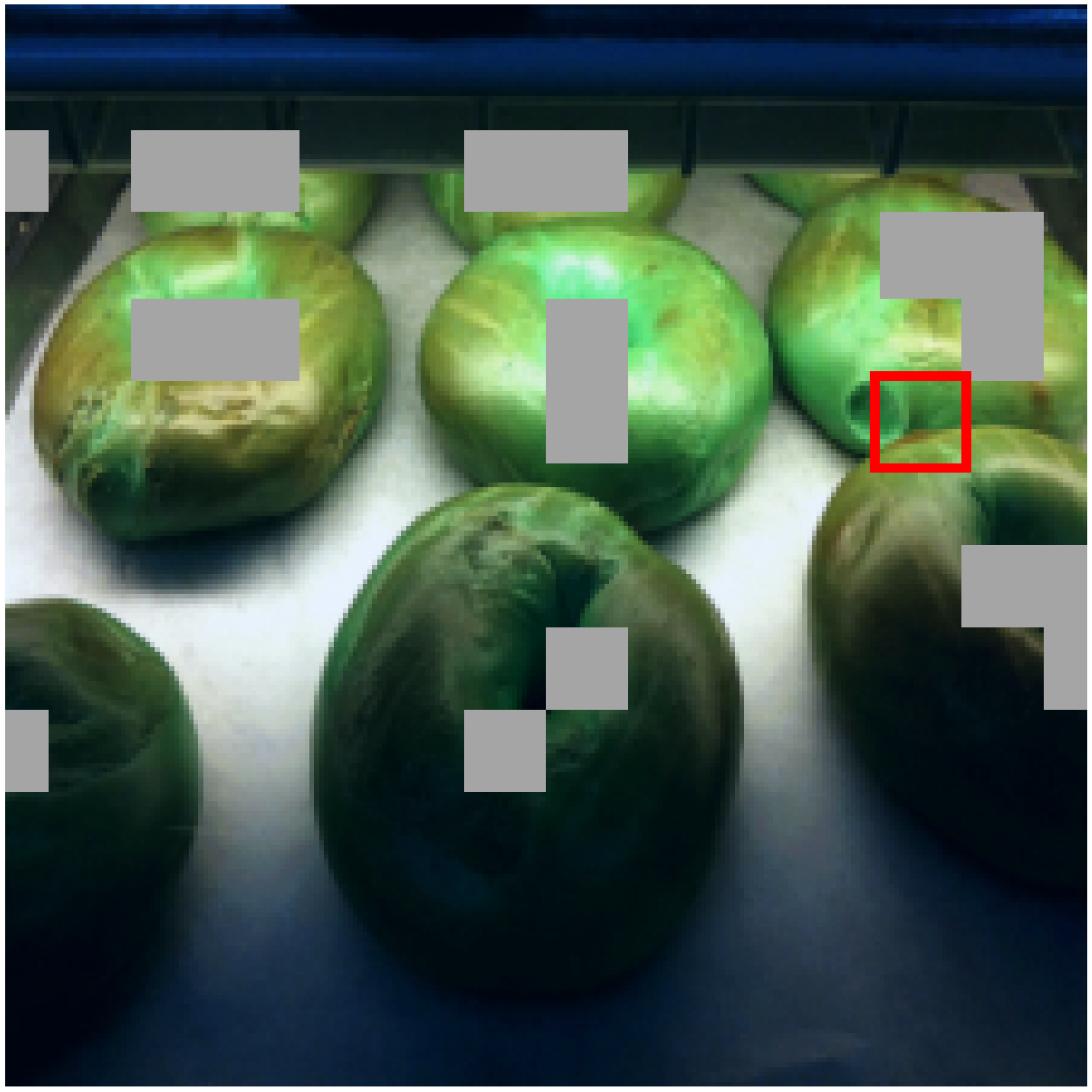}                          &
	\fig[.120]{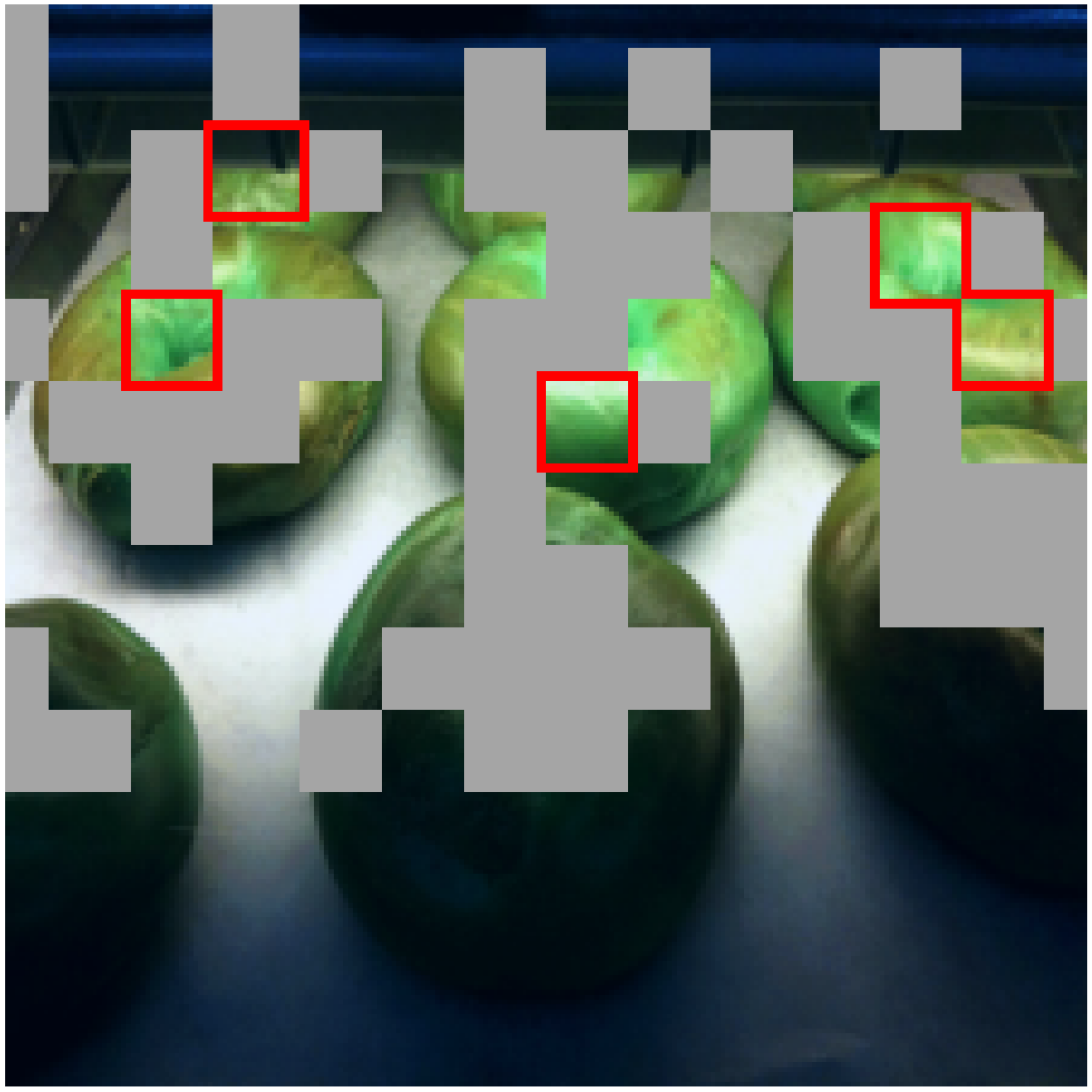}                          &
	\fig[.120]{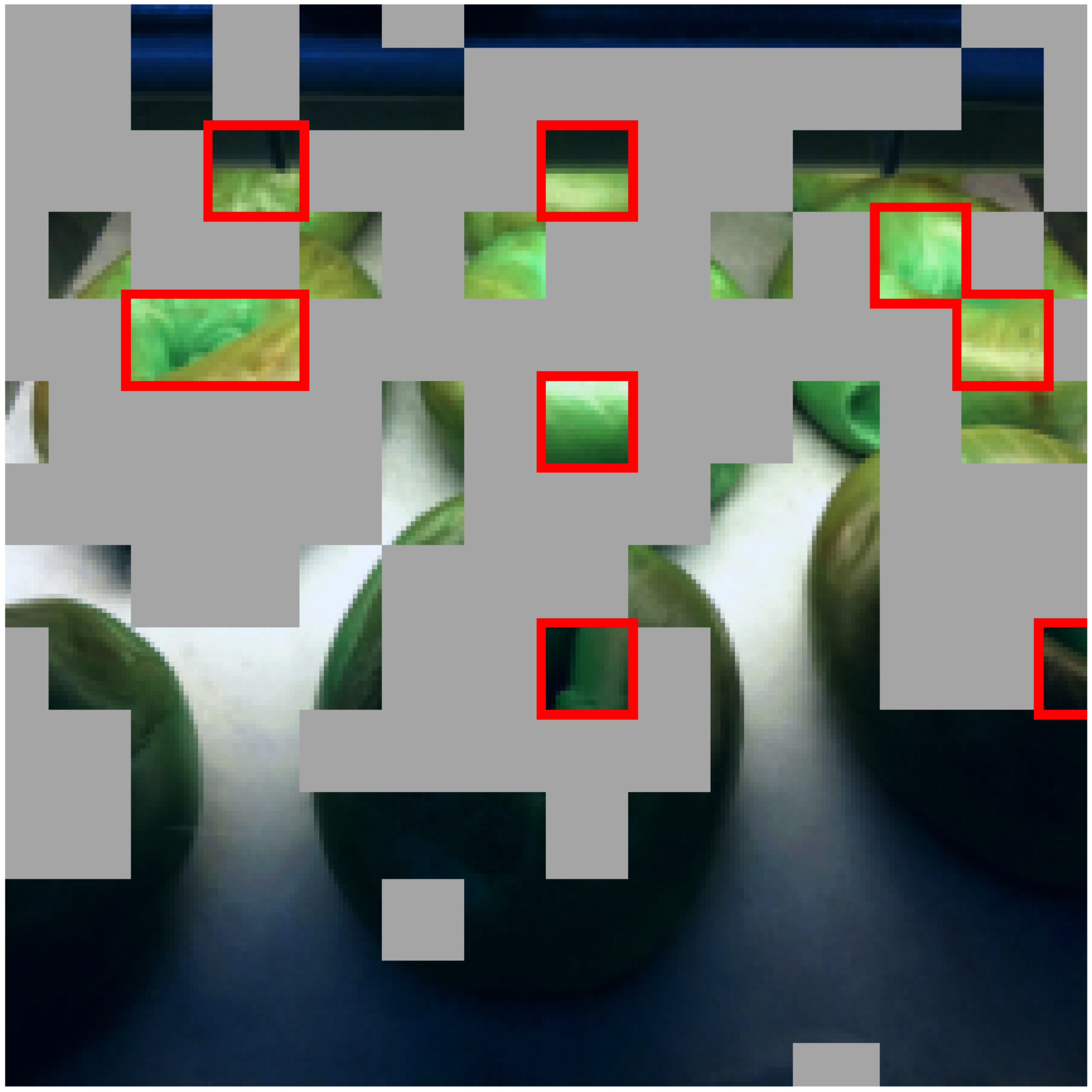}                          \\

	& & & & & & & & \\
	\midrule
	& & & & & & & & \\

	\raisebox{15pt}{Block-Wise}  &
	\fig[.120]{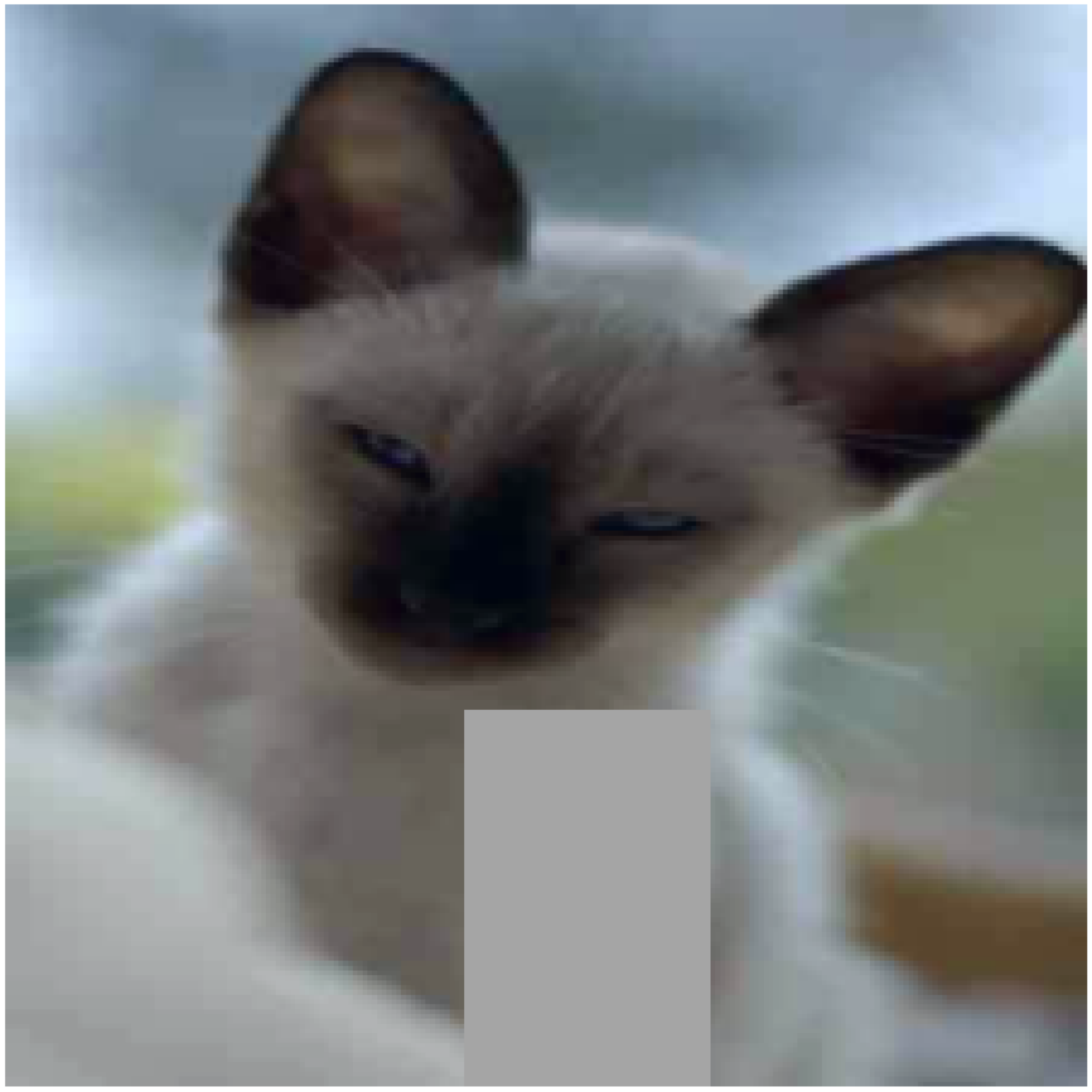}               &
	\fig[.120]{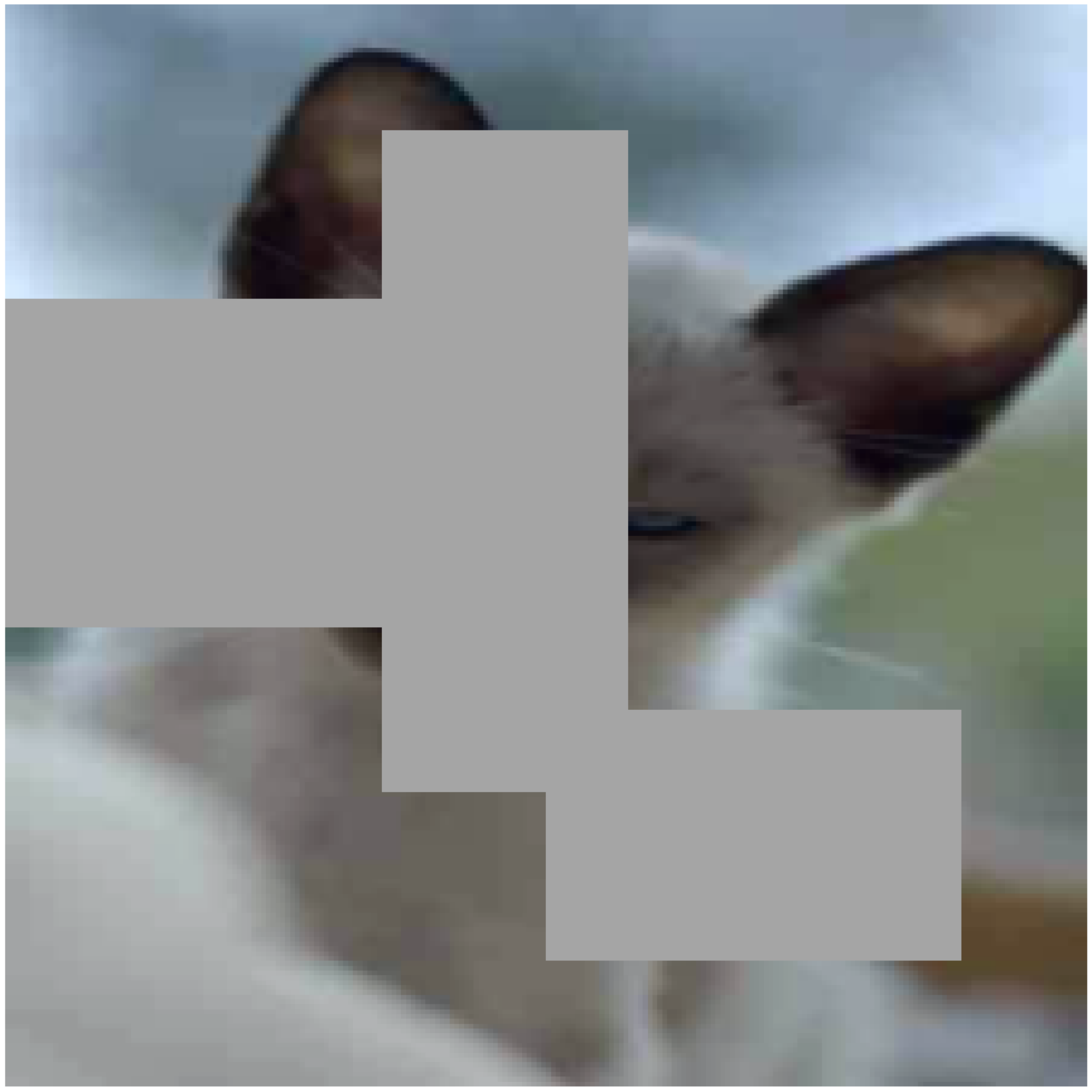}               &
	\fig[.120]{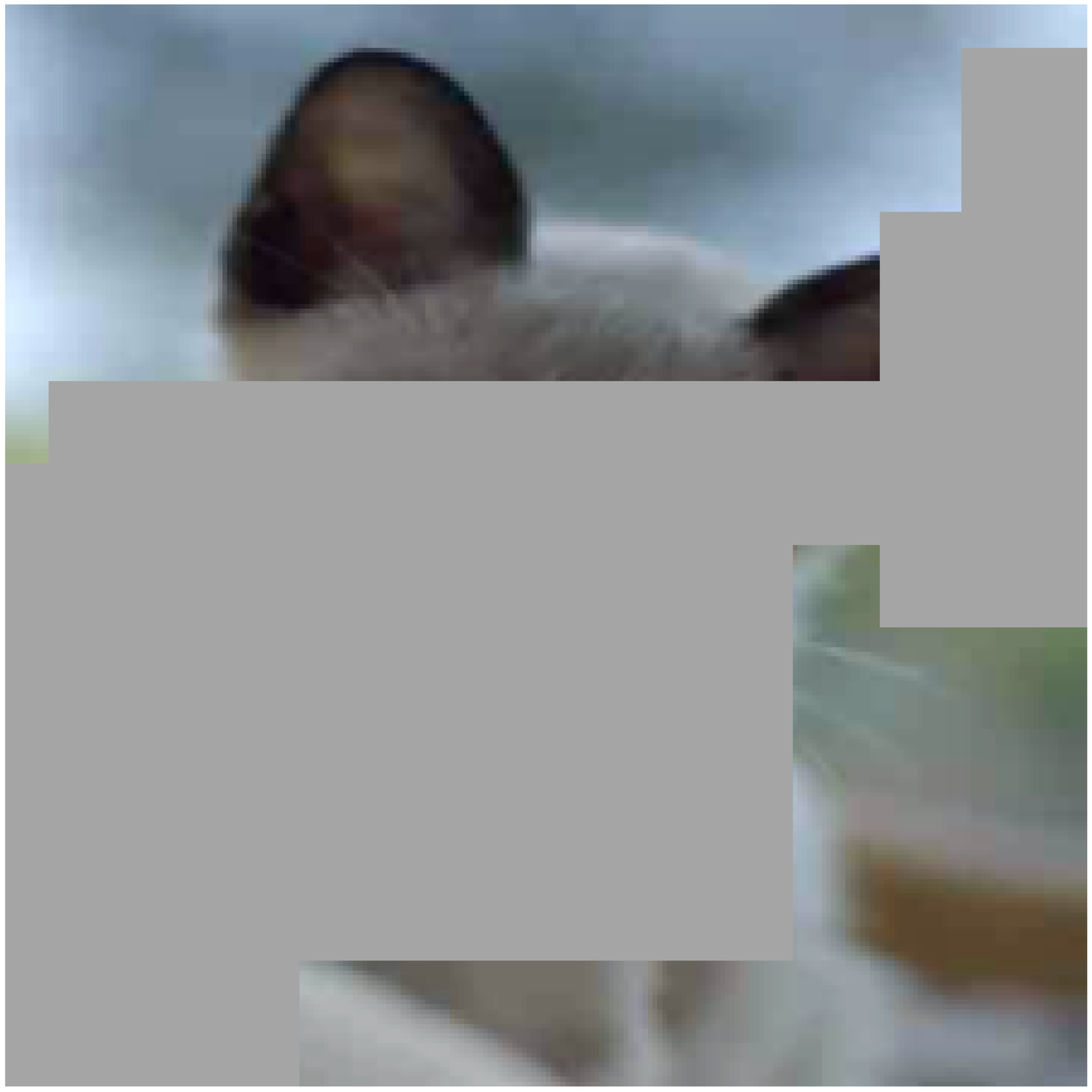}               &

    & &

	\fig[.120]{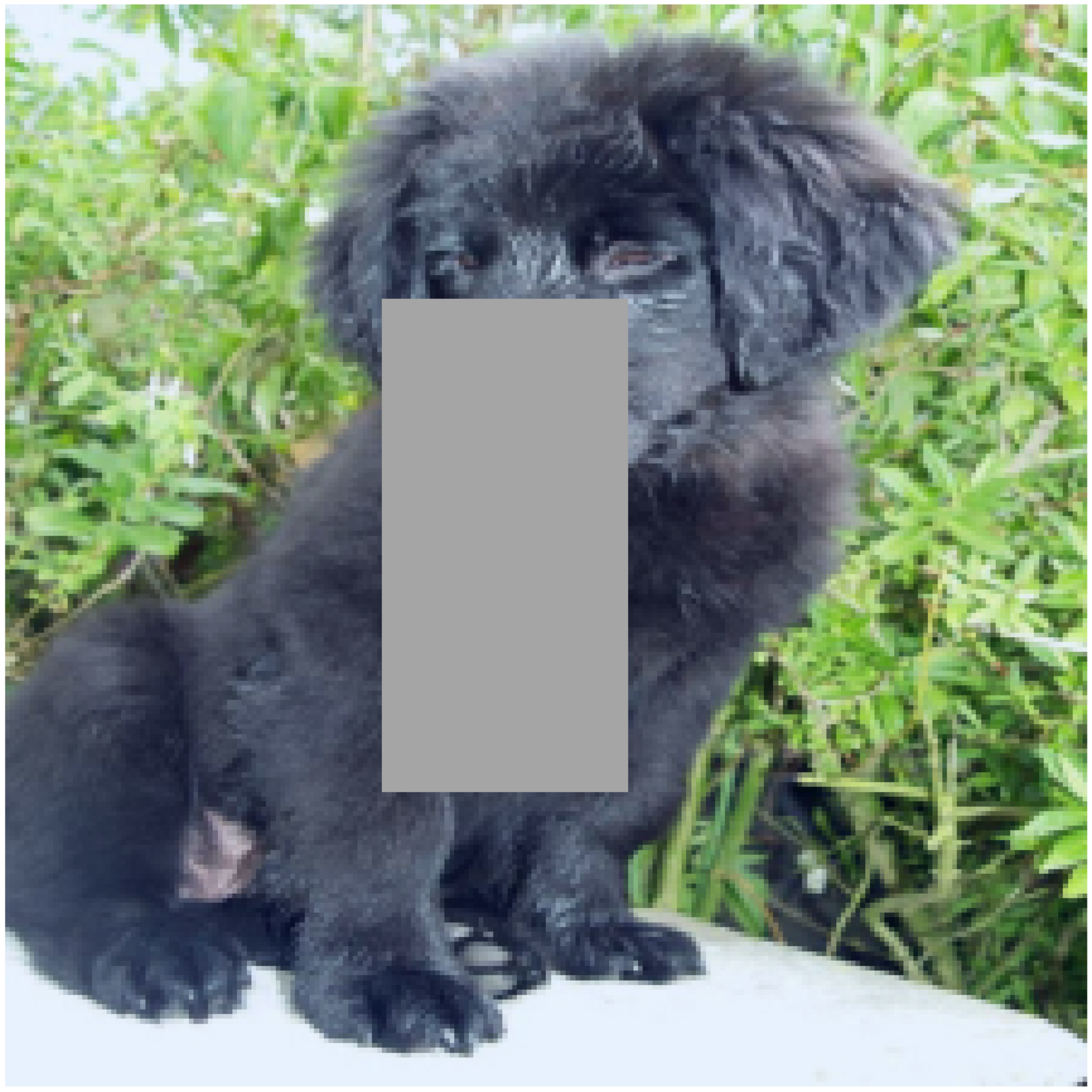}               &
	\fig[.120]{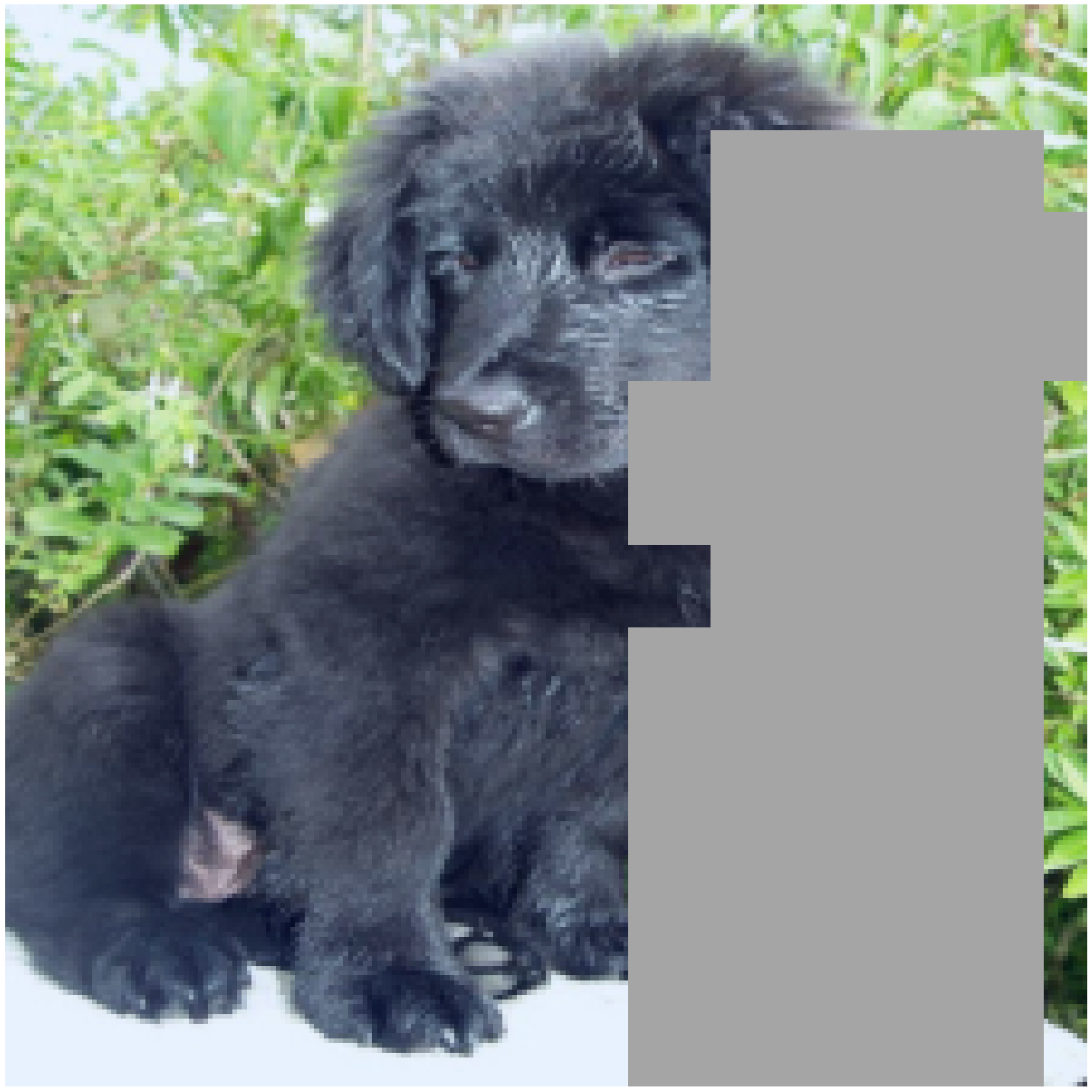}               &
	\fig[.120]{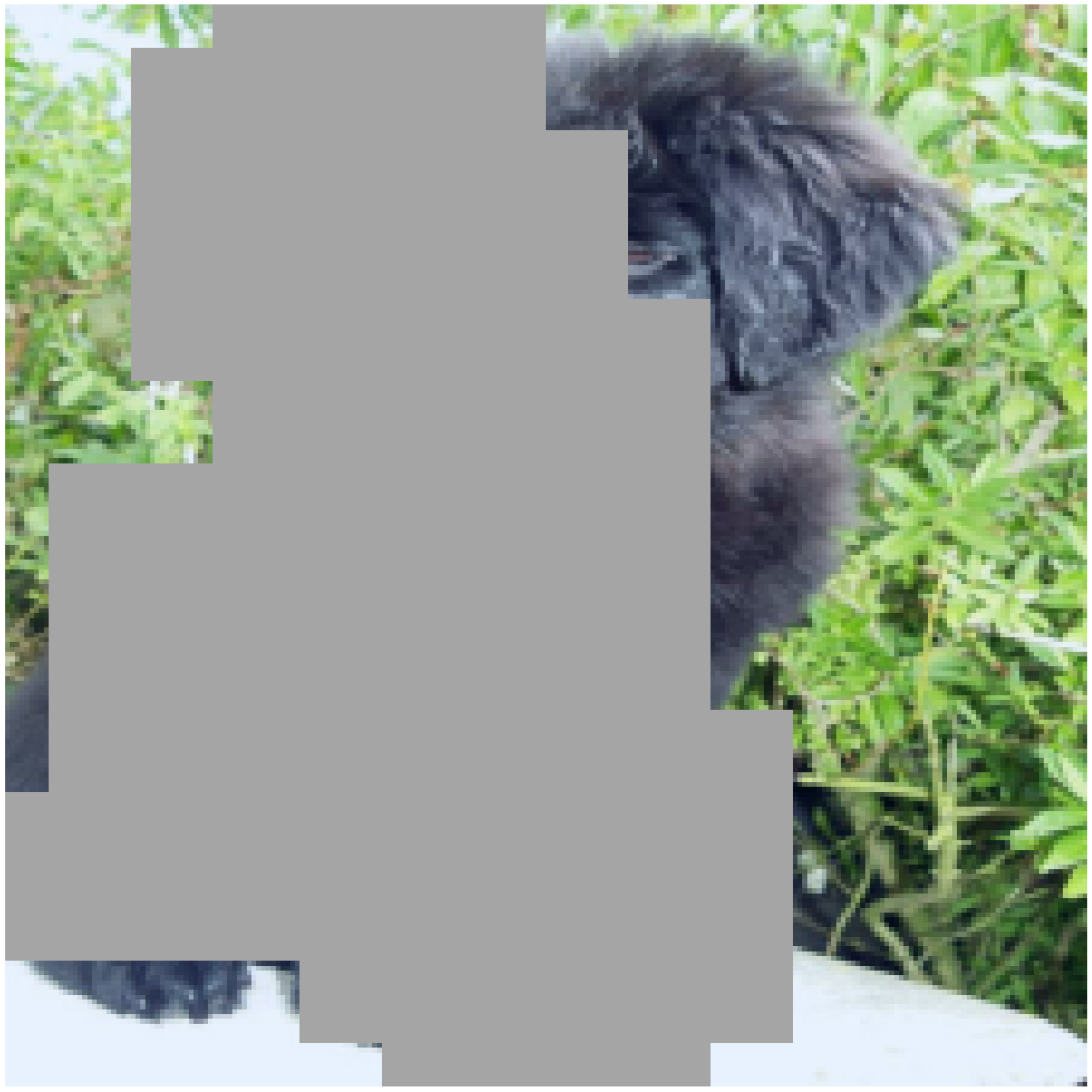}               \\

	\raisebox{15pt}{Random} &
	\fig[.120]{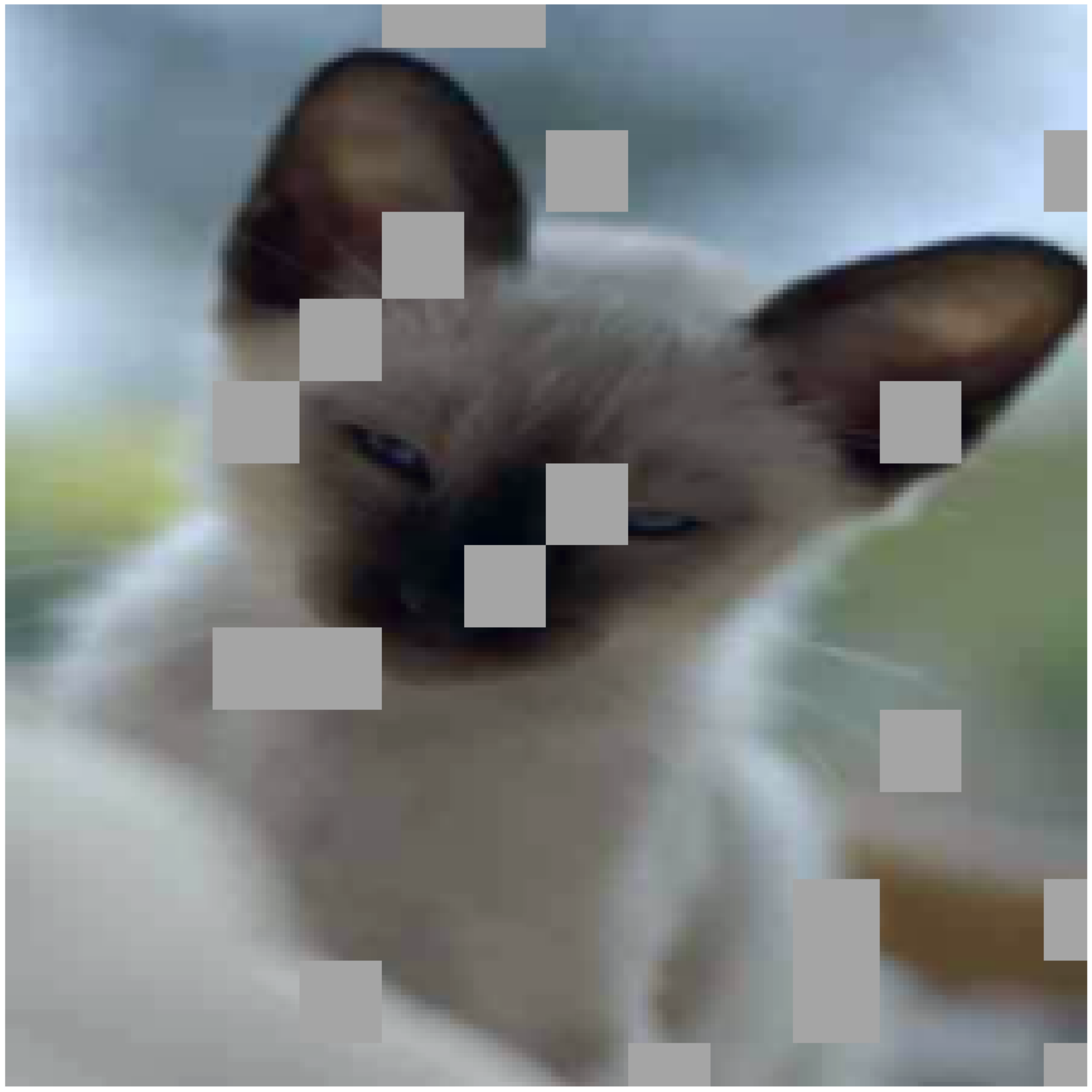}                          &
	\fig[.120]{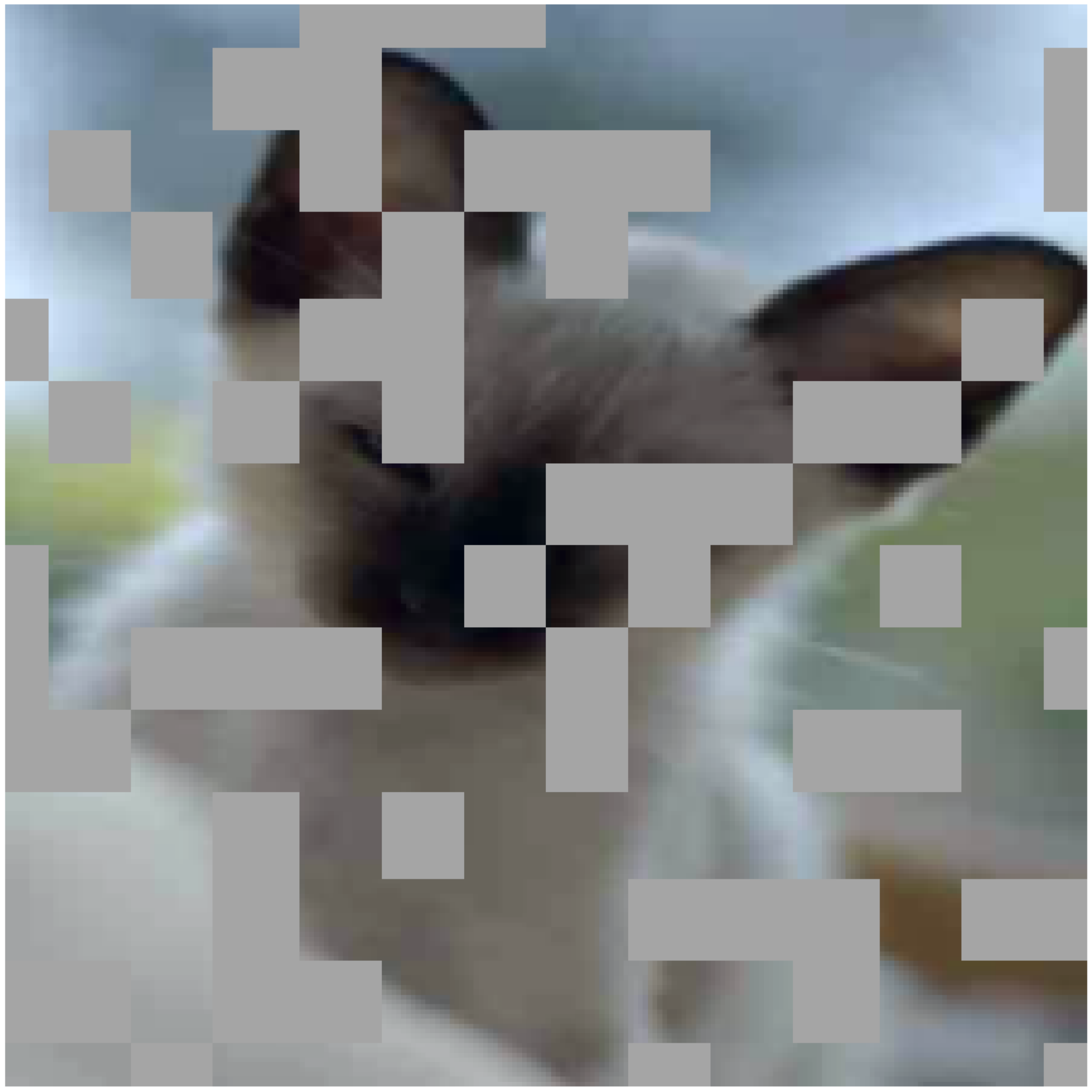}                          &
	\fig[.120]{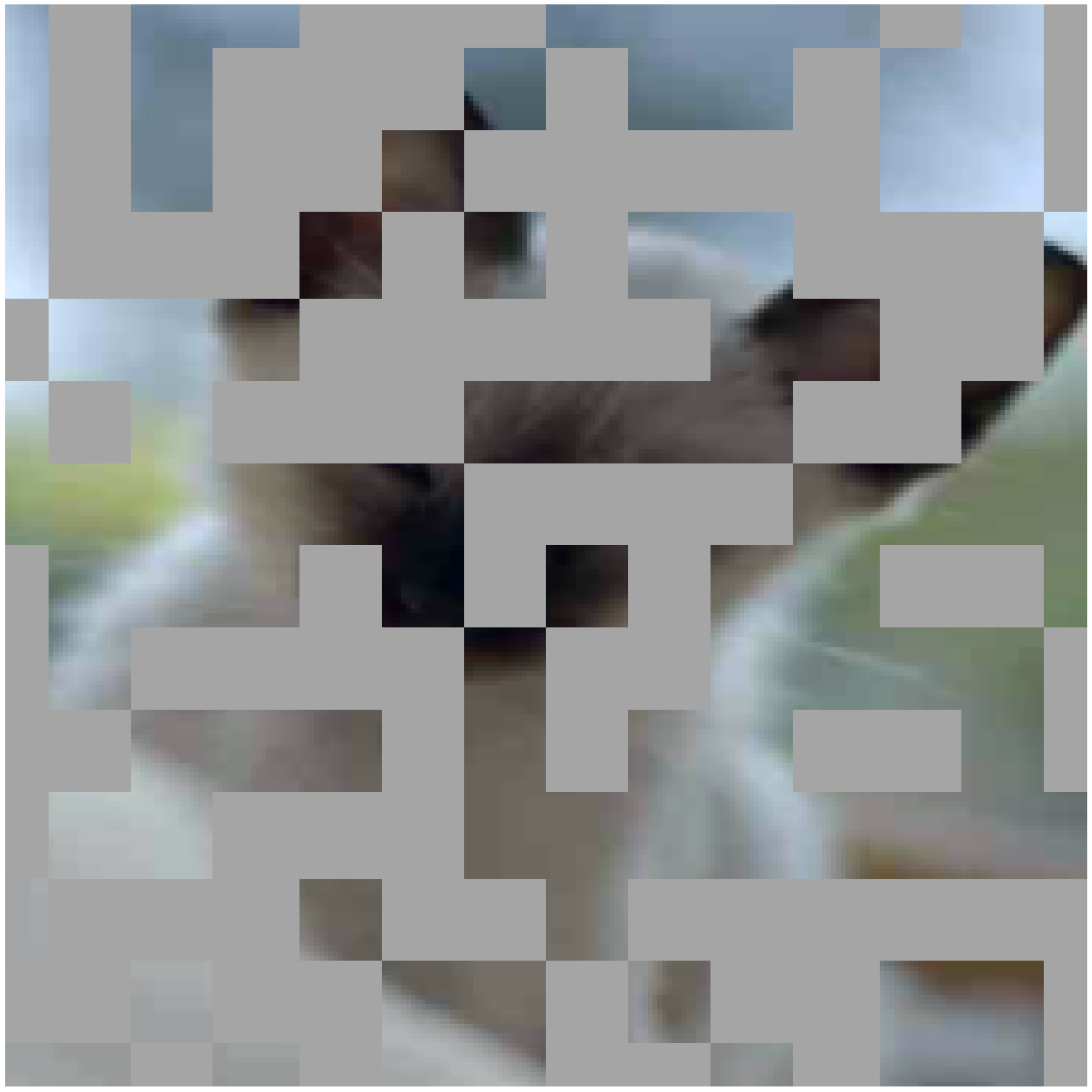}                          &

	& &

	\fig[.120]{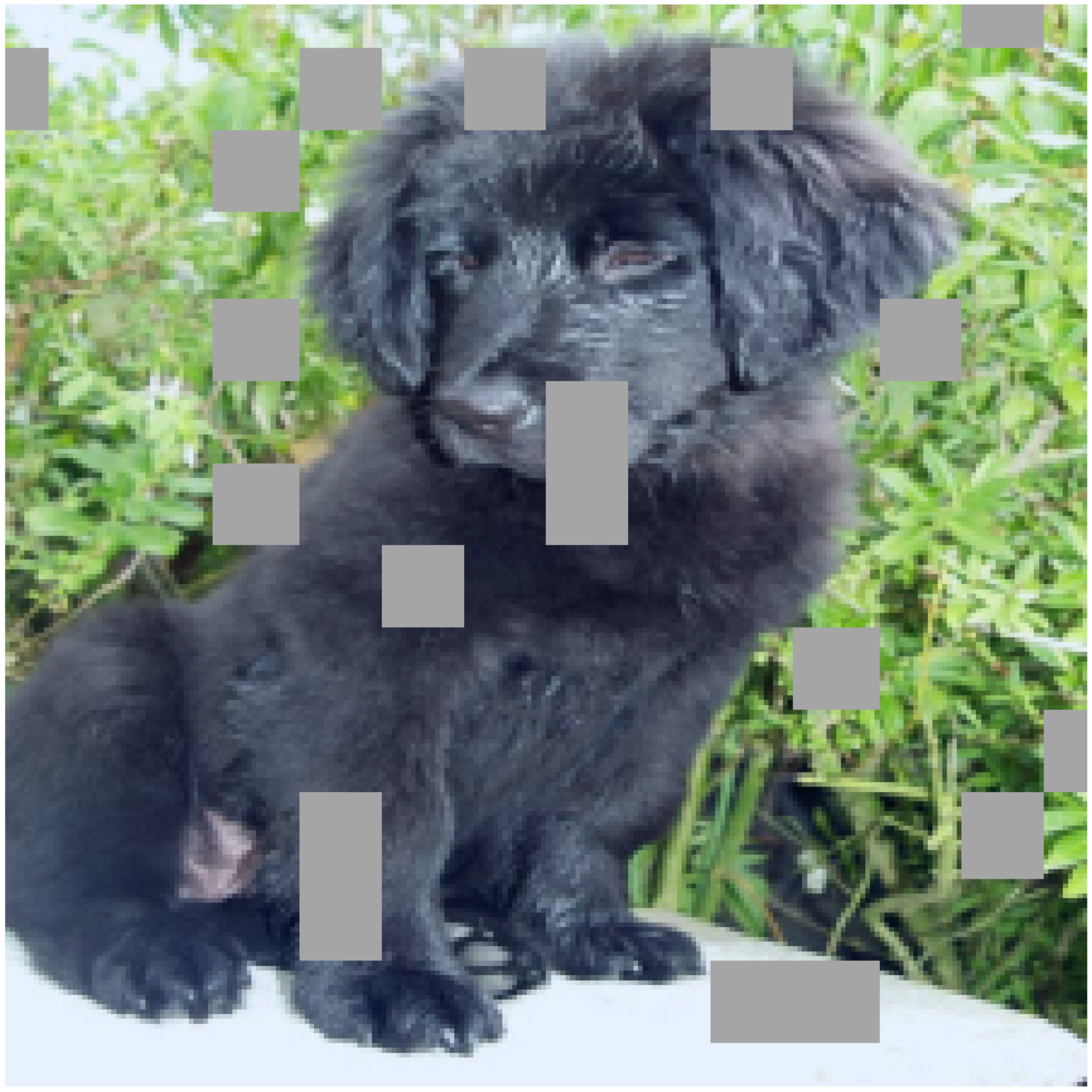}                          &
	\fig[.120]{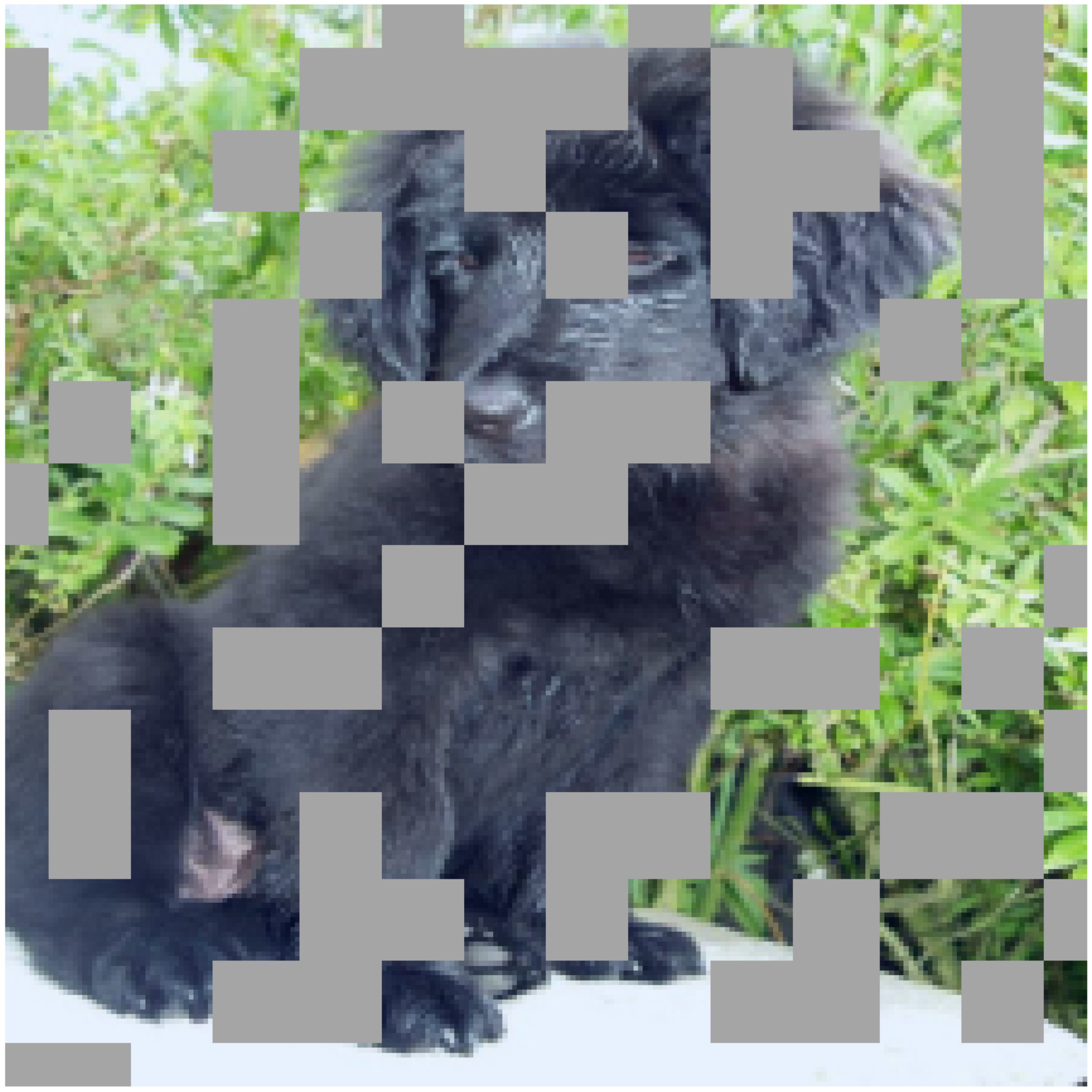}                          &
	\fig[.120]{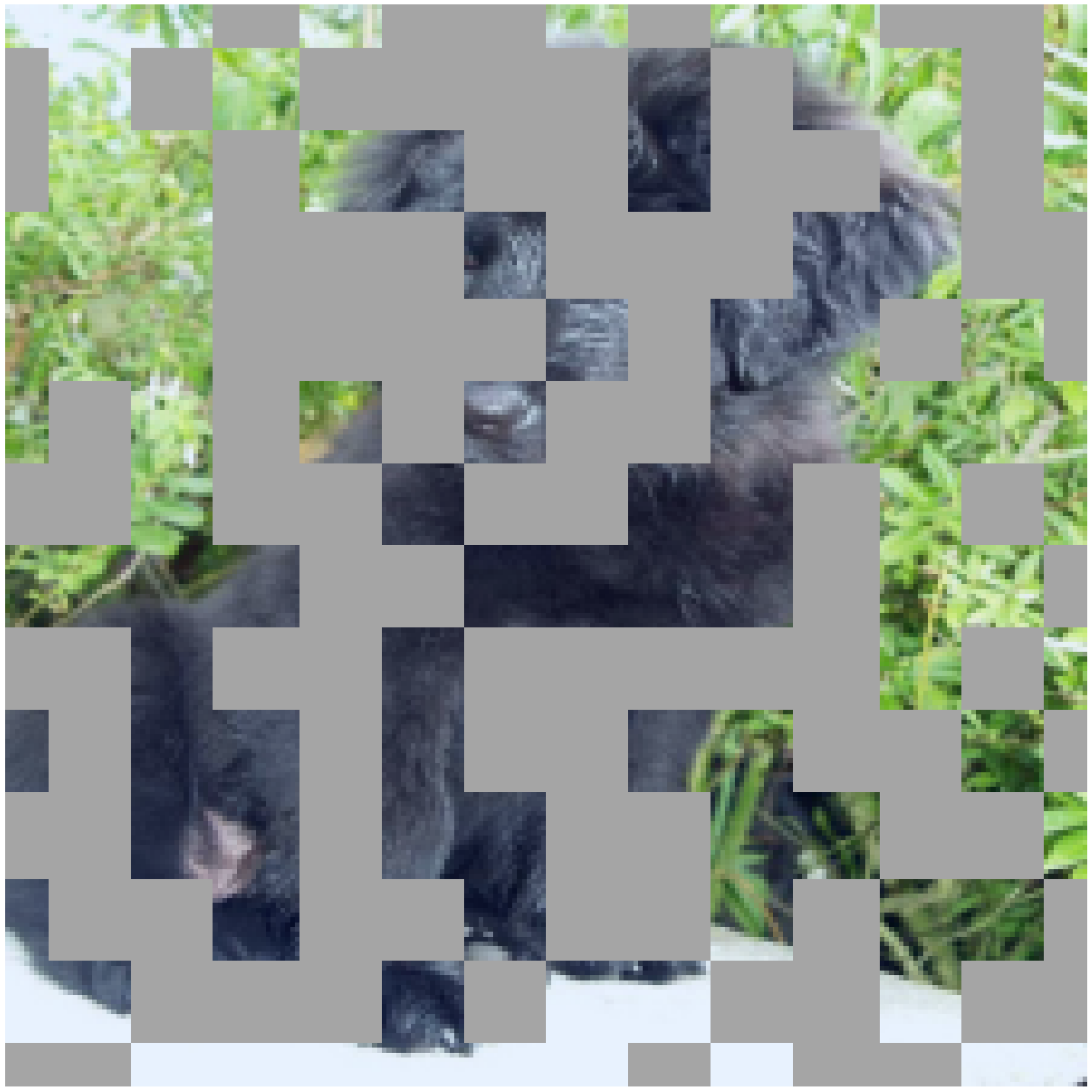}                          \\

	\raisebox{15pt}{\ours-High} &
	\fig[.120]{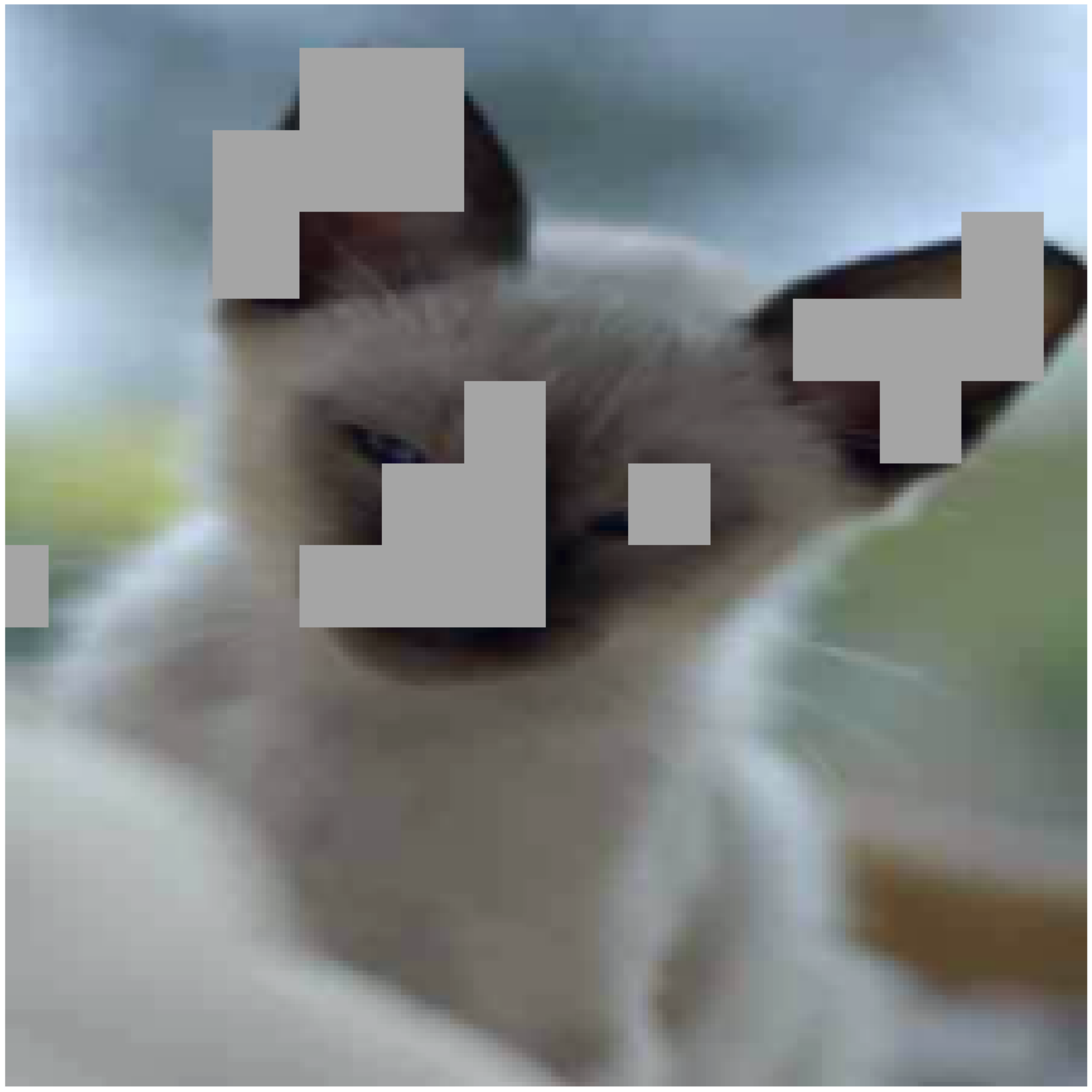}                          &
	\fig[.120]{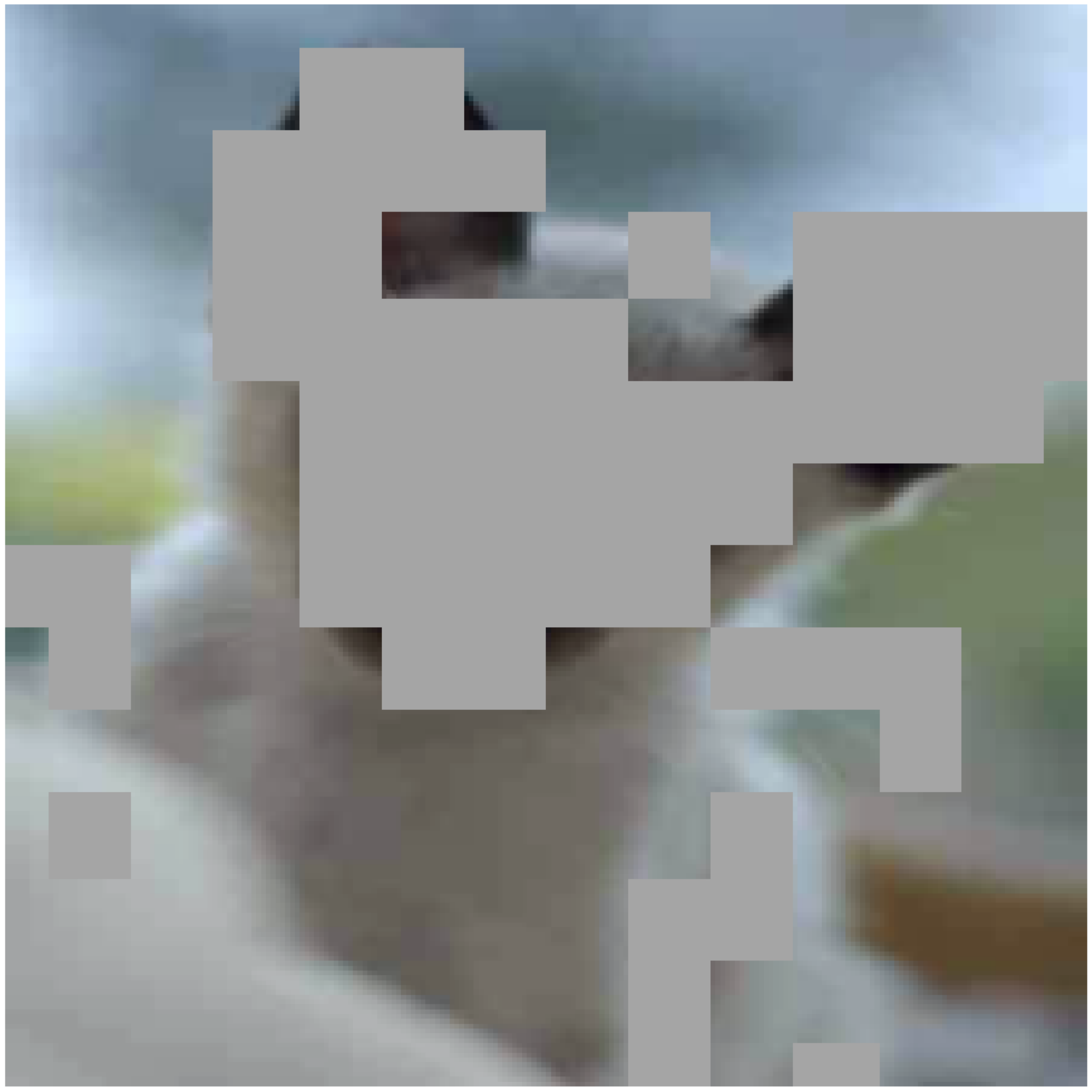}                          &
	\fig[.120]{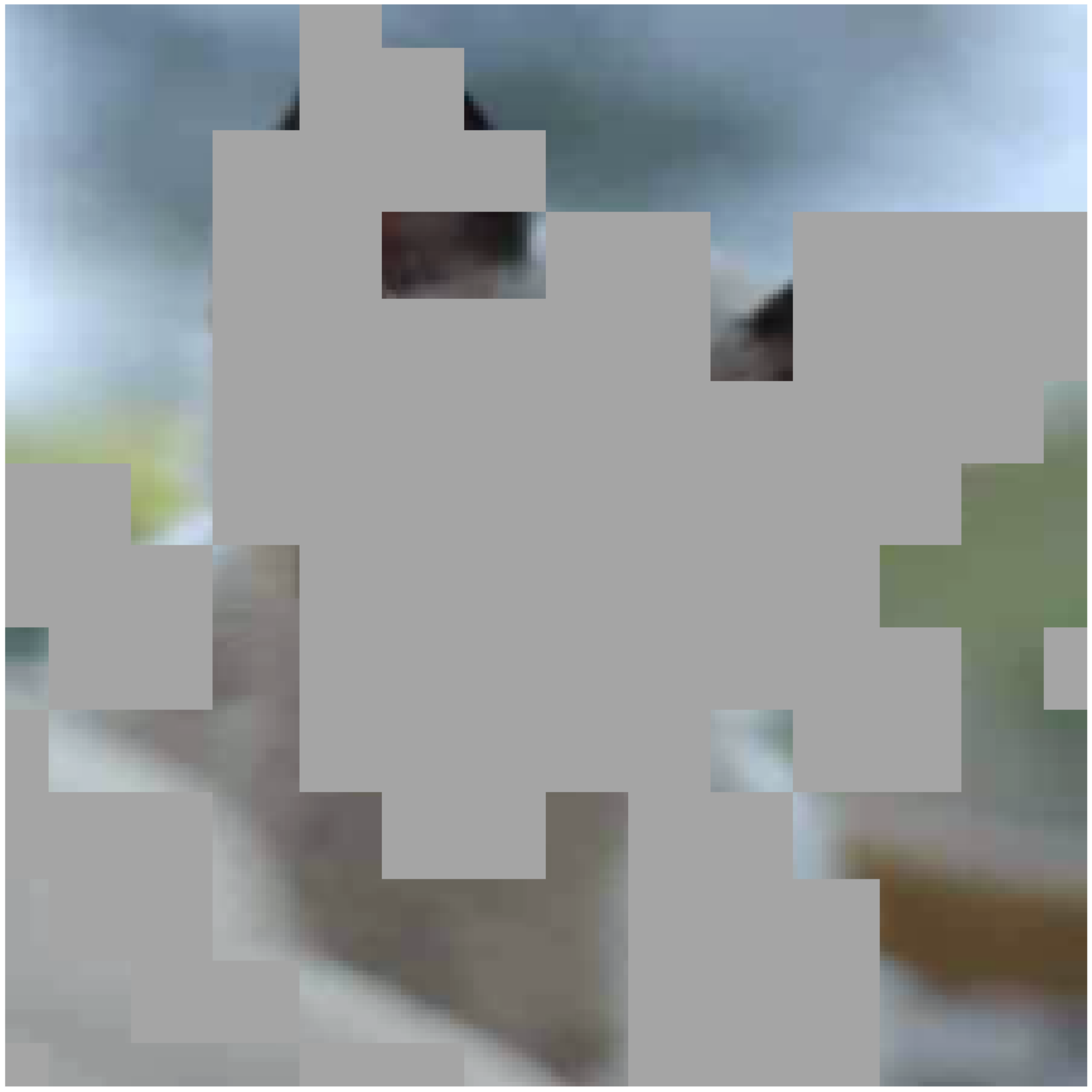}                          &

	& &

	\fig[.120]{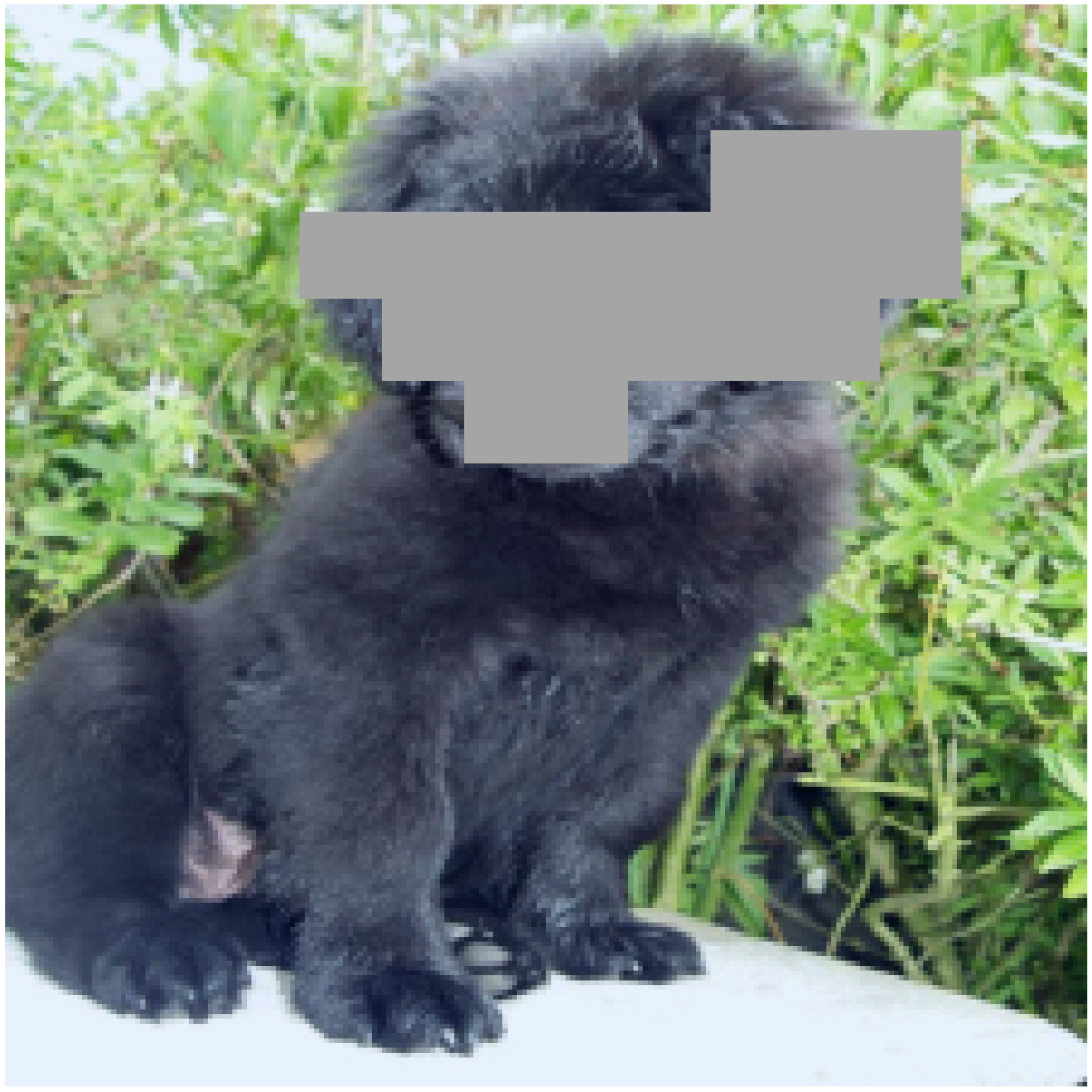}                          &
	\fig[.120]{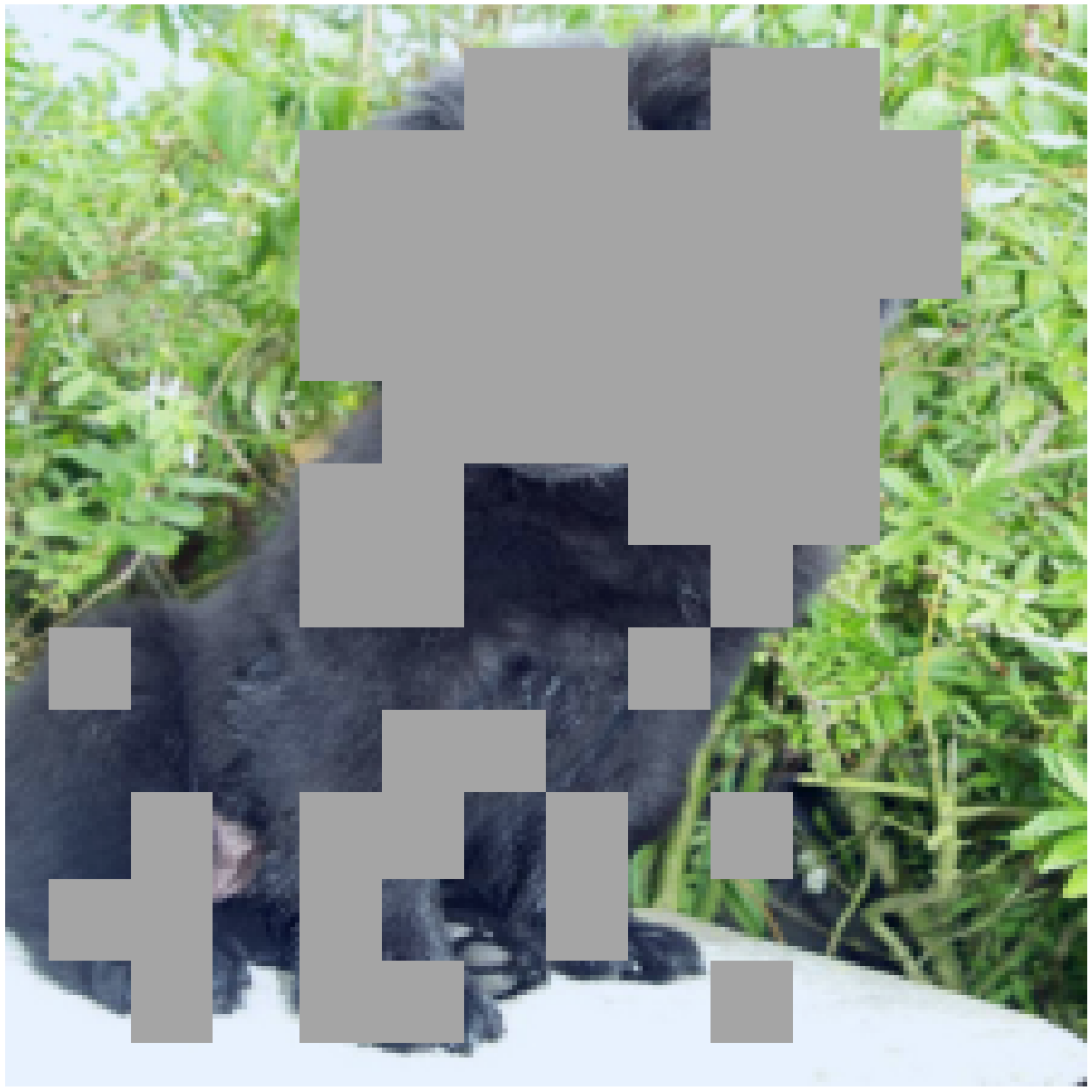}                          &
	\fig[.120]{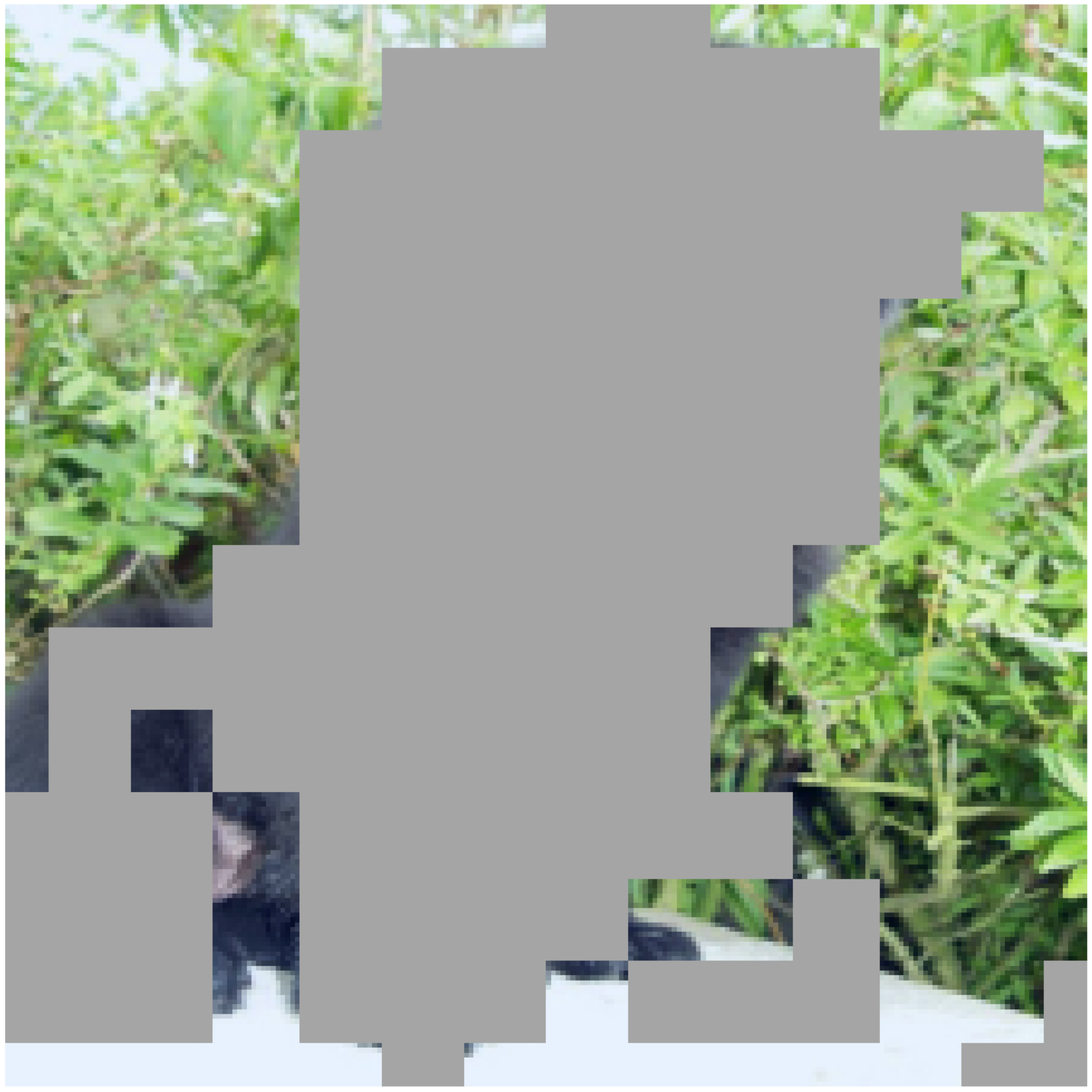}                          \\

	\raisebox{15pt}{\ours-Hint} &
	\fig[.120]{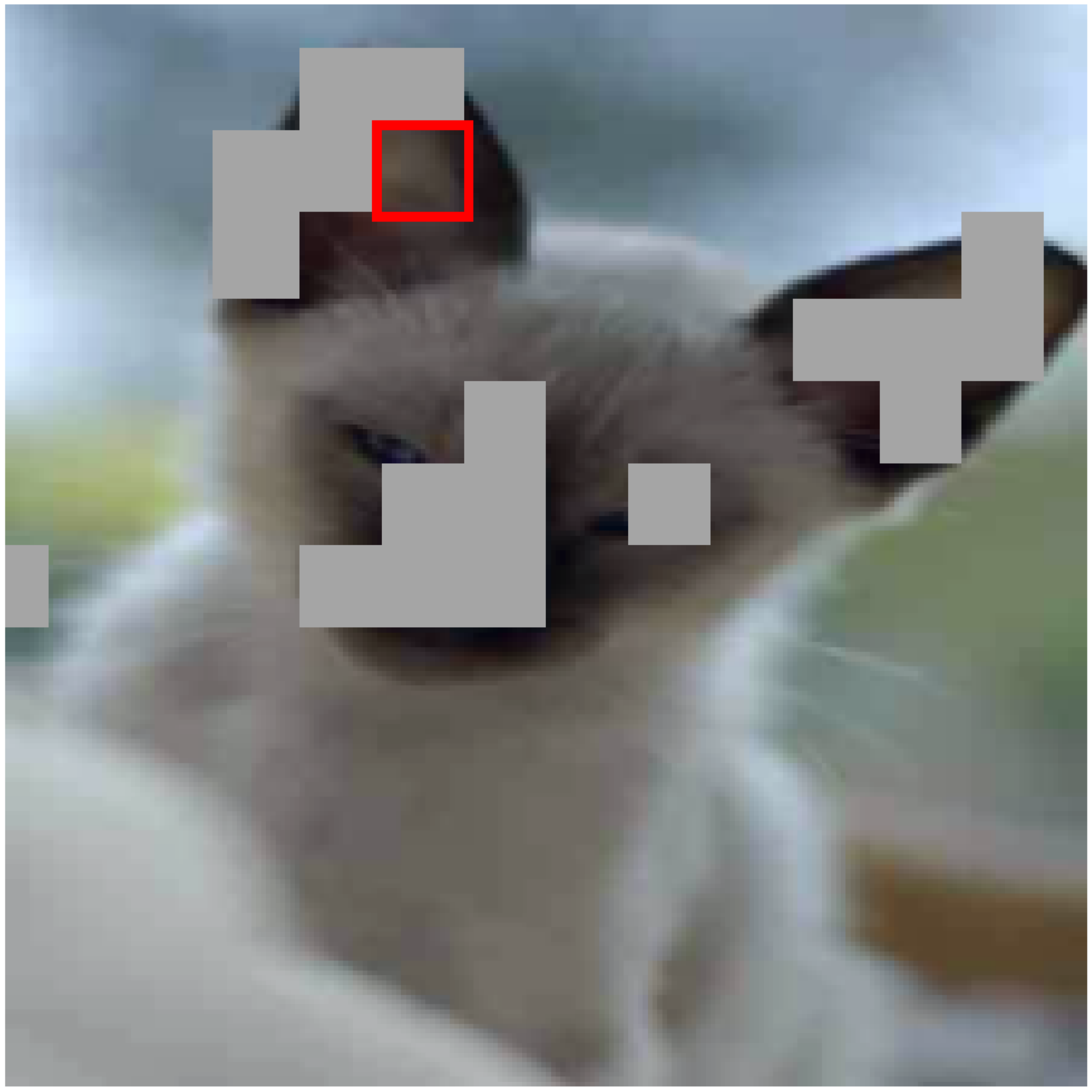}                          &
	\fig[.120]{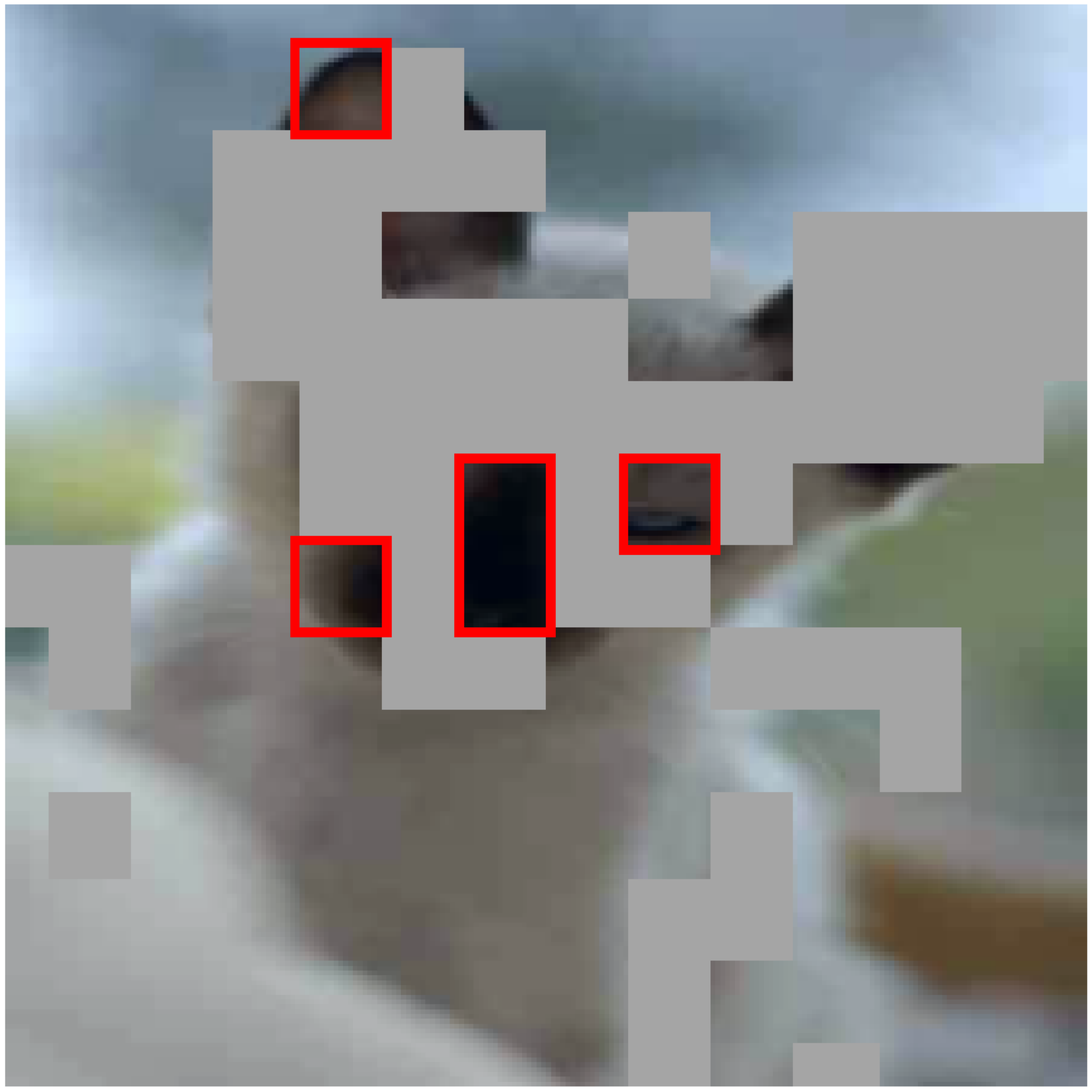}                          &
	\fig[.120]{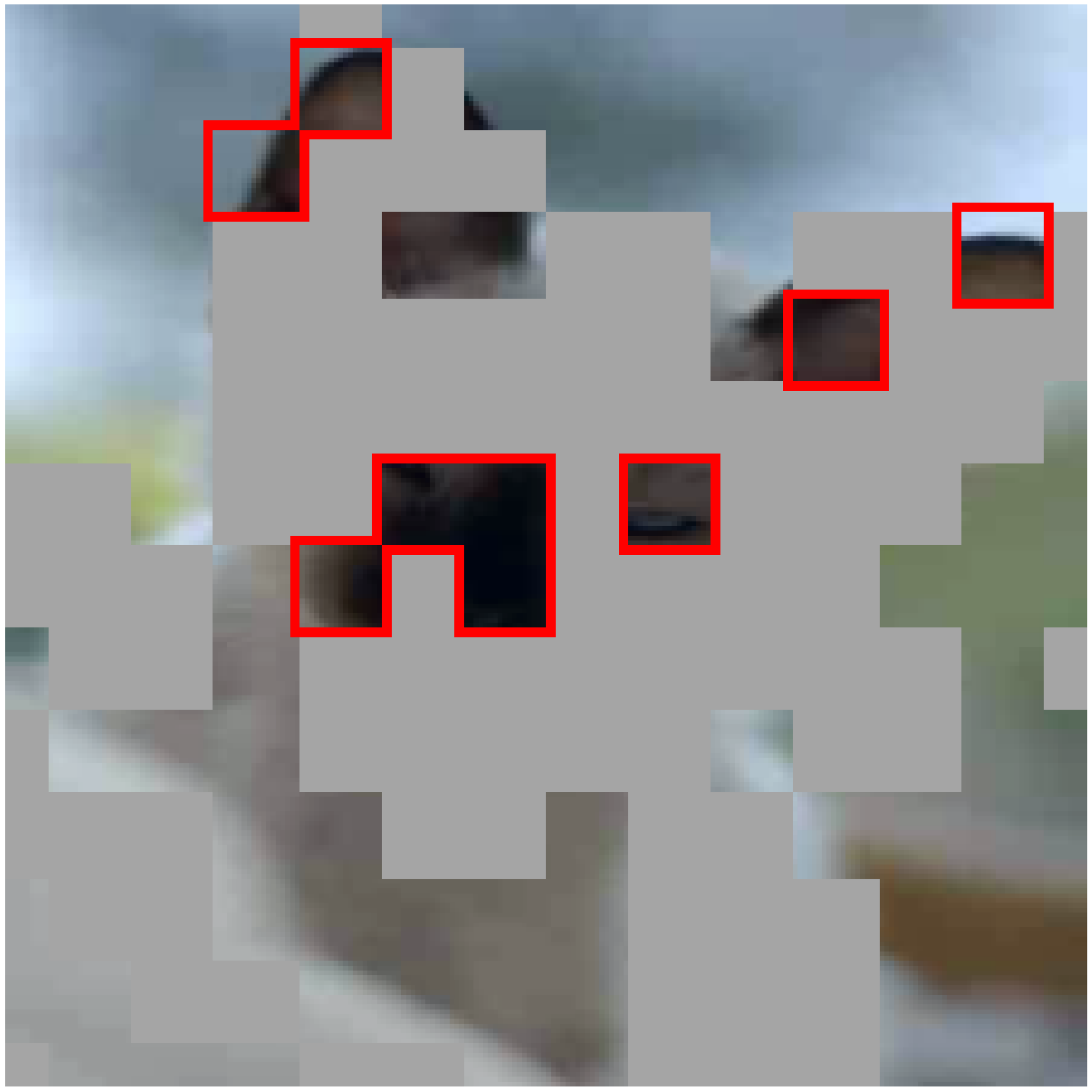}                          &

	&  &

	\fig[.120]{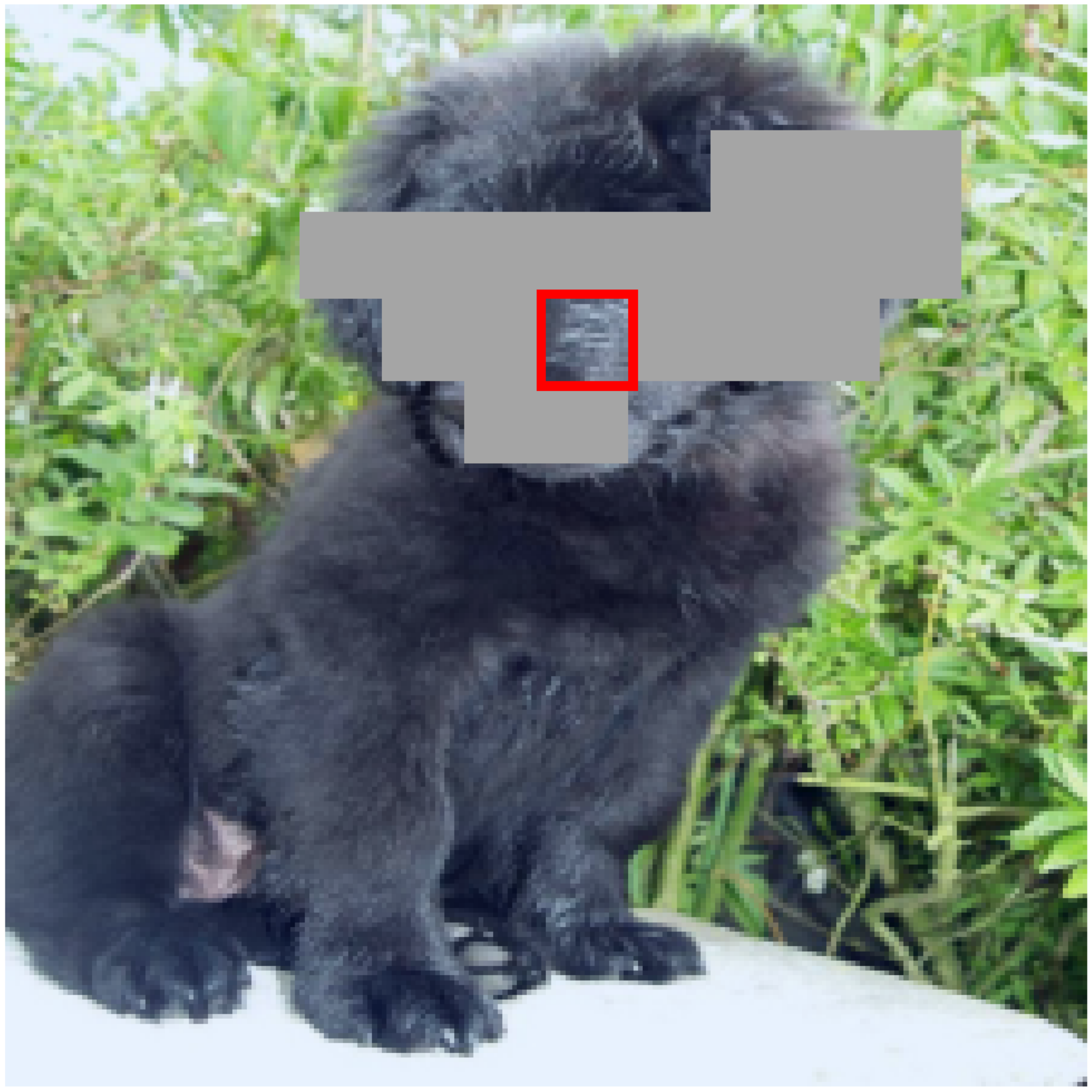}                          &
	\fig[.120]{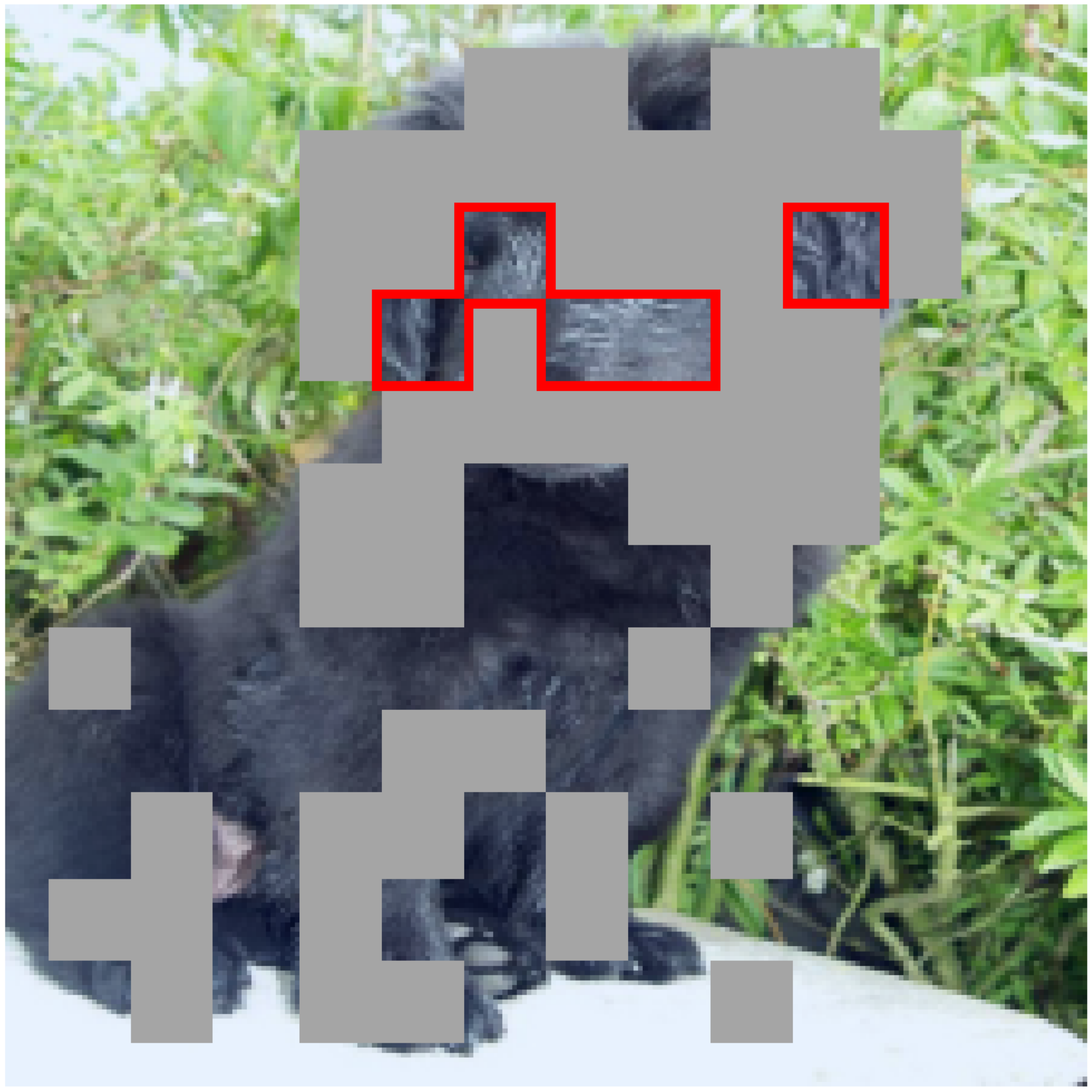}                          &
	\fig[.120]{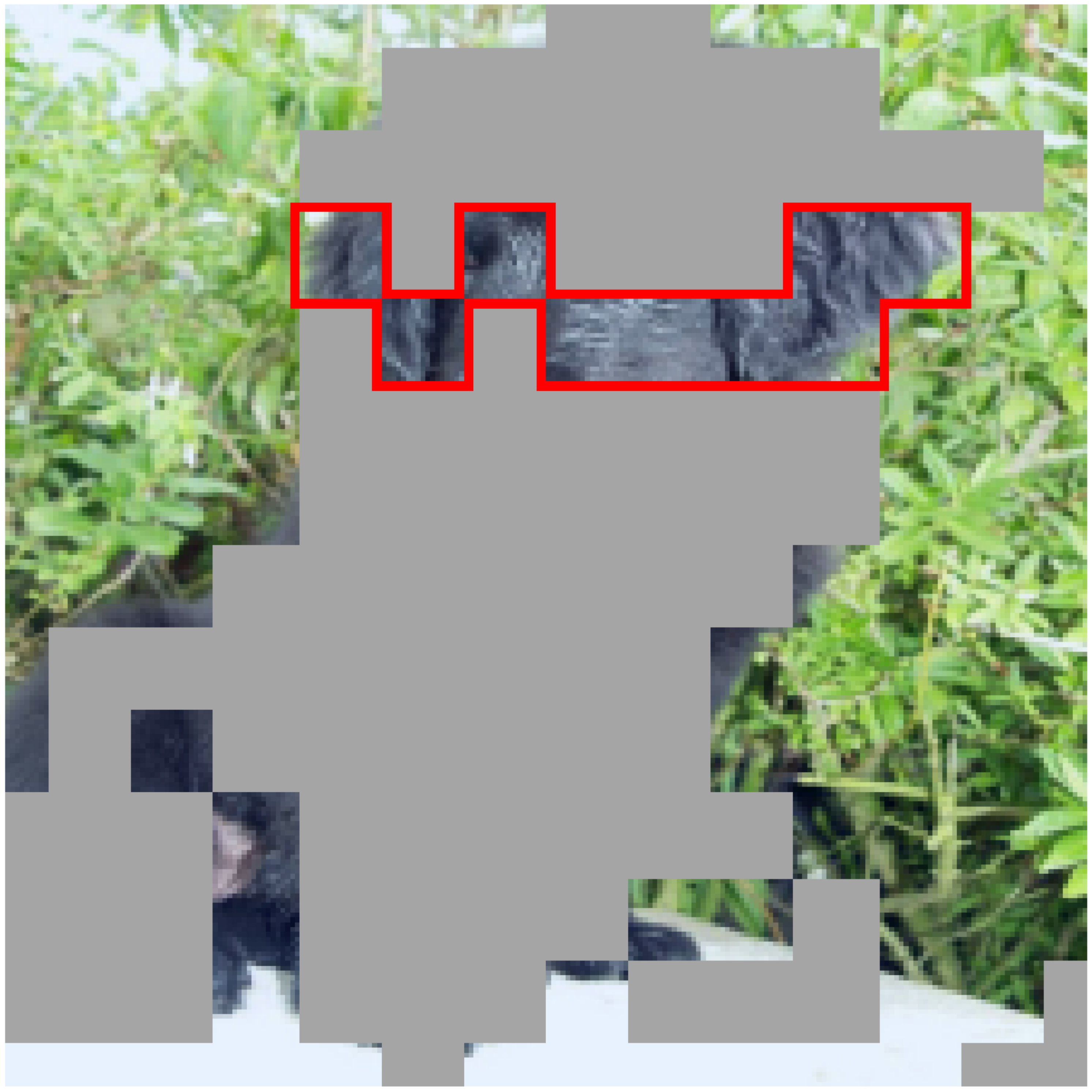}                          \\
\end{tabular}
\caption{Illustration of different masking strategies \vs mask ratio $r$ (\%) (part 1). We compare random Block-Wise masking (BEiT~\cite{beit}) with Random masking (SimMIM~\cite{simmim}), \ours-High and \ours-Hint. Our \ours-High uses the attention map arising in the encoder to hide patches, while \ours-Hint reveals very salient patches to leave hints about the identity of the masked object.}
\label{fig:MoreStrategies2}
\end{figure}

\begin{figure}[t]
\centering
\tiny
\centering
\setlength{\tabcolsep}{1.5pt}
\begin{tabular}{ccccp{0.15cm}|p{0.15cm}ccc}
    &
    10\%               &
	30\%               &
	50\%               &

    & &

    10\%               &
	30\%               &
	50\%               \\

	\raisebox{15pt}{Block-Wise}  &
	\fig[.120]{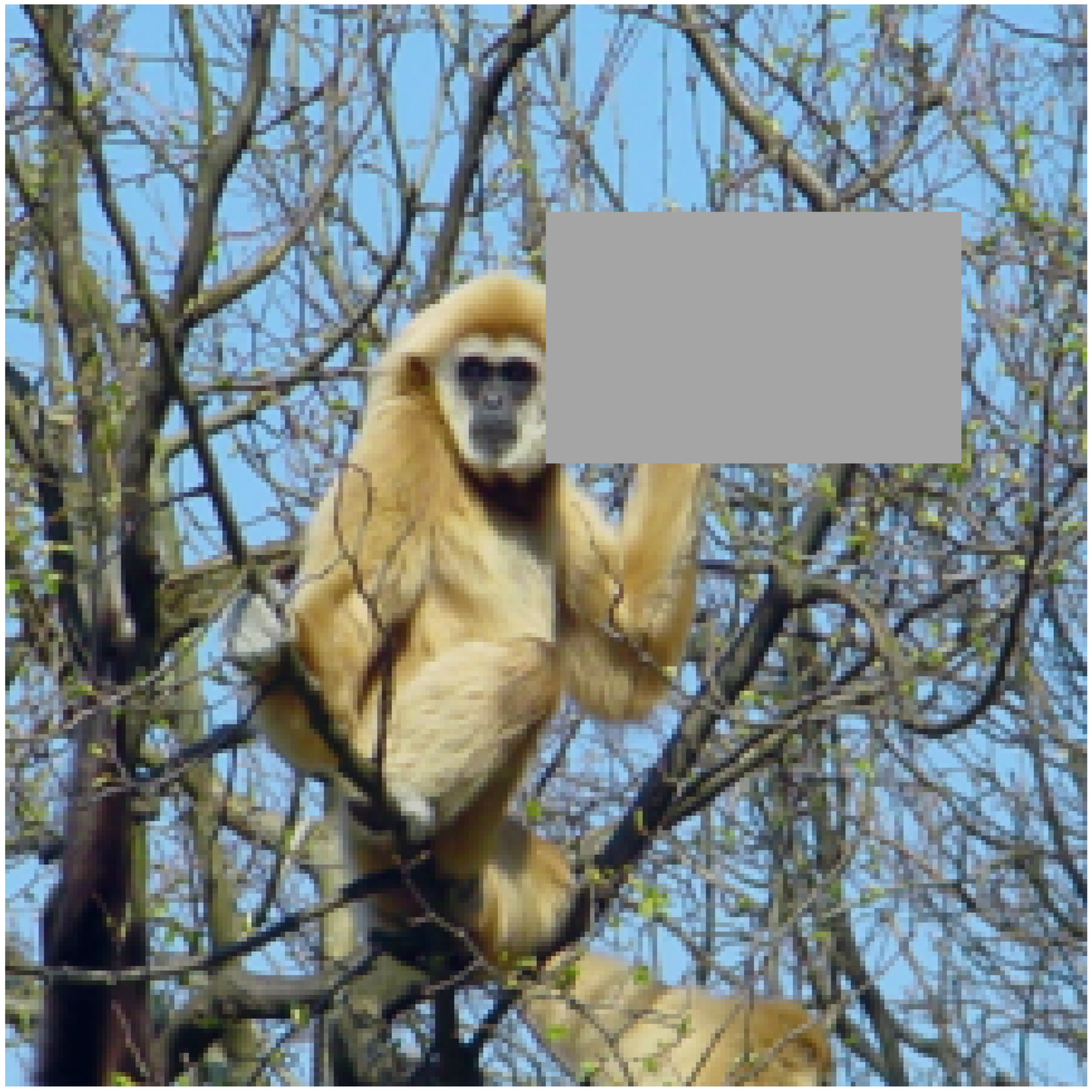}               &
	\fig[.120]{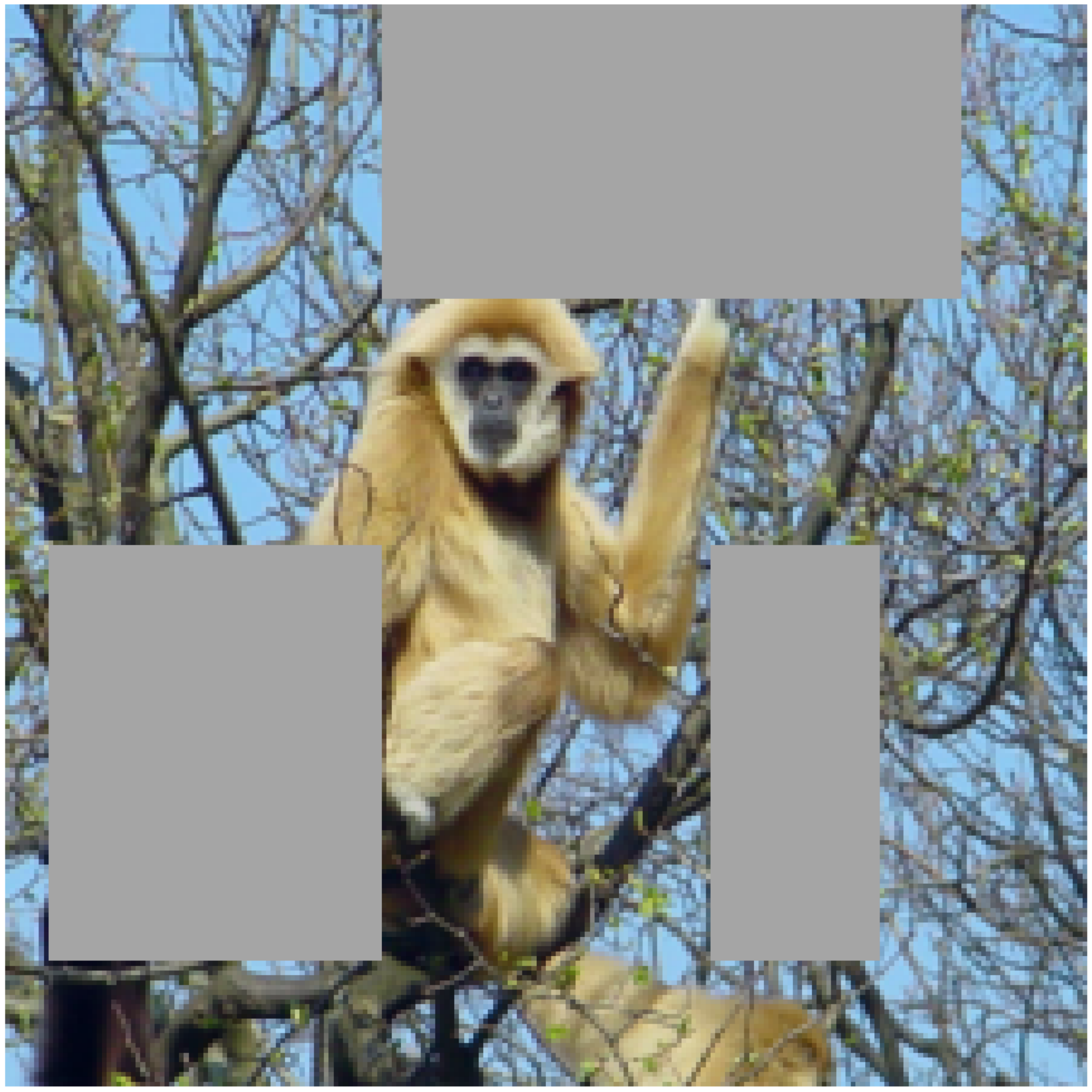}               &
	\fig[.120]{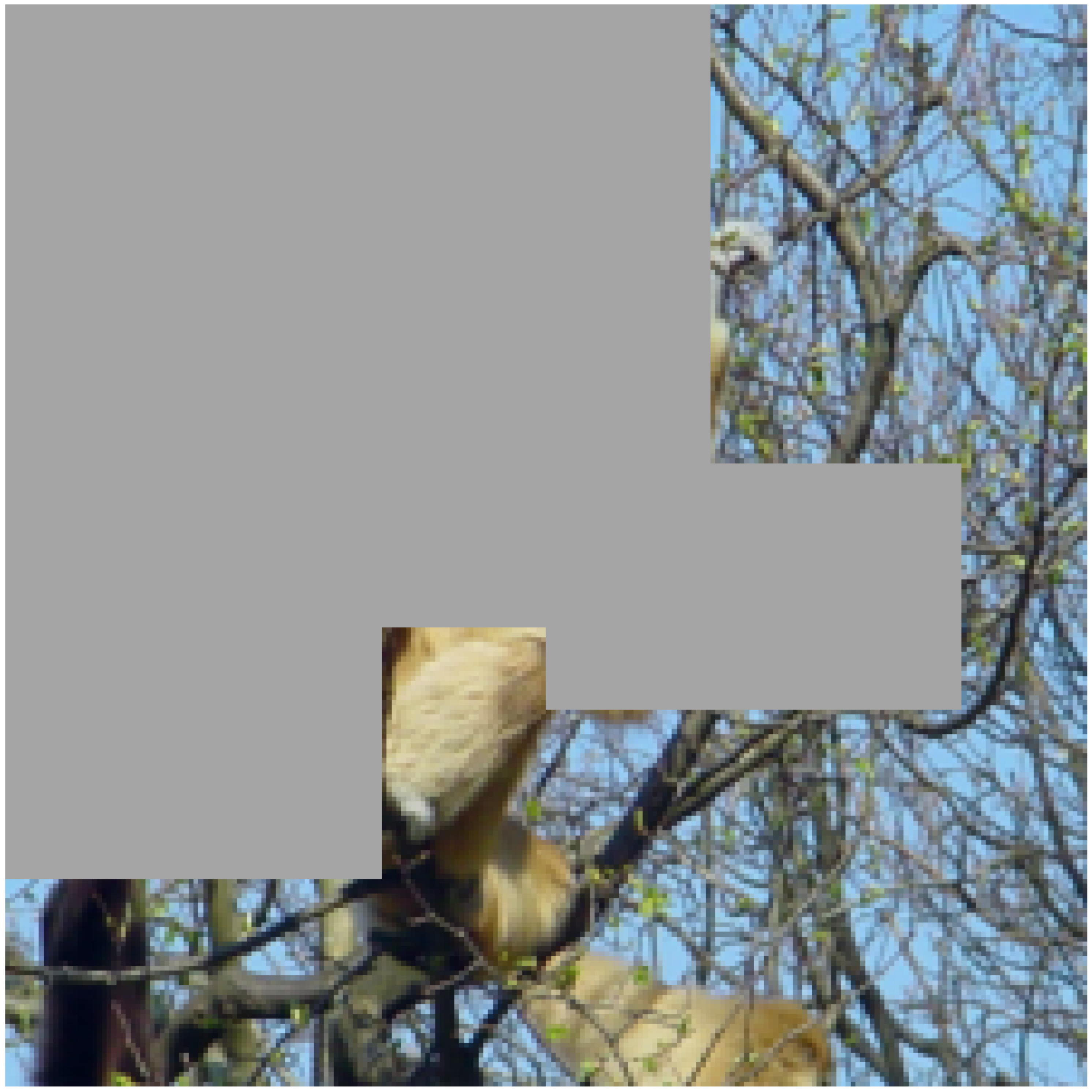}               &

    & &

	\fig[.120]{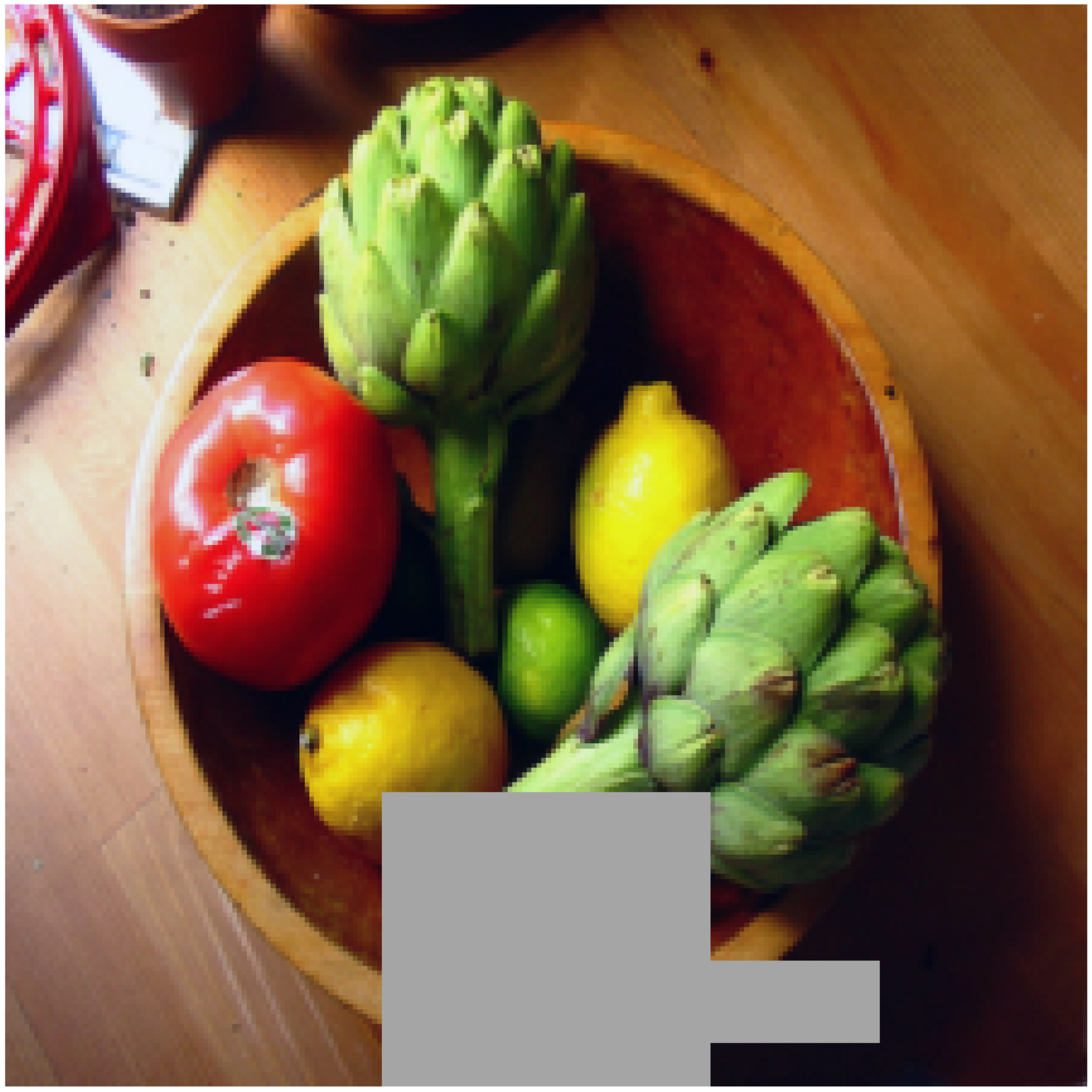}               &
	\fig[.120]{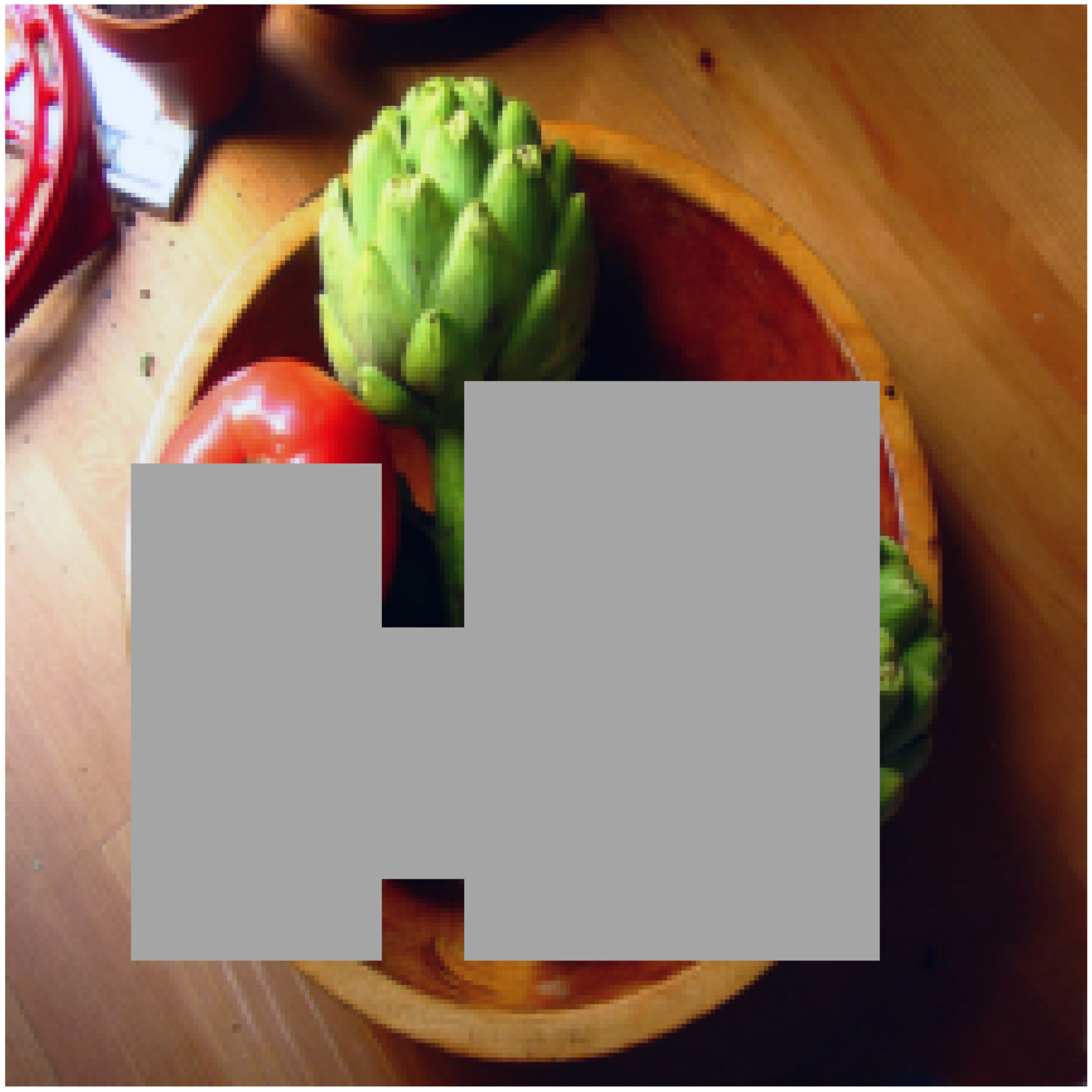}               &
	\fig[.120]{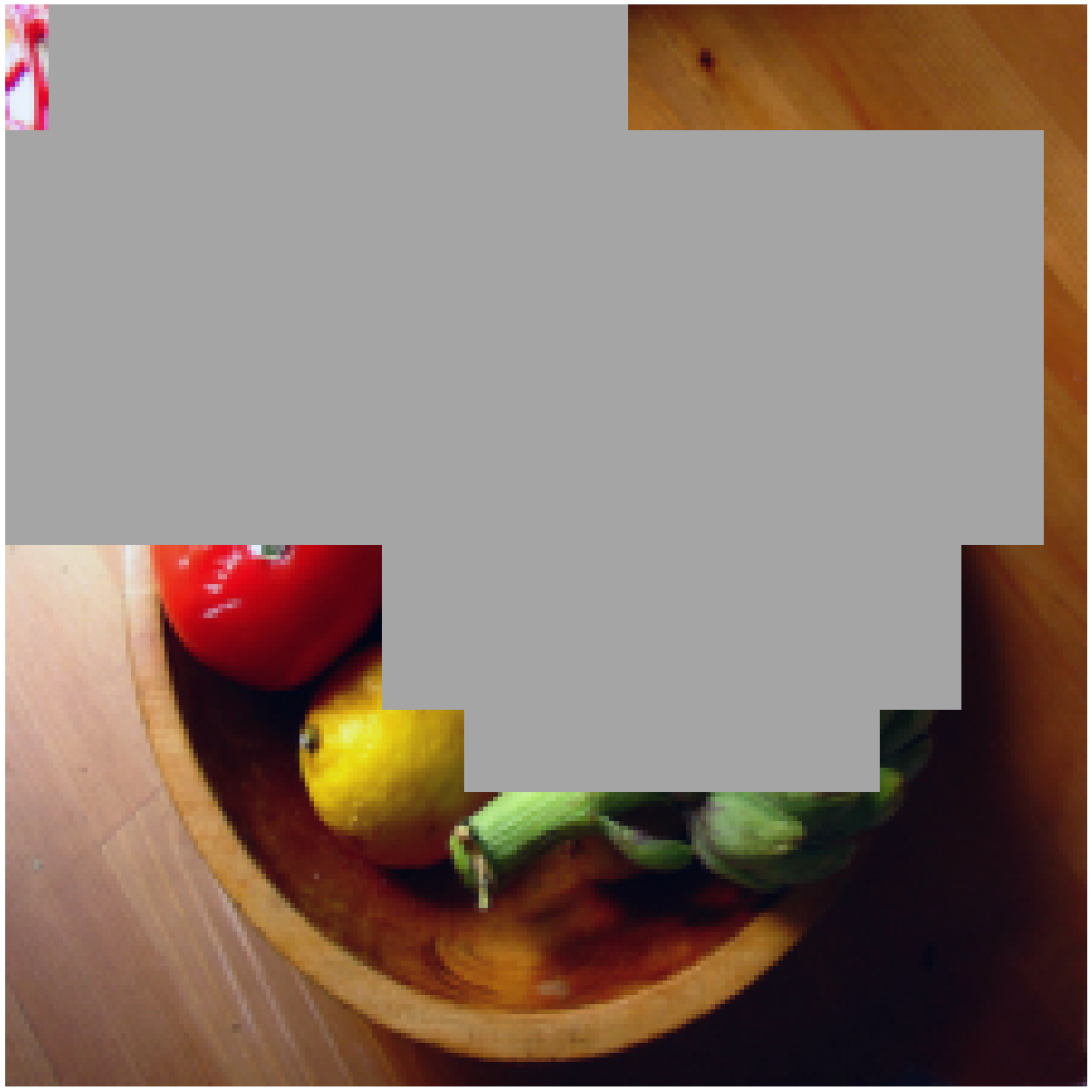}               \\

	\raisebox{15pt}{Random} &
	\fig[.120]{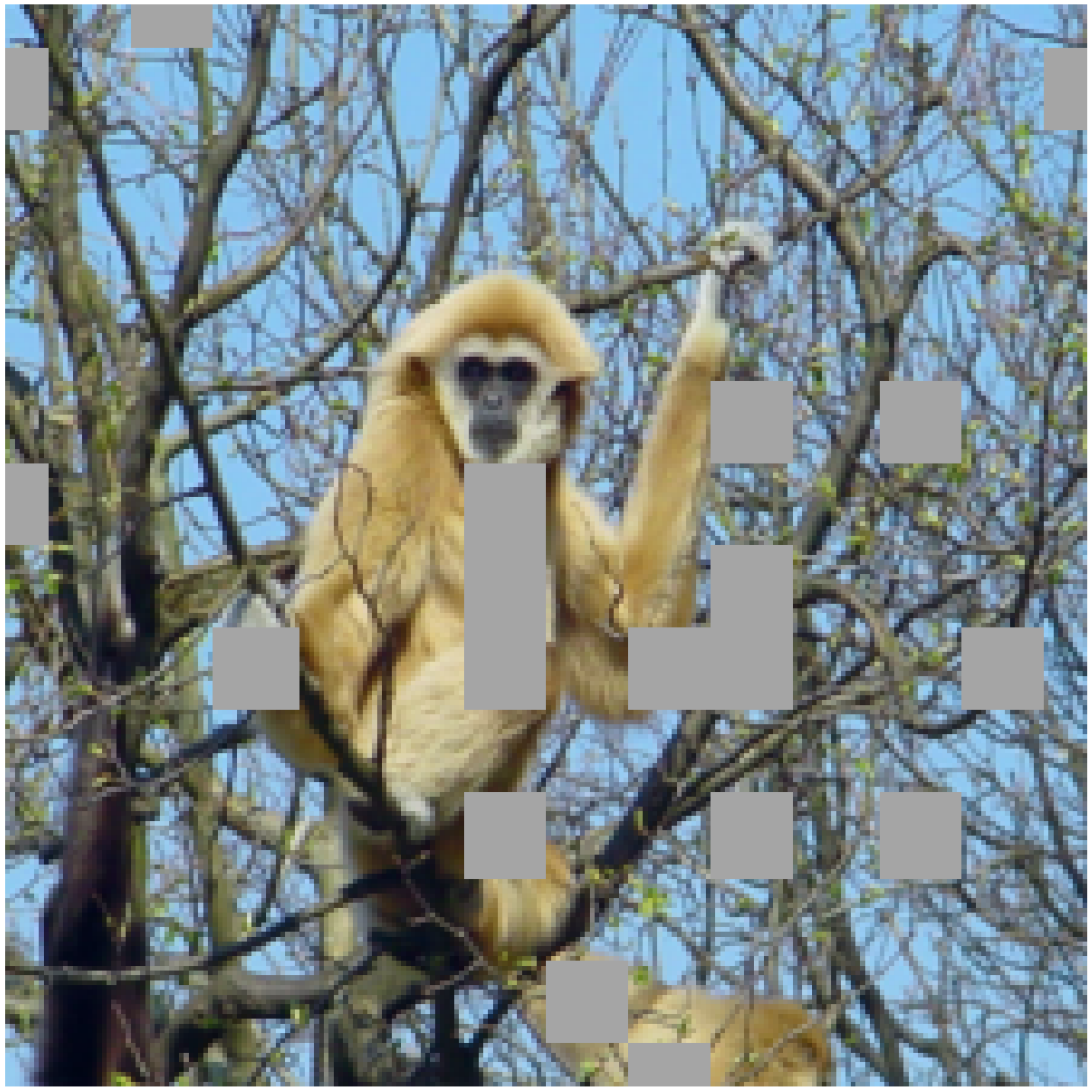}                          &
	\fig[.120]{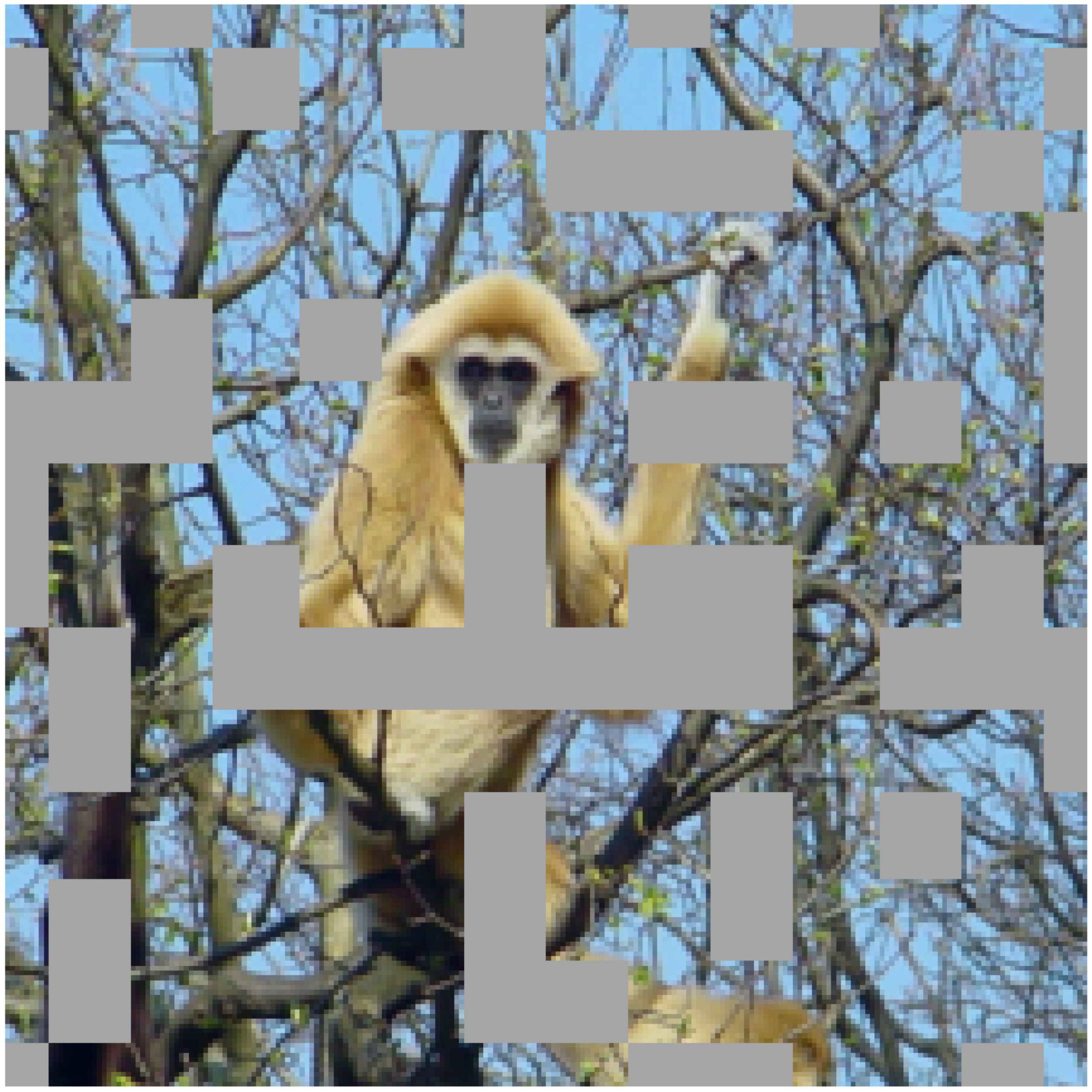}                          &
	\fig[.120]{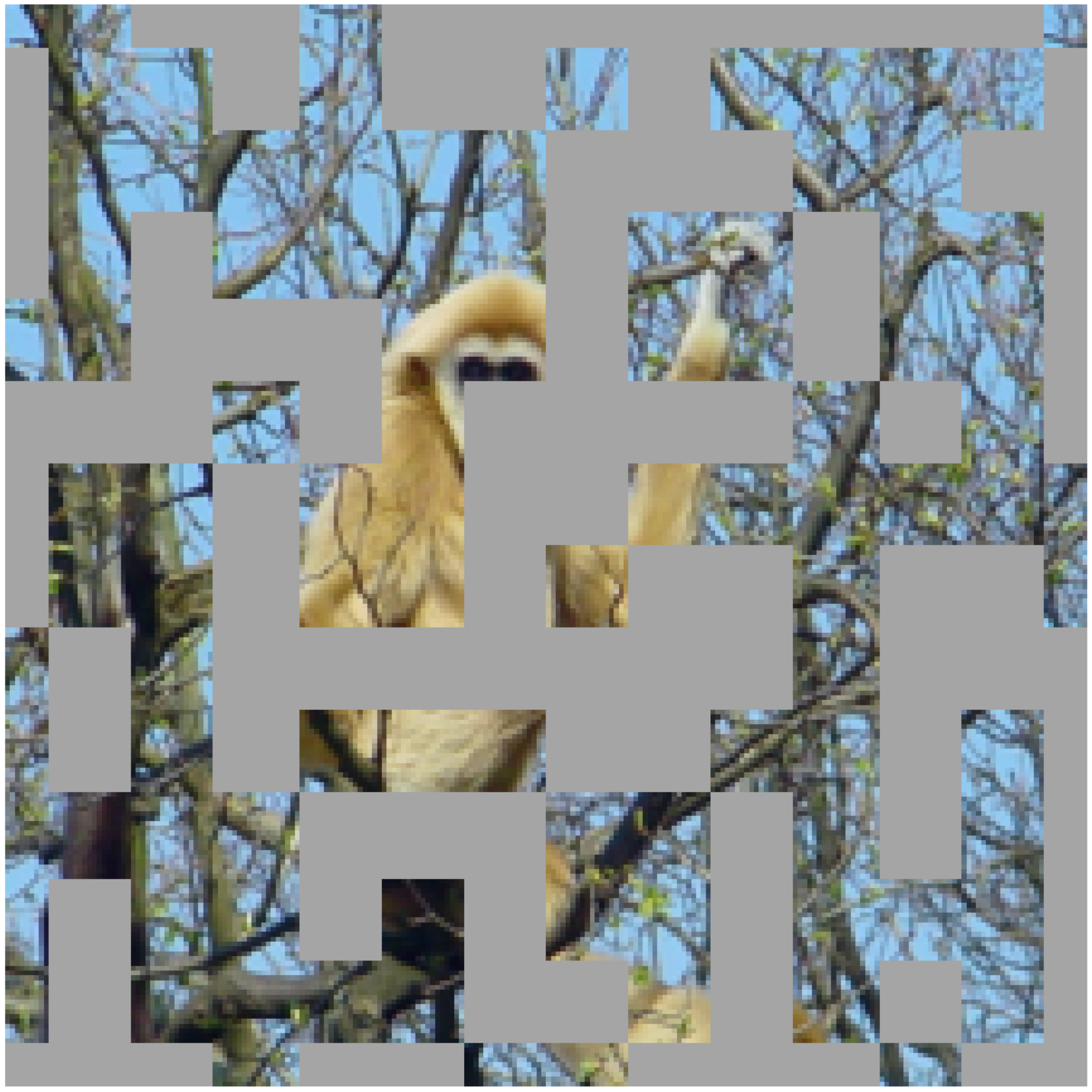}                          &

	& &

	\fig[.120]{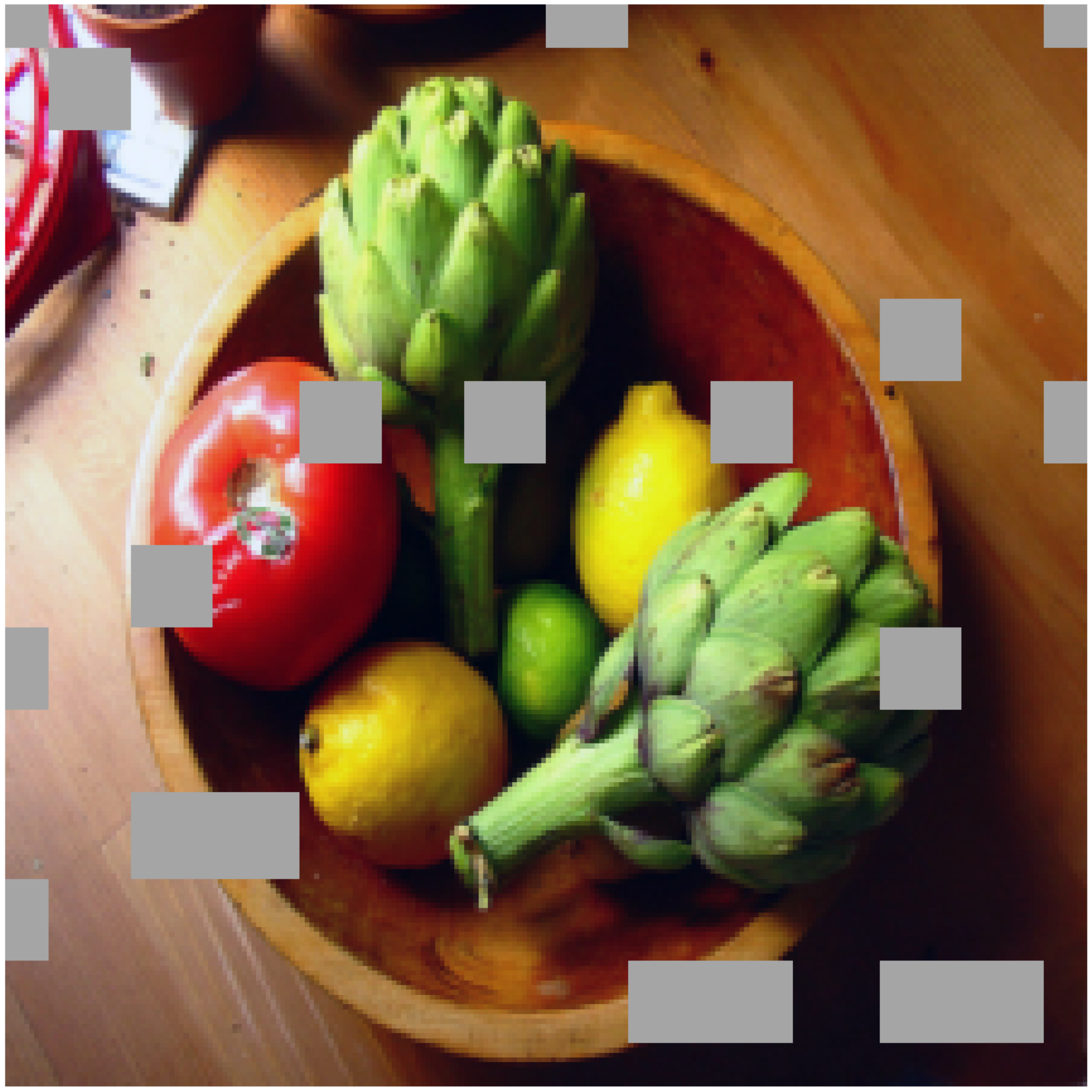}                          &
	\fig[.120]{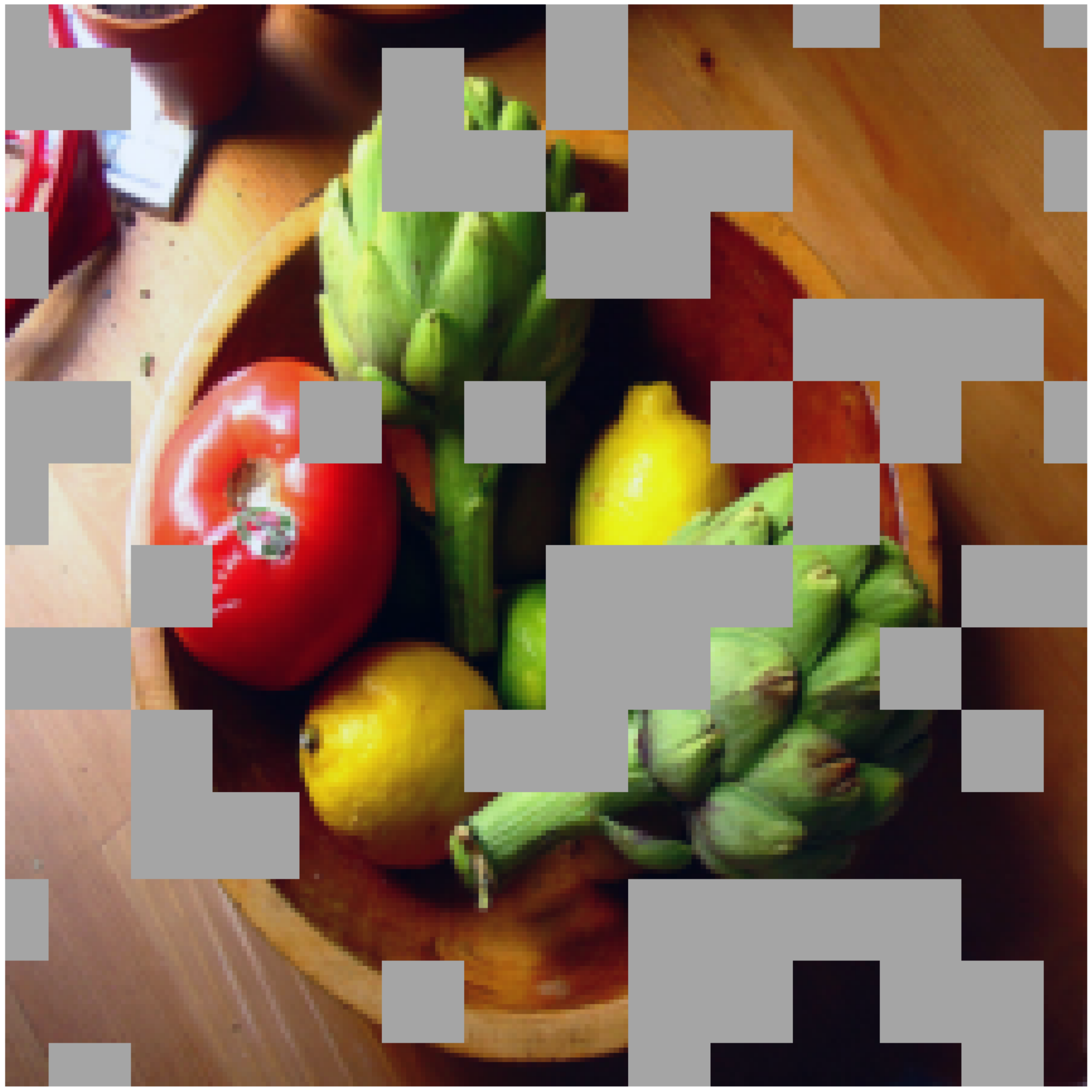}                          &
	\fig[.120]{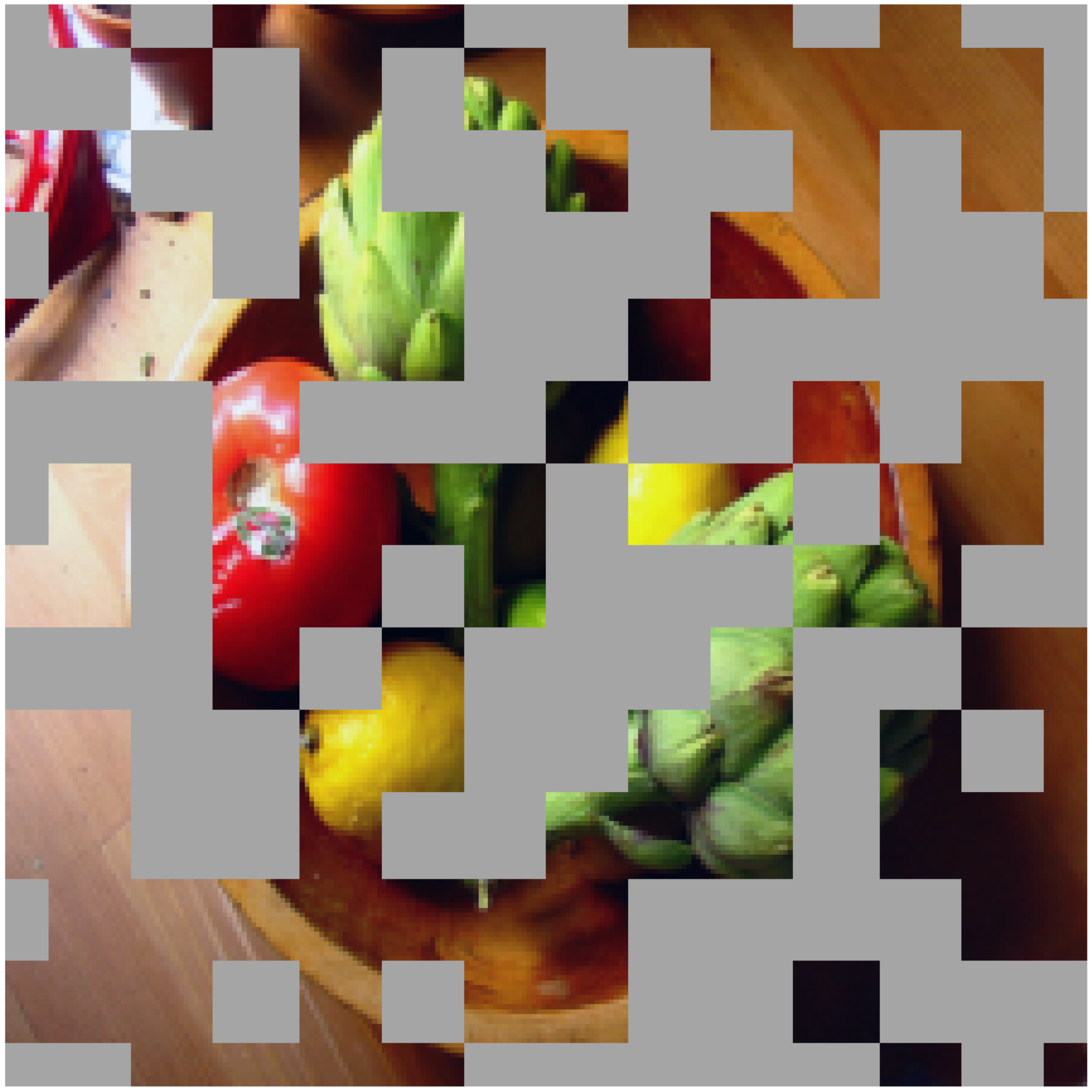}                          \\

	\raisebox{15pt}{\ours-High} &
	\fig[.120]{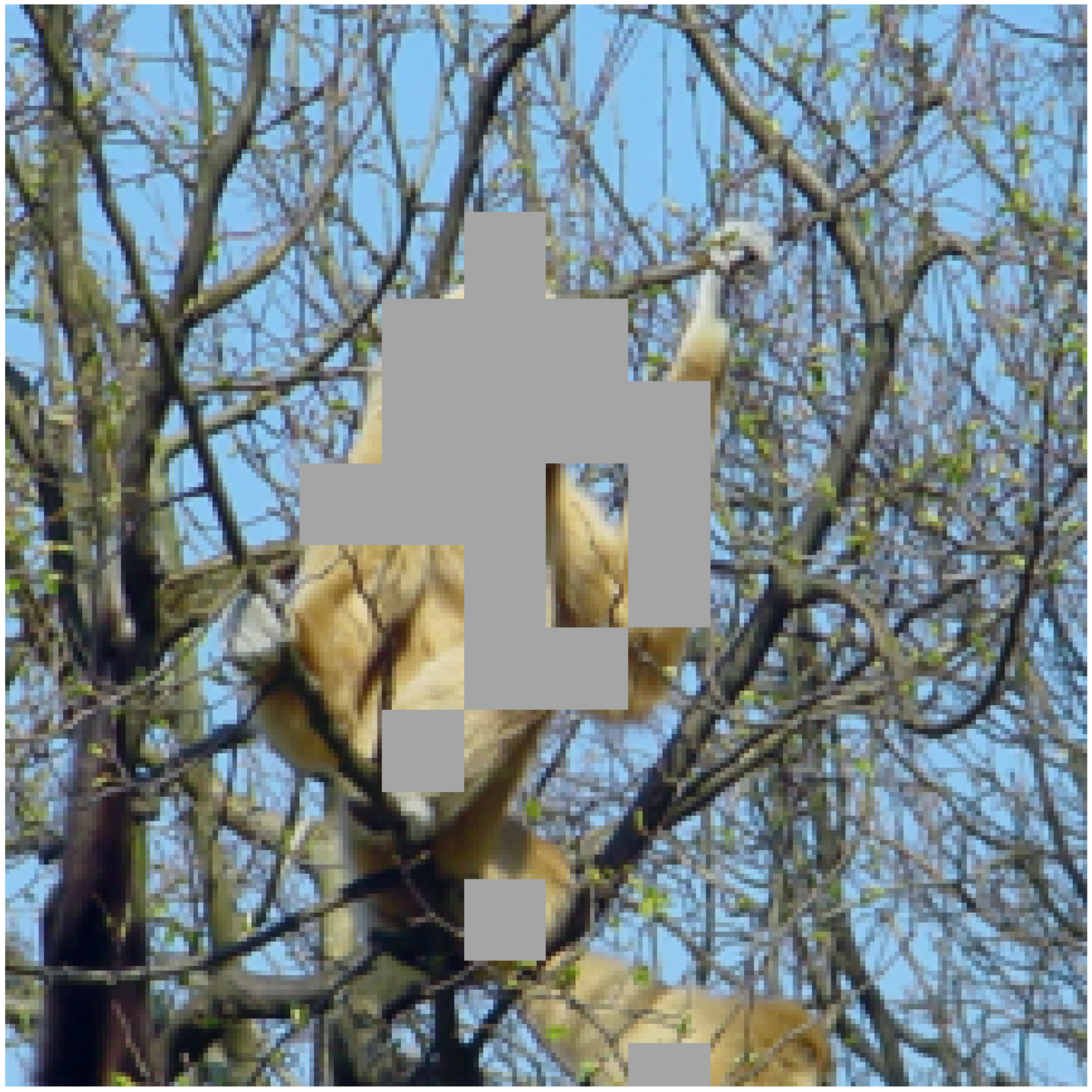}                          &
	\fig[.120]{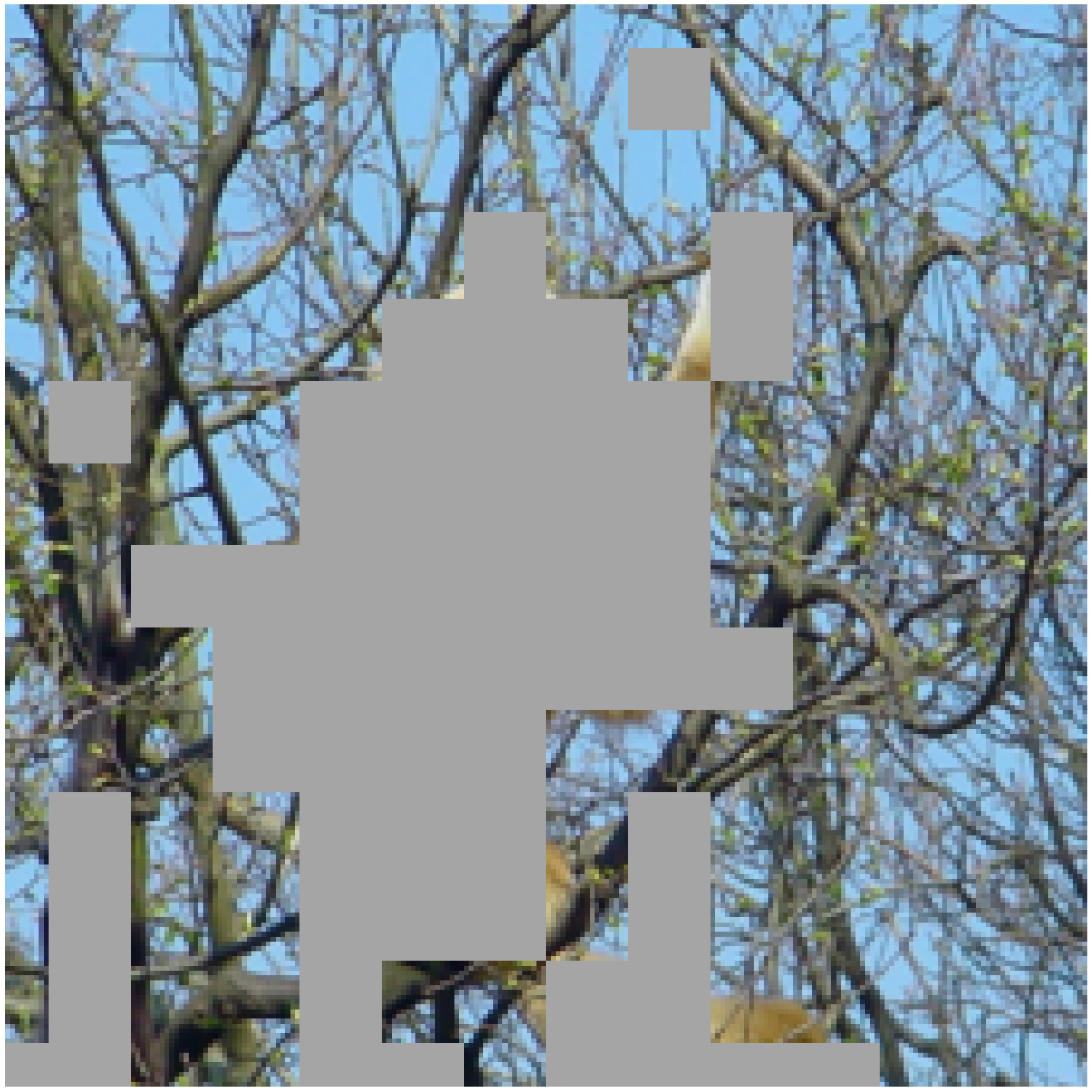}                          &
	\fig[.120]{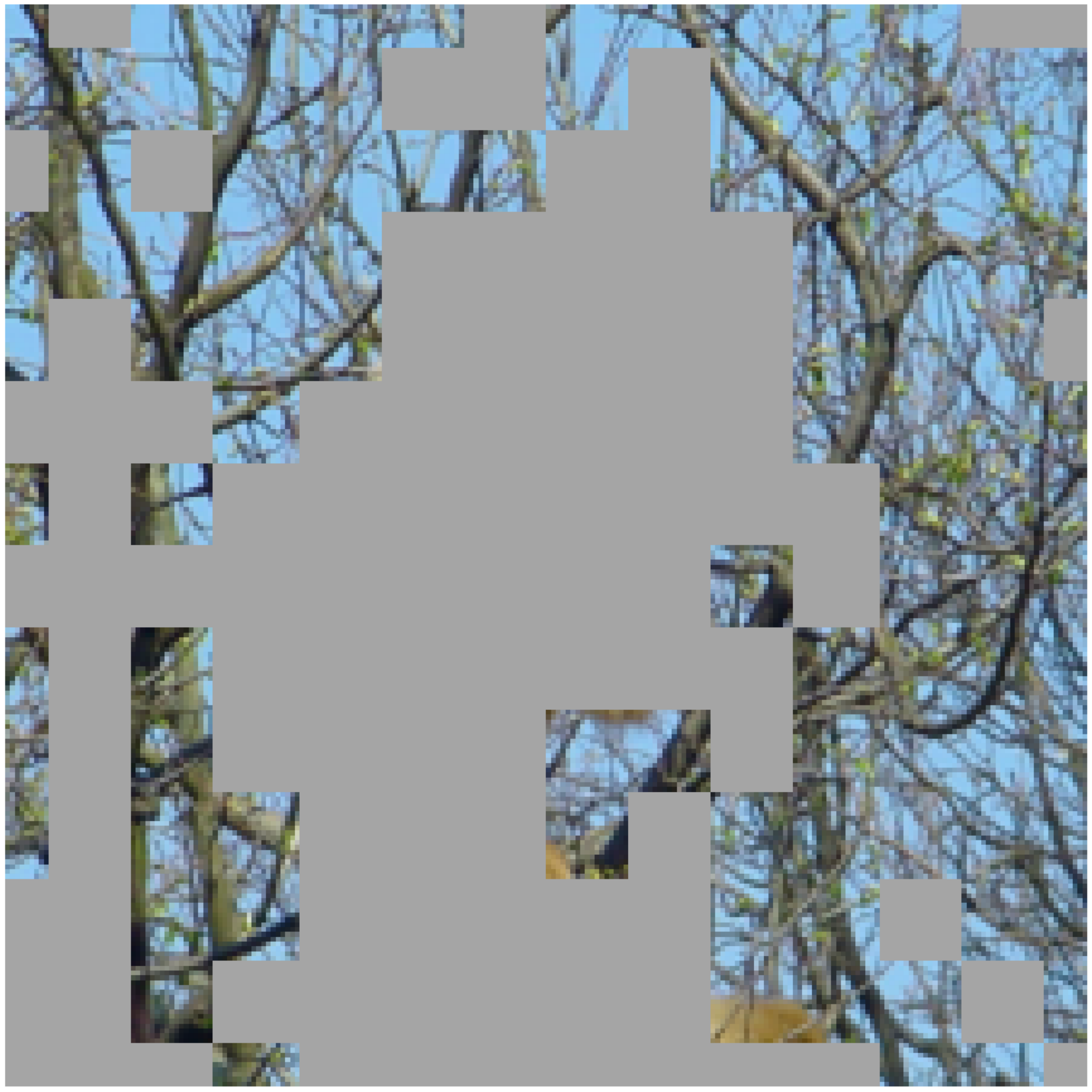}                          &

	& &

	\fig[.120]{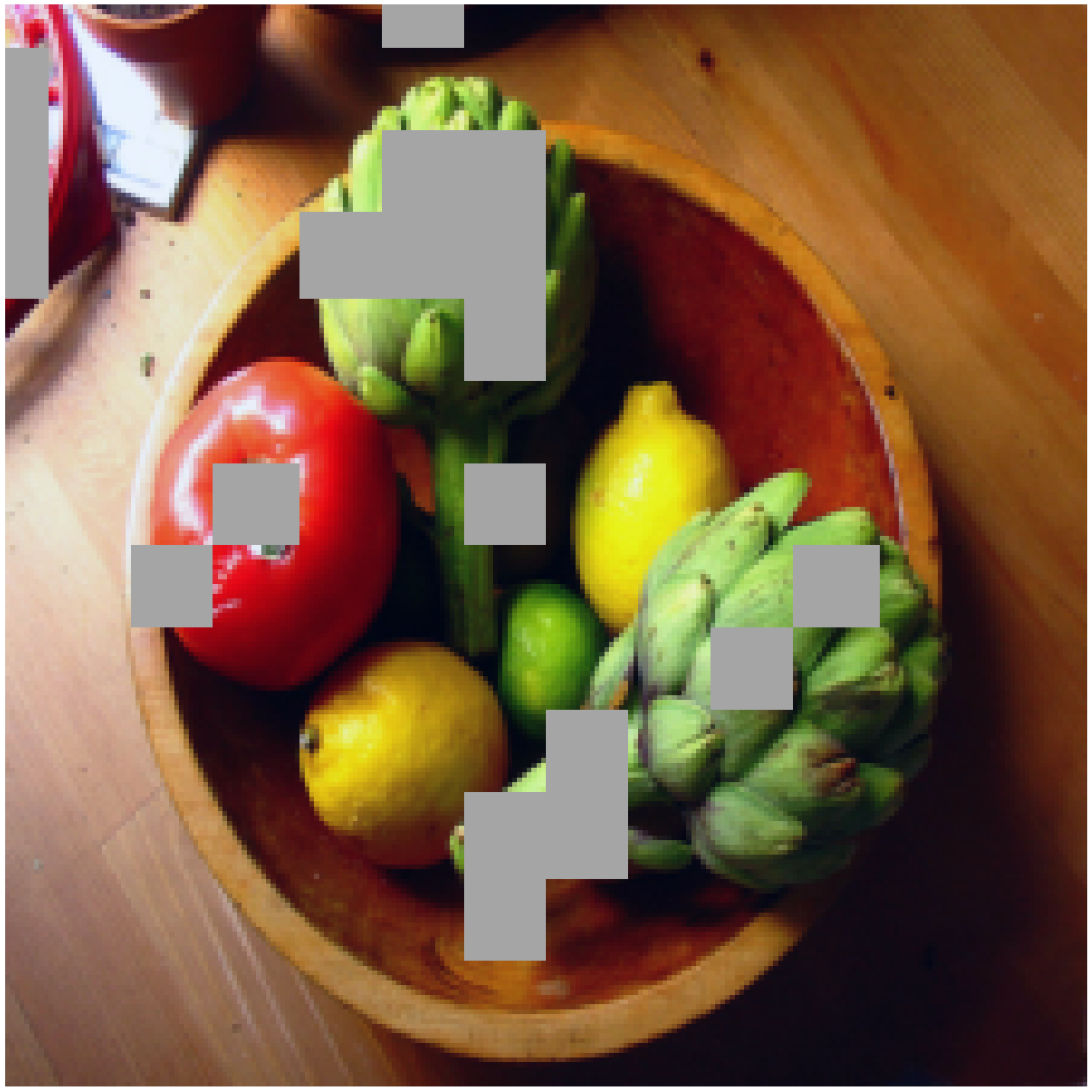}                          &
	\fig[.120]{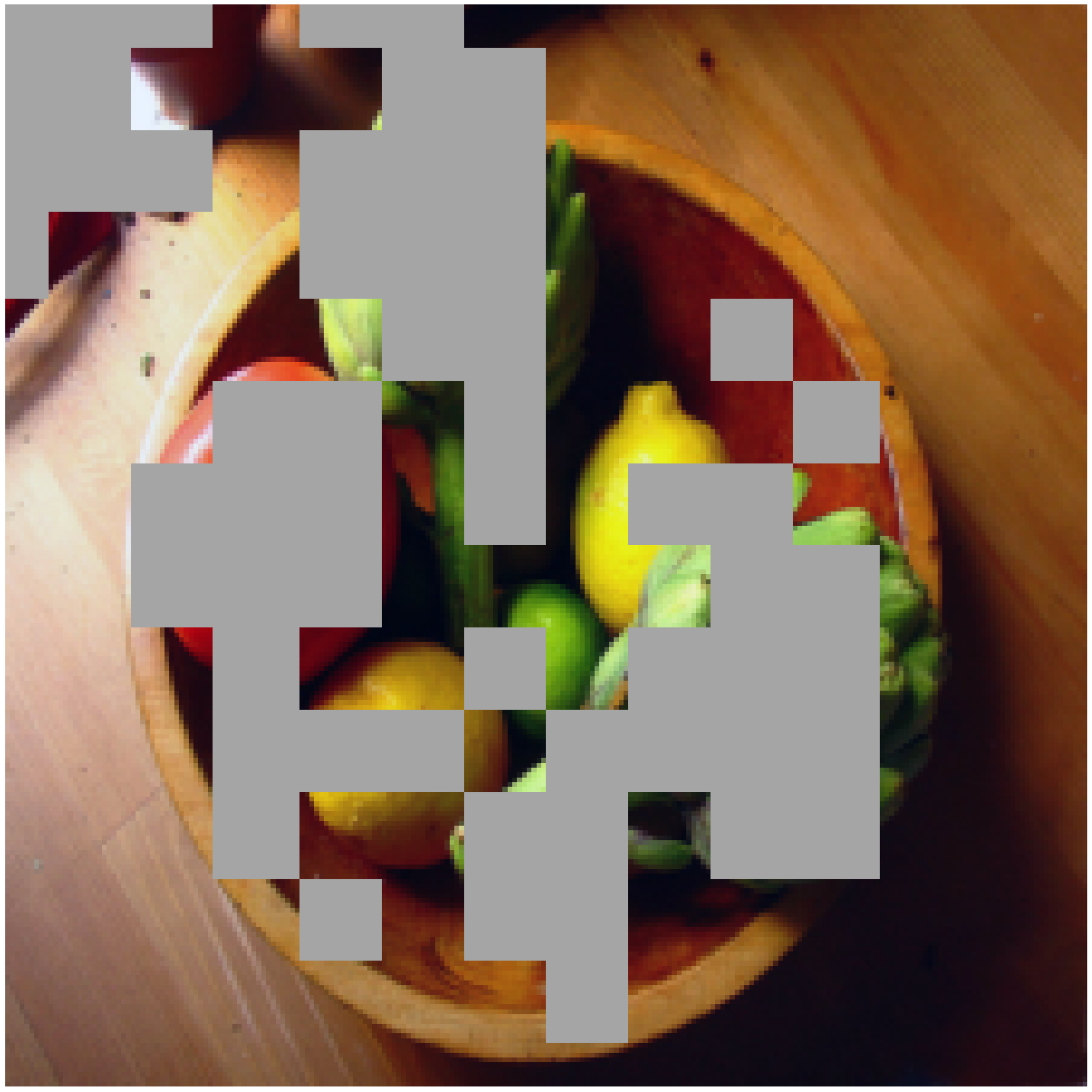}                          &
	\fig[.120]{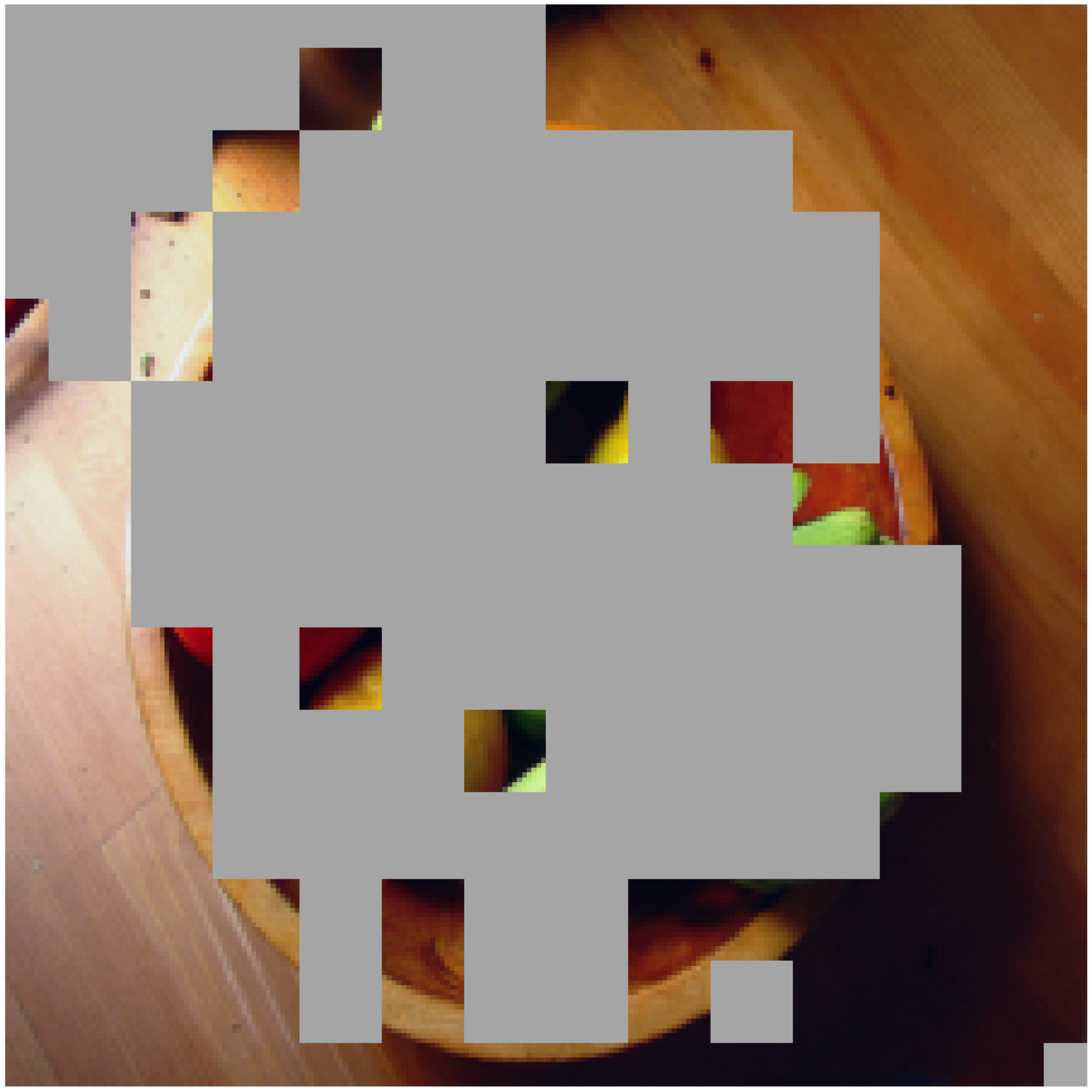}                          \\

	\raisebox{15pt}{\ours-Hint} &
	\fig[.120]{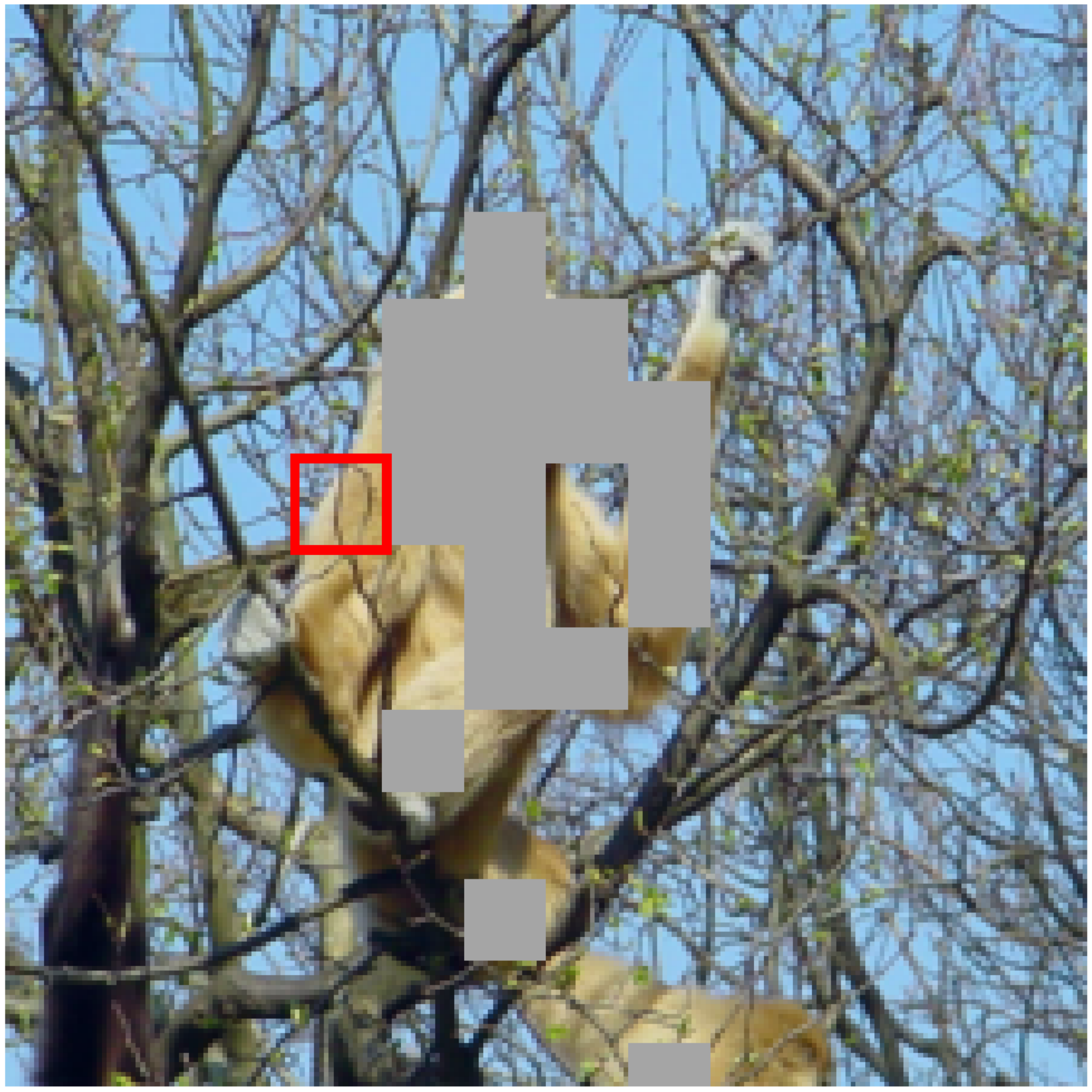}                          &
	\fig[.120]{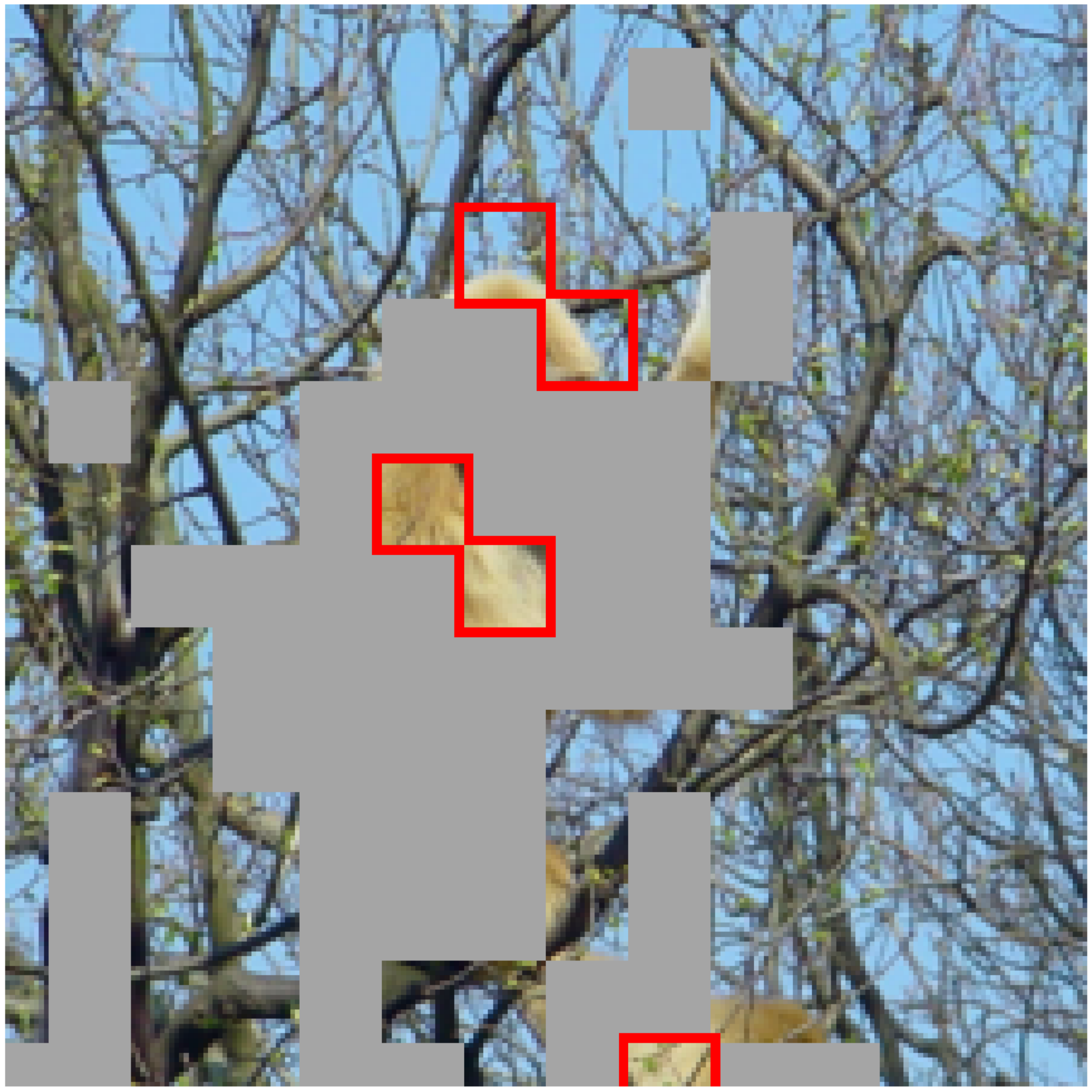}                          &
	\fig[.120]{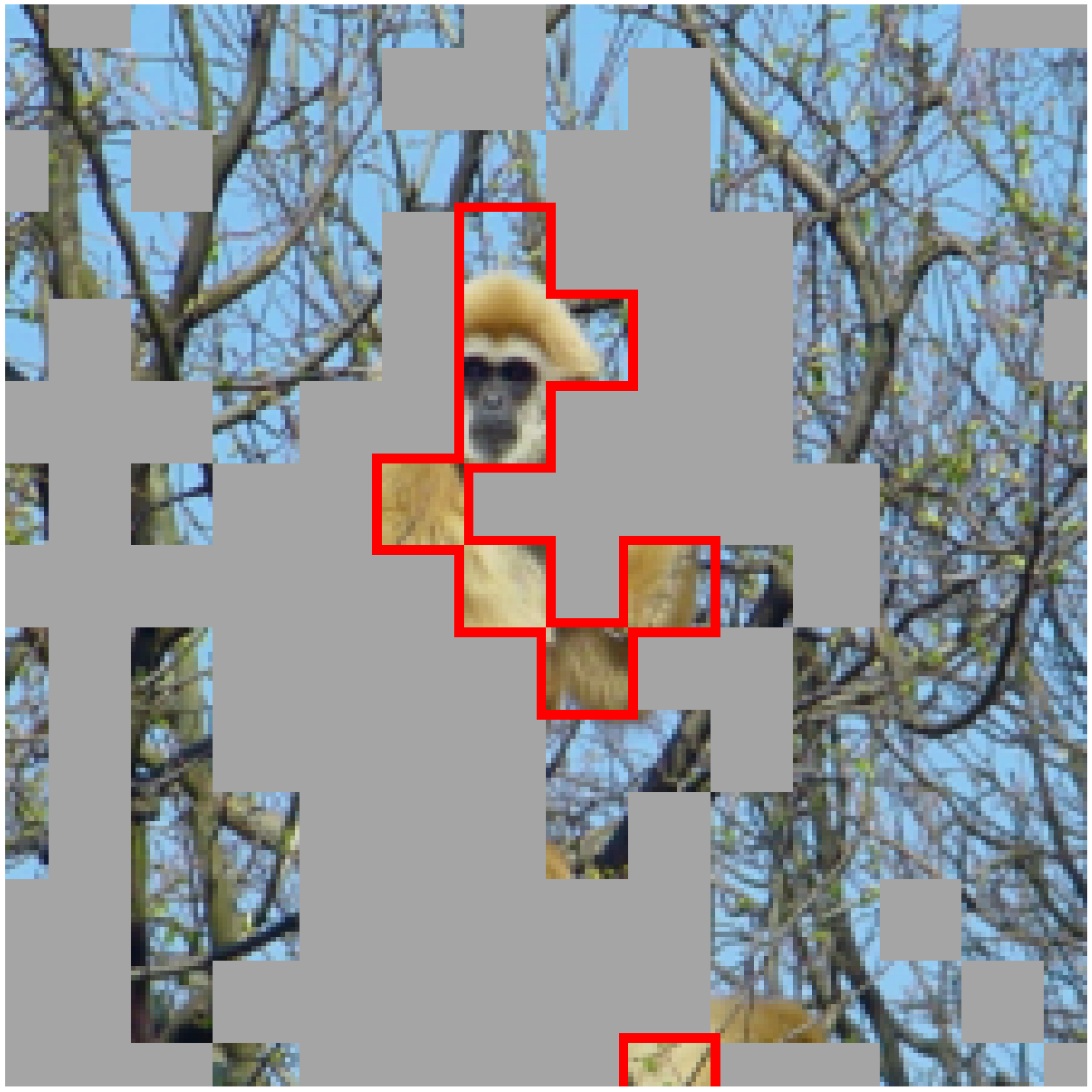}                          &

	&  &

	\fig[.120]{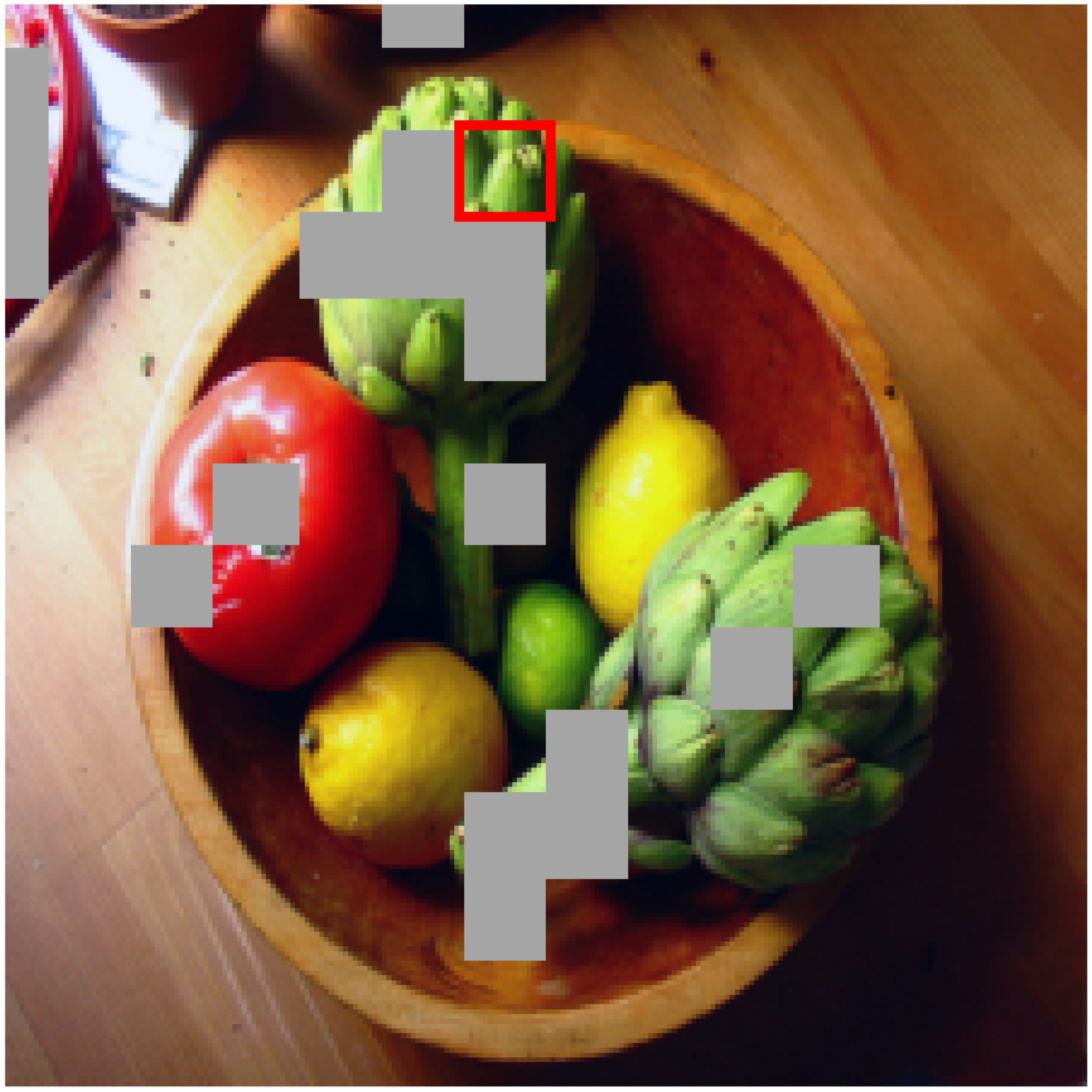}                          &
	\fig[.120]{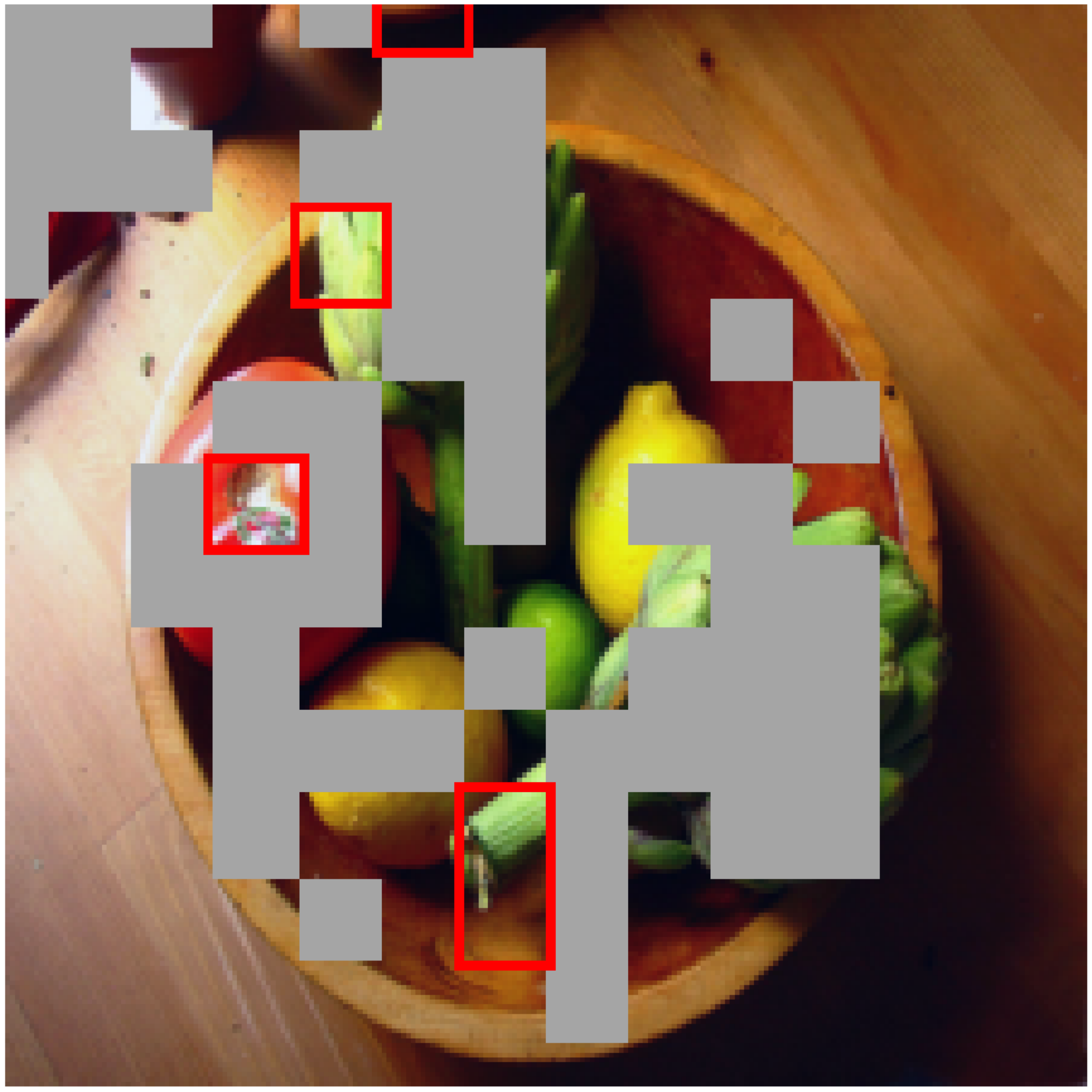}                          &
	\fig[.120]{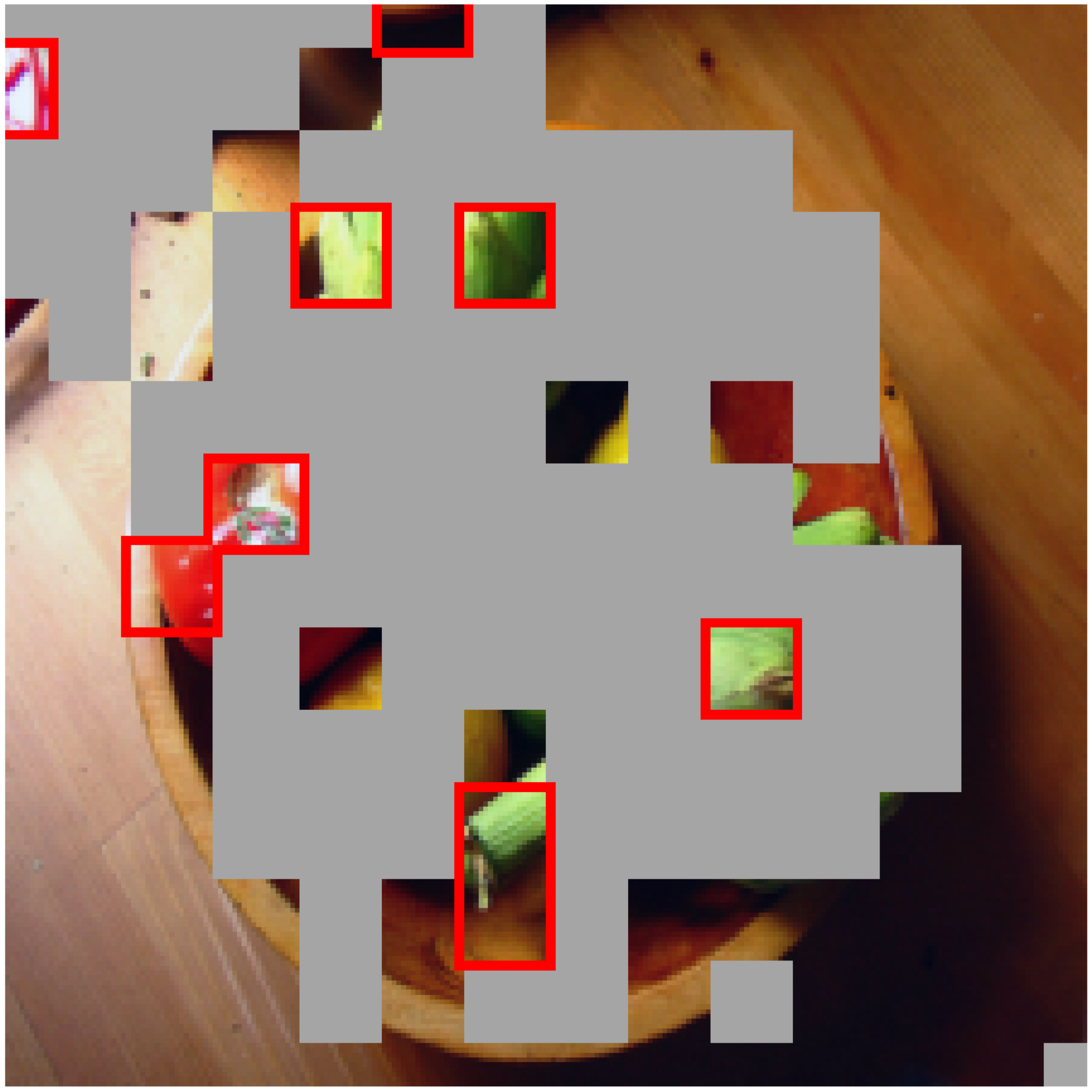}                          \\

	& & & & & & & & \\
	\midrule
	& & & & & & & & \\

	\raisebox{15pt}{Block-Wise}  &
	\fig[.120]{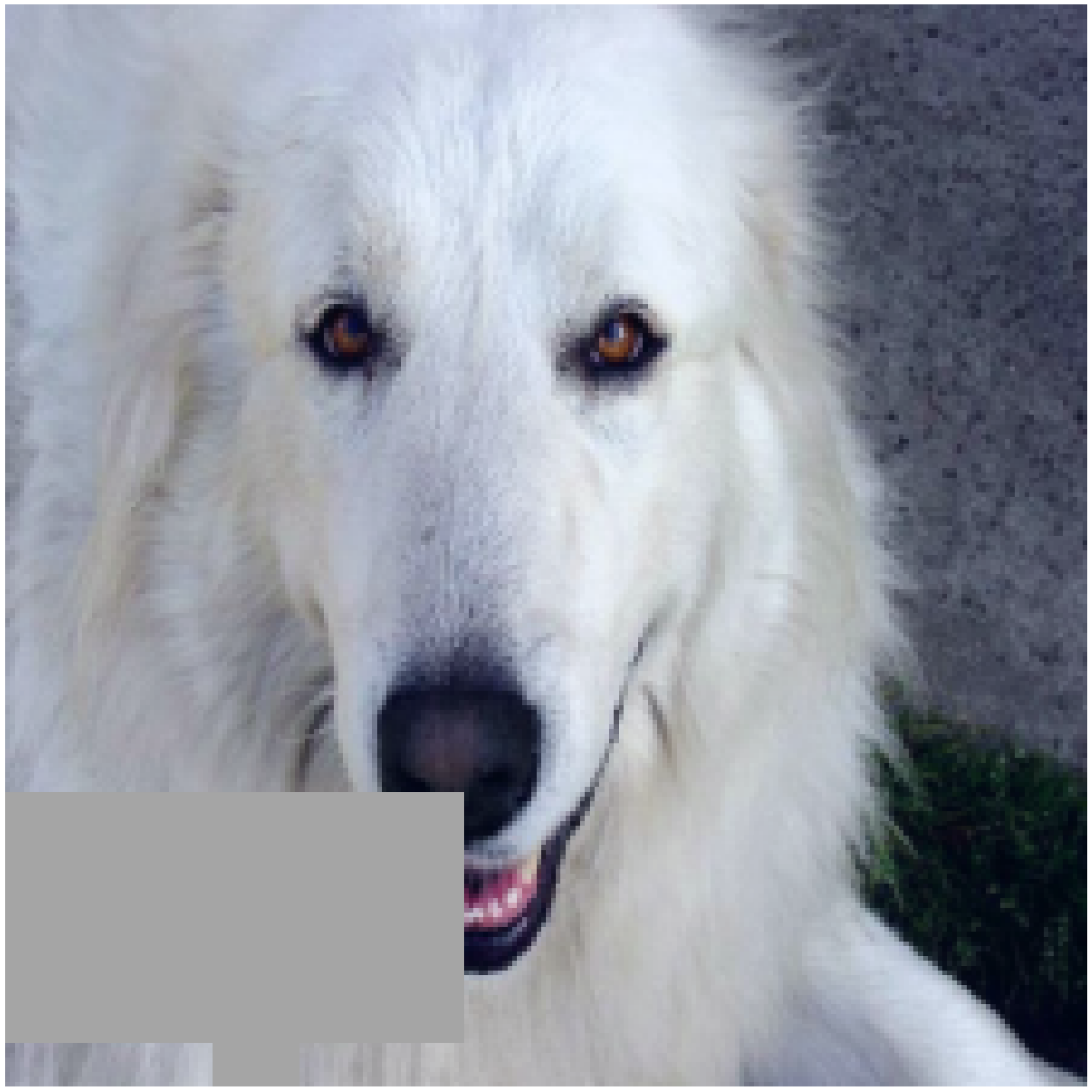}               &
	\fig[.120]{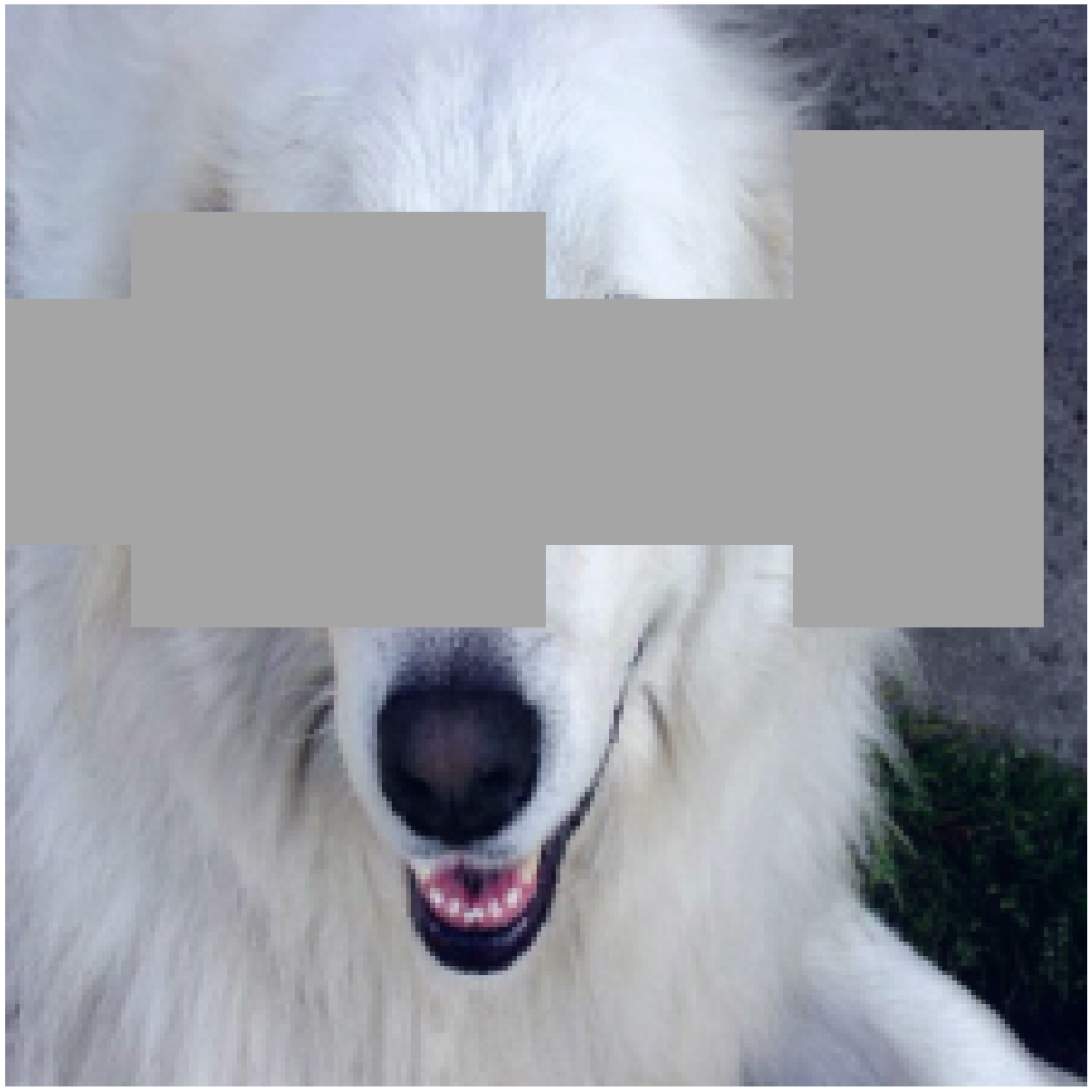}               &
	\fig[.120]{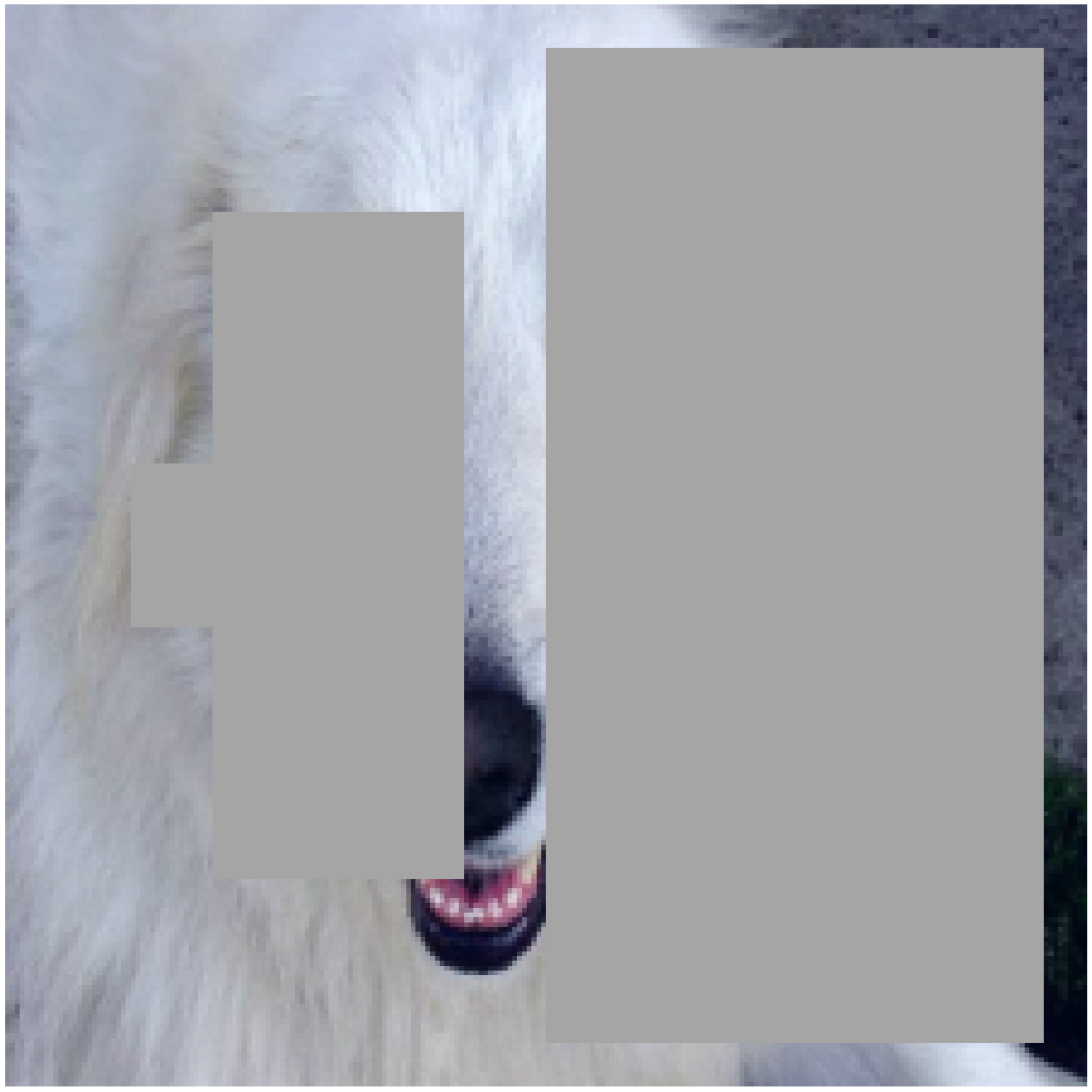}               &

    & &

	\fig[.120]{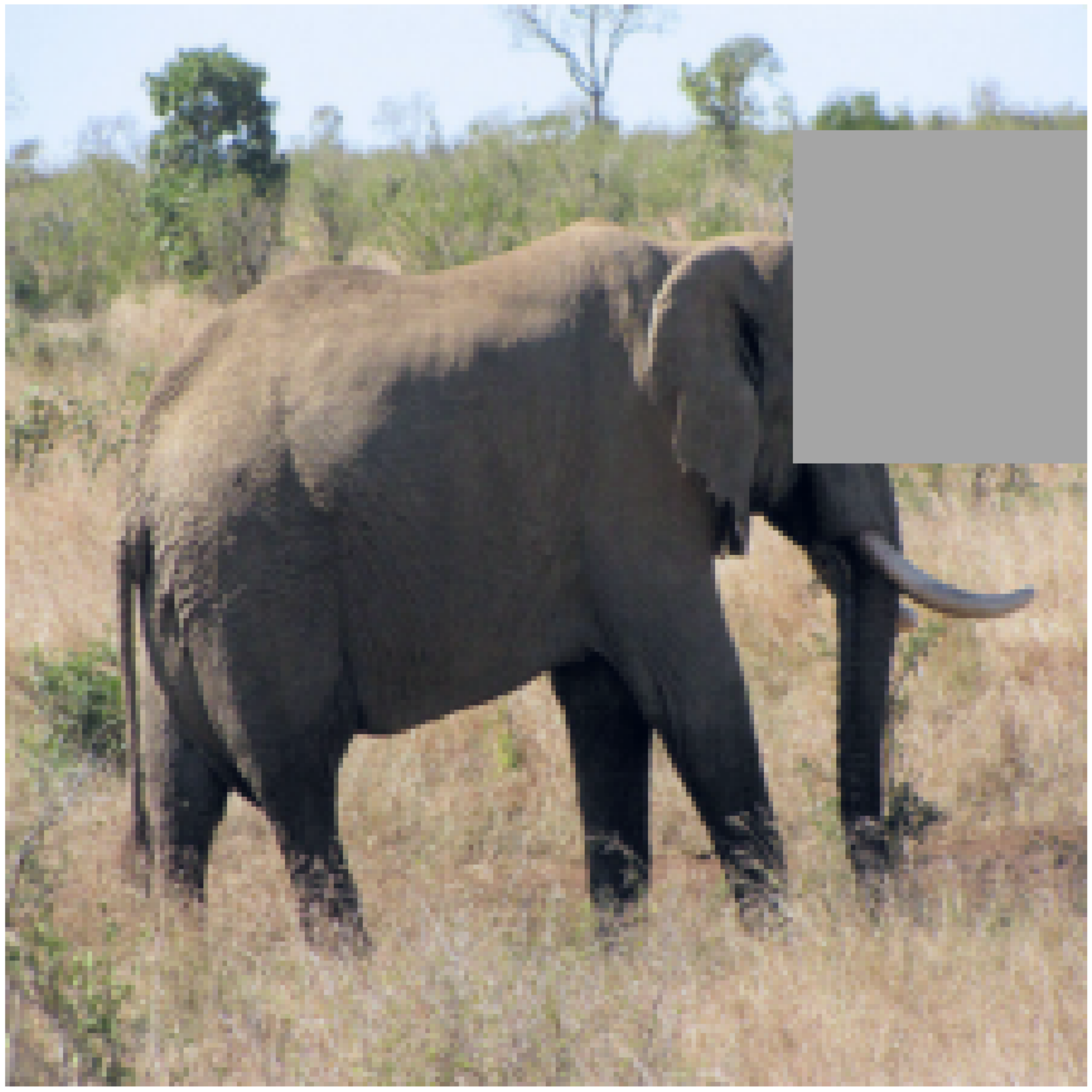}               &
	\fig[.120]{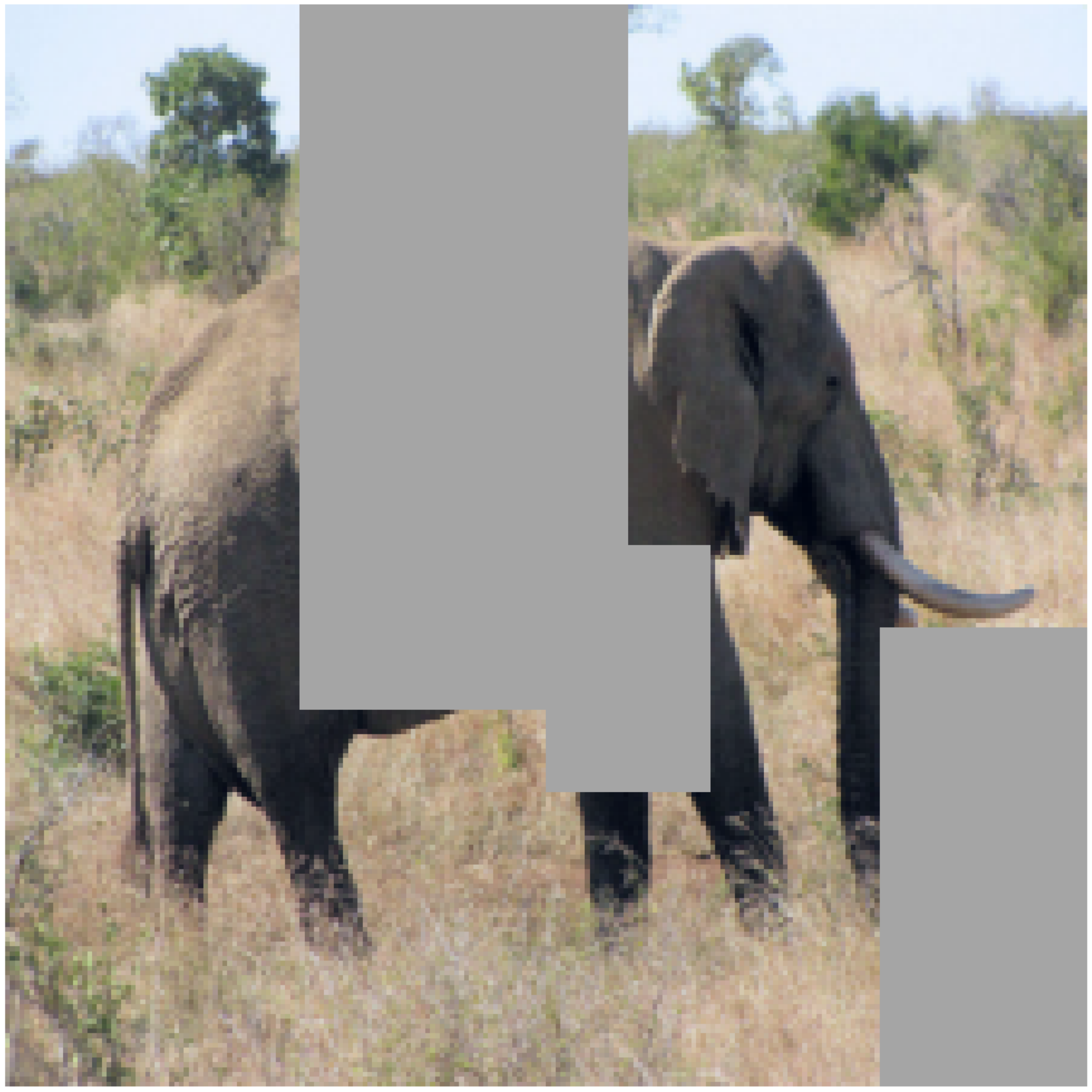}               &
	\fig[.120]{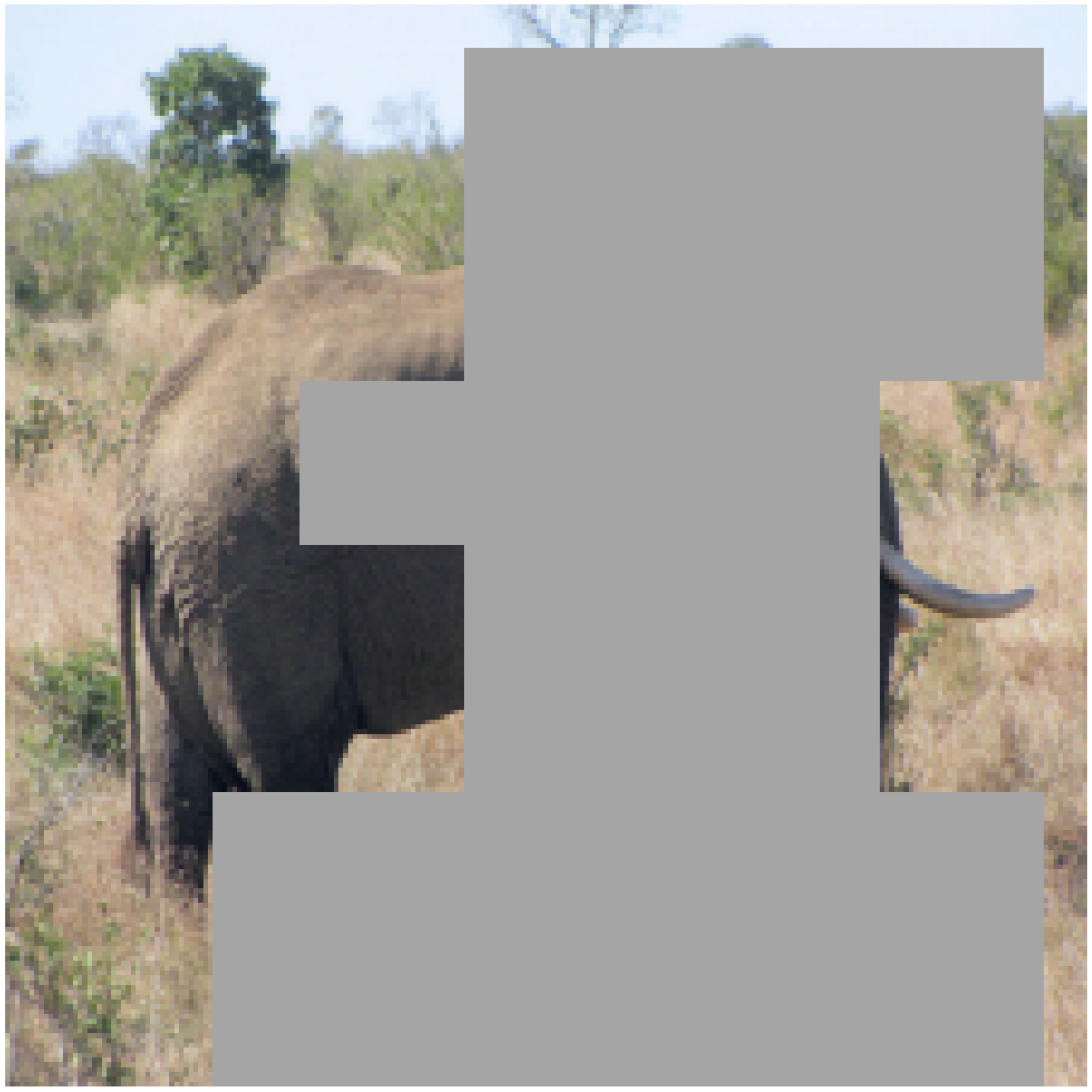}               \\

	\raisebox{15pt}{Random} &
	\fig[.120]{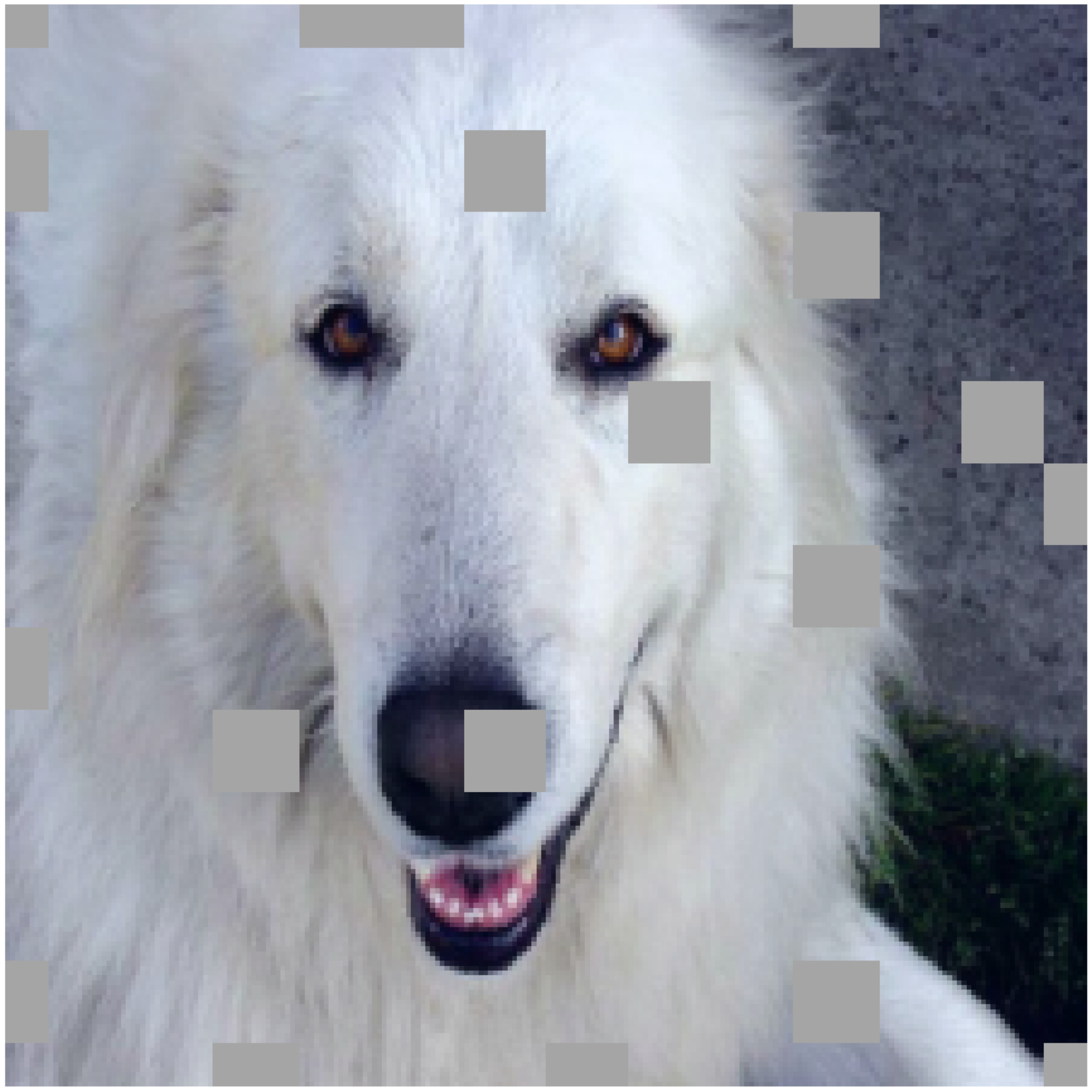}                          &
	\fig[.120]{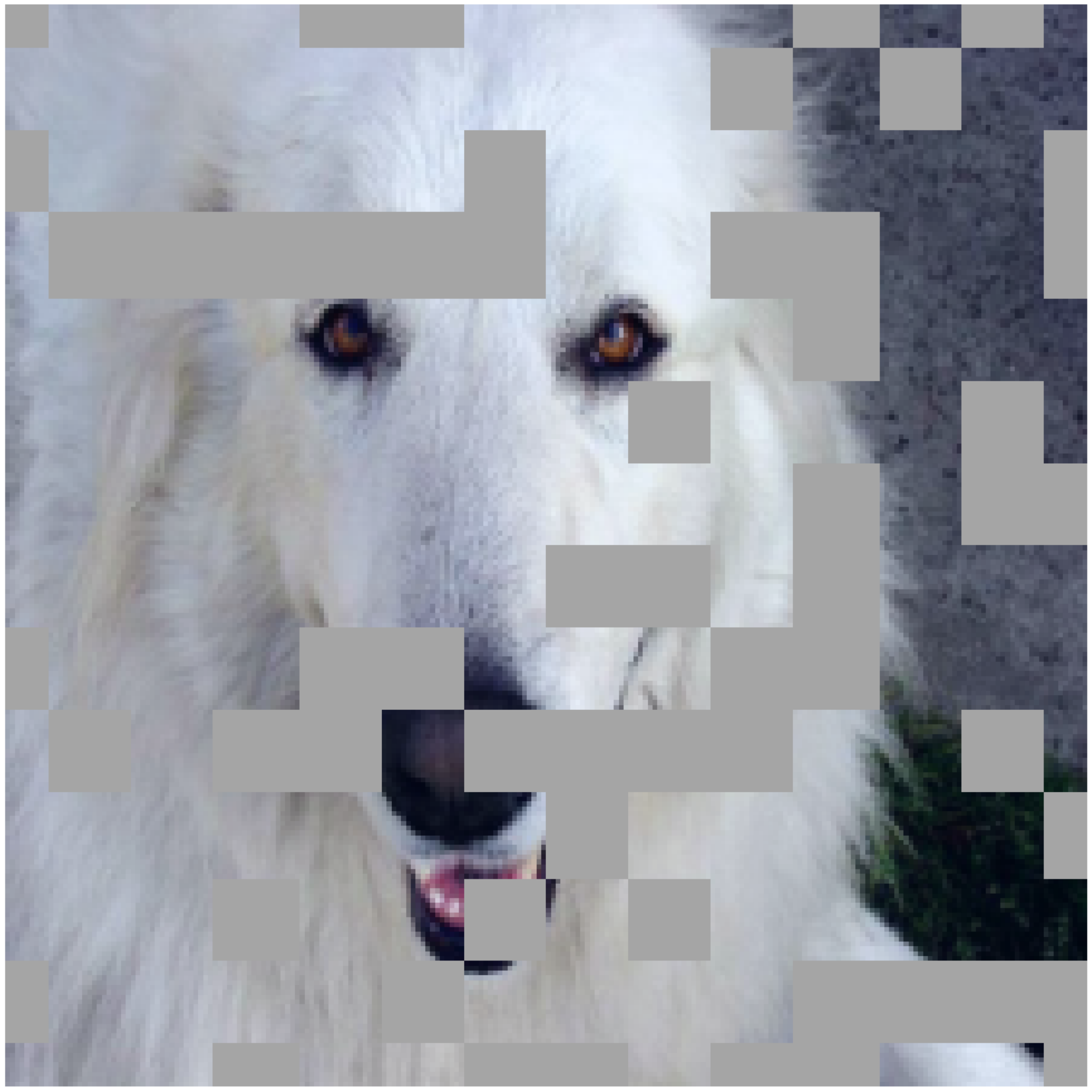}                          &
	\fig[.120]{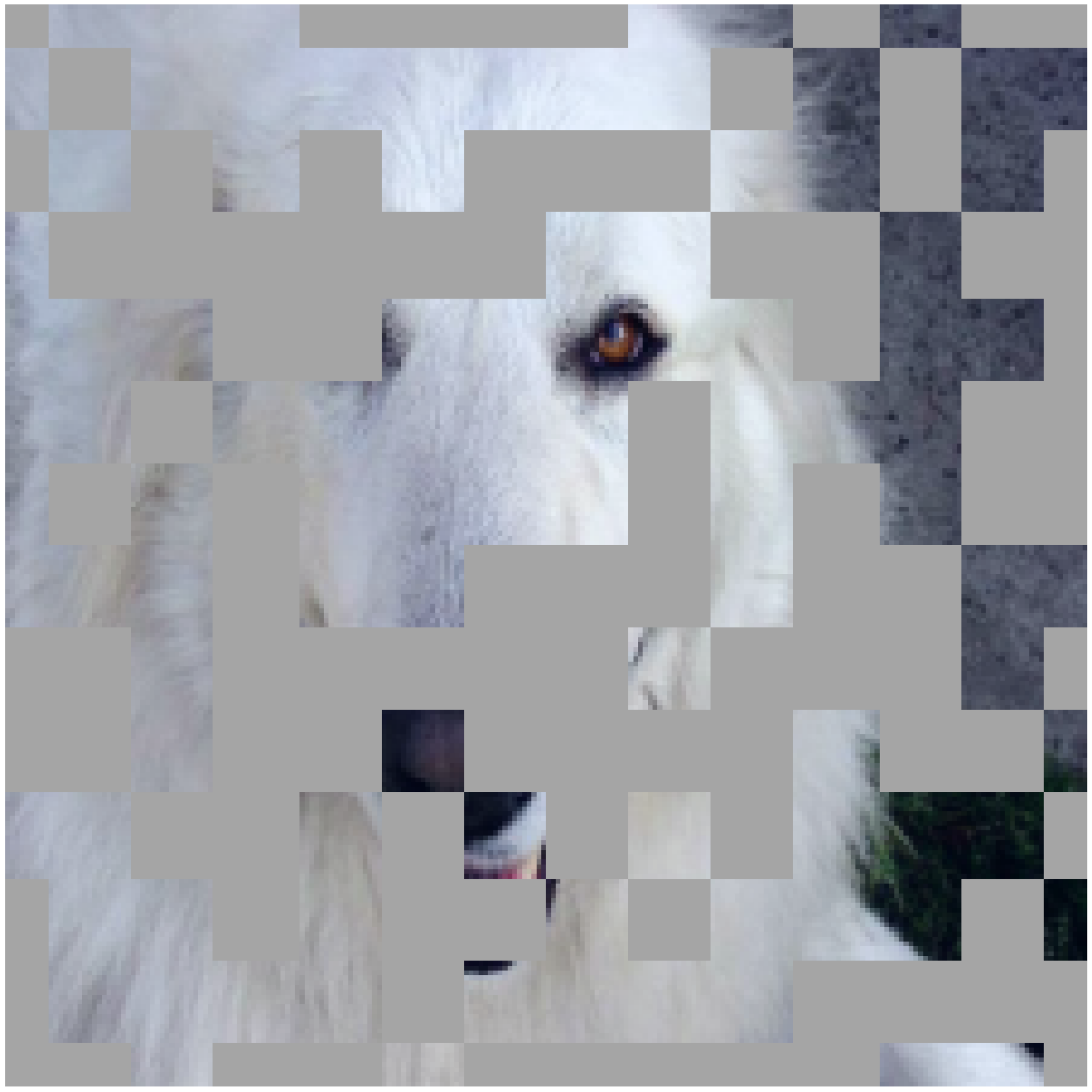}                          &

	& &

	\fig[.120]{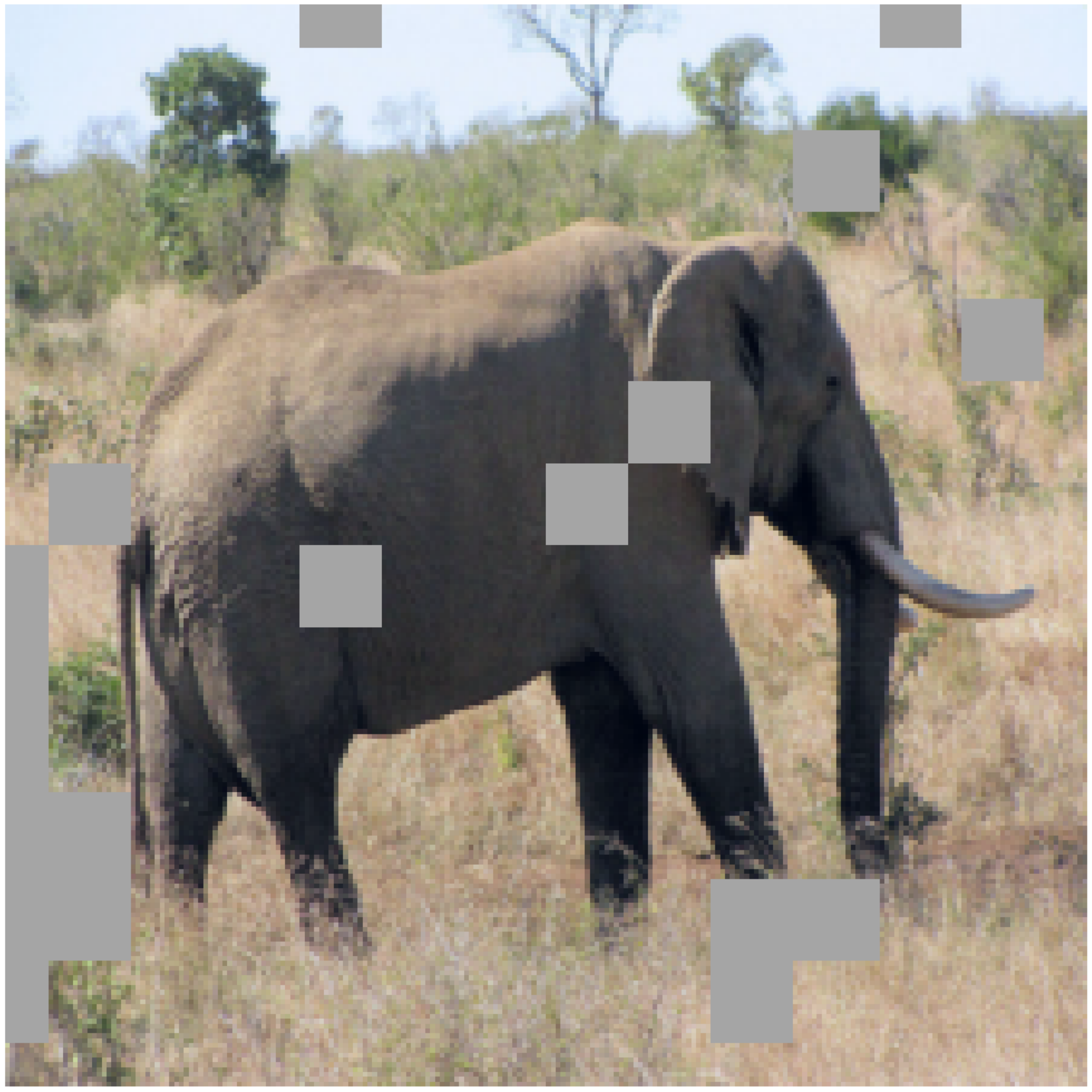}                          &
	\fig[.120]{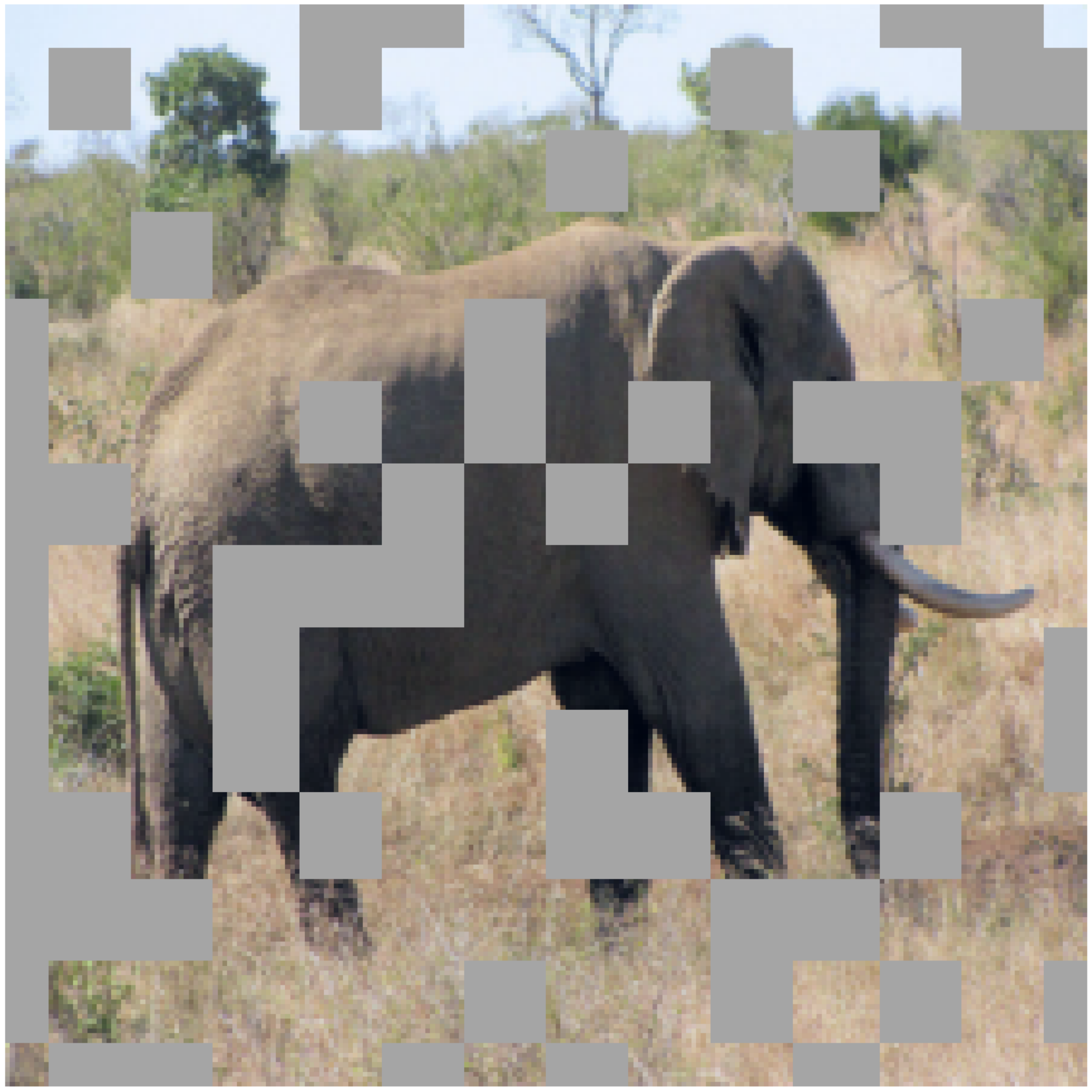}                          &
	\fig[.120]{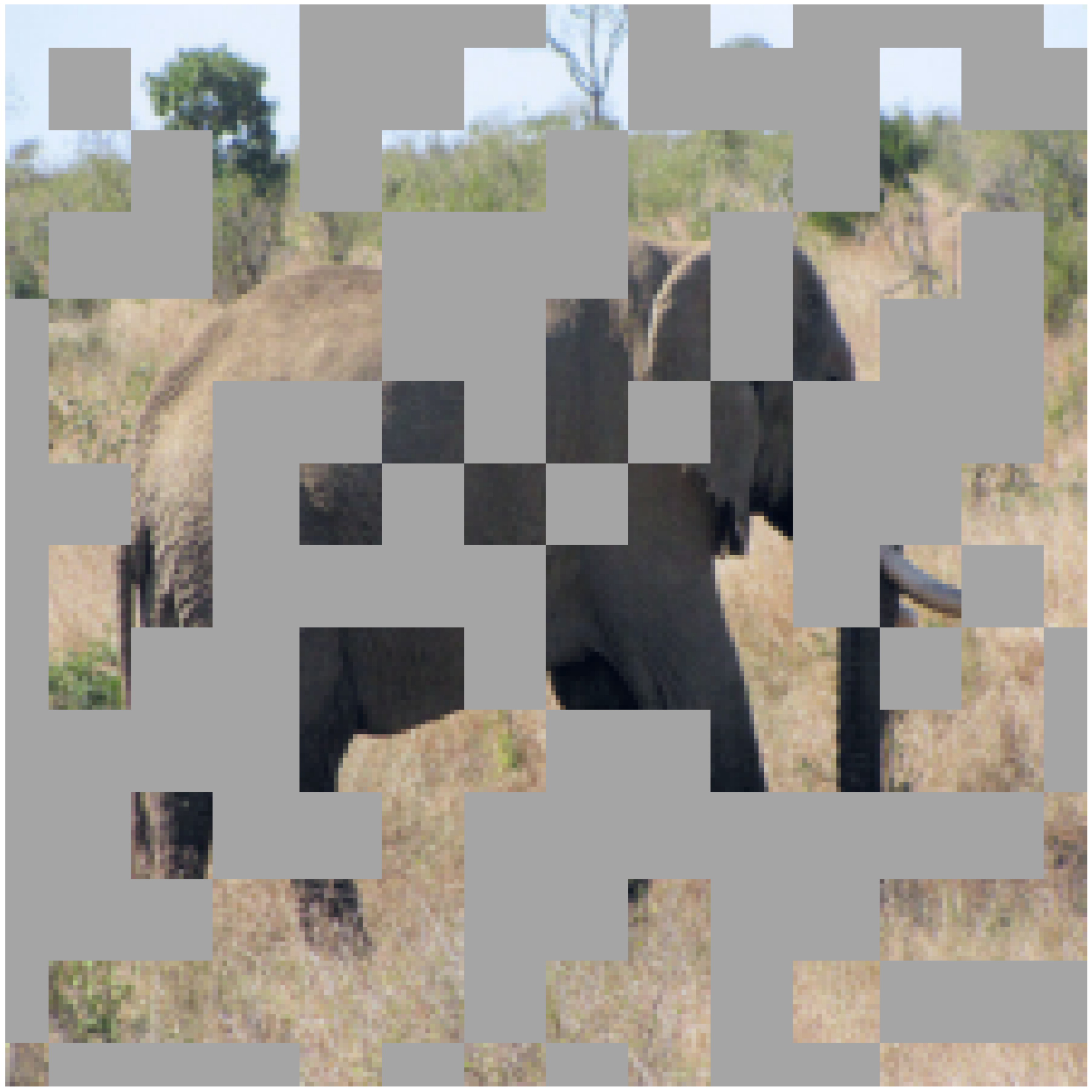}                          \\

	\raisebox{15pt}{\ours-High} &
	\fig[.120]{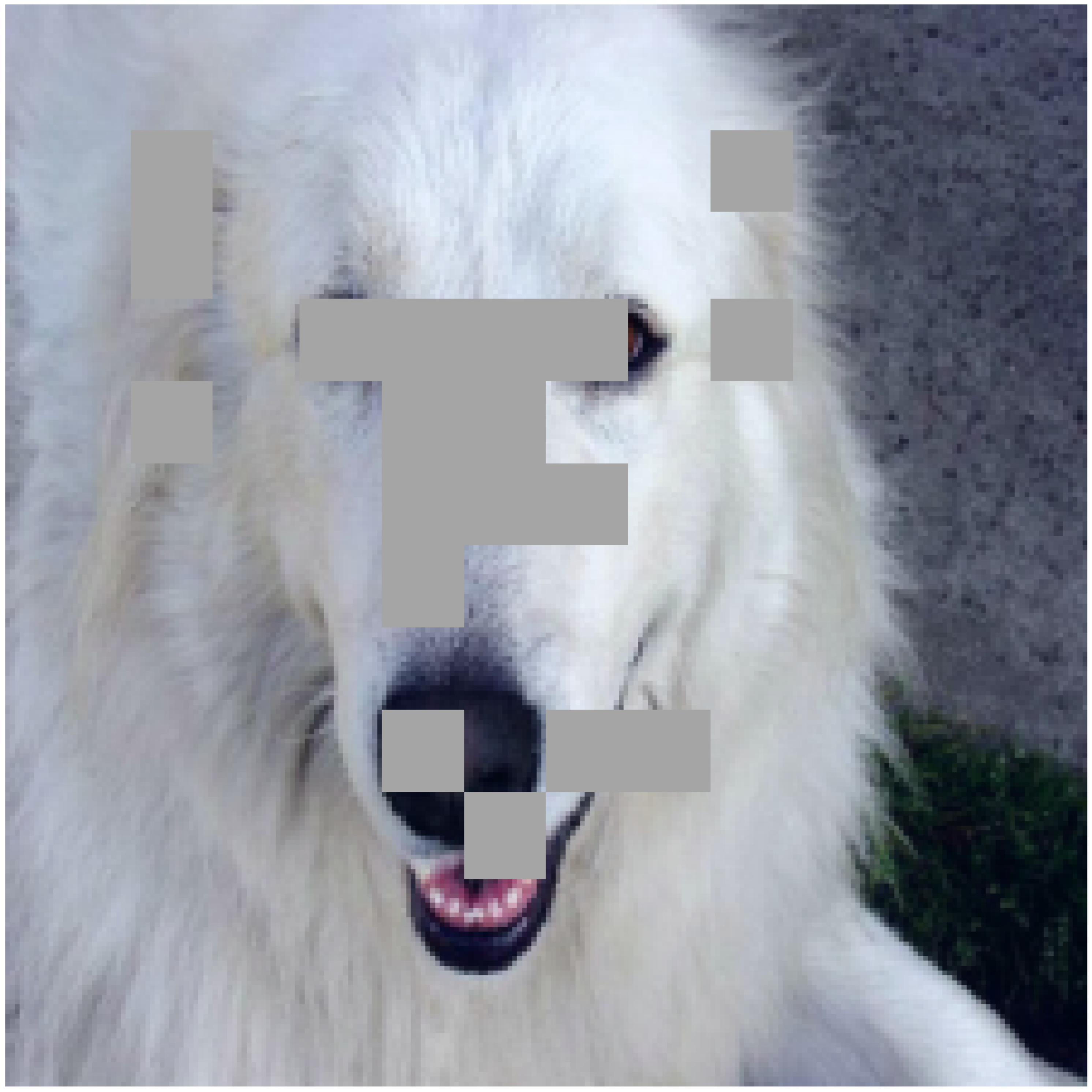}                          &
	\fig[.120]{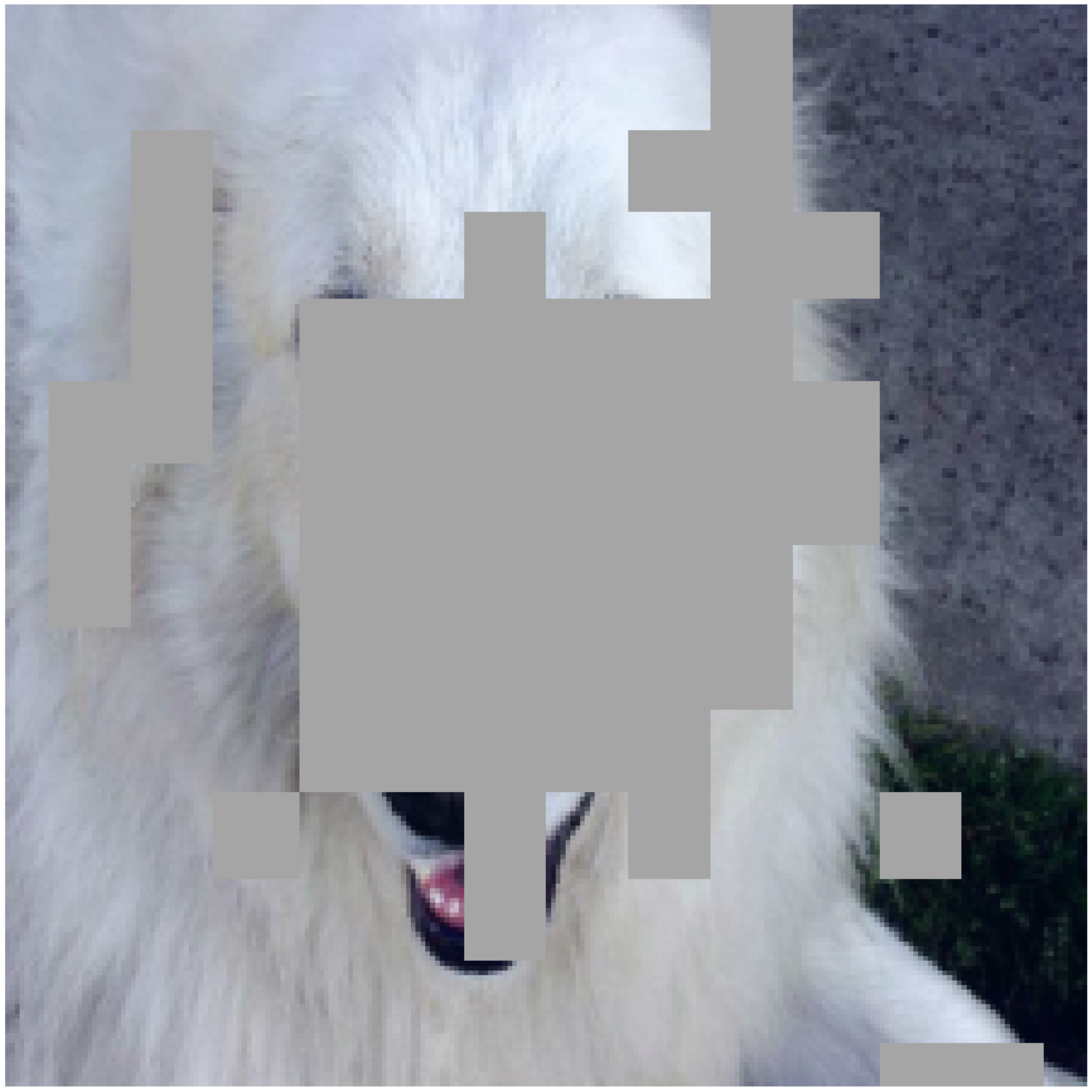}                          &
	\fig[.120]{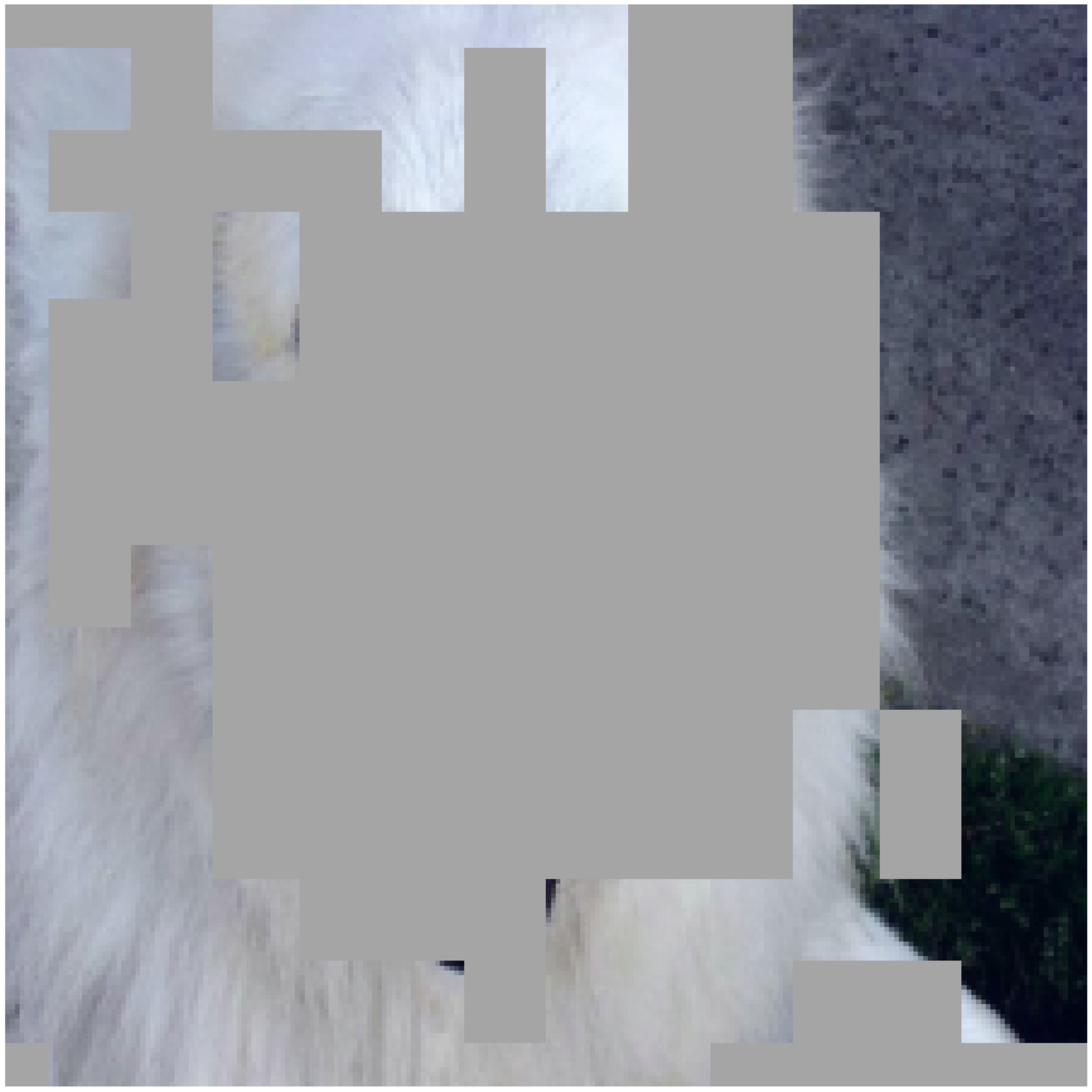}                          &

	& &

	\fig[.120]{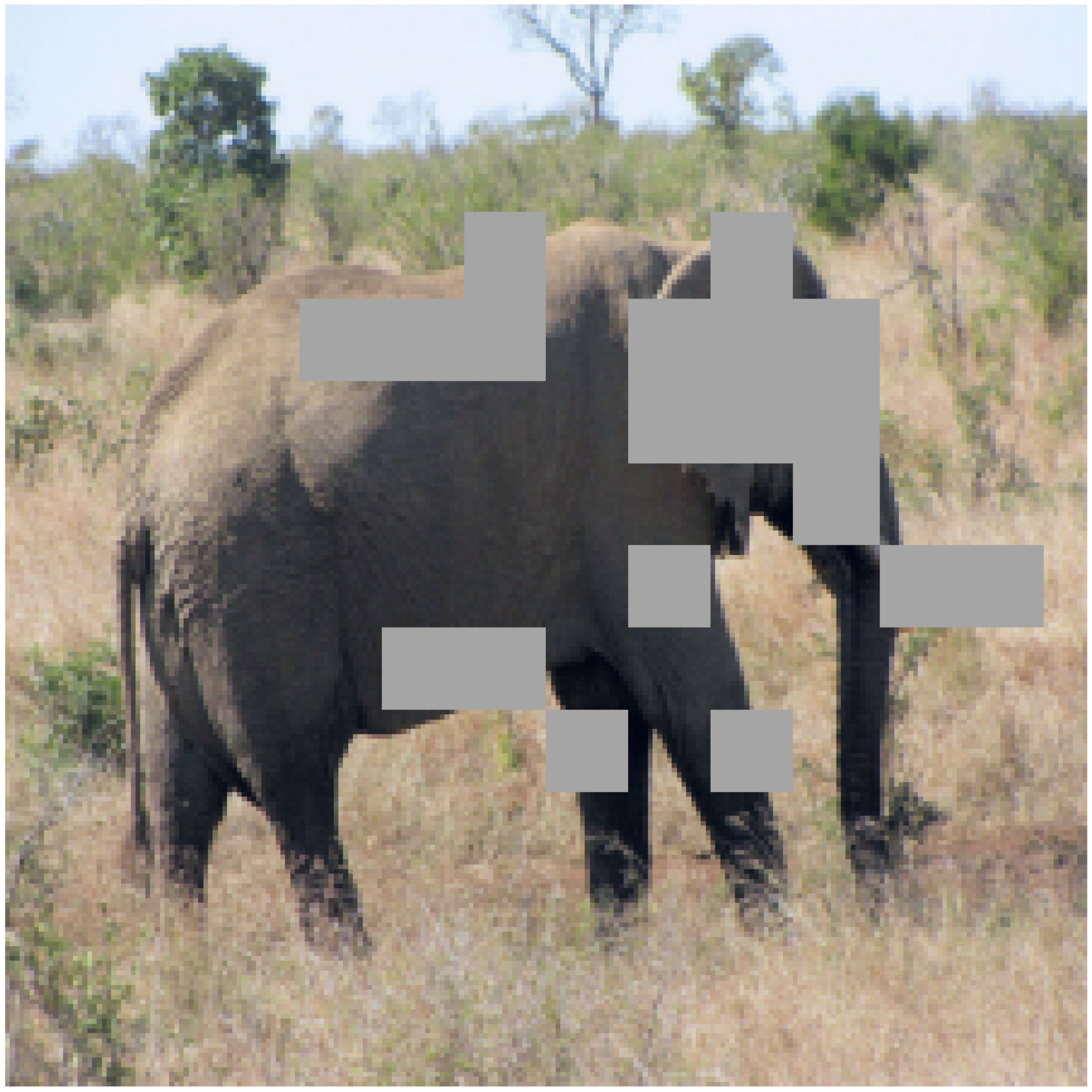}                          &
	\fig[.120]{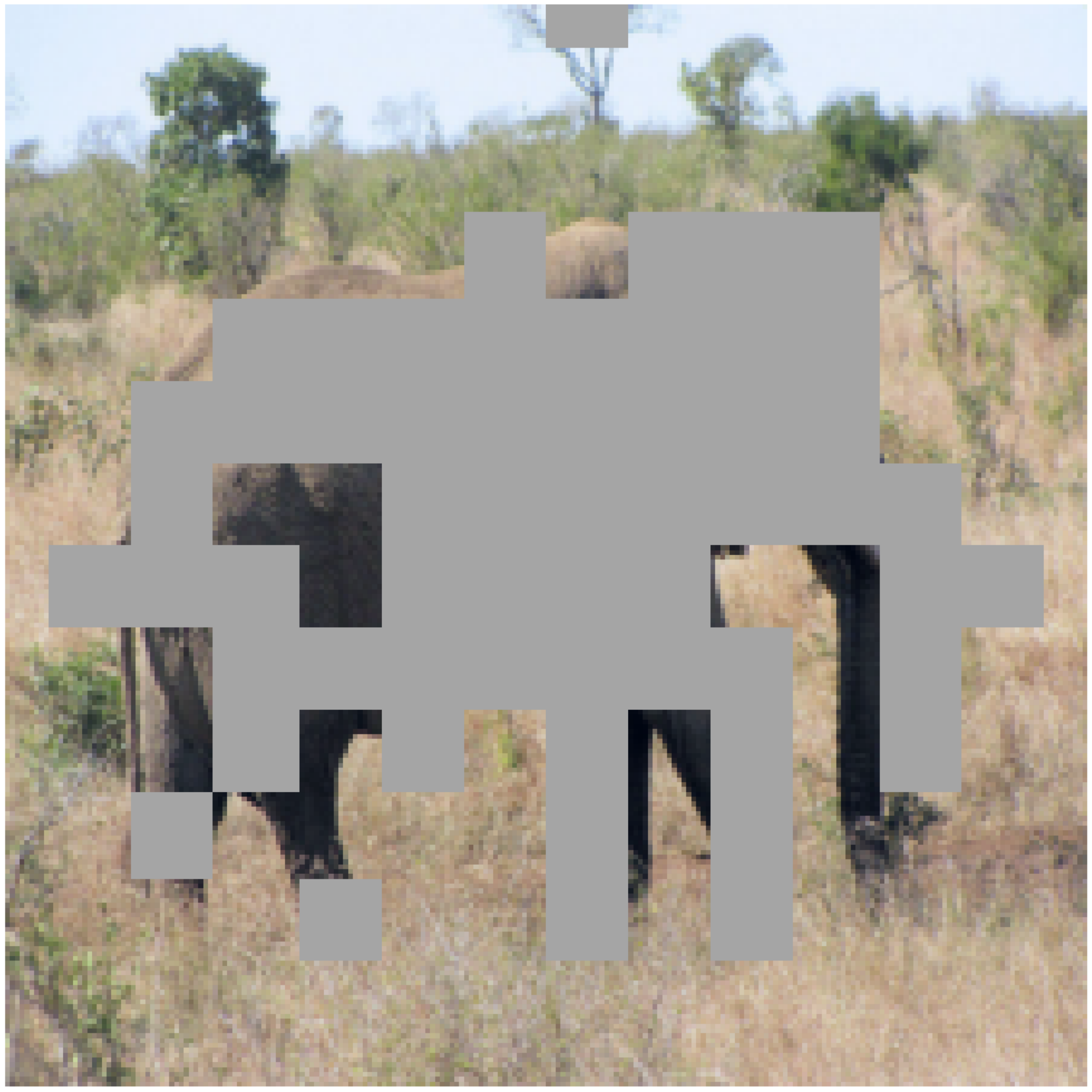}                          &
	\fig[.120]{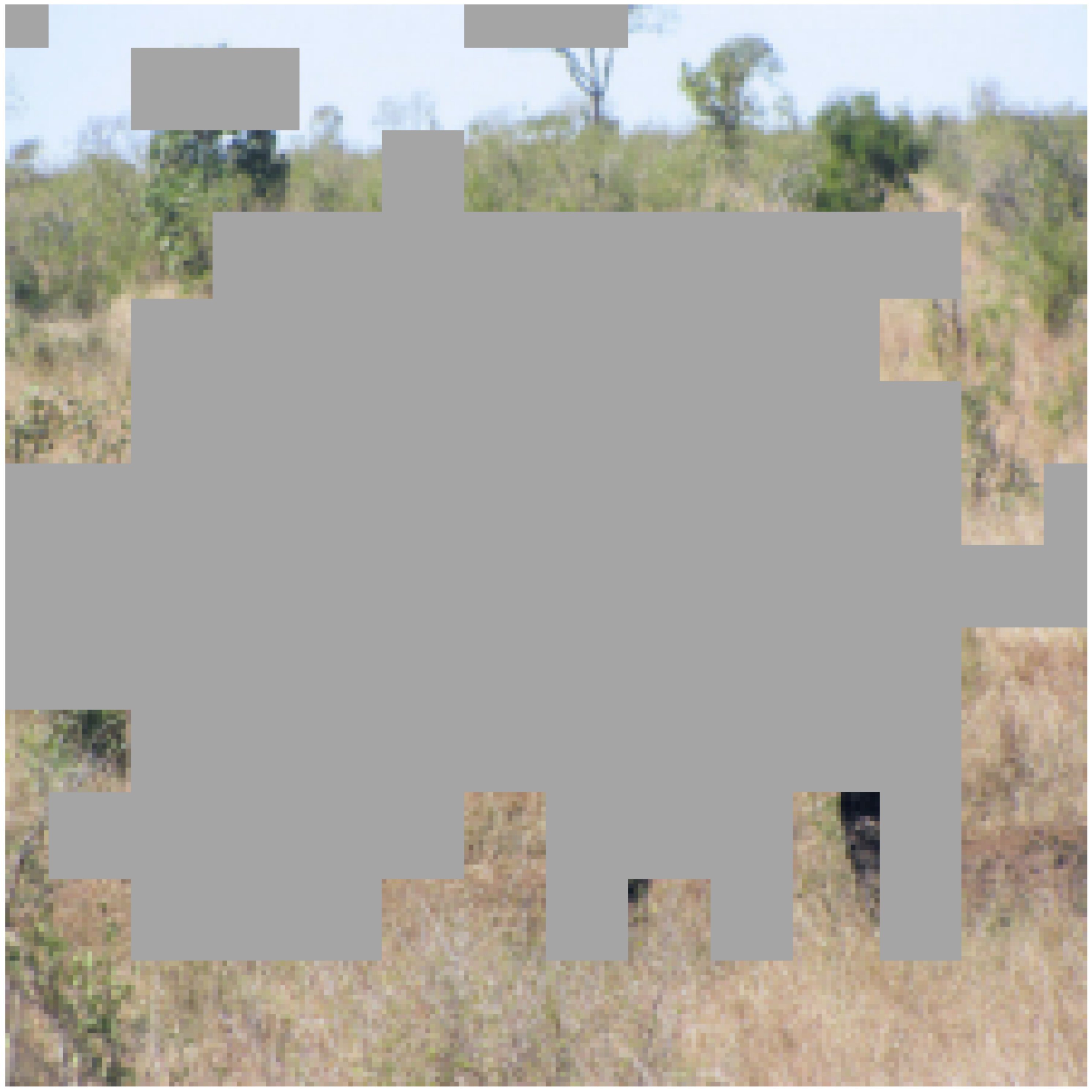}                          \\

	\raisebox{15pt}{\ours-Hint} &
	\fig[.120]{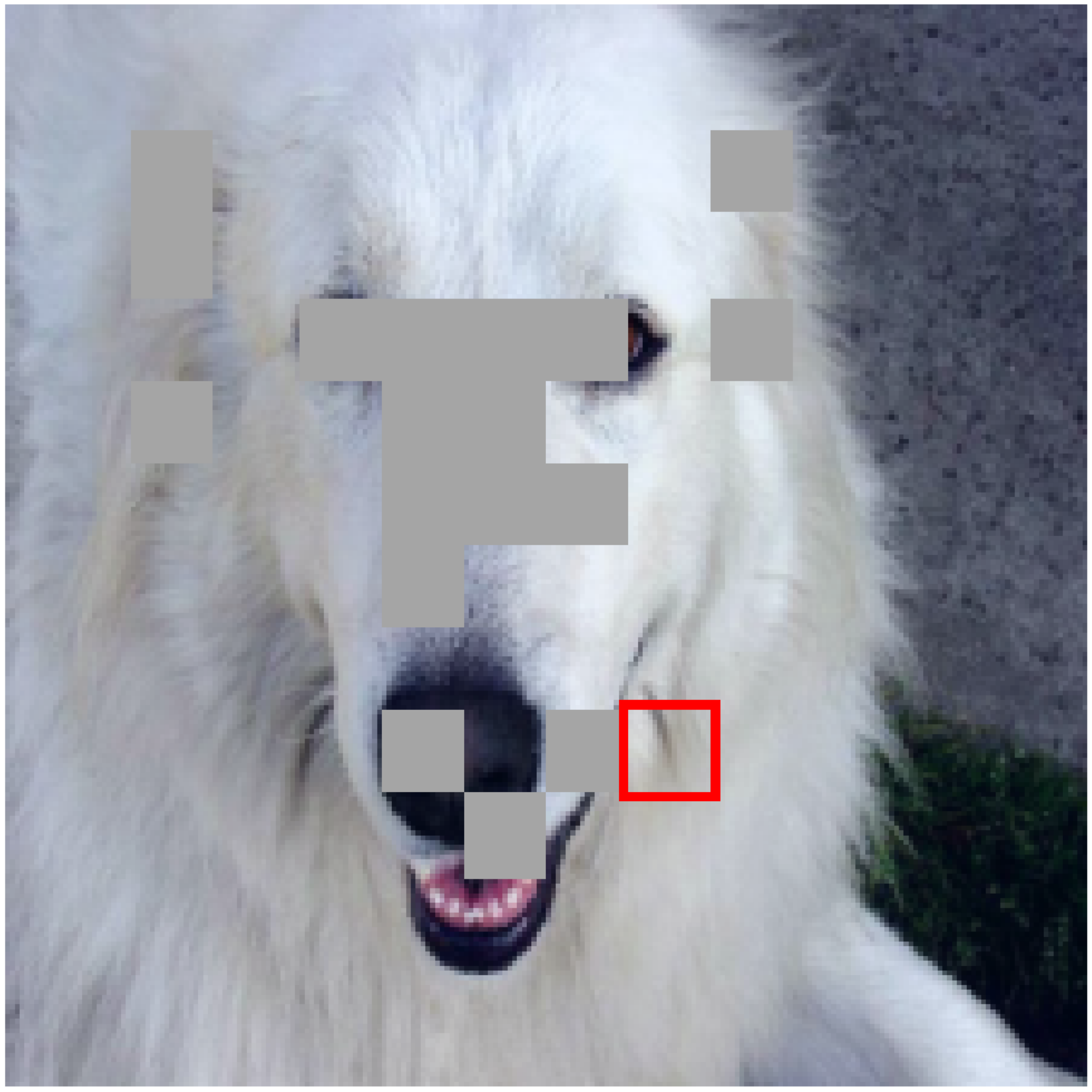}                          &
	\fig[.120]{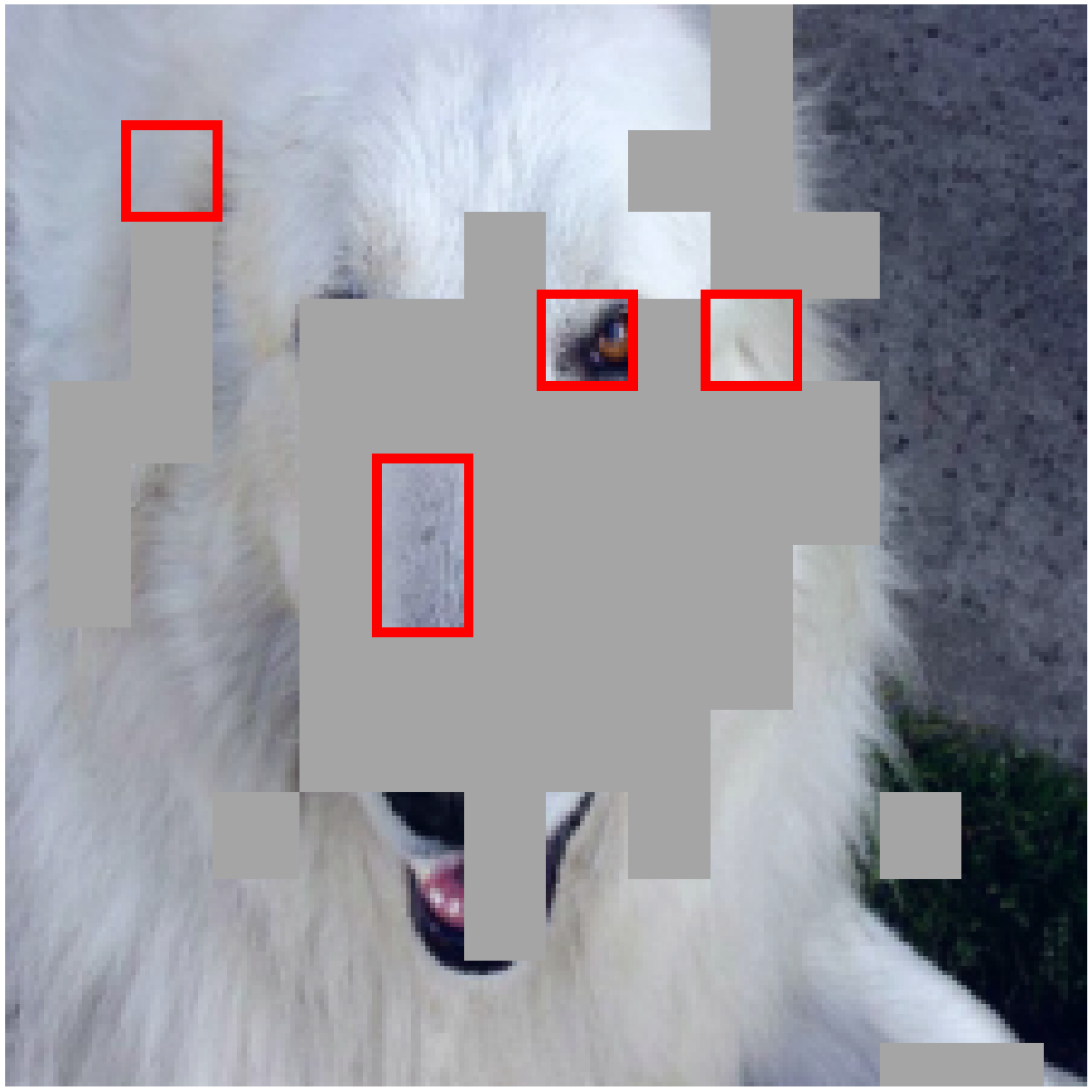}                          &
	\fig[.120]{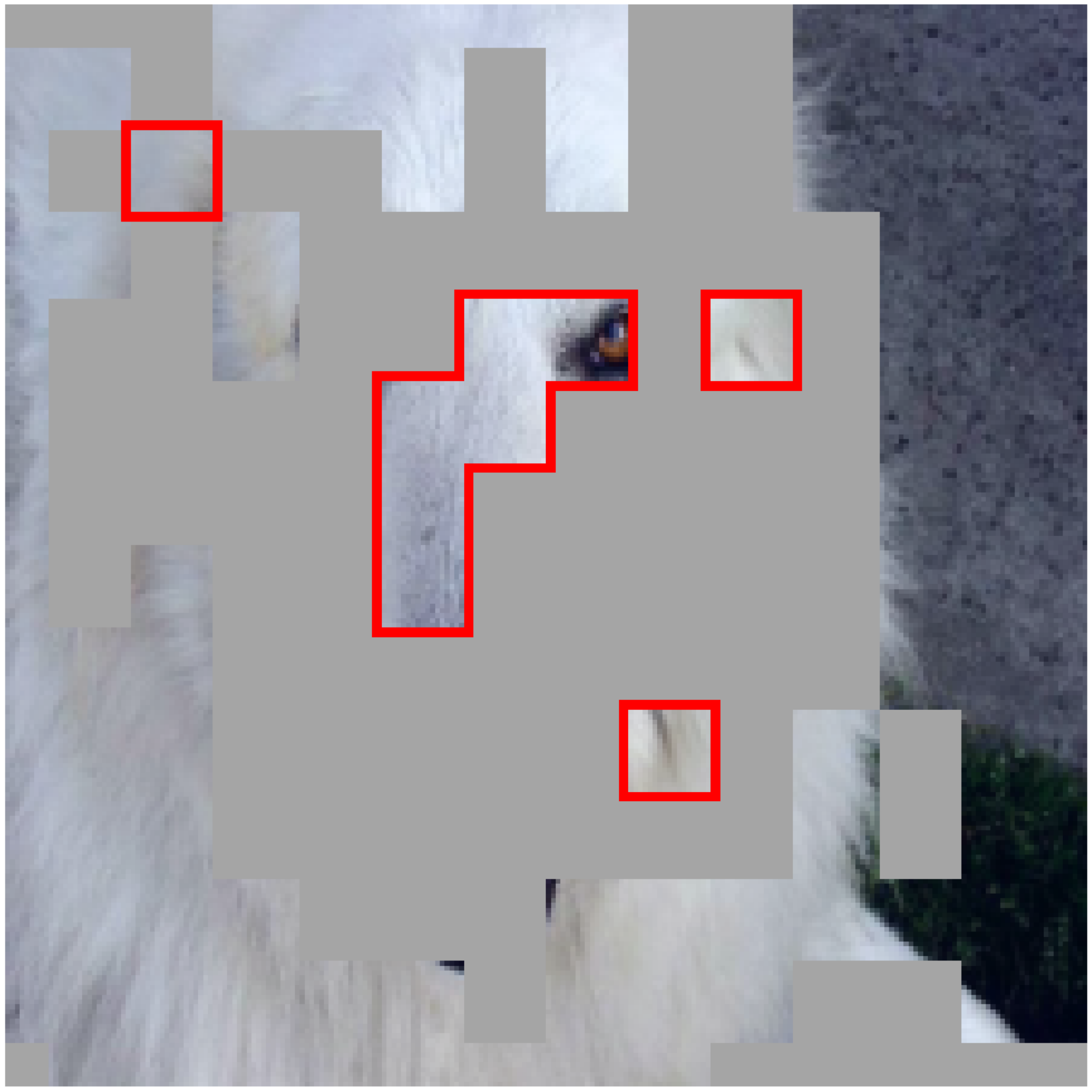}                          &

	&  &

	\fig[.120]{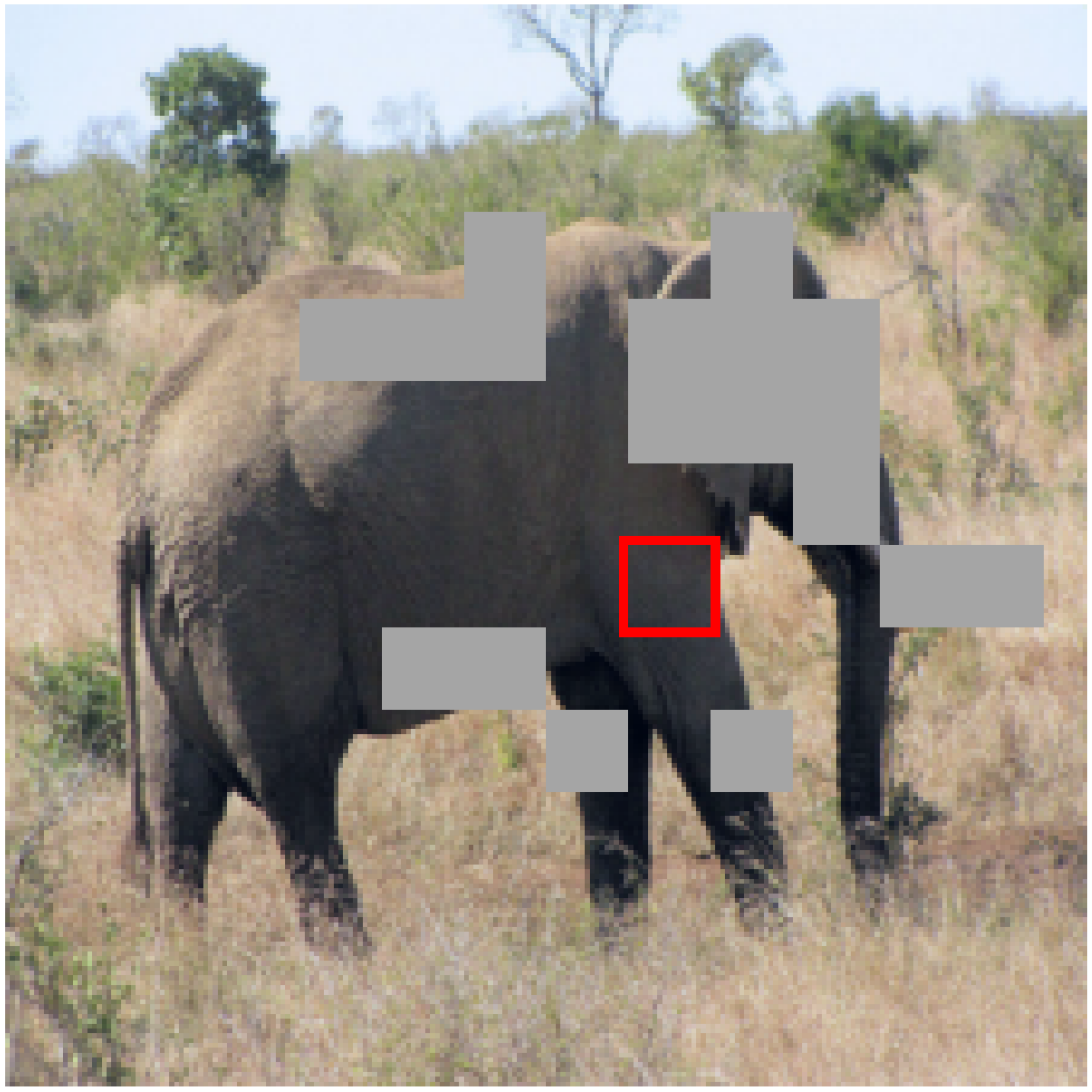}                          &
	\fig[.120]{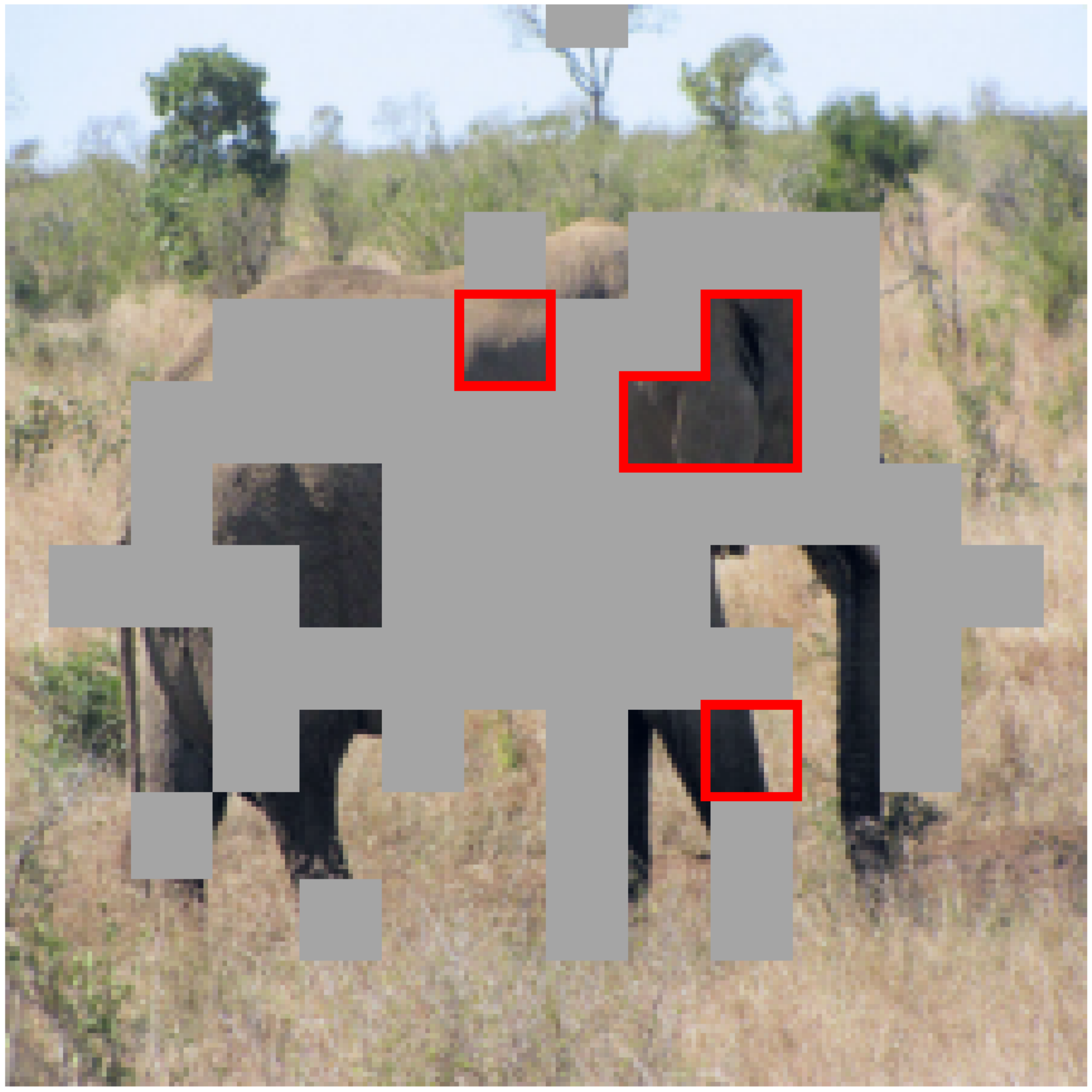}                          &
	\fig[.120]{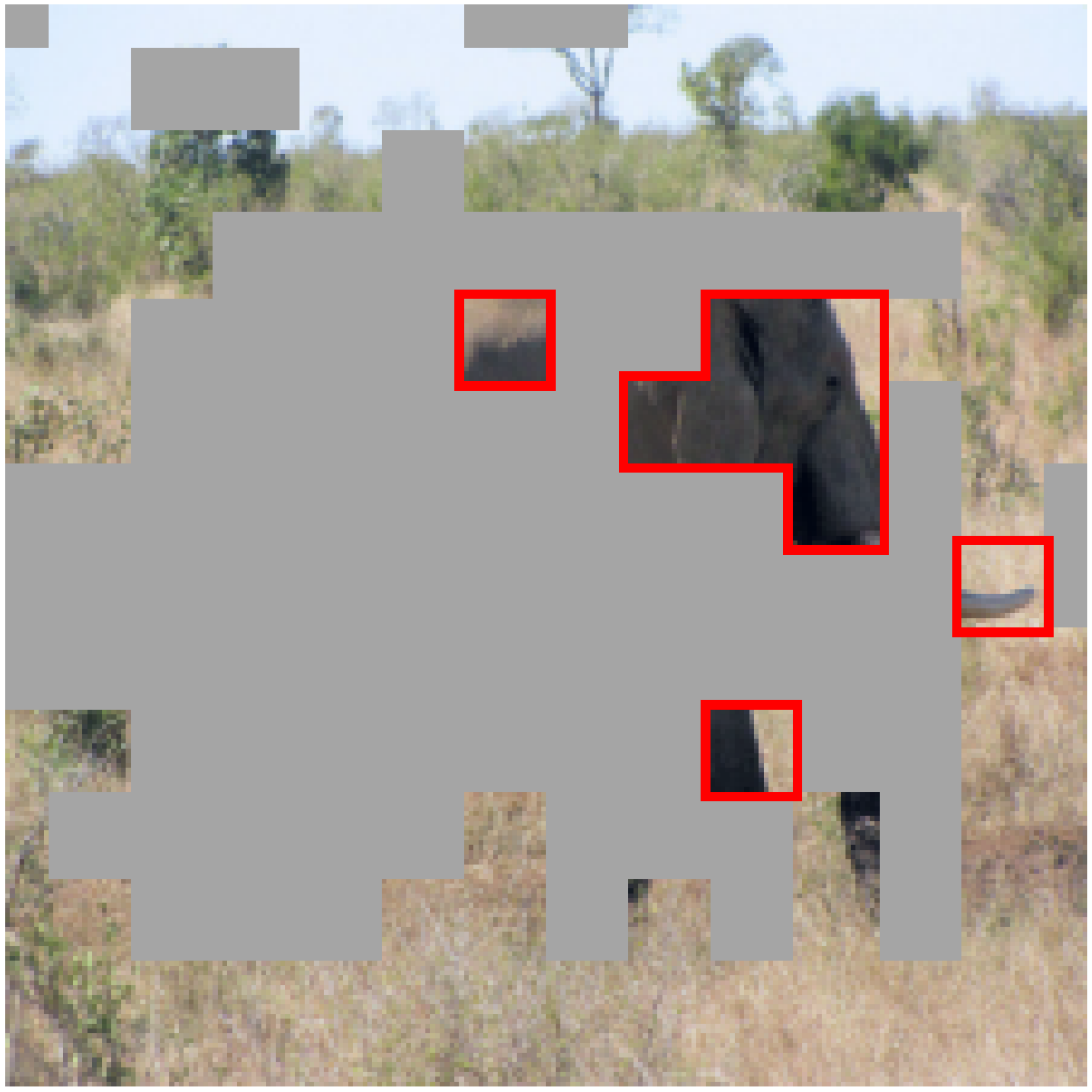}                          \\
\end{tabular}
\caption{Illustration of different masking strategies \vs mask ratio $r$ (\%) (part 2). We compare random Block-Wise masking (BEiT~\cite{beit}) with Random masking (SimMIM~\cite{simmim}), \ours-High and \ours-Hint. Our \ours-High uses the attention map arising in the encoder to hide patches, while \ours-Hint reveals very salient patches to leave hints about the identity of the masked object.}
\label{fig:MoreStrategies3}
\end{figure}